\documentclass[
    10pt,
    twocolumn,
    letterpaper,
    pagebackref,
    breaklinks,
    colorlinks,
    allcolors=cvprblue
]{article}

\usepackage{cvpr}              

\usepackage{url}            
\usepackage{booktabs}       
\usepackage{amsfonts}       
\usepackage{nicefrac}       
\usepackage{microtype}      

\usepackage{graphicx}
\usepackage{doi}
\usepackage{caption}
\usepackage{array}
\usepackage{longtable}
\usepackage{subcaption}
\usepackage[table,x11names,svgnames]{xcolor}
\usepackage{multirow}
\usepackage{array}
\usepackage{amssymb}
\usepackage{verbatim}
\usepackage{multicol}
\usepackage{makecell}
\usepackage{pdflscape}
\usepackage{rotating}
\usepackage{listings}
\usepackage{caption}
\usepackage{tabularx}
\usepackage{threeparttable} 
\usepackage{siunitx}
\usepackage{hyperref}
\usepackage{float}

\usepackage{hyperref}
\usepackage{xurl} 

\hypersetup{
    linkcolor = {RoyalBlue4},
    citecolor = {RoyalBlue4},
}

\newcolumntype{C}[1]{>{\centering\arraybackslash}m{#1}}

\usepackage{minitoc}
\usepackage[toc,page,header]{appendix}

\definecolor{cvprblue}{rgb}{0.21,0.49,0.74}
\usepackage{booktabs}
\usepackage{amssymb}
\def\paperID{*****} 
\def\confName{CVPR}
\def\confYear{2026}

\title{The \textit{DeepSpeak} Dataset}

\author{%
Sarah Barrington$^{1}$, Maty Bohacek$^{2}$, Hany Farid$^{1}$
\vspace{2pt}\\
$^1$University of California, Berkeley, $^2$Stanford University
}

\begin{document}
\maketitle

\doparttoc 
\faketableofcontents

\maketitle

\begin{abstract}
    Deepfakes represent a growing concern across domains such as disinformation, fraud, and non-consensual media. In particular, the rise of video conference and identity-driven attacks in high-stakes scenarios--such as impostor hiring--demands new forensic resources. Despite significant efforts to develop robust detection classifiers to distinguish the real from the fake, commonly used training datasets remain inadequate: relying on low-quality and outdated deepfake generators, consisting of content scraped from online repositories without participant consent, lacking in multimodal coverage, and rarely employing identity-matching protocols to ensure realistic fakes. To overcome these limitations, we present the DeepSpeak dataset, a diverse and multimodal dataset comprising over 100 hours of authentic and deepfake audiovisual content, specifically focused on the challenging and diverse ``talking heads'' context. We contribute: i) more than 50 hours of real, self-recorded data collected from 500 diverse and consenting participants, ii) more than 50 hours of state-of-the-art audio and visual deepfakes generated using 14 video synthesis engines and three voice cloning engines, and iii) an embedding-based, identity-matching approach to ensure the creation of convincing, high-quality identity face swaps that realistically simulate adversarial deepfake attacks. We also perform large-scale evaluations of state-of-the-art deepfake detectors and show that, without retraining, these detectors fail to generalize to this DeepSpeak dataset, highlighting the importance of a large and diverse dataset containing deepfakes from the latest generative-AI tools.

    \vspace{1em}
    \noindent\small{\url{https://github.com/hfaridlab/deepspeak}}
    
\end{abstract}

\section{Introduction}
 
Today, generative-AI is capable of creating hyper-realistic images~\cite{nightingale2022ai}, voices~\cite{barrington2025people}, and videos~\cite{groh2022deepfake} of people talking or doing just about anything. These technologies hold the promise to both revolutionize many industries while also amplifying the spread and belief in dangerous lies and conspiracies~\cite{chesney2019deep,vaccari2020deepfakes}, interfering with elections~\cite{ferrara2024charting,stockwell2024ai}, super-charging small- and large-scale fraud~\cite{bateman2022deepfakes}, and -- seemingly unable to escape its roots -- continue to be used in the creation of non-consensual intimate imagery (NCII)~\cite{ding2025malicious,viola2023designed}.

Scalable, generalizable, and accurate detection of deepfakes has, therefore, become a pressing problem with deep social, political, and economic implications. At the same time, the nascent digital-forensics community has struggled with the lack of large-scale, high-quality, up-to-date, and ethically collected datasets for training and evaluation. 

\begin{figure*}[t!]
\centering
\includegraphics[width=\linewidth]{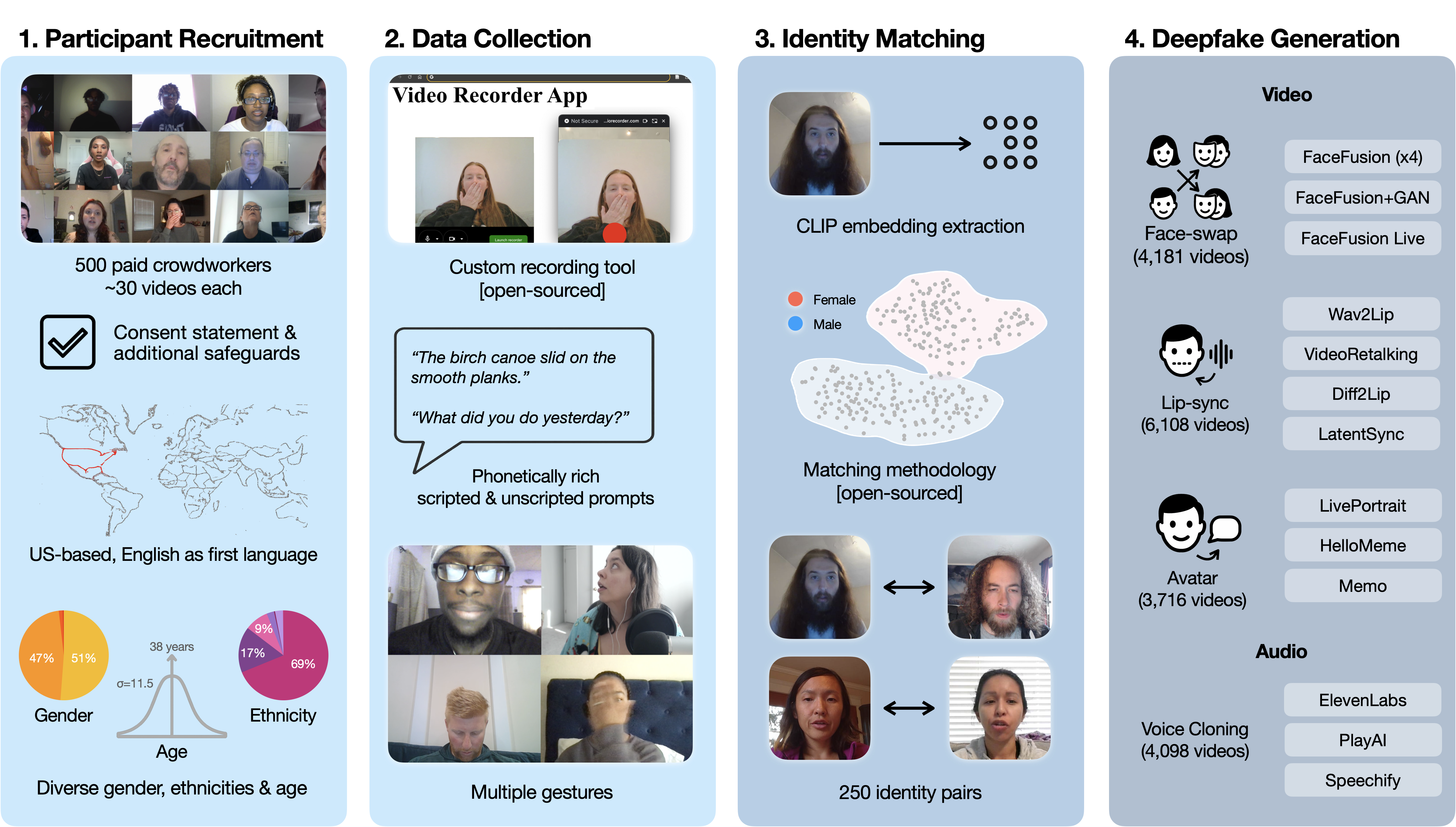} 
\caption{An overview of the \textit{DeepSpeak} Dataset sourced from a diverse selection of consenting participants using a custom-built data collection methodology. The dataset also comprises deepfakes generated from 14 video and three audio deepfake methods using facial identity matching to improve the realism of the generated deepfakes.}
\label{fig:deepspeak_overview}
\end{figure*}

In this work, we introduce an audio and video dataset designed to aid the digital-forensic, computer-vision, and broader AI-safety communities. This dataset consists of 100 hours of real and deepfake video of people talking and gesturing. The real videos were self-recorded with consent from the participants using their own hardware, ensuring a wide range of recording environments, hardware variability and identities, a crucial component for the development of robust detectors. These deepfakes consist of avatar deepfakes (from three generators), face-swap deepfakes (with multiple variants from three generators), lip-sync deepfakes (from four generators), and audio deepfakes (from three generators) spliced into a subset of the lip-sync deepfakes. Table~\ref{tab:datasets} presents a comparison between our dataset and recent datasets released over the past seven years.

We focus exclusively on ``talking heads'' in which one person is talking or gesturing in a typical video conferencing setup. We focus on this context because of the increasing prevalence of distinct harms that have emerged from both offline and real-time deepfake identity impersonations (including impostor hiring, fraud, and disinformation). This scenario presents unique challenges concerning identity verification and liveness detection, as compared to the more common analysis of text- or image-to-video content. The ``talking heads'' scenario shifts the question from only a `real v. fake' or `identity' question to both one of realism {\em and} identity. Our effort provides a dataset to simultaneously address both of these questions. Existing datasets in this domain do not reflect the breadth of ours, nor additionally, the current quality and diversity of deepfake generators, while bearing ethical, practical, and legal shortcomings (see Table~\ref{tab:datasets}). Specifically, existing datasets largely comprise low-quality, outdated deepfake generators, where the underlying data was scraped without participant consent. Moreover, these datasets do not include all types of deepfake generators and attack settings. A more comprehensive review of other datasets is included in Appendix~\ref{app:prior_work}.

\begin{table*}[t]
\centering
\resizebox{\textwidth}{!}{
\setlength{\tabcolsep}{3pt}
\begin{tabular}{rcrrcccccccr}
    \hline \noalign{\vskip 0.5ex}
         & \textbf{Release} & \textbf{Unique}     & \textbf{Original}   &         &                &           &        & \textbf{Fake} &    \textbf{Conversational}    &  \textbf{Identity}      & \textbf{Deepfake} \\
    \textbf{Name} & \textbf{Year}    & \textbf{Identities} & \textbf{Footage}    & \textbf{Consent} & \textbf{Faceswap} & \textbf{Lipsync} & \textbf{Avatar} & \textbf{Audio} & \textbf{Webcam} & \textbf{Matching} & \textbf{Footage} \\
    &  &  &  &  &  &  &  &  &  &  &  \\
    \hline
    FaceForensics   & 2018 & NA & 1{\small,}004 & N & ? & - & - & - & - & - & 2{\small,}008 \\
    FaceForensics++ & 2019 & NA & 1{\small,}000 & Partial & $\checkmark$ & - & - & - & - & - & 4{\small,}000 \\
    DFDC            & 2020 & 3{\small,}426 & 23{\small,}654 & Y & $\checkmark$ & - & - & - & Partial & $\checkmark$ & 104{\small,}500 \\
    DFD             & 2019  &  28 &  363           & Y & $\checkmark$ & - & - & - & $\checkmark$ & - & 3{\small,}068 \\
    Celeb-DF        & 2020 & 59 & 590            & N & $\checkmark$ & - & - & - & - & - & 5{\small,}049 \\
    AVDeepfake1M    &  2024  &  2{\small,}068 &  286{\small,}721           & N & - & $\checkmark$ & - & $\checkmark$ & - & - & 860{\small,}039 \\
    FakeAVCeleb     & 2021 & 600 & 570           & N & $\checkmark$ & $\checkmark$ & - & $\checkmark$ & - & - & 25{\small,}000 \\
    Deepfake-Eval-2024 &  2025  &  NA &  --        & N & $\checkmark$ & - & - & $\checkmark$ & Partial & - & -- \\
    LAV-DF          & 2022 & 153 & 36{\small,}431 & N & - & $\checkmark$ & - & $\checkmark$ & - & - & 99{\small,}873 \\
    DF40            & 2024 & NA & NA             & N & $\checkmark$ & $\checkmark$ & $\checkmark$ & - & - & - & 100,000+ \\
    NVFAIR            & 2025 & 161 & NA             & Y & - & $\checkmark$ & - & - & $\checkmark$ & - & 650,000 \\
    Polyglotfake & 2024 & NA & 766 & N & - & $\checkmark$ & - & $\checkmark$ & - & - & 14,472 \\
    Illusion & 2025 & NA & 139,740 & N & $\checkmark$ & - & - & $\checkmark$ & - & $\checkmark$ & 1,232,246 \\
    DF-Platter & 2023 & 454 & 764 & N & $\checkmark$ & - & - & - & - & - & 132,496 \\
    \hline
    \textbf{DeepSpeak (Ours)} & 2025 & 500 & 16{\small,}043 & Y & $\checkmark$ & $\checkmark$ & $\checkmark$ & $\checkmark$ & $\checkmark$ & $\checkmark$ & 14{\small,}005 \\
    \hline
\end{tabular}
}
\caption{A comparison of forensic-themed public datasets. Although not the most informative metric, we report original and deepfake footage as number of videos for consistency with previous published datasets (NA: not available). ``Fake Audio'' refers to speech synthesized by AI-enabled voice cloning.}
\label{tab:datasets}
\end{table*}

To remedy these shortcomings, our work makes the following contributions:
\begin{itemize}
    \item \textbf{Documentation and Release of DeepSpeak.} We introduce a methodology for the large-scale collection of real video recordings, self-submitted by a diverse selection of consenting participants (Section~\ref{sec:data-collection}), along with the procedures used to generate corresponding deepfake video and audio (Sections~\ref{sec:audio-gen}~and~\ref{sec:video-gen}). 
    \item \textbf{Data Collection Tool.} We provide a codebase for a web-based application designed to facilitate participant-led remote data collection. When used in conjunction with our collection survey, the collected data is phonetically rich and diverse in terms of speech content, video durations, gestures, and includes both scripted and unscripted segments. 
    \item \textbf{Method for Identity Matching.} We devise a method for matching participants based on their visual features to create more convincing face-swap deepfakes, consistent with real-world deepfake attacks (Section~\ref{sec:id-mathching}). 
    \item \textbf{Large-scale Benchmarking and Generalization Study.} We perform large-scale evaluations of state-of-the-art deepfake detectors across audio, visual and multimodal detectors and show that they fail to accurately distinguish between real and fake audio and video when trained on other datasets (Table~\ref{tab:datasets})(Section~\ref{sec:experiments}). These evaluations highlight the importance of a large and diverse dataset containing deepfakes from the latest generative-AI tools.
\end{itemize}
\nocite{dufour2019dfd}

\section{Data Collection}
\label{sec:data-collection}

The data collection was performed in four steps. Data collection for \textit{DeepSpeak} was determined to qualify for exempt status by UC Berkeley Office for Protection of Human Subjects (OPHS).

\vspace{-1.0em}

\paragraph{Participant Recruitment.} Participants were crowd-sourced through the Prolific research recruitment platform. Participants were asked to give their consent for including their recordings, without any other identifying information, in a public dataset. Details of the consent statement can be found in Appendix~\ref{app:survey-materials}. A total of 500 participants were selected from a stratified sample ensuring equal distribution of gender, and with all participants reported as being native English speakers and U.S. residents, with demographics as follows (some participants identified with more than one race/ethnicity): 
\begin{itemize}
\item {\bf Age}:  Range = 18-75 years, Mean = 38 years; standard deviation = 11.5 years
\item {\bf Gender}: 256 male, 235 female, 7 non-binary, 2 not provided
\item {\bf Race/Ethnicity}: 362 White/Caucasian, 87 Black/African American, 45 Asian, 14 American Indian/Alaska Native, 2 Native Hawaiian/Other Pacific Islander, 15 other, 1 prefer not to say. 
\end{itemize}
\vspace{-1.0em}

\paragraph{Survey.} The data collection survey was designed to capture both speech and visual actions. For speech, it included phonetically rich audio data spanning varied audio durations with both scripted vs. conversational-style responses. Each participant was instructed to record themselves responding to between 32 and 35 separate prompts. Participants were paid \$7 for their time. The first two prompts were used for voice-clone training data (see Section~\ref{sec:audio-gen}). The remaining prompts were divided into four categories: (1) 10 standardized scripted responses in which each participant read the same prompt; (2) 10 randomized scripted responses in which participants read a randomized prompt; (3) 10 unscripted responses in which participants responded to questions; and (4) between 5-8 actions in which participants performed simple actions. Scripted responses were generated using transcripts of the TIMIT dataset, a linguistics research dataset consisting of utterances from 462 real female and male American-English speakers. See Appendix~\ref{app:prompts} for the full list of prompts and scripts used. 

\vspace{-1.0em}

\paragraph{Data Collection tool.} Both audio and video were recorded using a custom-built Google Chrome web application. The JavaScript and Python repository for this web application is available at \url{https://github.com/hfaridlab/deepspeak}. Details of the encoding and data pre-processing associated with the tool can be found in the Appendix~\ref{app:data_collection}.

\vspace{-1.0em}

\paragraph{Validation.} Participants were given written and visual instructions to allow them to practice recording themselves and test their hardware. Participants were asked to adhere to a series of recording conditions intended to improve consistency within the overall dataset. We manually removed any invalid responses from the final dataset that did not meet these requirements. The details of this can be found in the Appendix~\ref{app:data_collection}.

\section{Identity Matching}
\label{sec:id-mathching}

During manual inspection of the collected data, we observed that, albeit diverse in age, gender, and ethnicity, our collected data contains many individuals with similar facial and vocal features. In order to exploit this feature of the dataset and create more compelling deepfakes, each identity in the dataset was paired with another, perceptually similar one. The code for producing this visual matching, as well as the resulting visual pairs is open-sourced at~\url{https://github.com/hfaridlab/deepspeak}.

\vspace{-1.0em}

\begin{figure}[t]
    \centering
    \includegraphics[width=1.0\linewidth]{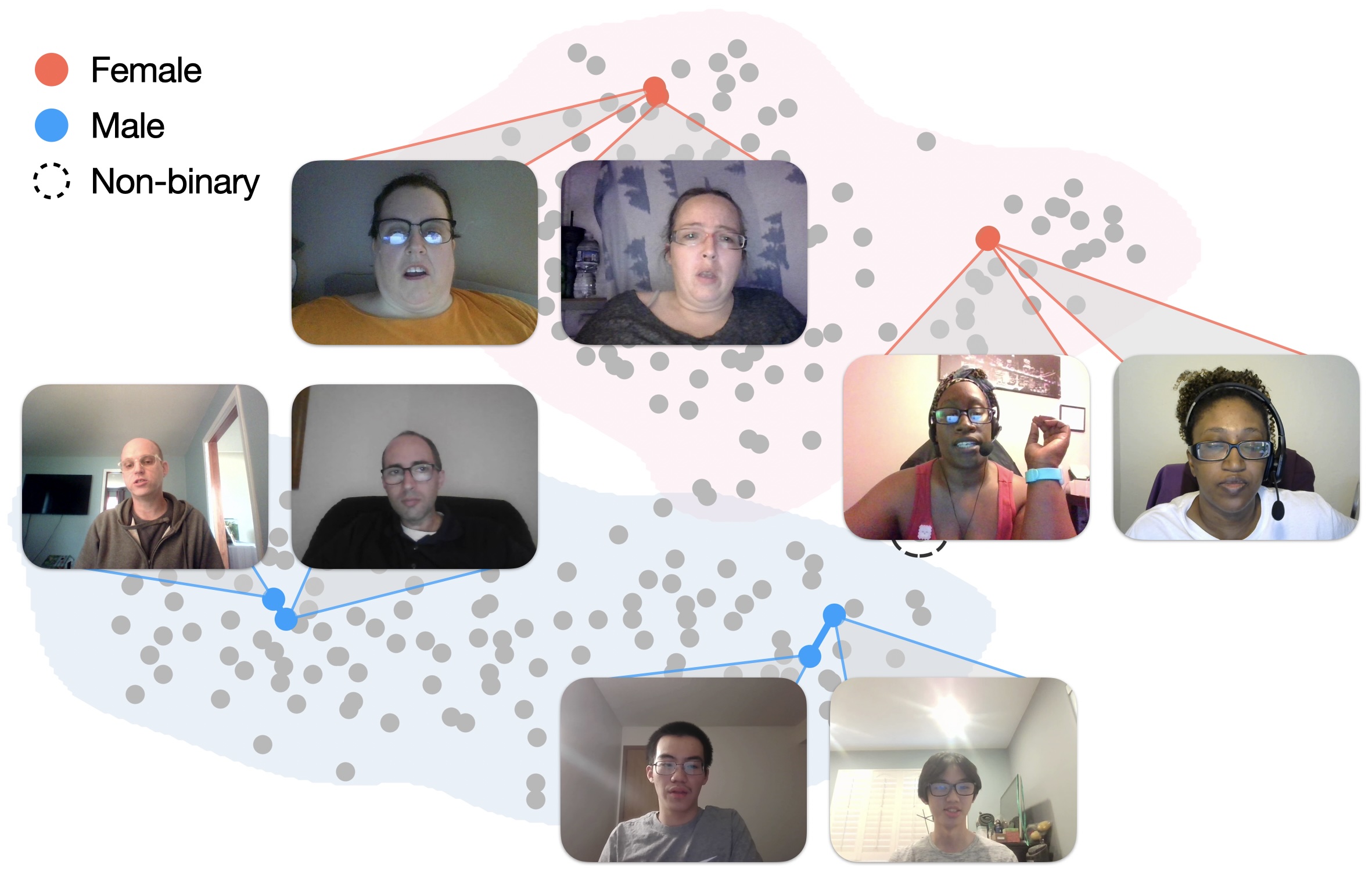}
    \caption{A t-SNE visualization of CLIP embeddings from real participant's videos. The four highlighted pairs correspond to identities with maximal similarity as measured by the cosine distance between CLIP embeddings. Perceptually similar identities cluster in this t-SNE representation. The red/blue color coding corresponds to people who identify as female/male, which also clusters in this t-SNE representation.}
    \label{fig:identity_matching_visual}
\end{figure}

\paragraph{Visual Matching.} Each identity is first represented by the average CLIP embedding\footnote{\url{https://github.com/OpenAI/CLIP}}~\cite{radford2021learning} extracted from five random video frames (filtered for low-quality frames, see Section~\ref{sec:validation}). Shown in Figure~\ref{fig:identity_matching_visual} is a t-SNE visualization of a subset of these embeddings. Comparing this representation against the self-reported demographic information reveals that these  CLIP embeddings cluster based on gender, ethnicity and facial similarity.
%
%


For each identity, a unique matched identity is assigned using the agglomerative clustering algorithm with cosine distance and cluster size constraint from the scikit-learn library\footnote{\url{https://scikit-learn.org/stable/modules/generated/sklearn.cluster.AgglomerativeClustering.html}}. Additional examples of visual pairs are shown in Appendix~\ref{app:visual_identity_pairing_examples}. This approach was adopted instead of a more traditional biometric matching like ArcFace~\cite{deng2019arcface} because we observed, during manual review, better qualitative matching for women and people of color. We also found that our CLIP-based matching outperforms ArcFace in terms of the Frechet inception distance (FID) between the matched face pairs by $4\%$ ($214$ vs $222$), and the LPIPS distance between the matched face pairs by $6\%$ ($0.61$ vs $0.65$). For both FID and LPIPS, a smaller value corresponds to higher perceptual similarity.

With this identity matching, averaged across all videos, DeepSpeak achieves an average FID of $238$ and LPIPS of $0.46$. Compared to baselines datasets, this is $36\%$ better than DF40 ($325$ FID and $0.68$ LPIPS), $33\%$ better than FaceForensics ($317$ and $0.74$), and $71\%$ better than DFDC ($408$ and $0.70$).

%
%
%
%

\section{Audio Generation}
\label{sec:audio-gen}

Participants were first asked to record themselves reading 10 consecutive phonetically-rich sentences, sourced from List 1 of the standard Harvard Sentences~\cite{rothauser1969ieee}, a collection of sentences representing best practice for standardized evaluation of speech processing and audio quality in controlled settings. Participants were then asked to repeat the standard elicitation paragraph from the Speech Accent Archive, a phonetically comprehensive passage comprising a breadth of vowels and consonants~\cite{saa}. These two scripted responses were used for the purpose of voice cloning, and had an average length of 30 seconds. 

Using each participant's cloned voice, a synthetic audio was created in their voice saying the same thing as in the original audio/video. For the unscripted responses, the original audio was transcribed using OpenAI's Whisper, and for the scripted responses, we assumed that the participant correctly read the script. These text transcriptions were then provided to each voice cloning generators' API to generate matching synthetic voices. 

Voice clones were generated using three commercial cloning and Text-to-Speech (TTS) services: ElevenLabs, PlayAI and Speechify. The details of API end points used, alongside parameters, can be found in Appendix~\ref{app:audio_gen_methods}. 

\section{Video Generation}
\label{sec:video-gen}

We generated three types of video deepfakes: face swap, lip sync, and avatar, each of which is described next. The resulting dataset is randomly split into 80/20 training/testing splits with no overlap in facial or voice identities. A breakdown of the resulting dataset's statistics, including the total file size (GB), file counts (N), and video length (hrs) are included in Appendix~\ref{app:vid_gen_methods}.

\begin{figure}[t]
    \centering
    \includegraphics[width=\linewidth]{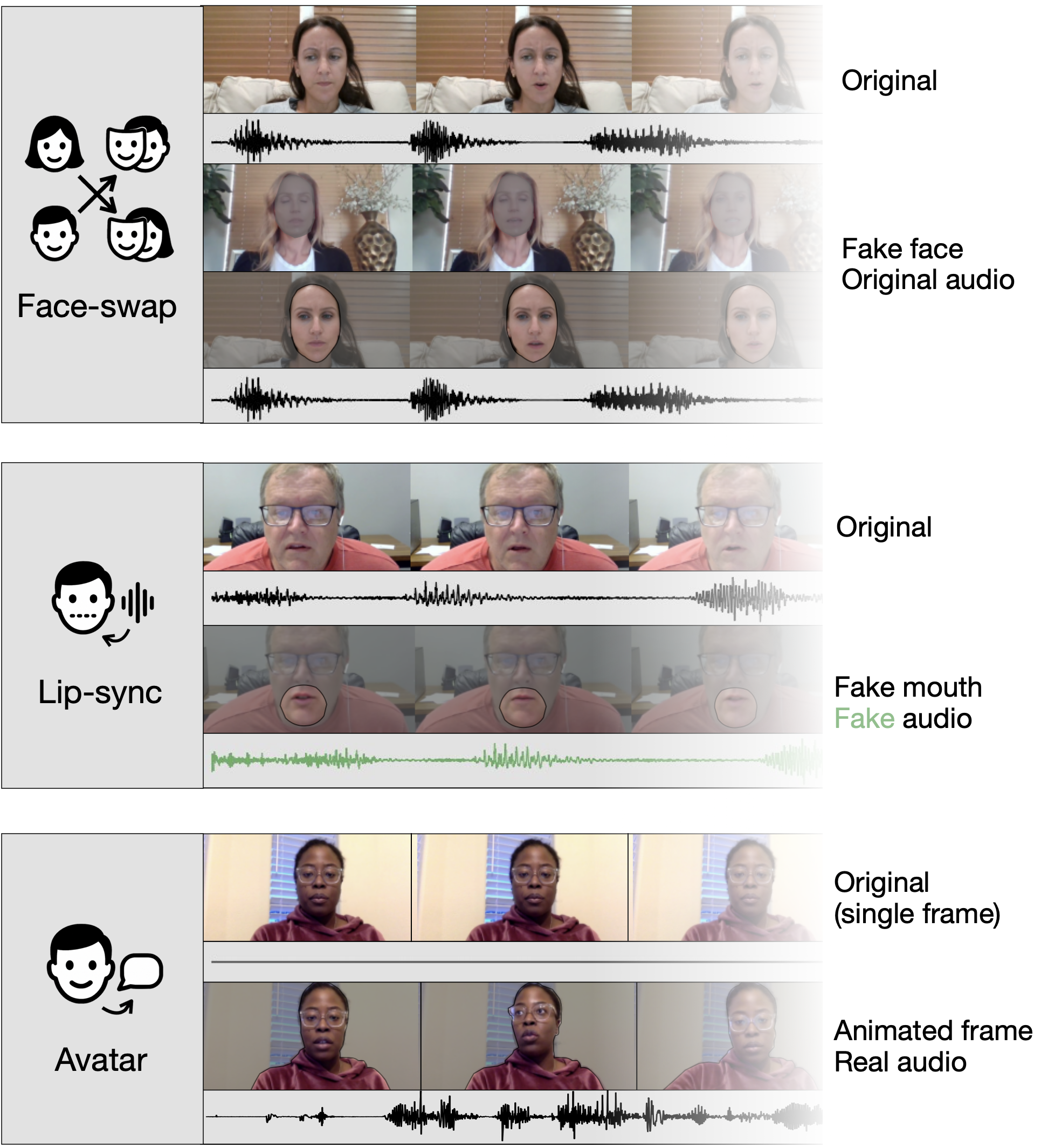}
    \caption{The {\em DeepSpeak} dataset consists of face-swap, lip-sync, and avatar deepfakes.}
    \label{fig:deepfake_generation}
\end{figure}

\subsection{Generation}

\paragraph{Face-Swap.} Face-swap deepfakes are created by replacing -- eyebrow to chin and cheek to cheek -- the original identity in a video with a new identity. We swapped faces of identity pairs identified through the visual matching (see Section~\ref{sec:id-mathching}). This ensured that the swapped identities were perceptually similar to begin with, which made for more compelling deepfakes. This resembles conventional practices of in-the-wild deepfake production, where actors are chosen based on their similarity to the target identity.

An overview of face-swap deepfake generation is shown in Figure~\ref{fig:deepfake_generation}, row one. To generate a face swap, the video of the original identity and a single frame of the matched identity are provided to the face-swap synthesis engine. The single frame is initially chosen to be the fifth frame in a randomly selected video of the matched identity. We found that if the eyes are closed in the matched face, the resulting face-swap deepfake suffered in quality. As such, we used MediaPipe~\cite{lugaresi2019mediapipe} to extract facial features and ensured that the distance between the top and bottom eyelid landmarks was greater than a specified threshold. If this constraint failed, then the tenth frame was selected for consideration; this process was repeated, skipping five frames each time, until a suitable frame was found. We used seven face-swap methods, as detailed in Appendix~\ref{app:vid_gen_methods}.

\vspace{-1.0em}

\paragraph{Lip-Sync.} Whereas the face-swap deepfake replaces an entire face with a new identity, a lip-sync deepfake modifies the mouth region to be consistent with a different audio track. An overview of lip-sync deepfake generation is shown in row two of Figure~\ref{fig:deepfake_generation}.  Given an original video and associated audio, we create two types of lip-sync deepfakes: (1) a lip-sync deepfake with an audio of the same identity extracted from a different video (i.e.,~the audio and video are now mismatched); and (2) a lip-sync deepfake with an AI-generated voice of the same identity (Section~\ref{sec:audio-gen}) with a transcript taken from a different video. Four methods of lip-sync deepfakes were employed, as further described in Appendix~\ref{app:vid_gen_methods}.

\vspace{-1.0em}

\paragraph{Avatar.} An overview of avatar deepfake generation is shown in Figure~\ref{fig:deepfake_generation}, row three. Avatar deepfakes animate the head and lip movements of a static image to match a target video or audio track. Unlike face-swap and lip-sync deepfakes, which modify an existing video, avatar deepfakes generate movement from a single static image. Avatar deepfakes were created using three methods further described in Appendix~\ref{app:vid_gen_methods}. LivePortrait and HelloMeme take as input a single image of a person to animate with a video (and associated audio) that drives this animation. For these two generators, the avatar deepfakes contain only real audio from the original driving video. Memo takes as input a single image of a person to animate with only an audio that drives this animation. In this case, the audio can be either real or fake.

\subsection{Validation}
\label{sec:validation}

During manual inspection of the generated videos, we identified multiple types of failures, including deepfake engines (1) producing corrupted faces with consistently closed eyes or mouths, (2) generating malformed avatars with distorted facial or upper body structure, (3) failing to apply any changes and yielding back the original video, (4) modifying only parts of the video, (5) producing empty output consisting of with black frames, among others. To prevent failed deepfakes, we designed a suite of input and output detectors to filter undesired features. This filtering code is open-sourced at~ \url{https://github.com/hfaridlab/deepspeak}. The details of this filtering can be found in Appendix~\ref{app:validation}.

\section{Experiments}
\label{sec:experiments}

We conducted a series of baseline experiments on \textit{DeepSpeak} for the tasks of audio and video deepfake detection. The code for these experiments, including data pre-processing, is open-sourced at \url{https://github.com/hfaridlab/deepspeak}. The experiments were conducted on NVIDIA A100 GPUs over the course of approximately four weeks (see Appendix~\ref{app:compute} for details pertaining compute resources).

\subsection{Video Deepfake Detection}


\paragraph{Baselines.} Both classic- and deep-learning methods for deepfake video detection can be categorized by the scrutinized signal deemed to discriminate the real from the fake, with most performing (1) spatial-domain analysis, (2) frequency-domain analysis, or (3) cross-modal temporal coherence analysis. To capture the breadth of the existing approaches, we evaluate state-of-the-art methods representing these distinct lines of work. The first evaluated architecture is a frequency-based method FreqNet~\cite{cai2021freqnet}. The second, spatial-domain, architecture is GenConViT~\cite{wodajo2023deepfake} (with ED and VAE variants). The third, multi-modal, architecture is LipFD~\cite{liu2024lips} designed to detect misalignments between the visual and vocal stream of lip-sync deepfakes.

For each of these four architectures, we evaluated three model variants: (1) the pretrained model released alongside the respective publication (trained on a different, non-DeepSpeak dataset), (2) the model trained from scratch on DeepSpeak, and (3) the model, starting with the pre-trained weights, fine-tuned on DeepSpeak. A total of $12$ models were evaluated.

\vspace{-1.0em}

\paragraph{Experimental Setup.} To perform inference, training, and fine-tuning of the included architectures, we used the official code repositories released alongside the respective publications. To make the results comparable despite the differing number of parameters of these architectures, we used default hyperparameters when possible, with a simple search over learning rates (see Appendix~\ref{app:exp_setup_hyperparameters} for details).

Each model is evaluated against the testing split of its architecture's original dataset and DeepSpeak. The original dataset refers to the dataset used for the pretrained model in the respective publication: for FreqNet, it is a custom GAN-generated dataset compiled by its authors~\cite{cai2021freqnet}; for GenConViT ED and VAE, it is Celeb-DF 2~\cite{li2020celeb}; and for LipFD, it is AVLips~\cite{liu2024lips}. The accuracy on the real and fake class is reported separately, along with the overall F1 score.

\vspace{-1.0em}

\paragraph{Results.} Shown in Table~\ref{tab:video_baseline_results} are the results of the pretrained models and models trained from scratch on DeepSpeak. All four evaluated architectures follow the same pattern: they perform reasonably well on the testing splits of their original training datasets but fail to generalize to DeepSpeak. The same trend holds when models are trained on DeepSpeak and evaluated on the original dataset. Notably, even on the original testing sets, class bias was evident—for example, GenConViT attained an accuracy of $98.2\%$ on fake but only $56.7\%$ on real, while LipFD showed the opposite pattern, scoring $97.9\%$ on real versus $69.1\%$ on fake.

Also shown in last six columns of Table~\ref{tab:video_baseline_results} are the results of the fine-tuned models (labeled Original + DeepSpeak). While some models, such as GenConViT ED and VAE, achieved performance on DeepSpeak comparable to training from scratch (F1 score above $0.9$), this came at the cost of a sharp drop in performance on the original testing set, where F1 scores fell below $0.2$. LipFD was able to fine-tune on DeepSpeak while maintaining comparable performance on the original testing set (both with F1 scores around $0.7$), though it should be noted that the model exhibits a strong bias toward the fake class.

\begin{table*}[t]
\centering
\addtolength{\tabcolsep}{-0.2em}
\begin{tabularx}{\linewidth}{lcccccccccccccccccc}
\toprule
 & \multicolumn{6}{c}{\textbf{Original}} & \multicolumn{6}{c}{\textbf{DeepSpeak}} & \multicolumn{6}{c}{\textbf{Original + DeepSpeak}} \\
\cmidrule(lr){2-7}\cmidrule(lr){8-13}\cmidrule(lr){14-19}
& \multicolumn{3}{c}{\textbf{Original}} & \multicolumn{3}{c}{\textbf{DeepSpeak}} & \multicolumn{3}{c}{\textbf{Original}} & \multicolumn{3}{c}{\textbf{DeepSpeak}} & \multicolumn{3}{c}{\textbf{Original}} & \multicolumn{3}{c}{\textbf{DeepSpeak}} \\
\cmidrule(lr){2-4}\cmidrule(lr){5-7}\cmidrule(lr){8-10}\cmidrule(lr){11-13}\cmidrule(lr){14-16}\cmidrule(lr){17-19}
\textbf{Method} & Real & Fake & F1 & Real & Fake & F1 & Real & Fake & F1 & Real & Fake & F1 & Real & Fake & F1 & Real & Fake & F1\\
\midrule
FN      & 97.1 & 88.3 & 0.9 & 65.3 & 15.4 & 0.2 & 34.4 & 26.6 & 0.6 & 77.3 & 69.9 & 73.6 & 50.5 & 14.1 & 0.3 & 74.2 & 66.1 & 0.7 \\
GC-ED   & 57.3 & 98.2 & 0.7 & 88.5 & 39.1 & 0.7 & 2.8 & 100 & 0.1 & 90.5 & 90.7 & 0.9 & 7.9 & 100 & 0.2 & 91.7 & 78.2 & 0.9 \\
GC-VAE  & 56.7 & 98.2 & 0.7 & 88.5 & 39.2 & 0.7 & 4.5 & 100 & 0.1 & 91.1 & 96.4 & 0.9 & 9.0 & 99.7 & 0.2 & 93.0 & 89.6 & 0.9 \\
LFD     & 97.9 & 69.1 & 0.8 & 98.8 & 3.5 & 0.1 & 7.30 & 88.7 & 0.7 & 71.8 & 77.1 & 0.8 & 2.8 & 97.8 & 0.7 & 28.2 & 96.6 & 0.7 \\
\bottomrule
\end{tabularx}
\caption{\textbf{Video deepfake detection accuracies (\%)} of four state-of-the-art architectures: FreqNet (FN), GenConViT ED (GC-ED), GenConViT VAE (GC-VAE), and LipFD (LFD). The heading in the first row corresponds to the dataset on which each model was trained, and the heading in the second row corresponds to the dataset on which each model is evaluated against.}
\label{tab:video_baseline_results}
\end{table*}

\subsection{Audio Deepfake Detection}
\label{sec:audio_experiments}


\paragraph{Baselines.} We evaluated the performance of two model architecture types on the DeepSpeak dataset, consistent with recent literature: (i) a foundation model, and (ii) a raw waveform model. Foundation models use a pretrained model to extract embeddings from the input waveform, which are then passed to a classifier. Three state-of-the-art models were selected: TitaNet~\cite{koluguri2022titanet,barrington2023single}, Wav2Vec-XLSR~\cite{cozzolino2024,babu22wav2vec2xlsr}, and LAION-CLAP~\cite{cozzolino2024,wu2023laionclap}. For each embedding type, both linear and non-linear classifiers were tested. Raw waveform models operate directly on the audio waveform. Three leading models were chosen: AASIST~\cite{jung2022aasist}, RawNet2~\cite{jung2022aasist,tak21rawnet}, and RawGAT-ST~\cite{jung2022aasist,tak21rawgat}.

For both architectures, we evaluated two versions of each model: (1) a pretrained model trained on a dataset other than DeepSpeak, and (2) a model trained from scratch on DeepSpeak. In the case of foundation models, the foundation model used to extract embeddings remained pretrained, while the downstream classifiers were trained from scratch. In total, 18 models were evaluated. A summary is provided in Table~\ref{tab:audio_detection_results}.

\vspace{-1.0em}

\paragraph{Experimental Setup.} Pretrained raw waveform model weights were sourced directly from the AASIST implementation of AASIST, RawNet2, and RawGAT-ST~\footnote{\url{https://github.com/clovaai/aasist}}. Default configuration were used for each model, as detailed in the Appendix~\ref{app:exp_setup_hyperparameters}. For retraining these models from scratch on DeepSpeak, the same configurations and architectures were maintained, with DeepSpeak training data replacing ASVSpoof. For foundation models, classifiers (both logistic regression and random forest) were trained using balanced datasets with embeddings extracted from the training sets of either ASVSpoof (for Wav2Vec-XLSR and LAION-CLAP) or TIMIT-ElevenLabs (for Titanet). Embeddings from the DeepSpeak test dataset split were used for evaluation. Both linear (logistic regression) and non-linear (random forest) classifiers were tested for each embedding type. No cross-validation or hyperparameter tuning was performed for either the pretrained or from-scratch models.

Each model is evaluated against the testing split of its architecture's original dataset and DeepSpeak. The original dataset corresponds to the one used for pretraining in the respective publications. For AASIST, RawNet2, and RawGAT-ST, this dataset is ASVSpoof (as implemented in~\cite{jung2022aasist}), and for TitaNet-based embeddings approaches, this dataset is TIMIT-ElevenLabs~\cite{barrington2023single}. Since prior literature on detection using Wav2Vec-XLSR and LAION-CLAP largely focusses on training-free methods~\cite{cozzolino2024}, we trained our own benchmarks  on ASVSpoof for consistency and because it serves as one of the most comprehensive and widely used benchmarking datasets. Performance metrics are reported for both the original dataset’s test set and the DeepSpeak test set. As shown in Table~\ref{tab:audio_detection_results}, accuracies for both real and fake classes are presented separately, along with overall accuracy to account for class imbalance (since fake audio only occurs in lip-sync deepfakes, representing a subset of the full dataset), and the error rate (EER).

\vspace{-1.0em}

\paragraph{Results.} 

\begin{table*}[t]
\scriptsize
\centering
\resizebox{\textwidth}{!}{%
    \addtolength{\tabcolsep}{0.1em} 
    \begin{tabular}{l l *{12}{c}} 
    \toprule
    & & \multicolumn{6}{c}{\textbf{Original}} & \multicolumn{6}{c}{\textbf{DeepSpeak}} \\
    \cmidrule(lr){3-8} \cmidrule(lr){9-14}
    & & \multicolumn{3}{c}{\textbf{Original}} & \multicolumn{3}{c}{\textbf{DeepSpeak}} & \multicolumn{3}{c}{\textbf{Original}} & \multicolumn{3}{c}{\textbf{DeepSpeak}} \\
    \cmidrule(lr){3-5} \cmidrule(lr){6-8} \cmidrule(lr){9-11} \cmidrule(lr){12-14}
    \textbf{Model} & \textbf{Clf} & Real & Fake & F1 & Real & Fake & F1 & Real & Fake & F1 & Real & Fake & F1 \\
    \midrule
    Titanet (FM) & LR & 99.4 & 100.0 & 1.00 & \textbf{10.0} & \textbf{97.4} & \textbf{0.18} & 61.6 & 98.8 & 0.75 & 91.3 & 89.1 & 0.95 \\
                & RF & 99.8 & 100.0 & 1.00 & \textbf{54.2} & \textbf{64.3} & \textbf{0.68} & 74.8 & 83.1 & 0.71 & 96.3 & 79.3 & 0.97 \\
    Wav2Vec2-xlsr (FM) & LR & 79.7 & 82.7 & 0.48 & \textbf{1.3} & \textbf{97.0} & \textbf{0.03} & 7.6 & 83.8 & 0.06 & 76.8 & 65.6 & 0.84 \\
                      & RF & 98.1 & 88.5 & 0.66 & \textbf{19.3} & \textbf{95.4} & \textbf{0.32} & 93.6 & 36.9 & 0.25 & 97.4 & 78.0 & 0.97 \\
    LAION-CLAP (FM) & LR & 92.9 & 92.0 & 0.71 & \textbf{33.8} & \textbf{76.6} & \textbf{0.49} & 90.3 & 53.3 & 0.30 & 93.7 & 91.9 & 0.96 \\
                    & RF & 93.1 & 90.5 & 0.67 & \textbf{65.3} & \textbf{68.3} & \textbf{0.77} & 93.9 & 56.7 & 0.33 & 95.8 & 89.6 & 0.97 \\
    AASIST (RW)     & -  & 99.5 & 99.5 & 0.98 & \textbf{60.1} & \textbf{60.2} & \textbf{0.72} & 73.1 & 73.1 & 0.36 & 98.8 & 98.8 & 0.99 \\
    RawNet2 (RW)    & -  & 99.0 & 99.0 & 0.95 & \textbf{54.4} & \textbf{54.4} & \textbf{0.67} & 69.4 & 69.3 & 0.32 & 94.1 & 94.3 & 0.96 \\
    RawGAT-ST (RW)  & -  & 99.1 & 99.1 & 0.96 & \textbf{57.7} & \textbf{57.7} & \textbf{0.70} & 75.0 & 75.0 & 0.38 & 96.8 & 96.8 & 0.98 \\
    \bottomrule
    \end{tabular}
} 
\caption{\textbf{Audio deepfake detection accuracies (\%)} of nine state-of-the-art models using two separate architectures (FM = Foundation Model, RW = Raw Waveform). For each training/testing combination, we report the real class accuracy, fake class accuracy, and F1 score. The Clf column indicates the type of classifier used for embedding-based models, either logistic regression (LR) or random forest (RF).} 
\label{tab:audio_baseline_results}
\end{table*}
\label{tab:audio_detection_results}

When trained and tested on DeepSpeak, raw waveform models perform well, with AASIST achieving 98.8\% accuracy - only 0.7 percentage points lower than its original ASVSpoof benchmark. Embedding-based models also show strong, though comparatively lower performance, with the best performing models being those trained on LAION-CLAP embeddings (see Table~\ref{tab:audio_detection_results}).

Pretrained models, however, do not generalize well to DeepSpeak data. AASIST remains the top-performing pretrained model, albeit with substantially lower performance when evaluated out-of-the-box on DeepSpeak data, dropping to an accuracy of 60.1\% and 60.2\% for real and fake. Pretrained embedding-based models also show substantially lower performance when evaluated on DeepSpeak data, alongside notable class imbalances (see Table~\ref{tab:audio_detection_results}). 

These results suggest that feature representations learned directly from raw waveform inputs may be more resilient to domain shift in DeepSpeak data than those extracted from foundation embeddings-based models.  In all cases, pretrained models are insufficient for accurately distinguish real from fake audio. 

This pattern for audio detection models is similar to video detection models: (1) these models struggle with out-of-domain data; but (2) these models can improve with appropriate training.

\subsection{Combined Audio-Visual Detection}

While most deepfake detection methods treat audio and video independently, several joint approaches have also been proposed. Although multimodal detection is not the primary focus of this work, we confirm that the limited generalization observed in unimodal detection methods extends to a leading combined method exploiting audio-video mismatch~\cite{bohacekLIT}. This technique classifies a video as real/fake based on a normalized Levenshtein distance between an audio transcription and a video transcription (based on an automatic lip reading). At a threshold of 0.55 on this distance (where are distance of 0.0 corresponds to a perfect match between the audio and video, and a distance of 1.0 corresponds to a maximal mismatch), the accuracy across 2963 real and 2953 AI-generated videos is 65.5\% and 67.9\%. Accuracy across different types of AI-generated videos (face-swap, lip-sync, and avatar) ranged from a low of 43.4\% to a high of 99.0\%.

\section{Closing Thoughts}

\paragraph{Discussion.} Year after year, we see a dramatic rise in the number of deepfake generators and the quality of the fake audio and video. Given the pace at which deepfake technology is progressing, it is critical that evaluation datasets keep up with the latest technologies. This is made apparent by our evaluation of recent state-of-the-art deepfake detectors that struggle to generalize to the latest deepfake generators. To this end, our DeepSpeak dataset is partitioned into two parts (v1 and v2), containing a snapshot of the state of the art in deepfake generation in 2024 and 2025, respectively. We plan to release one to two new datasets each year to keep pace with these new threats.

\vspace{-1.0em}

\paragraph{Limitations.} \textit{DeepSpeak} captures the state of the art in deepfake generation at the time of publication, making it well-suited for developing and evaluating detection methods for current and emerging deepfake engines. However, as generative AI evolves rapidly, it is essential to recognize the dataset's limitations and the potential for future expansion.

Due to the lack of high-quality open-source deepfake engines for non-English languages, DeepSpeak currently includes participants speaking English. As high-quality multilingual engines become available, we will need to expand DeepSpeak to include additional languages.

Currently, open-source deepfake engines operate on the video level, which is reflected in DeepSpeak---every video in the dataset is either entirely real or entirely fake. Once targeted manipulation (i.e.,~changing only some words in the video) improves, we will include them in future versions.

Lastly, to date, all of the DeepSpeak video generators are based on open-source models and not on commercially available models. As we have done with commercial audio generators, we will seek to establish relationships with commercial video generators to allow for large-scale video generation of commercial offerings.

\vspace{-1.0em}

\paragraph{Ethical Considerations:} Too many other datasets in media forensics and computer vision have adopted a ``scrape and distribute, ask questions later'' approach. We take issue with this both from the perspective of participant consent and intellectual property. 

While we don't object to the development of deepfake generators, we will not knowingly license \textit{DeepSpeak} for this purpose. Our rationale here is that the harms that are coming from deepfakes are not insignificant and we simply don't want to be contributing to a plethora of online harms.

\vspace{-1.0em}

\paragraph{Conclusion.} Our motivation for creating this dataset is to support the media-forensics research community and the development and refinement of techniques to detect deepfake audio, image, and video. The world of generative AI and media forensics is fast moving. It is, therefore, important that shared datasets be regularly updated to keep up with the latest trends. To this end, we expect to release updates to this dataset once to twice a year. To help serve the community better, we welcome feedback, comments, requests for future releases of this dataset at \url{https://github.com/hfaridlab/deepspeak}.

\section*{Acknowledgments} We are grateful to David Chan for his many insightful comments and suggestions that significantly improved the quality of this paper. This work was supported by Google/YouTube and the University of California Noyce Initiative. We are grateful to ElevenLabs (\url{https://elevenlabs.io}) and PlayAI (\url{https://play.ai/}) for granting us API access for voice generation. 

{
    \small
    \bibliographystyle{ieeenat_fullname}
    \bibliography{main}
}

\clearpage

\appendix
\newpage




\newpage
\onecolumn

\addcontentsline{toc}{section}{Appendix}
{%
  \hypersetup{
    linkcolor=black,
    citecolor=black,
    filecolor=black,
    urlcolor=black
}
  \part{Appendix}%
  \parttoc%
}

\newpage
\section{Release and Usage Information}
\label{app:release_info}

We released the DeepSpeak dataset in three separate batches: versions 1.0, 1.1 and 2.0. Additional real data was collected between versions 1.x and 2.0 (220 and 280 identities respectively). The training and testing splits of each version are detailed in Table \ref{tab:summary_of_versions}. Furthermore, different generation engines were used between versions 1.x and 2.0 to reflect the current state-of-the-art methods at the time of release. The details of which engines were used in each version are shown in Table \ref{tab:version_info}. Version 1.1 corrects for minor errors from version 1.0. As such, we recommend combining versions 1.1 and 2.0 when creating the complete dataset.

\begin{itemize}
    \item Version 1.0: \url{https://huggingface.co/datasets/faridlab/deepspeak_v1} 
    \item Version 1.1: \url{https://huggingface.co/datasets/faridlab/deepspeak_v1_1} 
    \item Version 2.0: \url{https://huggingface.co/datasets/faridlab/deepspeak_v2}
\end{itemize}

\begin{table}[H]
    \centering
    \resizebox{\textwidth}{!}{
    \begin{tabular}{c|ccc|cc|cc}
                & \multicolumn{3}{c|}{{\bf Total}} & \multicolumn{2}{c|}{{\bf Train}} & \multicolumn{2}{c}{{\bf Test}} \\
        \textbf{Ver} & size (GB) & size (N) & size (hrs) & real (N [hrs]) & fake (N [hrs]) & real (N [hrs]) & fake (N [hrs]) \\
        \hline 
        1.0 & 40 & 13{\small,}025 & 44.3 & 
            4{\small,}902 [13.9] & 5{\small,}300 [21.0] & 
            1{\small,}324 [3.7] & 1{\small,}499 [5.8] \\
        1.1 & 46 & 13{\small,}463 & 
            48.0 & 5{\small,}251 [16.8] & 5{\small,}299 [21.0] & 
            1{\small,}416 [4.4] & 1{\small,}497 [5.8] \\
        2.0 & 124 & 16{\small,}585 & 52.7 &
            7{\small,}513 [23.6] & 5{\small,}793 [18.6] &
            1{\small,}863 [5.8] & 1{\small,}416 [4.6] \\
        \hline
    \end{tabular}
    }
    \vspace{0.25cm}
    \caption{A breakdown of the total size (gigabytes (GB), number of files (N), and length in hours (hrs)) of each version of the DeepSpeak dataset.}
    \label{tab:summary_of_versions}
\end{table}

\begin{table}[H]
    \centering
    \begin{tabular}{c|p{2.8cm}|p{2.8cm}|p{2.8cm}|p{2.8cm}}
    
    \textbf{Ver} & \textbf{Audio} & \textbf{Face-swap} & \textbf{Lip Sync} & \textbf{Avatar} \\
    \hline
    1.x & 
    \begin{tabular}[t]{@{}l@{}}ElevenLabs\end{tabular} &
    \begin{tabular}[t]{@{}l@{}}FaceFusion\\FaceFusion\\  \ \ + \textit{GAN}\\FaceFusion Live\end{tabular} &
    \begin{tabular}[t]{@{}l@{}}Wav2Lip\\VideoRetalking\end{tabular} & ---
    \\
    \hline
    2.0 & 
    \begin{tabular}[t]{@{}l@{}}ElevenLabs\\PlayAI\\Speechify\end{tabular} &
    \begin{tabular}[t]{@{}l@{}}INSwapper\\INSwapper\\ \ \ + \textit{CodeFormer}\\SimSwap\\SimSwap\\ \ \ + \textit{RestoreFormer}\end{tabular} &
    \begin{tabular}[t]{@{}l@{}}Diff2Lip\\LatentSync\end{tabular} &
    \begin{tabular}[t]{@{}l@{}}LivePortrait\\HelloMeme\\Memo\end{tabular} \\
    \hline
    \end{tabular}
    \vspace{0.3cm}
    \caption{An overview of deepfake generation engines used in each release version of the dataset.}
    \label{tab:version_info}
\end{table}

\newpage
\section{Related Work}
\label{app:prior_work}


\subsection{Video}

Table~\ref{tab:datasets} presents an overview of existing video deepfake datasets. None of these datasets contain all types of deepfakes (face-swap, lip-sync, avatar, and audio); the majority did not obtain consent from the individuals featured. Notable datasets are further detailed below.

\paragraph{DFDC.} Released in 2020 as part of the DeepFake Detection Challenge, DFDC~\cite{dolhansky2020deepfake} contains $128{\small,}154$ face-swap deepfakes of $3{\small,}426$ paid, consenting actors. While the dataset brought significant attention to the problem of deepfake detection, it also sparked controversy due to failure cases (e.g., videos where the face-swap failed but were still labeled as deepfakes) and inconsistent annotations. Today, only a small subset of the labels is publicly available.

\paragraph{Celeb-DF.} While most prior datasets were scripted and studio-recorded, Celeb-DF~\cite{li2020celeb} aimed to mimic the in-the-wild nature of deepfake detection. It includes $590$ real and $5{\small,}639$ deepfake videos of $59$ individuals who did not provide consent for such use. These were celebrities, mostly taken from YouTube videos. Like DFDC, the deepfakes were generated using face-swap models, but unlike DFDC, Celeb-DF's annotations are consistent and fully available.

\paragraph{DF40.} DF40~\cite{yan2024df40} contains over $100{,}000$ deepfake videos spanning various types (lip-sync, face-swap, and avatar), featuring individuals who did not provide consent for inclusion. Identity and real video statistics are not reported. In creating the dataset, the authors collected some new data and repurposed content from Celeb-DF, FFHQ, and other datasets.


\subsection{Audio}

Most existing audio deepfake and spoofing datasets are single-modal, focusing exclusively on audio. The key datasets in this area are outlined below.

\paragraph{ASVSpoof.} ASVspoof~\cite{asvspoof} is considered one of the most popular audio spoofing datasets, and is commonly used for training and evaluating deepfake detection models. There have been multiple releases of this dataset (including 2019 and 2021), released alongside the ASVspoof challenges for each corresponding year and updated in accordance with new tools and generation methods. The dataset contains three categories of data: Physical Access (pertaining to audio undergoing physical attack methods such as replay attacks), Logical Access (pertaining to audio created by Text-To-Speech and voice conversion systems), and Deepfake Audio (as with Logical Access, but with generalized compression and codec variation).

While ASVspoof is considered a popular benchmarking dataset, it poses three main issues that we sought to address through releasing DeepSpeak: (1) the TTS and VC systems do not leverage state-of-the-art commercial platforms that are popular with real-world adversaries; (2) there are few real speakers used in the training and validation sets (for example, 20 speakers in ASVspoof 2021); and (3) the audio is not paired with video. 

\paragraph{WaveFake.} WaveFake~\cite{frank2021wavefake} contains approximately 196 hours of both real and fake audio. The dataset is primarily based on the LJSPEECH dataset~\cite{ljspeech17} (a public English speech corpus), alongside the JSUT dataset~\cite{sonobe2017jsut} (a Japanese speech corpus). Both of these datasets include audio clips recorded by a single female speaker. As such, WaveFake poses two additional issues that we sought to address through releasing DeepSpeak: (1) only comprising two speaker identities; (2) providing largely scripted audios rather than conversational; (3) generation methods that are no longer considered state-of-the-art (including MelGAN and HiFi-Gan methods). 

\paragraph{FakeAVCeleb.} FakeAVCeleb~\cite{khalid2021fakeavceleb} is a multi-modal deepfake dataset. By way of fake audios, the dataset only comprises one generation method using a real-time voice cloning tool (SV2TTS~\cite{sv2tts}, released in 2019). By contrast, DeepSpeak encompasses three more recently released state-of-the-art commercial voice clone and accompanying TTS methods.

\section{Data Collection Details}
\label{app:data_collection}

\subsection{Real Participants} 

Participants were asked to give their consent for including their recordings, without any other identifying information, in a public dataset. The precise consent language was: ``This dataset will be used for research purposes for detecting deepfakes. Please note that your recordings will be made public in a dataset, but no other identifying information will be shared outside of our research group. Please select the option below to consent to participate in this study.''  The complete introductory page, including the consent information presented to participants, is available in Appendix~\ref{app:survey-materials}.

\subsection{Survey}

For scripted responses, participants were asked to record themselves repeating a short script while looking into the camera. Scripted responses were obtained using transcripts of the TIMIT dataset~\cite{timit}. The TIMIT dataset consists of 462 real female and male American-English speakers, uttering a total of 1{\small,}718 short-to-medium length phonetically-rich sentences. Sentences of length less than the mean of 50 characters were removed. Ten sentences were then selected at random for the standardized scripts read by all participants. The remaining 728 sentences comprised the randomized scripts, for which each participant read a random sample of 10. 

By way of unscripted prompts, participants  responded to four open-ended unscripted questions and were asked to aim for a response that was close to 30 seconds in length. These were followed by quick-fire questions in which they repeated the question and provided a short response. Following the scripted and unscripted responses, participants were asked to perform simple actions using head and hand gestures. 

By way of action prompts, participants were asked to perform seven simple visual actions: (1) wave their hand in front of their face while counting 1,2,3; (2) look down and right, straight down, and down and left, each time holding and counting 1,2,3; (3) look up and right, straight up, and up and left, each time holding and counting 1,2,3; (4) lean towards the camera without counting); (5) pretending to yawn; (6) pretending to laugh for between 2 to 5 seconds; (7) clapping loudly three times, pausing for about 5 seconds between claps.  

The full list of all prompts, including scripted, unscripted and action-based, can be found in Appendix~\ref{app:prompts}.

\subsection{Data Collection Tool}
\begin{figure*}[t]
    \centering
    \includegraphics[width=\textwidth]{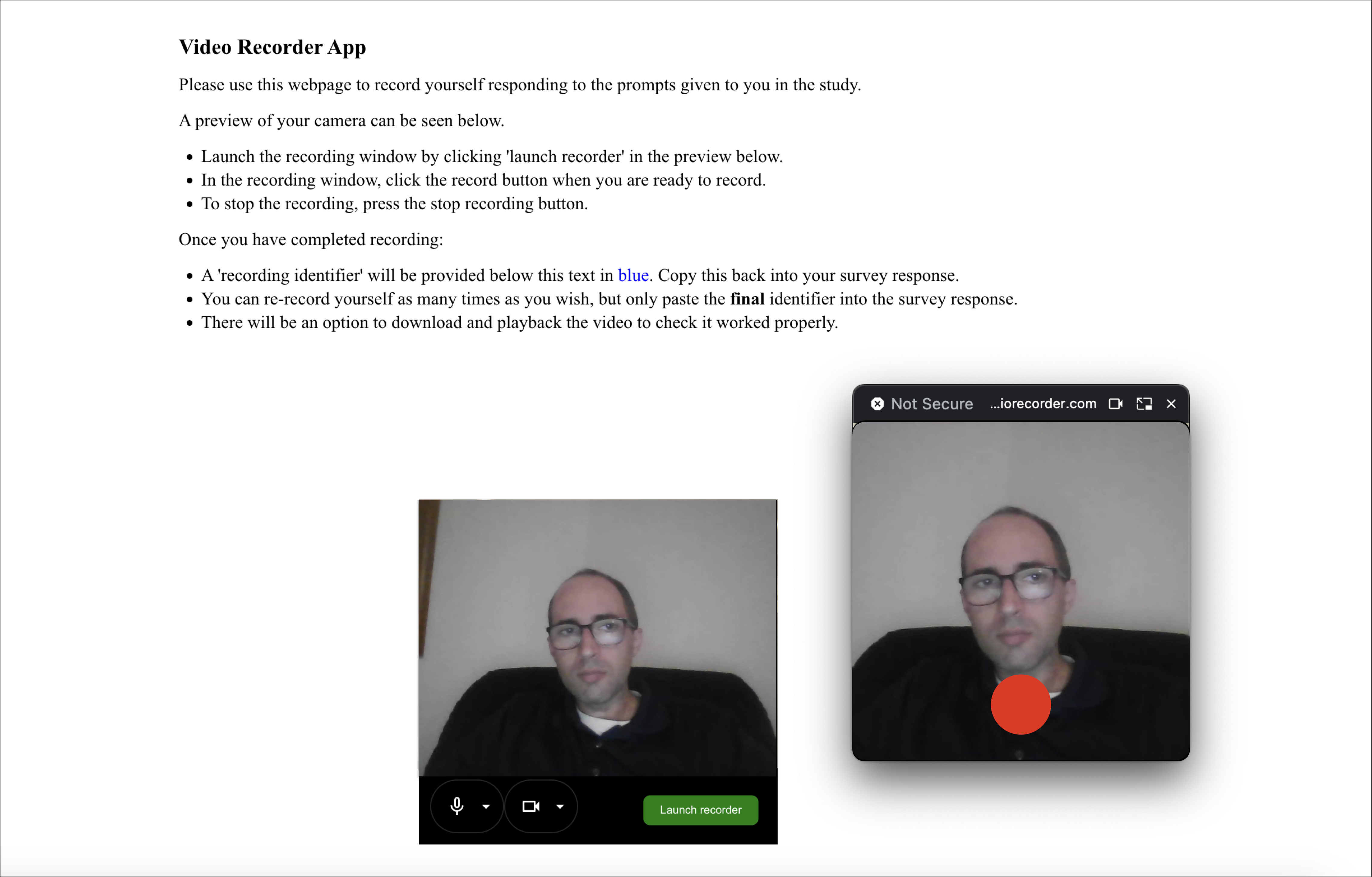}
    \caption{A screenshot of the custom-built recording tool used for real participant data collection. Participants were directed from the data collection survey to the URL hosting the tool. The landing page contained brief instructions for participants to use the recording functionality. Once recording was stopped, a unique identifier was shown to the participant pertaining to that specific recording. This was then input into the survey by the participant, and was used to match responses in the survey to recordings stored by the tool.}
    \label{fig:recorder_screenshot}
\end{figure*}

Both audio and video were recorded using a custom-built Google Chrome web application. Recordings were captured as .webm files with a bitrate of 8 megabits per second, using the Google VP9 codec for video compression. The target resolution was set at 1280 $\times$ 720 pixels, but users with limited bandwidth were able to record at lower resolution of 640 $\times$ 480 pixels. The JavaScript and Python repository for this web application is available at \url{https://github.com/hfaridlab/deepspeak}. A screenshot of the recording web application is shown in\ref{fig:recorder_screenshot}.

The audio/video recordings captured by the tool were then converted from their initial .webm format to .mp4 and organized by prompt type using the H.264 codec with the libx264 library (this final video conversion is not performed in version 2.0 in order to minimize re-encoding). While audio/video recordings for the first release, denoted by v1.x on Hugging Face (\url{https://huggingface.co/datasets/faridlab/deepspeak_v1_1}), were converted from their initial .webm format to .mp4 using the H.264 codec with the libx264 library, an FFmpeg ``copy'' command was used in the second release, denoted by v2.x on Hugging Face (\url{https://huggingface.co/datasets/faridlab/deepspeak_v2}), to avoid re-encoding and preserve recording quality. The tool output a unique identifier per participant recording that was later used to match recordings to survey responses. To generate video deepfakes, all source videos were re-encoded using FFmpeg with the H.264 codec at a constant rate factor of 18. The encoding present was set to slow, and the original audio stream was preserved without re-encoding.

\subsection{Real Video Validation}
Participants were given written and visual instructions to allow them to practice recording themselves and test their hardware. Participants were asked to adhere to a series of recording conditions intended to improve consistency within the overall dataset, including positioning themselves centrally within their web camera frame, sitting in a well lit room, not allowing other faces or people to be present in the frame, and minimizing background noise. We manually removed any invalid responses from the final dataset. This included ensuring that only a single person was visible in the video, the scene was reasonably well lit, and each submitted video clip contained a valid audio and visual track. A full list of these conditions is shown in the survey screenshots in Appendix~\ref{app:survey-materials}.

\section{Video Generation Methods}
\label{app:vid_gen_methods}

\subsection{Face-Swap}

\begin{itemize}

    \item[] {\bf FaceFusion} The first face-swap configuration invokes the FaceFusion~\cite{facefusion} library with its default parameters. FaceFusion, built on InsightFace~\cite{ren2023pbidr}, localizes the face in the original video, performs 3D landmark estimation of the face, and superimposes the appropriately transformed matched face. This is followed by a set of post-processing steps to hide edge artifacts and improve temporal consistency.
    \item[] {\bf FaceFusion + GAN:} The second face-swap configuration extends the above FaceFusion configuration by enhancing each generated frame using the CodeFormer GAN~\cite{zhou2022codeformer}. This model corrects rendering discrepancies--mostly misshapen teeth and lips--and increases the overall photorealism of the face.
    \item[] {\bf FaceFusion Live:} The third face-swap configuration wraps the same FaceFusion configuration in a simulated live input streaming environment. Because our hardware cannot generate video frames in real-time ($24$ to $30$ frames per second (fps)), the effective frame-rate of the generated video is decreased to approximately $12$ fps.
    \item[] {\bf INSwapper:} The first face-swap configuration invokes the INSwapper-128 model~\cite{inswapper} with default parameters. This model leverages InsightFace~\cite{ren2023pbidr} to localize 3D facial landmarks and superimpose a transformed face of the matched identity onto the source identity video. This version is most similar to FaceFusion in version 1.0.
    \item[] {\bf INSwapper + CodeFormer:} The second face-swap configuration extends the above INSwapper configuration by enhancing each frame of the generated deepfake using CodeFormer~\cite{zhou2022towards}. This model was trained to fix any misshapen teeth or lips and improve the overall photorealism of the generated face. This version is most similar to FaceFusion+GAN in version 1.0.
    \item[] {\bf SimSwap:} The third face-swap configuration invokes the SimSwap-256 model~\cite{chen2020simswap} with default parameters. Unlike INSwapper, which directly modifies the target face based on detected facial landmarks, SimSwap extracts a latent identity representation to guide the generation of new frames.
    \item[] {\bf SimSwap + RestoreFormer:} The fourth face-swap configuration extends the above SimSwap configuration by enhancing each frame of the generated deepfake using RestoreFormer++~\cite{wang2023restoreformer}. This model, trained on degraded photographs, corrects any structural deficiencies and enhances the photorealism of the generated frames.

\end{itemize}

\subsection{Lip-Sync}

\begin{itemize}

    \item[]{\bf Wav2Lip:} The first lip-sync configuration invokes the Wav2Lip model~\cite{prajwal2020lip} with its default parameters. Wav2Lip is a neural network trained in a GAN-like generator-discriminator fashion. The generator architecture is a LipGAN~\cite{kr2019towards} trained on the LRS2 dataset~\cite{afouras2018deep}.
    \item[] {\bf VideoRetalking:} The second lip-sync configuration invokes the VideoRetalking pipeline~\cite{cheng2022videoretalking} with its default parameters. The input video is passed through three neural networks: (1) a semantic-guided reenactment model, which stabilizes the expression in the video; (2) a lip-sync model, which renders a new mouth and chin area matching the audio; and (3) a face enhancer, which fixes rendering discrepancies. The specific models employed in the first two stages are L-Net and D-Net; the third model is the CodeFormer GAN~\cite{zhou2022codeformer}.
    \item[] {\bf Diff2Lip:} The first lip-sync configuration invokes the Diff2Lip model~\cite{mukhopadhyay2024diff2lip} with default parameters. After localizing the face in the video, Diff2Lip adds noise over the mouth region and proceeds as an audio-conditioned diffusion model. It was trained on the Voxceleb2~\cite{chung2018voxceleb2} and LRW~\cite{yang2019lrw} datasets.
    \item[] {\bf LatentSync:} The second lip-sync configuration invokes the LatentSync model~\cite{li2024latentsync} with default parameters. While LatentSync is, at its core, a diffusion model similar to Diff2Lip, it differs in two aspects. First, it encodes the audio constraint as Whisper embeddings~\cite{radford2023robust} instead of chunks of audio signal. Second, it uses a noise mask delineated by facial landmarks to exactly match the shape of the face as opposed to a rectangular bounding box. LatentSync was trained on the VoxCeleb2~\cite{chung2018voxceleb2} and HDTF~\cite{zhang2021flow} datasets.

\end{itemize}

\subsection{Avatar}

\begin{itemize}

     \item[] {\bf LivePortrait:} The first avatar configuration invokes the LivePortrait model~\cite{guo2024liveportrait} with its default parameters. LivePortrait is a multi-stage model that first extracts the identity and motion embeddings, which are warped into a joint embedding that is later decoded into pixel space. This model was trained on the VoxCeleb~\cite{nagrani2017voxceleb}, MEAD~\cite{wang2020mead}, RAVDESS~\cite{livingstone2018ryerson}, and AAHQ~\cite{liu2021blendgan} datasets.
     \item[] {\bf HelloMeme:} The second avatar configuration invokes the HelloMeme~\cite{zhang2024hellomeme} model with its default parameters. HelloMeme consists of three modules: a reference module, which extracts an identity embedding; a control module, which extracts information about the head and mouth shape; and a diffusion module, which generates the resulting video frames. This model was trained on the CelebV-HQ~\cite{xie2022vfhq} and VFHQ~\cite{xie2022vfhq} datasets.
     \item[] {\bf Memo:} The third avatar configuration invokes the Memo model~\cite{zheng2024memo} with its default parameters. Memo is a diffusion model that combines identity, voice, and emotion embeddings as diffusion constraints. It was trained on a compilation of the HDTF~\cite{zhang2021flow}, VFHQ~\cite{xie2022vfhq}, CelebV-HQ~\cite{zhu2022celebv}, MultiTalk~\cite{sung2024multitalk}, and MEAD~\cite{wang2020mead} datasets.
\end{itemize}

\section{Audio Generation Methods}
\label{app:audio_gen_methods}

Three voice clone providers were used to create AI voice clones of all 500 participants, and generate Text-to-Speech audio using these clones. 

Firstly, the ElevenLabs Create Voice endpoint (\url{https://elevenlabs.io/docs/api-reference/voices/add}) was used for voice clone generation, followed by the Create Speech API (\url{https://elevenlabs.io/docs/api-reference/text-to-speech/convert}) for speech synthesis. The \verb|eleven_multilingual_v2| model was used, with audio output returned in MP3 format at 44.1 kHz. The API was accessed in April 2024.

Secondly, the PlayAI Instant Voice Cloning endpoint (\url{https://docs.play.ai/reference/api-create-instant-voice-clone}) was used for voice clone generation. The \verb|pyht| Python package, with TTSOptions set to default parameters was used for speech synthesis, with audio output provided in MP3 format, at a sampling rate of 24 kHz. 

Finally, the Speechify v1 Voices API (\url{https://docs.sws.speechify.com/v1/api-reference/api-reference/tts/voices/create}) was used for voice clone generation. The Speech API endpoint (\url{https://docs.sws.speechify.com/v1/api-reference/api-reference/tts/audio/speech}) was used for speech synthesis. The \verb|simba-english| model was used, with output audio returned in WAV format at 48 kHz. 

API calls for version 1 were made in April 2024 (ElevenLabs only), and version 2 calls were made in November 2024.

\newpage
\section{Video Validation}
\label{app:validation}

While failure types 3, 4, and 5 are general deepfake engine errors that are not flagged by the engines themselves, we found that failure types 1 and 2 often stem from certain features (or lack thereof) in the input. To achieve good performance with face-swap deepfakes, for example, the inputted image needs to show the individual facing the camera, with open eyes and a closed mouth. The same applies to the first frame of videos used at input to avatar deepfakes.

The suite of input and output detectors to filter undesired features performs the following validation types:

\paragraph{Input Validation.} Before an image is used as input to a face-swap deepfake engine and before the first frame of a video is used for an avatar-based deepfake, the following criteria are validated. If any of the following criteria are not satisfied, a different image or video is chosen: (1) The participant's face is fully visible and positioned in the center of the frame, facing the camera. This is established by verifying that there are no hand occlusions over the paricipant's face and the overall position and orientation of the participant; (2) The participant's eyes are open. This is established based on the distance of the X and Y landmark coordinates; or (3) The participant's mouth is open. This is established based on the distance of the upper and lower lip landmark coordinates.
     
\paragraph{Output Validation.} For each generated video, the following features are validated. If any of these criteria are not satisfied, the video is dropped from the dataset: (1) (\textit{Lip-sync only}) The distance between spectrograms of the original video and the target audio are above a threshold; (2) (\textit{Face-swap only}) The protagonist's face is more similar to the face of the target identity over the original identity. This is validated by the cosine distance of CLIP embeddings and Structural similarity index measure (SSIM) of faces cropped using MediaPipe~\cite{lugaresi2019mediapipe}; or (3) No frames of the video are fully black.

\section{Video Statistics}
\label{app:vid_stats}

\begin{table}[h]
    \centering
    \caption{A breakdown of the total size (gigabytes (GB), number of files (N), and length in hours (hrs)) of each version of the DeepSpeak dataset.}
    \vspace{0.2cm}
    \resizebox{\textwidth}{!}{
\begin{tabular}{c|ccc|cc|cc}
    \hline \noalign{\vskip 0.5ex}
                & \multicolumn{3}{c|}{{\bf Total}} & \multicolumn{2}{c|}{{\bf Train}} & \multicolumn{2}{c}{{\bf Test}} \\
        \textbf{Version} & size (GB) & size (N) & size (hrs) & real (N [hrs]) & fake (N [hrs]) & real (N [hrs]) & fake (N [hrs]) \\
        \hline 
        v1 & 46 & 13{\small,}463 & 
            48.0 & 5{\small,}251 [16.8] & 5{\small,}299 [21.0] & 
            1{\small,}416 [4.4] & 1{\small,}497 [5.8] \\
        v2 & 124 & 16{\small,}585 & 52.7 &
            7{\small,}513 [23.6] & 5{\small,}793 [18.6] &
            1{\small,}863 [5.8] & 1{\small,}416 [4.6] \\
        \hline
        \textbf{Total} & \textbf{170} & \textbf{30{\small,}048} & \textbf{100.7} &
        \textbf{12{\small,}764 [40.4]} & \textbf{11{\small,}092 [39.6]} &
        \textbf{3{\small,}279 [10.2]} & \textbf{2{\small,}913 [10.4]} \\
        \hline
    \end{tabular}
    }
    \label{tab:vid_stats}
\end{table}
\newpage

\section{Deepfake Video Frame Examples}
\label{app:deepfake_video_frame_examples}

%
%
%
%


\begin{figure}[H]
    \centering
    \resizebox{0.92\textwidth}{!}{
    \begin{tabular}{c@{\hspace{0cm}}c@{\hspace{0cm}}c@{\hspace{0cm}}c}
        \includegraphics[width=0.24\linewidth]{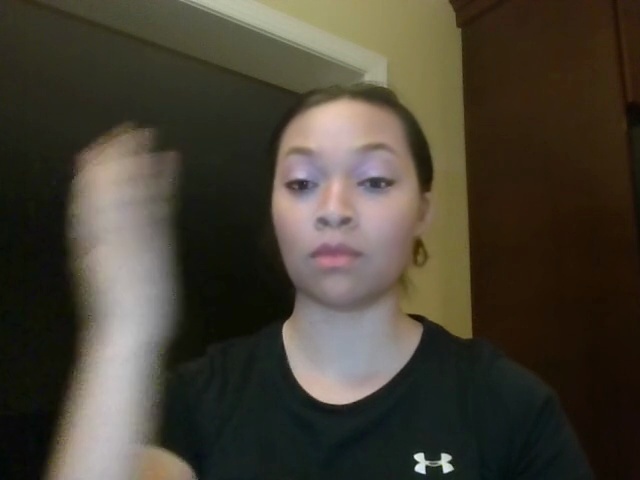} & 
        \includegraphics[width=0.24\linewidth]{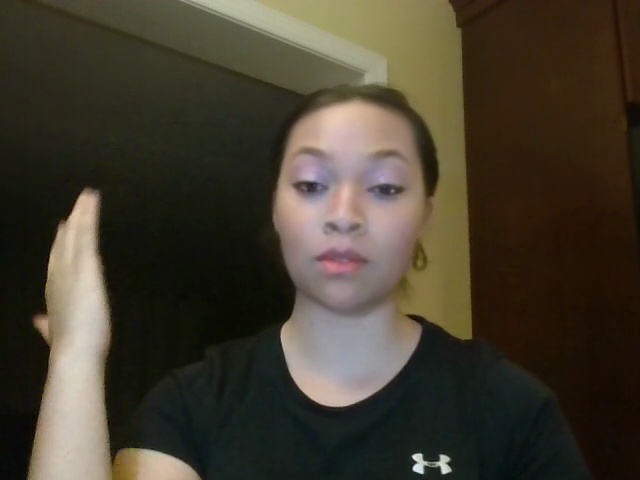} & 
        \includegraphics[width=0.24\linewidth]{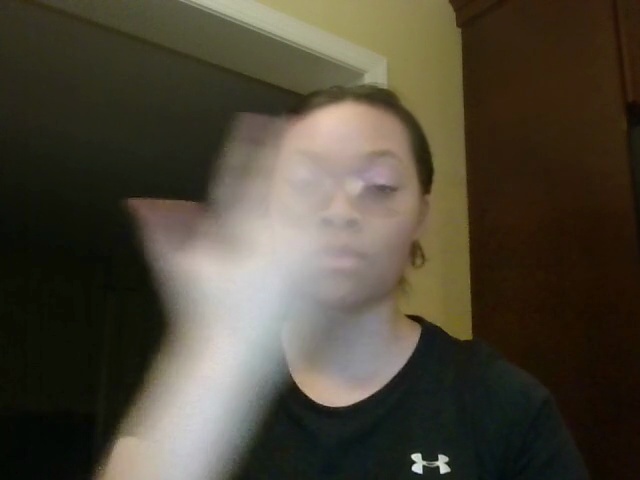} & 
        \includegraphics[width=0.24\linewidth]{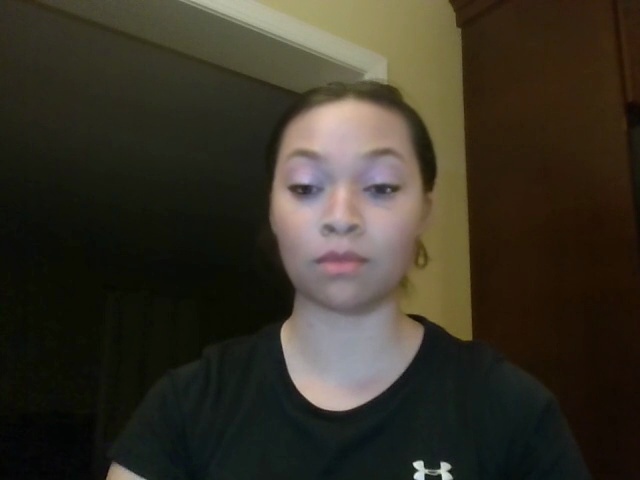} \\
        \includegraphics[width=0.24\linewidth]{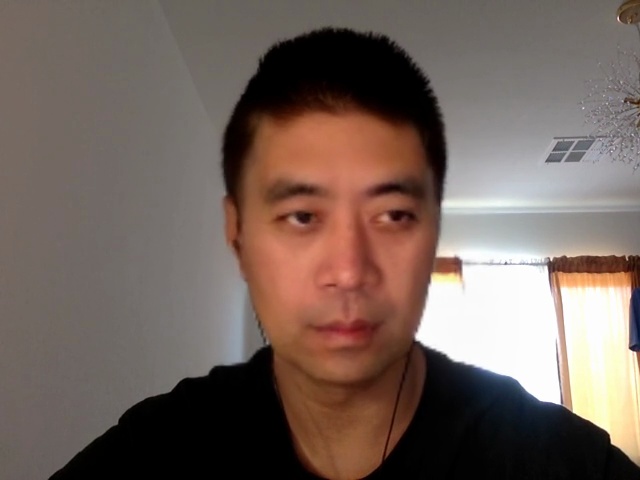} & 
        \includegraphics[width=0.24\linewidth]{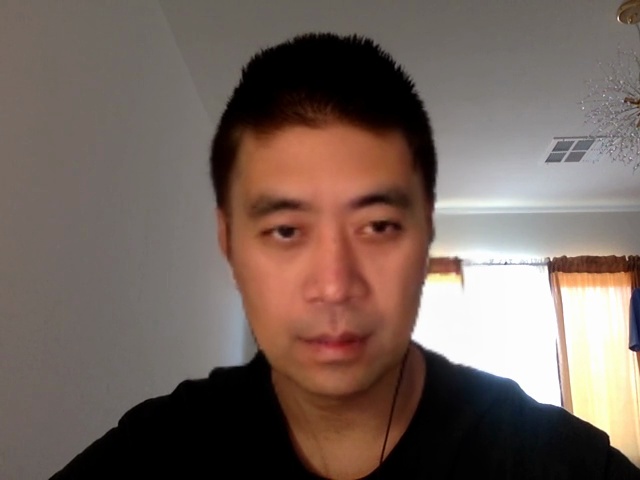} & 
        \includegraphics[width=0.24\linewidth]{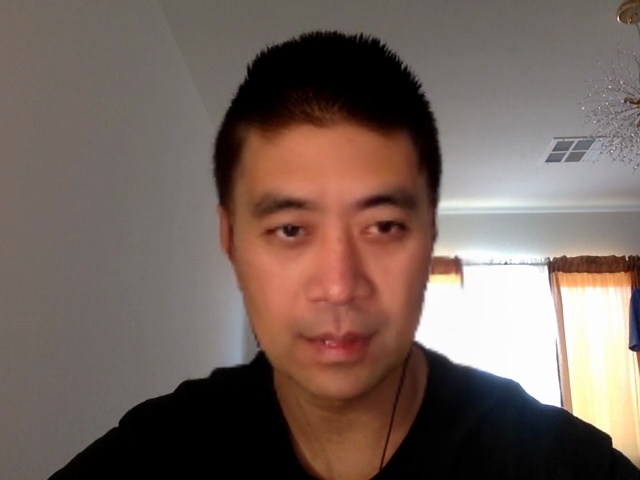} & 
        \includegraphics[width=0.24\linewidth]{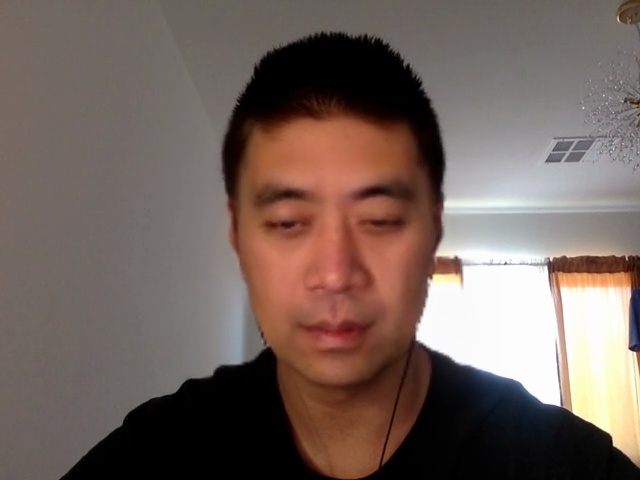} \\
        \includegraphics[width=0.24\linewidth]{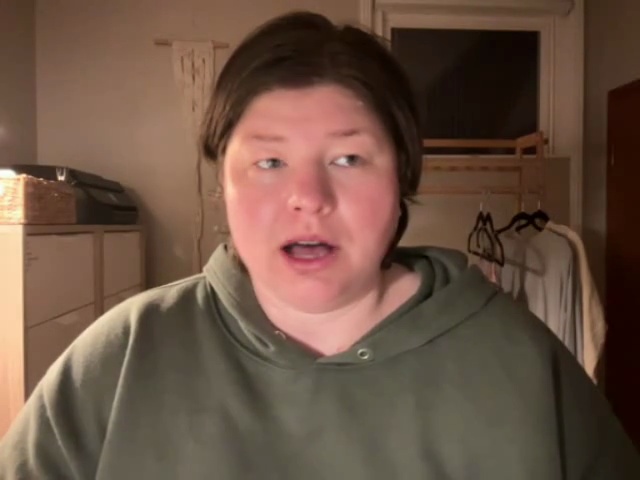} & 
        \includegraphics[width=0.24\linewidth]{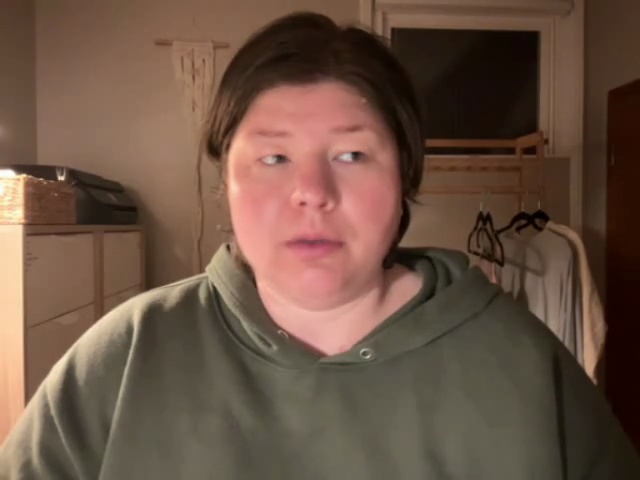} & 
        \includegraphics[width=0.24\linewidth]{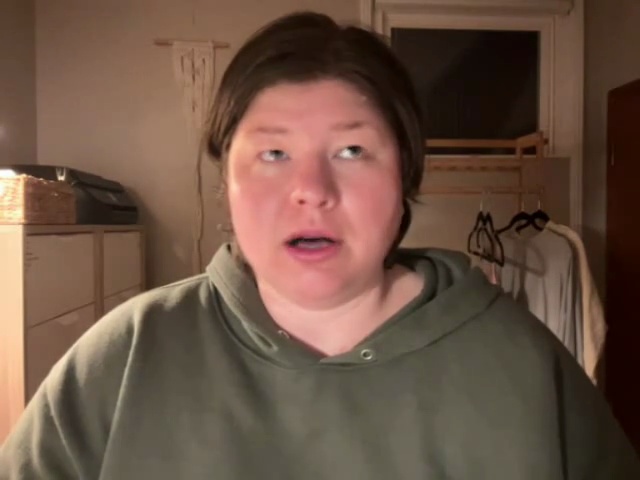} & 
        \includegraphics[width=0.24\linewidth]{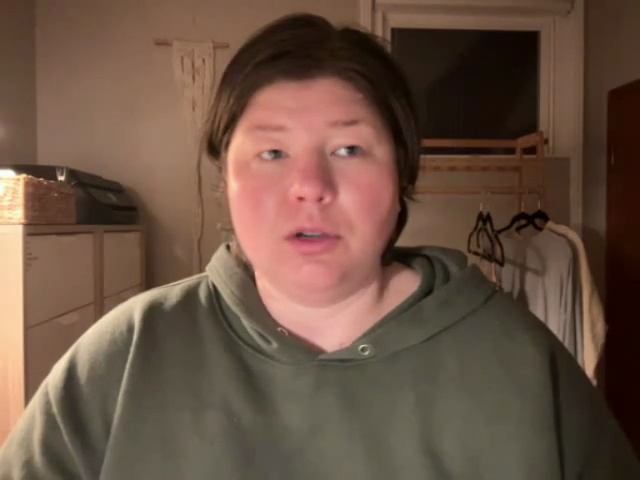}
    \end{tabular}
    }
    \caption{Video frames sampled from three representative examples of \textbf{face-swap} deepfakes generated using \textbf{FaceFusion}.}
    \label{fig:frame_examples_facefusion}
\end{figure}


\begin{figure}[H]
    \centering
    \resizebox{0.92\textwidth}{!}{
    \begin{tabular}{c@{\hspace{0cm}}c@{\hspace{0cm}}c@{\hspace{0cm}}c}
        \includegraphics[width=0.24\linewidth]{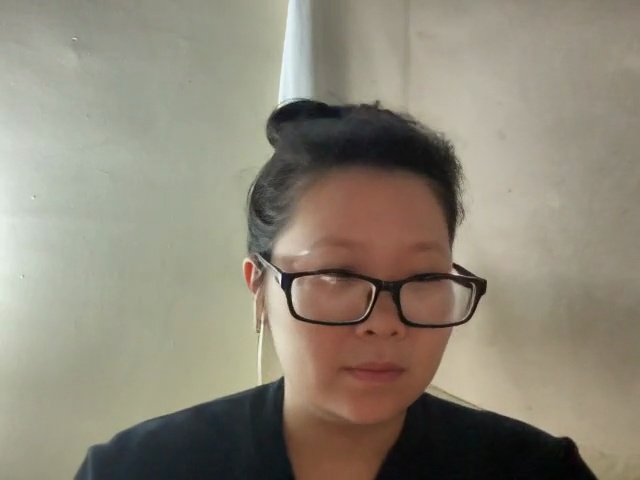} & 
        \includegraphics[width=0.24\linewidth]{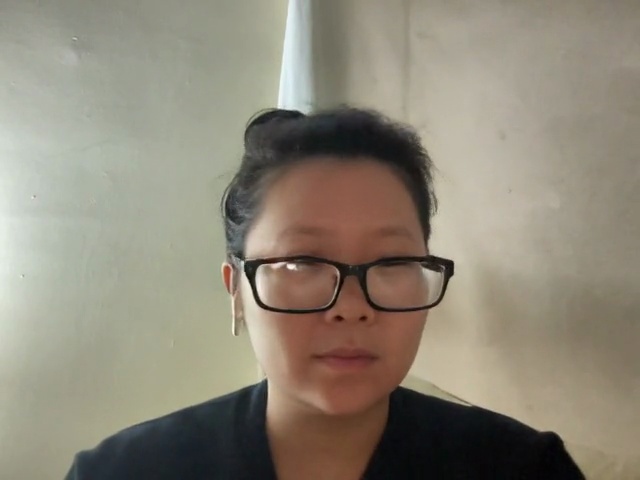} & 
        \includegraphics[width=0.24\linewidth]{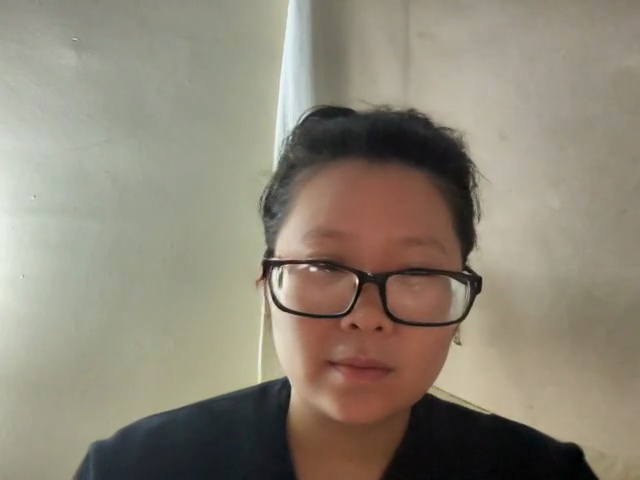} & 
        \includegraphics[width=0.24\linewidth]{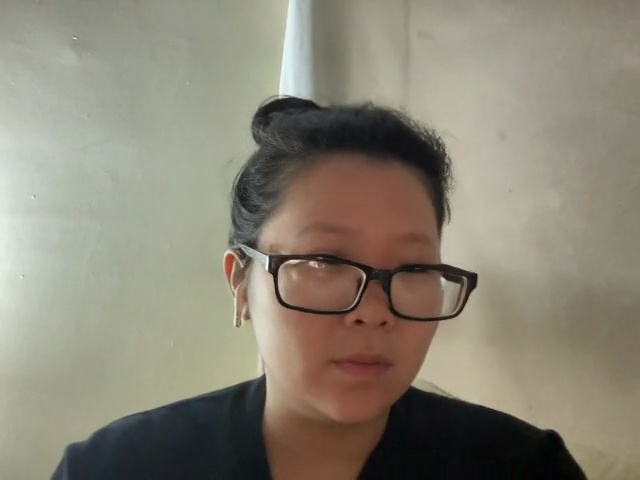} \\
        \includegraphics[width=0.24\linewidth]{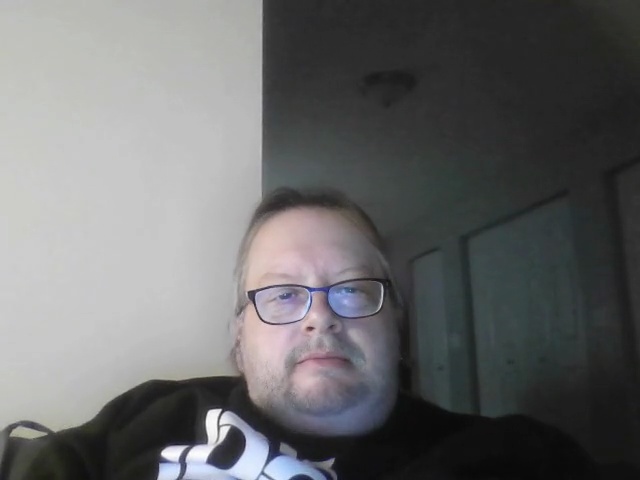} & 
        \includegraphics[width=0.24\linewidth]{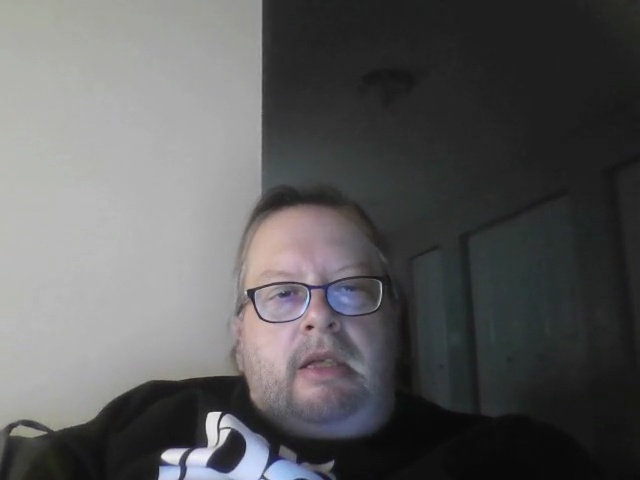} & 
        \includegraphics[width=0.24\linewidth]{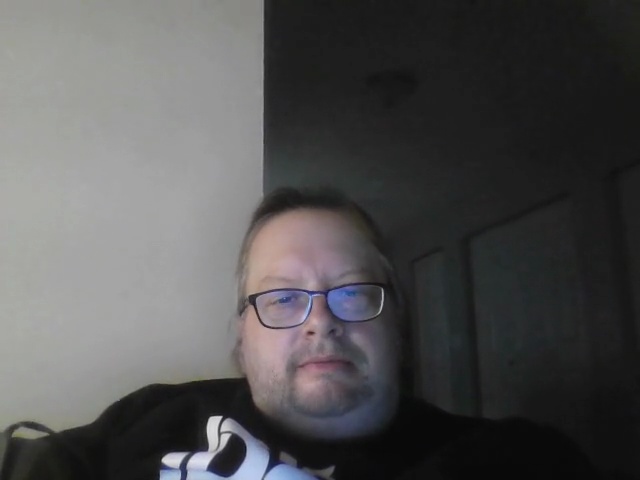} & 
        \includegraphics[width=0.24\linewidth]{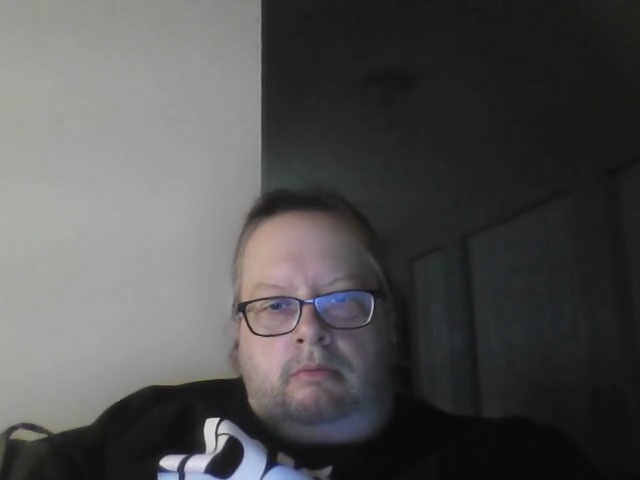} \\
        \includegraphics[width=0.24\linewidth]{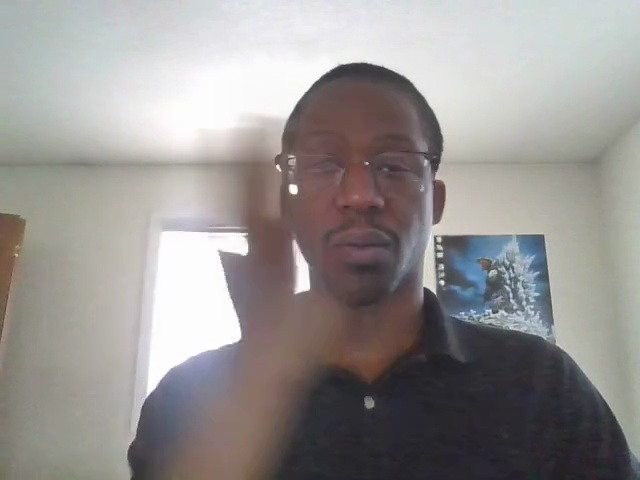} & 
        \includegraphics[width=0.24\linewidth]{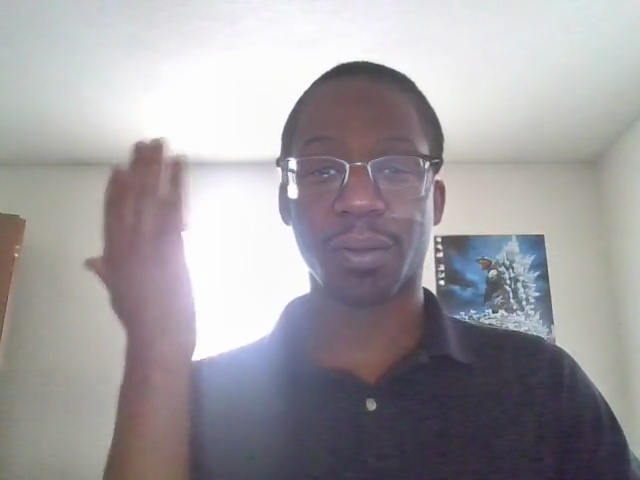} & 
        \includegraphics[width=0.24\linewidth]{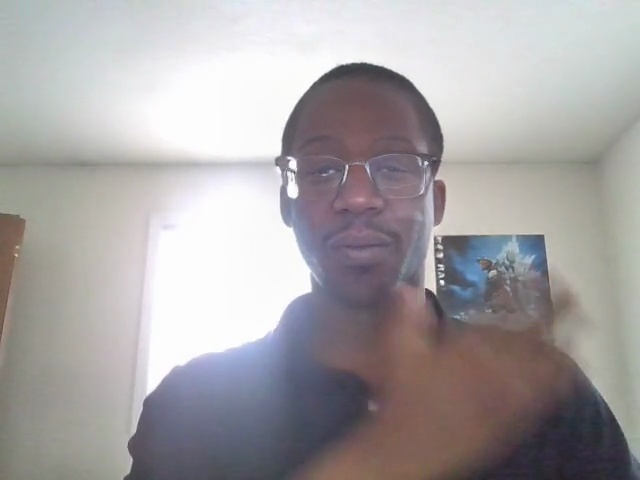} & 
        \includegraphics[width=0.24\linewidth]{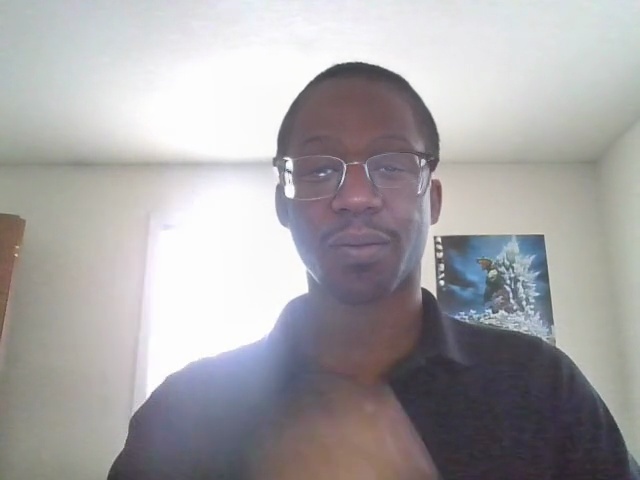}
    \end{tabular}
    }
    \caption{Video frames sampled from three representative examples of \textbf{face-swap} deepfakes generated using \textbf{FaceFusion + GAN}.}
    \label{fig:frame_examples_facefusion_gan}
\end{figure}

\newpage


\begin{figure}[H]
    \centering
    \resizebox{0.92\textwidth}{!}{
    \begin{tabular}{c@{\hspace{0cm}}c@{\hspace{0cm}}c@{\hspace{0cm}}c}
        \includegraphics[width=0.24\linewidth]{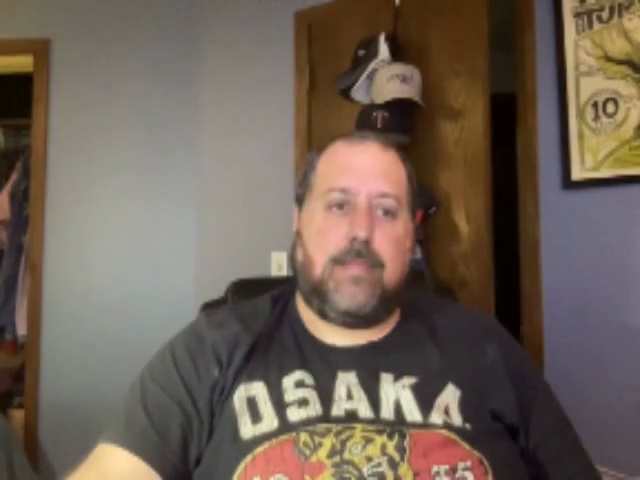} & 
        \includegraphics[width=0.24\linewidth]{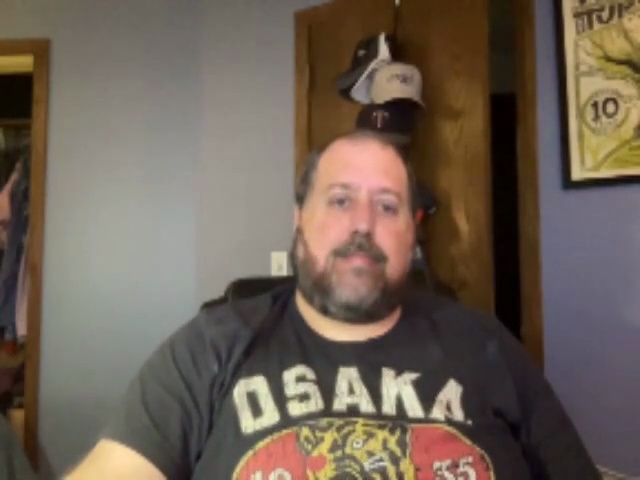} & 
        \includegraphics[width=0.24\linewidth]{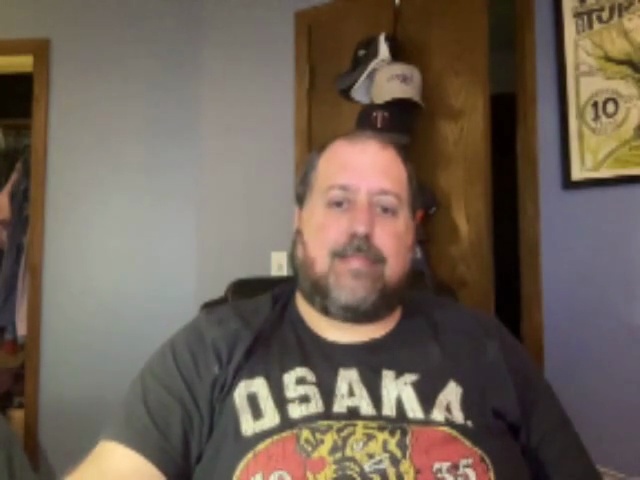} & 
        \includegraphics[width=0.24\linewidth]{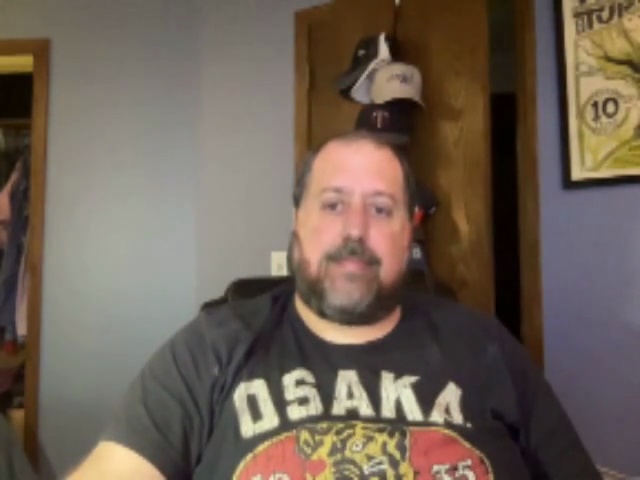} \\
        \includegraphics[width=0.24\linewidth]{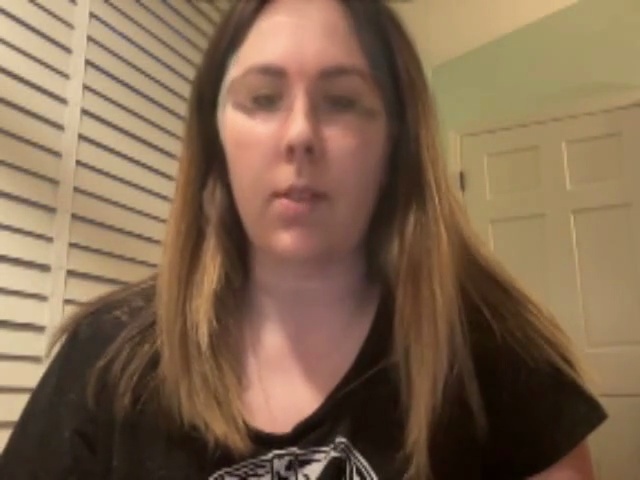} & 
        \includegraphics[width=0.24\linewidth]{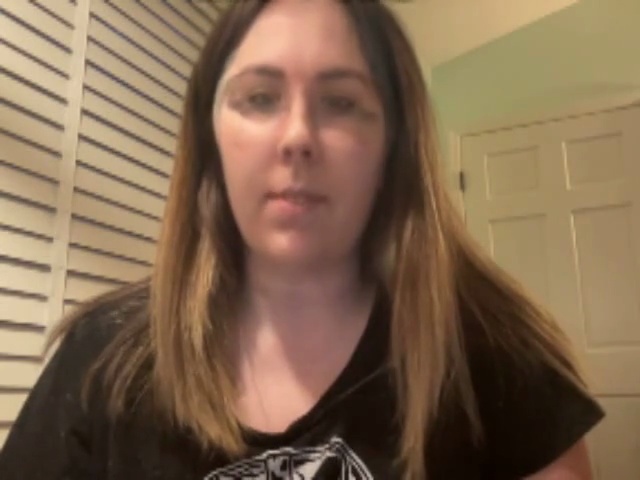} & 
        \includegraphics[width=0.24\linewidth]{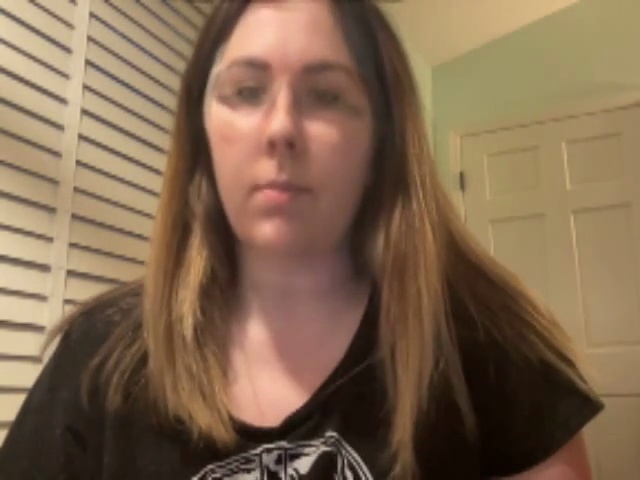} & 
        \includegraphics[width=0.24\linewidth]{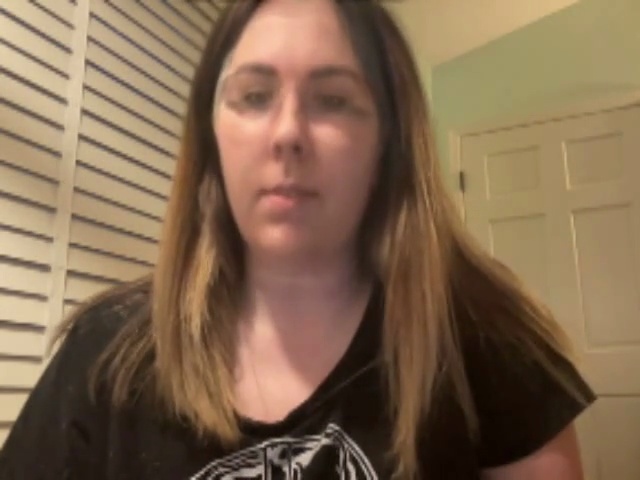} \\
        \includegraphics[width=0.24\linewidth]{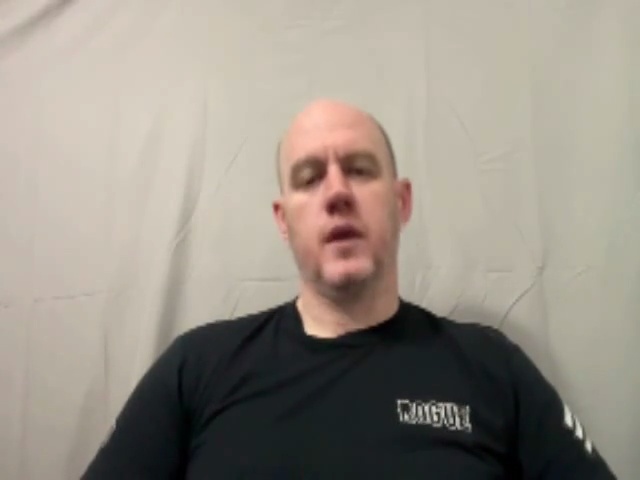} & 
        \includegraphics[width=0.24\linewidth]{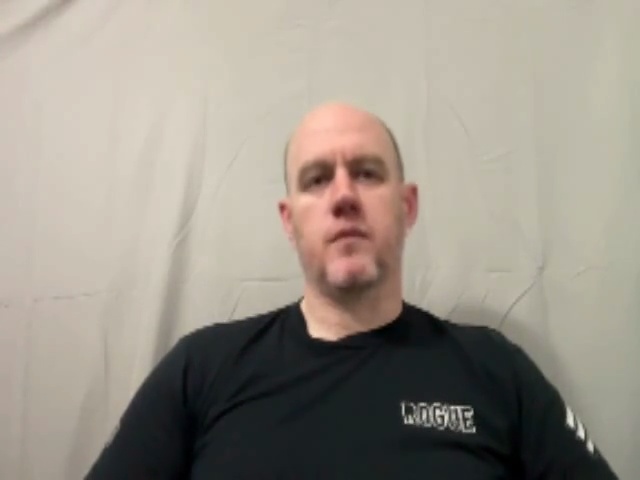} & 
        \includegraphics[width=0.24\linewidth]{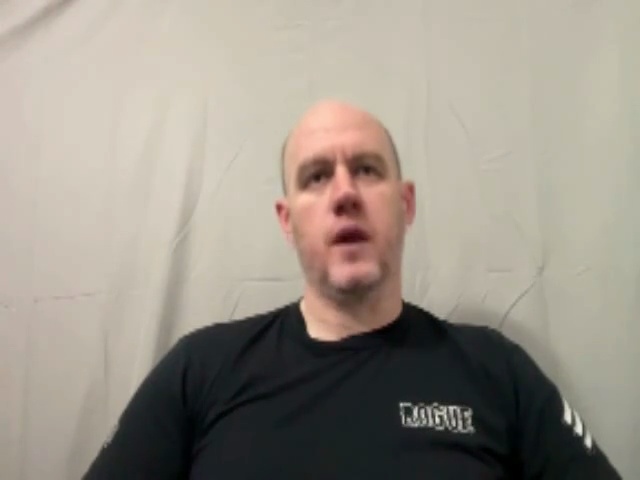} & 
        \includegraphics[width=0.24\linewidth]{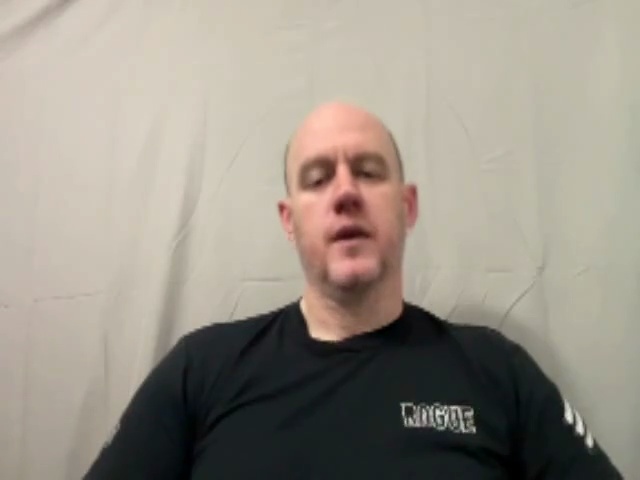}
    \end{tabular}
    }
    \caption{Video frames sampled from three representative examples of \textbf{face-swap} deepfakes generated using \textbf{FaceFusion Live}.}
    \label{fig:frame_examples_facefusion_live}
\end{figure}


\begin{figure}[H]
    \centering
    \resizebox{0.92\textwidth}{!}{
    \begin{tabular}{c@{\hspace{0cm}}c@{\hspace{0cm}}c@{\hspace{0cm}}c}
        \includegraphics[width=0.24\linewidth]{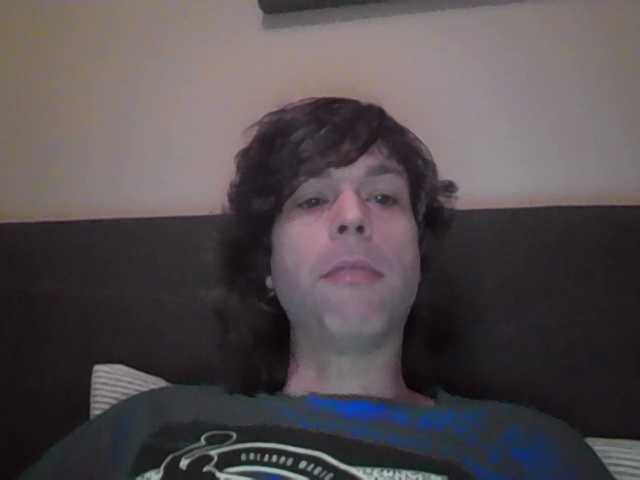} & 
        \includegraphics[width=0.24\linewidth]{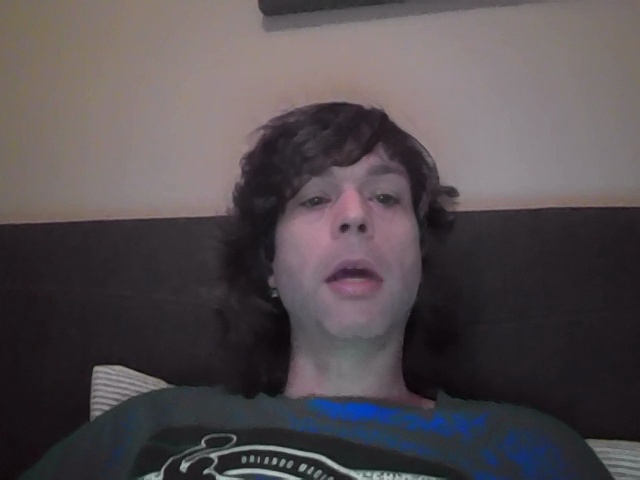} & 
        \includegraphics[width=0.24\linewidth]{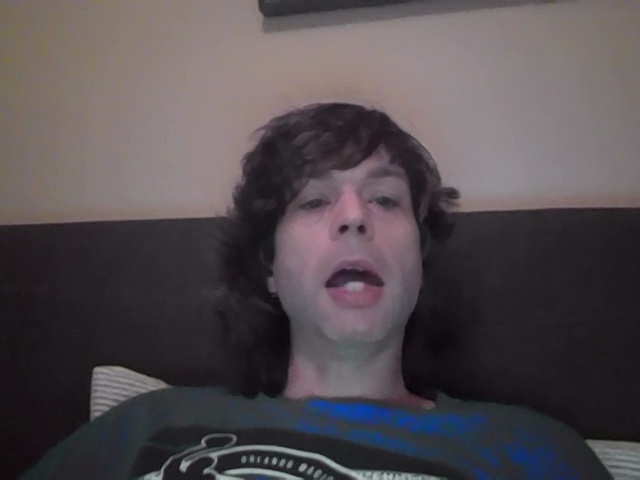} & 
        \includegraphics[width=0.24\linewidth]{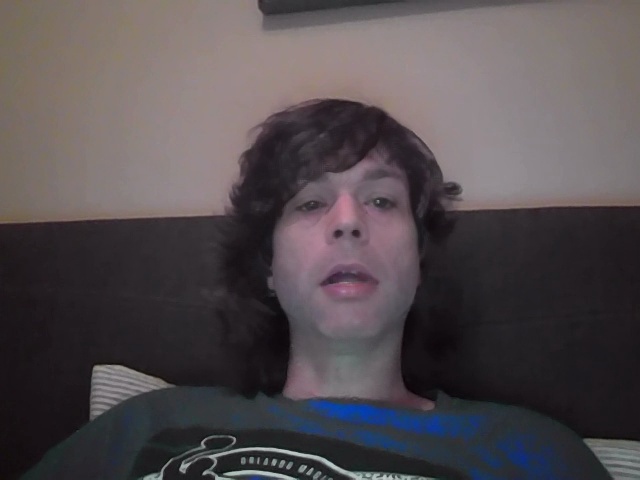} \\
        \includegraphics[width=0.24\linewidth]{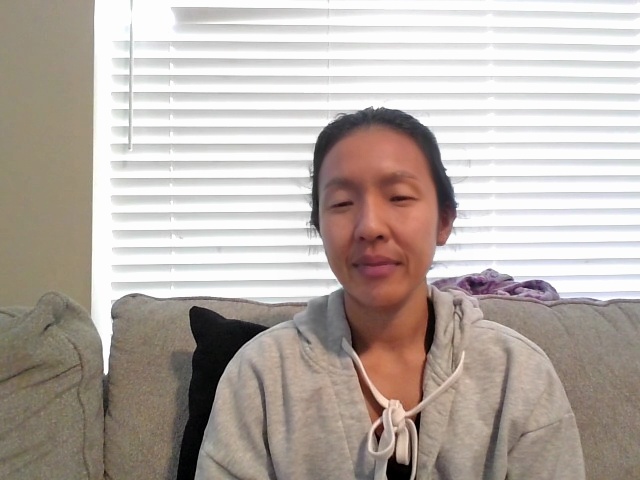} & 
        \includegraphics[width=0.24\linewidth]{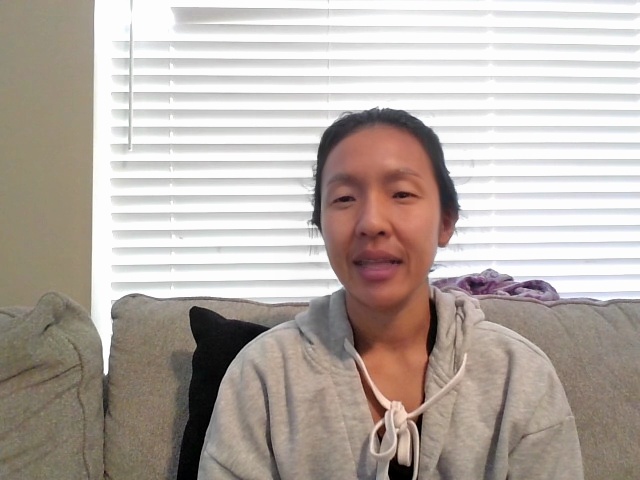} & 
        \includegraphics[width=0.24\linewidth]{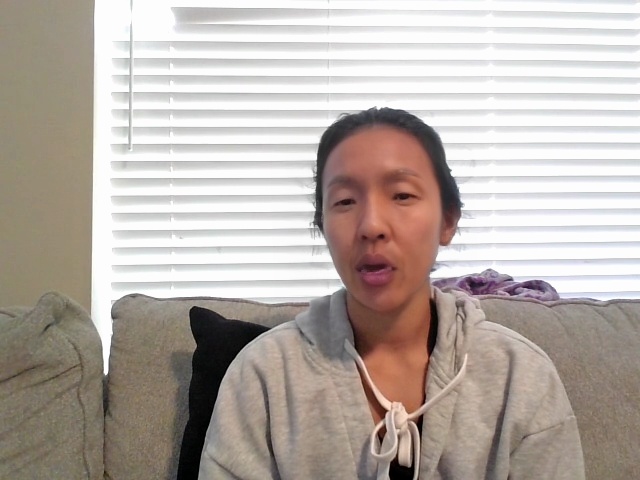} & 
        \includegraphics[width=0.24\linewidth]{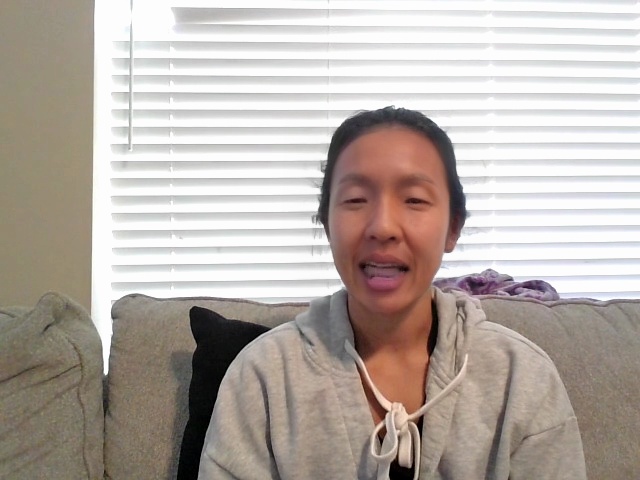} \\
        \includegraphics[width=0.24\linewidth]{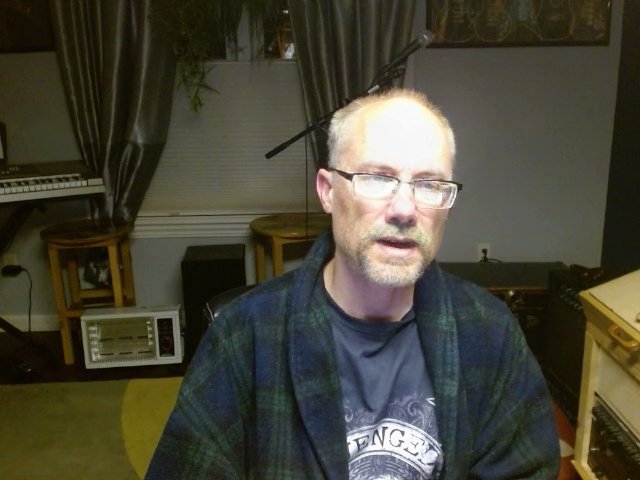} & 
        \includegraphics[width=0.24\linewidth]{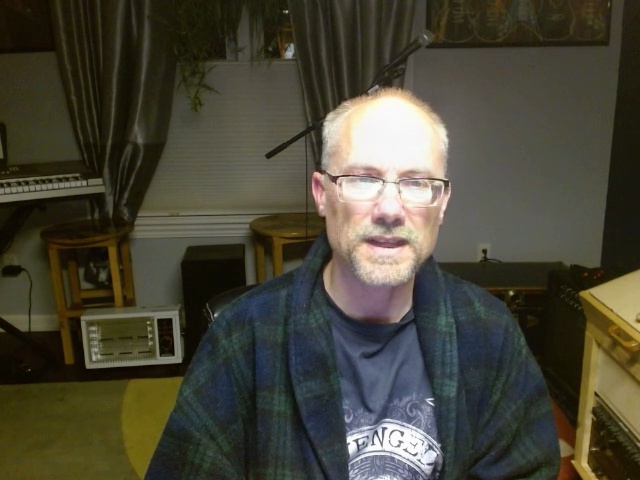} & 
        \includegraphics[width=0.24\linewidth]{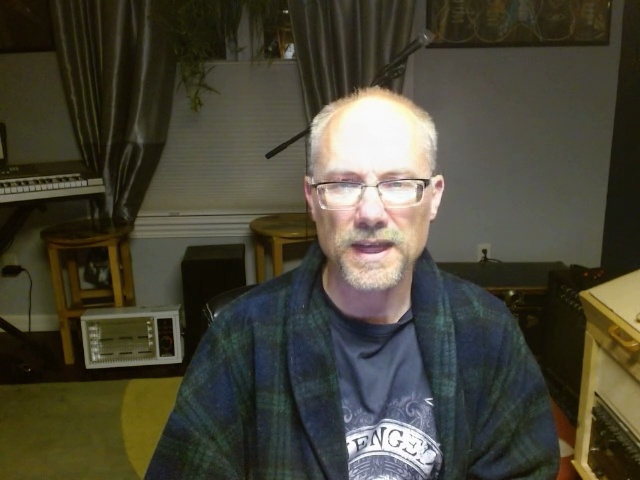} & 
        \includegraphics[width=0.24\linewidth]{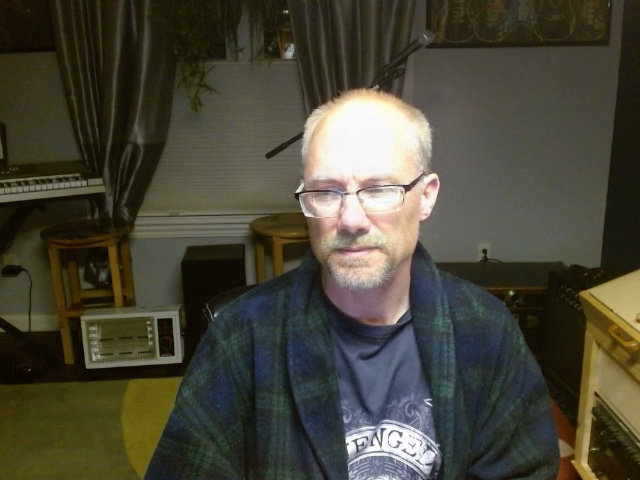}
    \end{tabular}
    }
    \caption{Video frames sampled from three representative examples of \textbf{face-swap} deepfakes generated using \textbf{INSwapper}.}
    \label{fig:frame_examples_inswapper}
\end{figure}

\newpage


\begin{figure}[H]
    \centering
    \resizebox{0.92\textwidth}{!}{
    \begin{tabular}{c@{\hspace{0cm}}c@{\hspace{0cm}}c@{\hspace{0cm}}c}
        \includegraphics[width=0.24\linewidth]{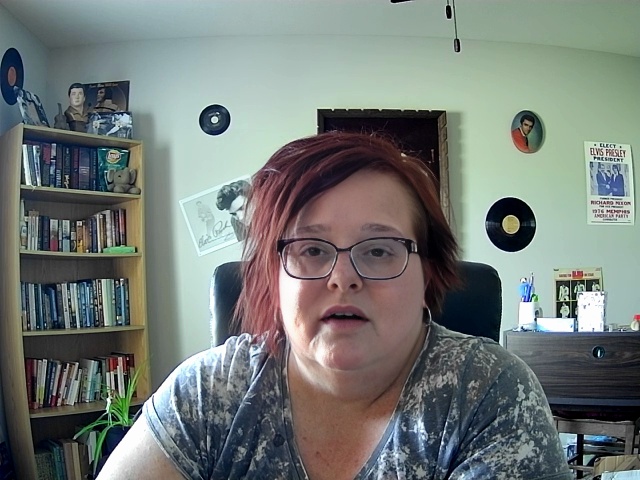} & 
        \includegraphics[width=0.24\linewidth]{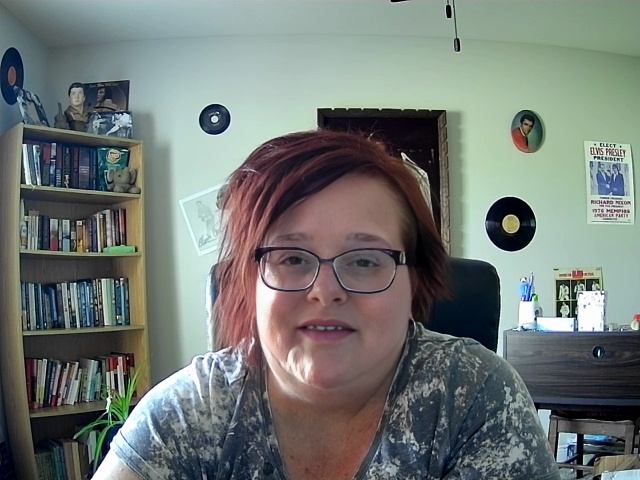} & 
        \includegraphics[width=0.24\linewidth]{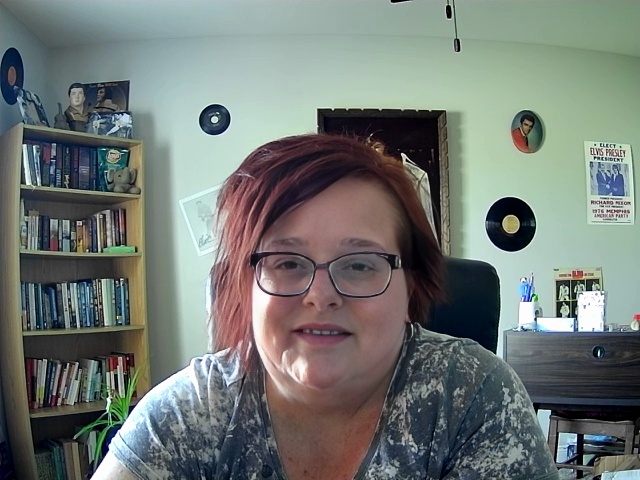} & 
        \includegraphics[width=0.24\linewidth]{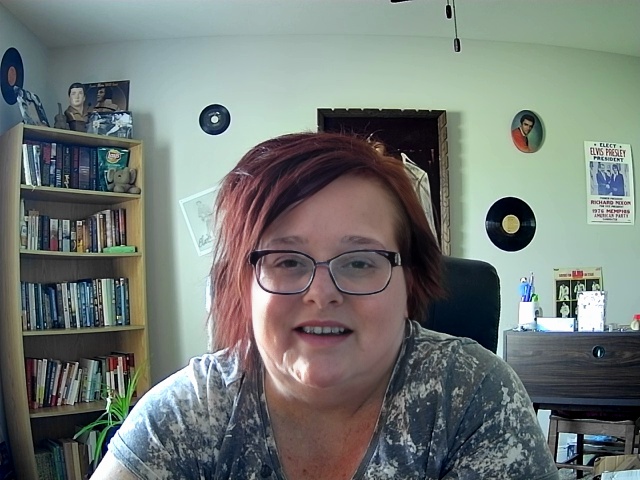} \\
        \includegraphics[width=0.24\linewidth]{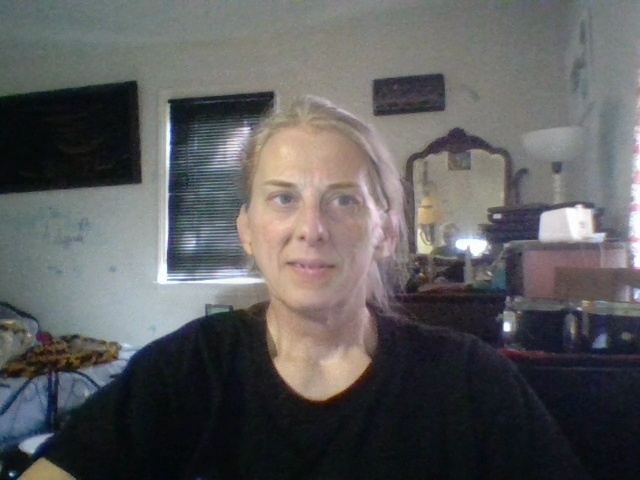} & 
        \includegraphics[width=0.24\linewidth]{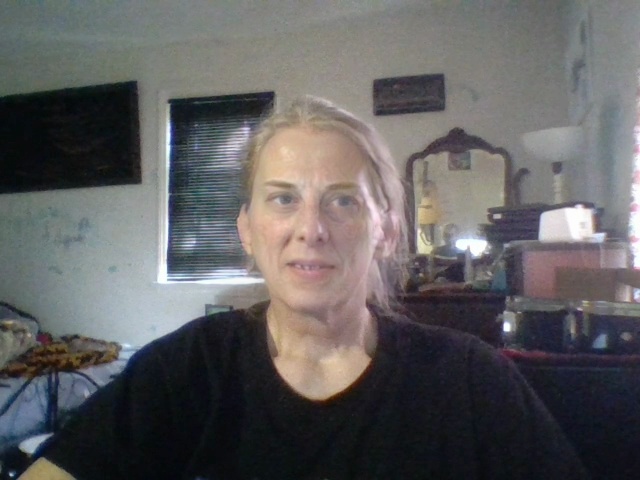} & 
        \includegraphics[width=0.24\linewidth]{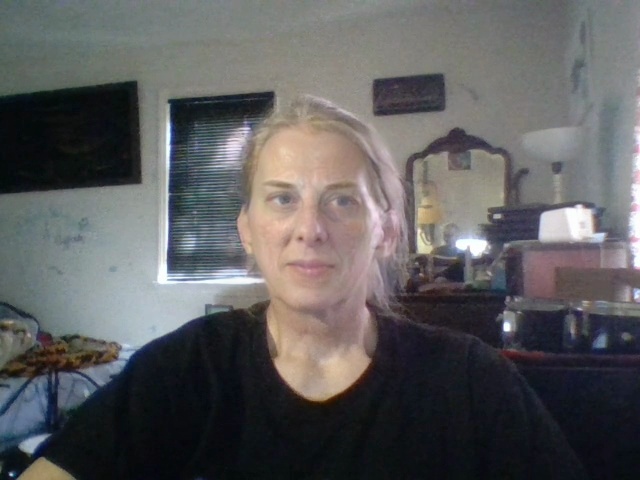} & 
        \includegraphics[width=0.24\linewidth]{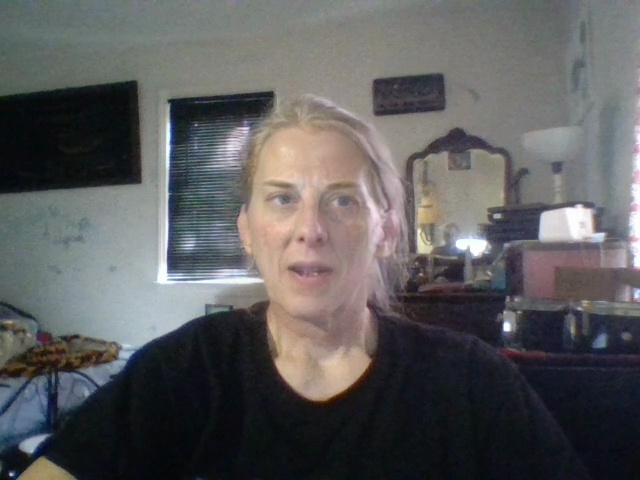} \\
        \includegraphics[width=0.24\linewidth]{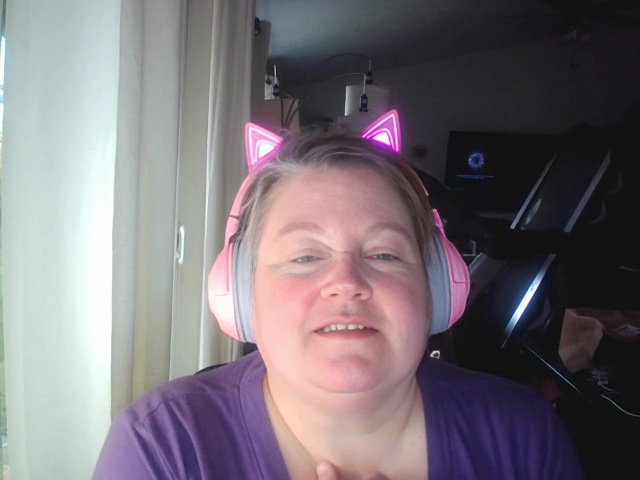} & 
        \includegraphics[width=0.24\linewidth]{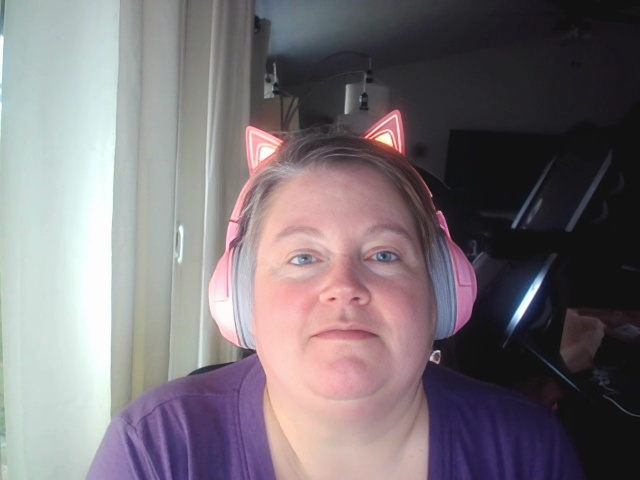} & 
        \includegraphics[width=0.24\linewidth]{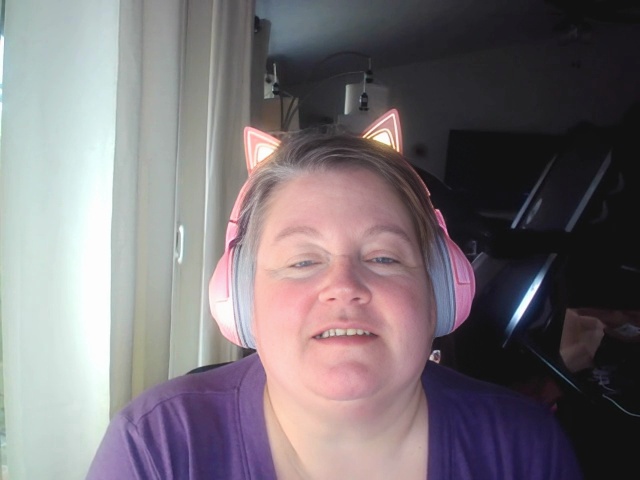} & 
        \includegraphics[width=0.24\linewidth]{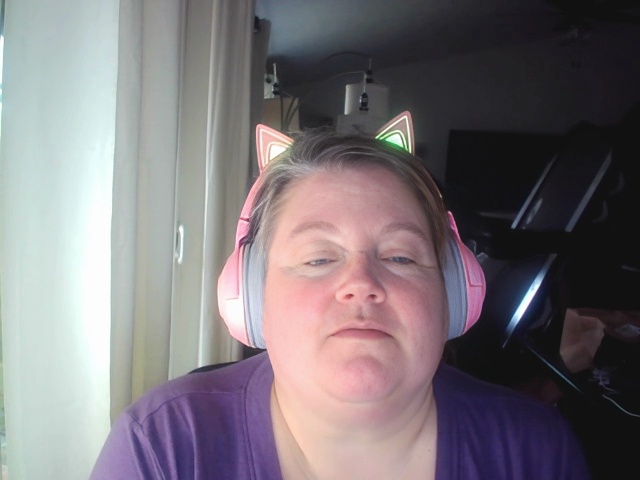}
    \end{tabular}
    }
    \caption{Video frames sampled from three representative examples of \textbf{face-swap} deepfakes generated using \textbf{INSwapper + CodeFormer}.}
    \label{fig:frame_examples_inswapper_codeformer}
\end{figure}


\begin{figure}[H]
    \centering
    \resizebox{0.92\textwidth}{!}{
    \begin{tabular}{c@{\hspace{0cm}}c@{\hspace{0cm}}c@{\hspace{0cm}}c}
        \includegraphics[width=0.24\linewidth]{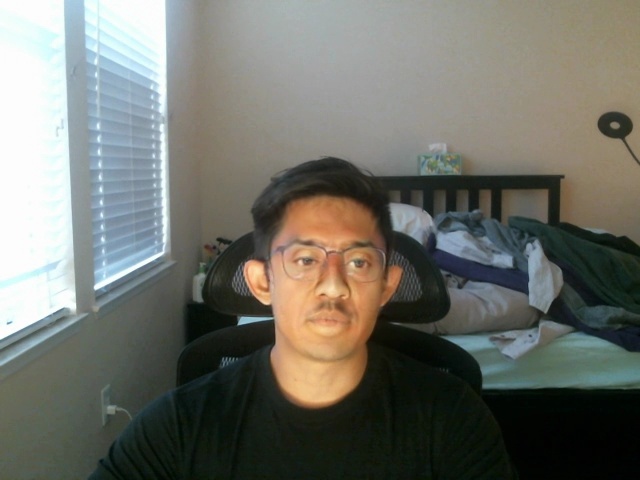} & 
        \includegraphics[width=0.24\linewidth]{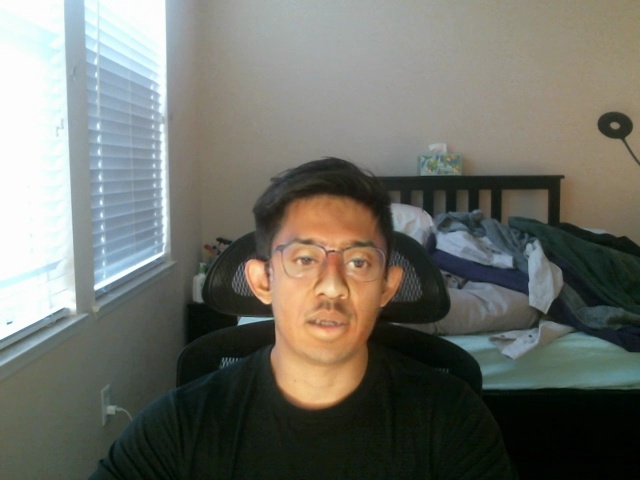} & 
        \includegraphics[width=0.24\linewidth]{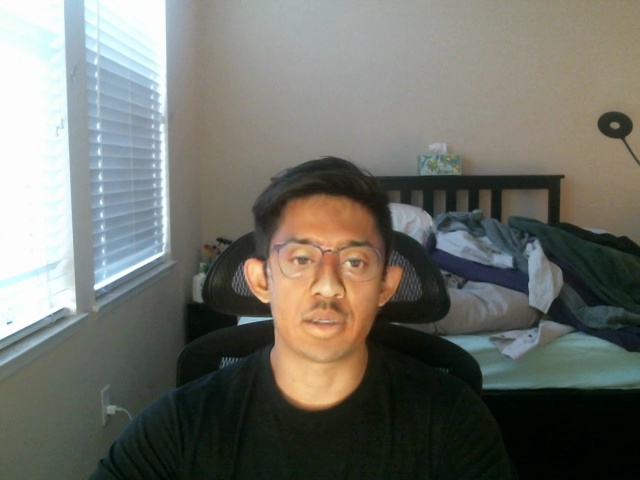} & 
        \includegraphics[width=0.24\linewidth]{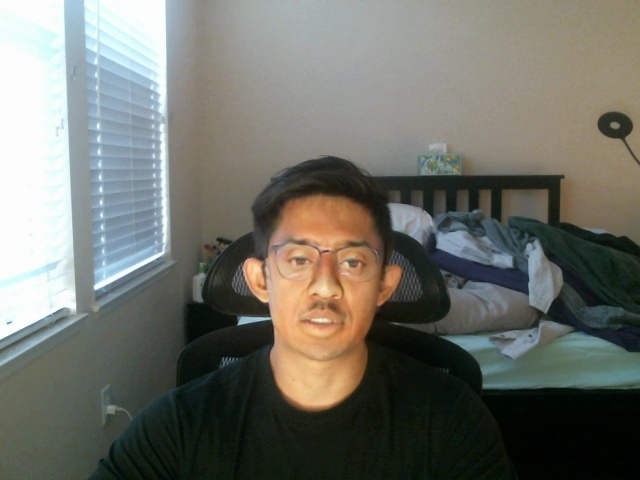} \\
        \includegraphics[width=0.24\linewidth]{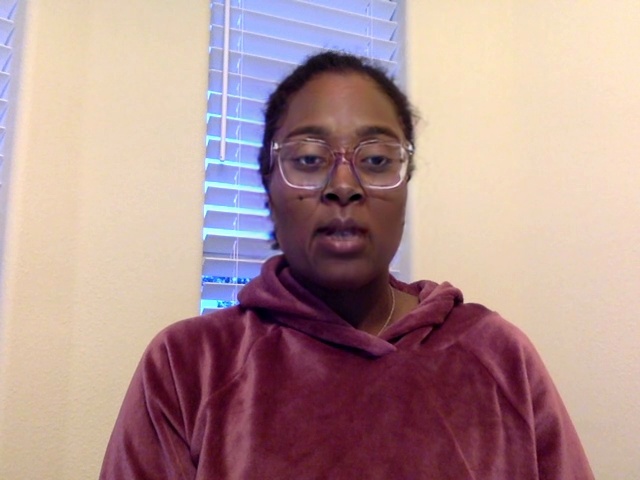} & 
        \includegraphics[width=0.24\linewidth]{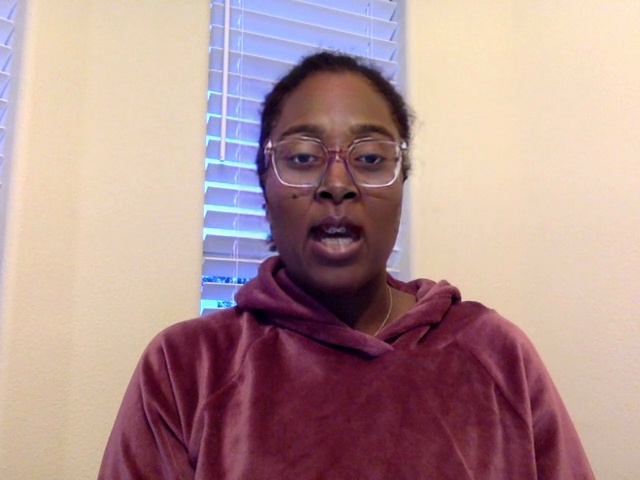} & 
        \includegraphics[width=0.24\linewidth]{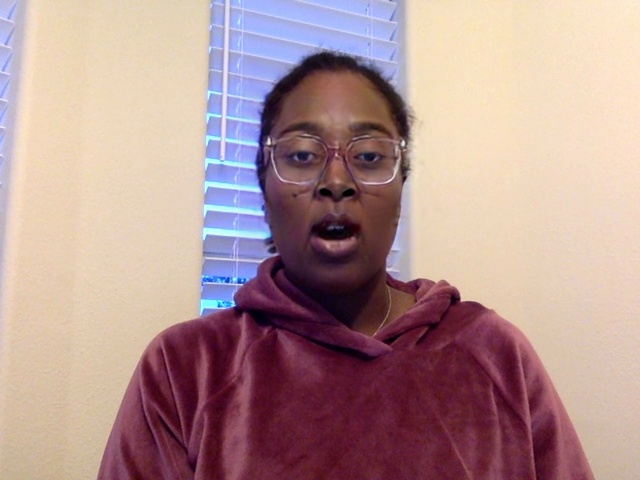} & 
        \includegraphics[width=0.24\linewidth]{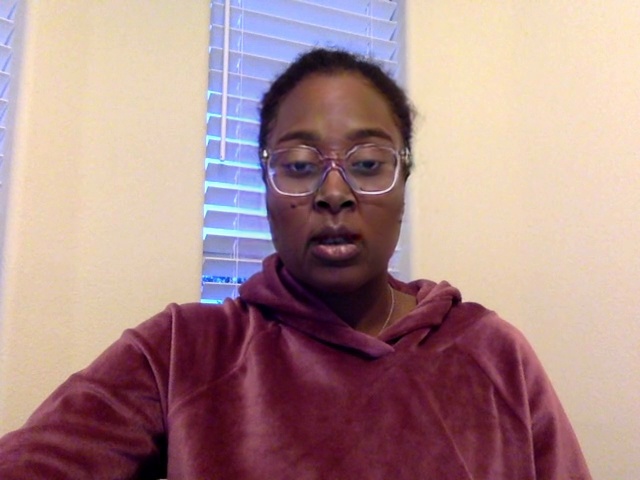} \\
        \includegraphics[width=0.24\linewidth]{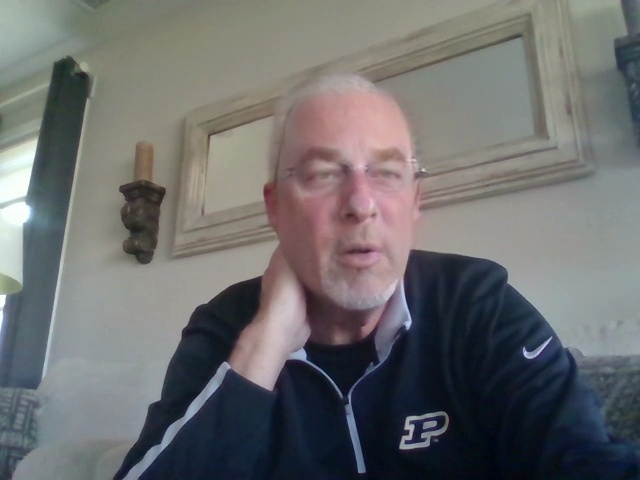} & 
        \includegraphics[width=0.24\linewidth]{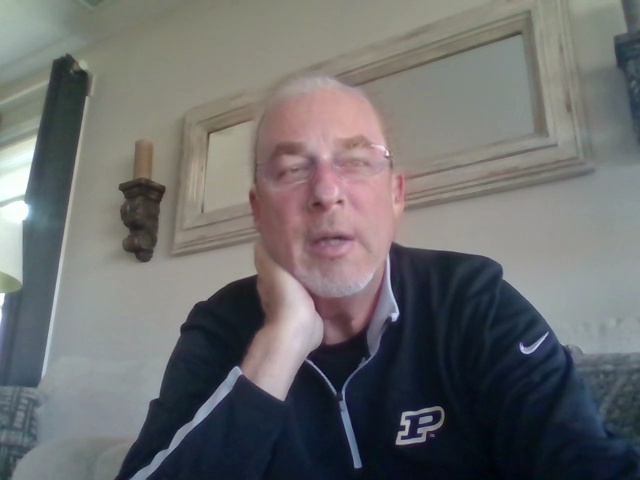} & 
        \includegraphics[width=0.24\linewidth]{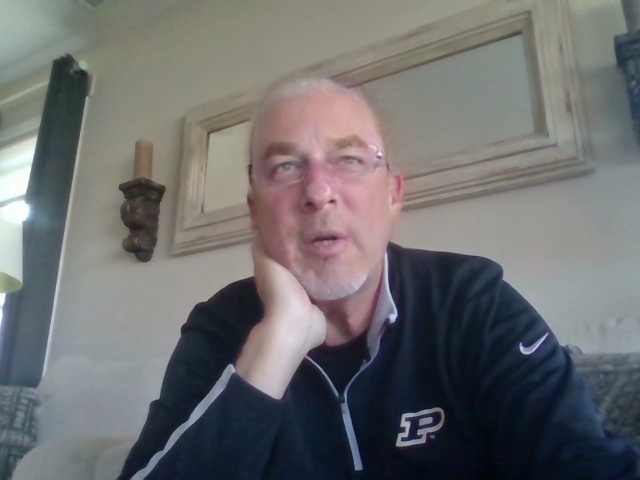} & 
        \includegraphics[width=0.24\linewidth]{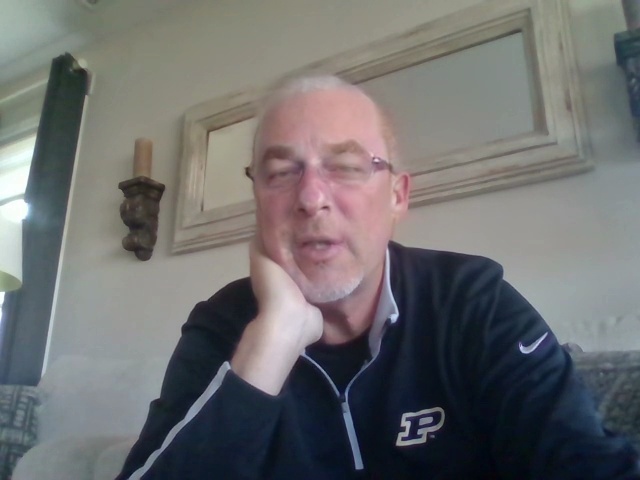}
    \end{tabular}
    }
    \caption{Video frames sampled from three representative examples of \textbf{face-swap} deepfakes generated using \textbf{SimSwap}.}
    \label{fig:frame_examples_simswap}
\end{figure}



\begin{figure}[H]
    \centering
    \resizebox{0.92\textwidth}{!}{
    \begin{tabular}{c@{\hspace{0cm}}c@{\hspace{0cm}}c@{\hspace{0cm}}c}
        \includegraphics[width=0.24\linewidth]{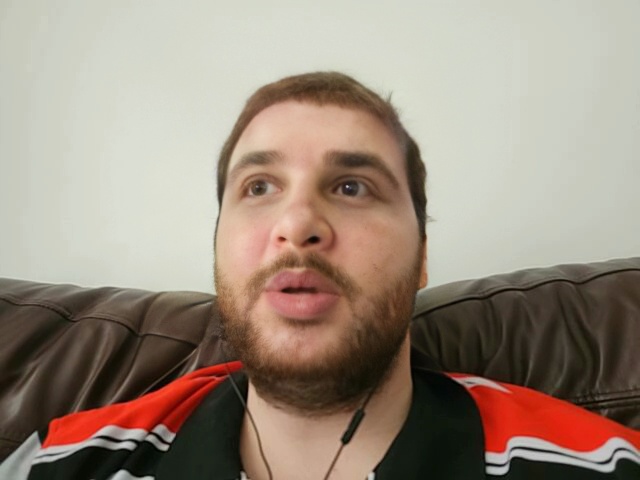} & 
        \includegraphics[width=0.24\linewidth]{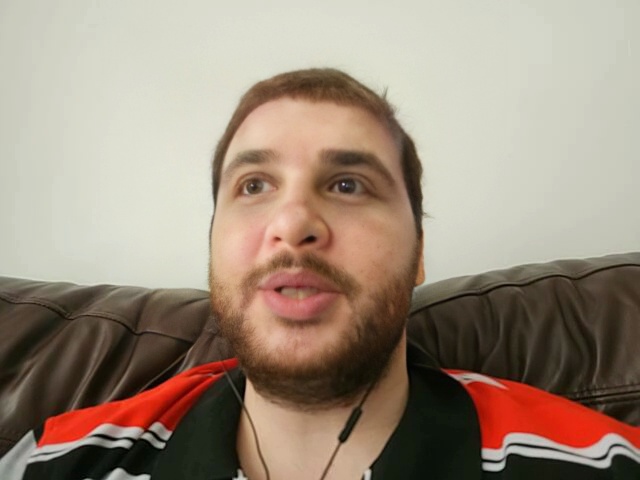} & 
        \includegraphics[width=0.24\linewidth]{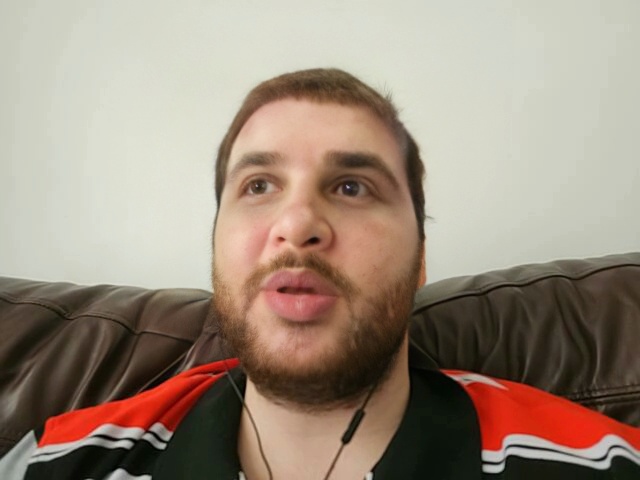} & 
        \includegraphics[width=0.24\linewidth]{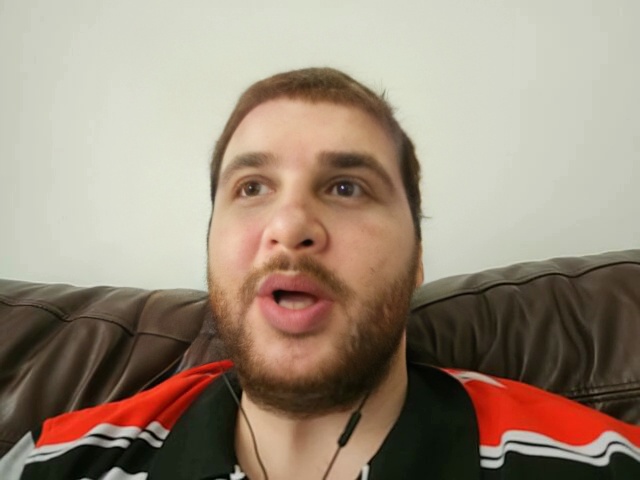} \\
        \includegraphics[width=0.24\linewidth]{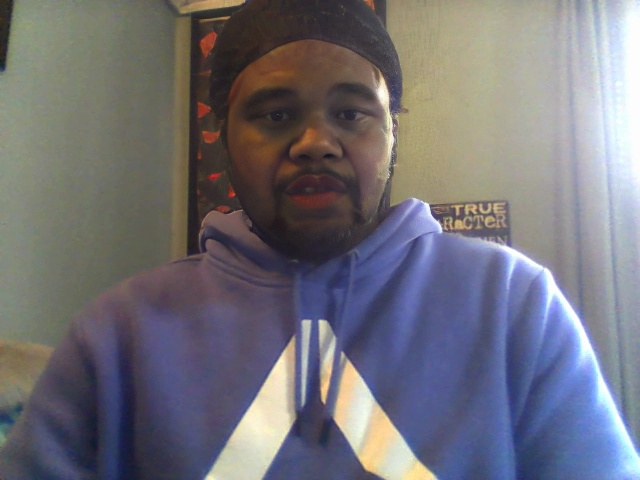} & 
        \includegraphics[width=0.24\linewidth]{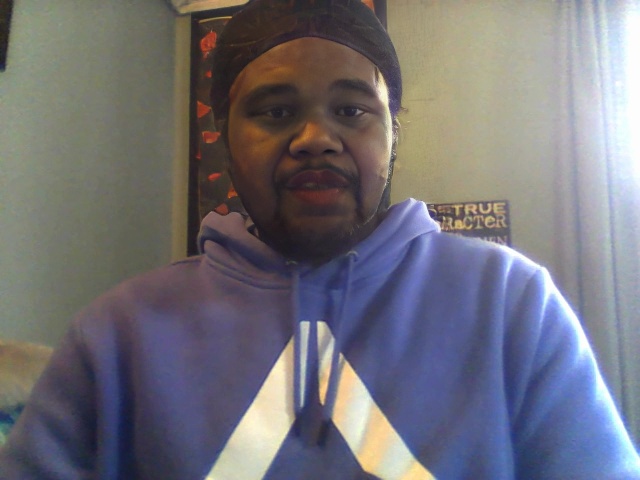} & 
        \includegraphics[width=0.24\linewidth]{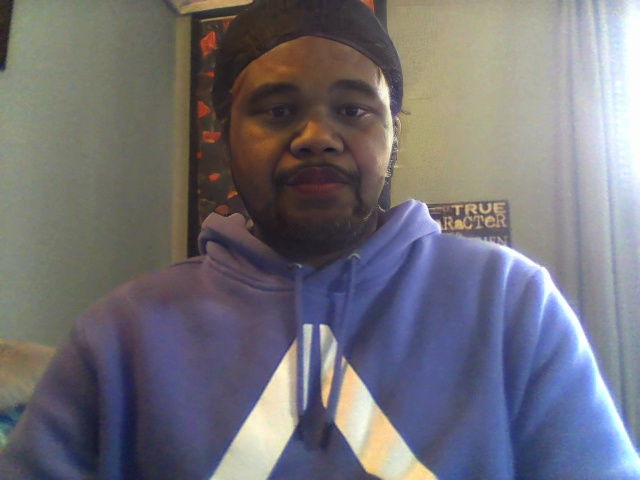} & 
        \includegraphics[width=0.24\linewidth]{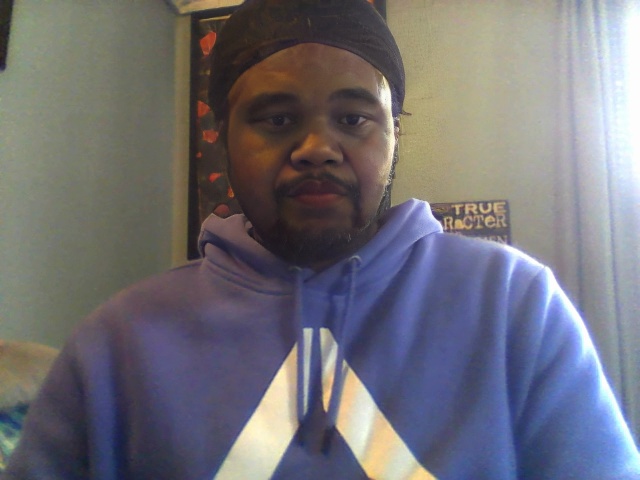} \\
        \includegraphics[width=0.24\linewidth]{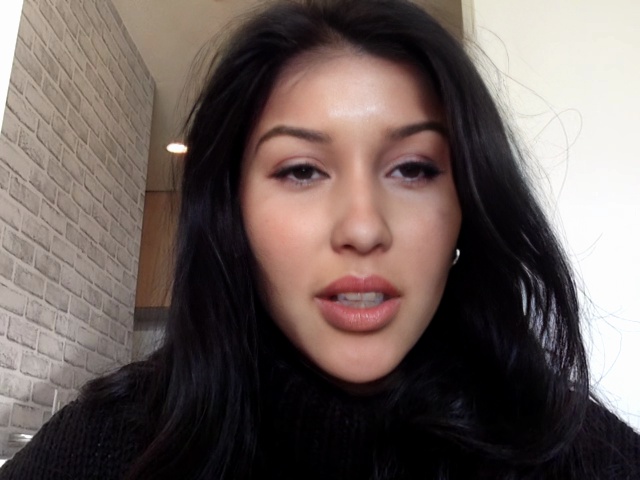} & 
        \includegraphics[width=0.24\linewidth]{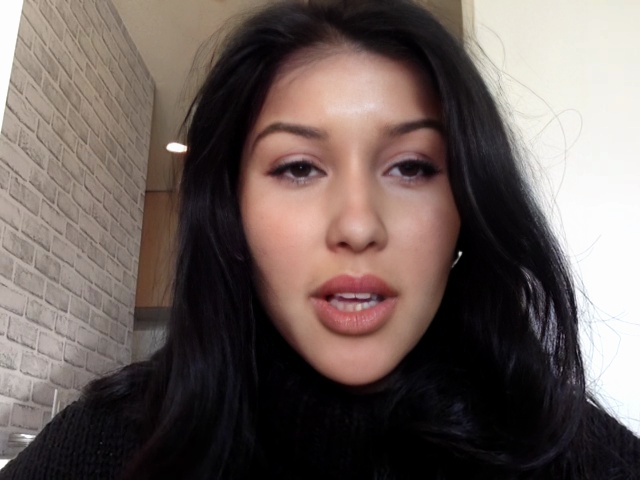} & 
        \includegraphics[width=0.24\linewidth]{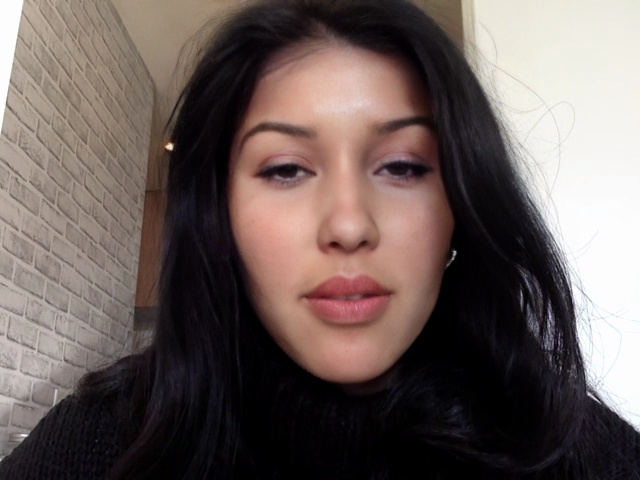} & 
        \includegraphics[width=0.24\linewidth]{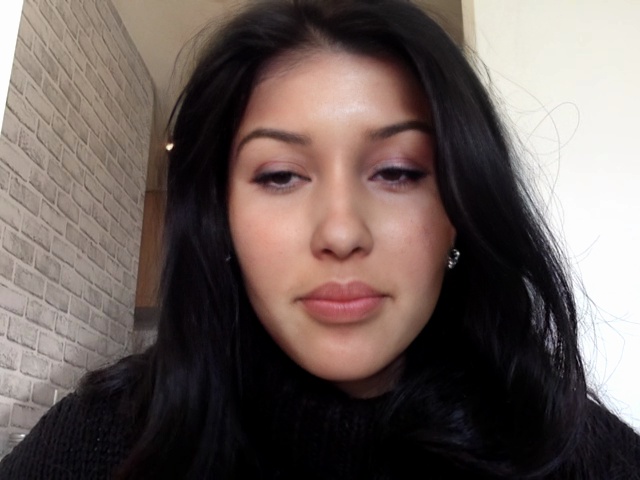}
    \end{tabular}
    }
    \caption{Video frames sampled from three representative examples of \textbf{face-swap} deepfakes generated using \textbf{SimSwap + RestoreFormer}.}
    \label{fig:frame_examples_simswap_restoreformer}
\end{figure}

%
%
%
%


\begin{figure}[H]
    \centering
    \resizebox{0.92\textwidth}{!}{
    \begin{tabular}{c@{\hspace{0cm}}c@{\hspace{0cm}}c@{\hspace{0cm}}c}
        \includegraphics[width=0.24\linewidth]{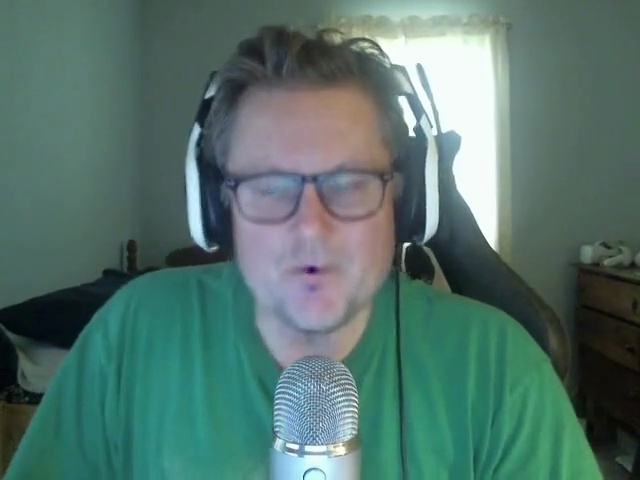} & 
        \includegraphics[width=0.24\linewidth]{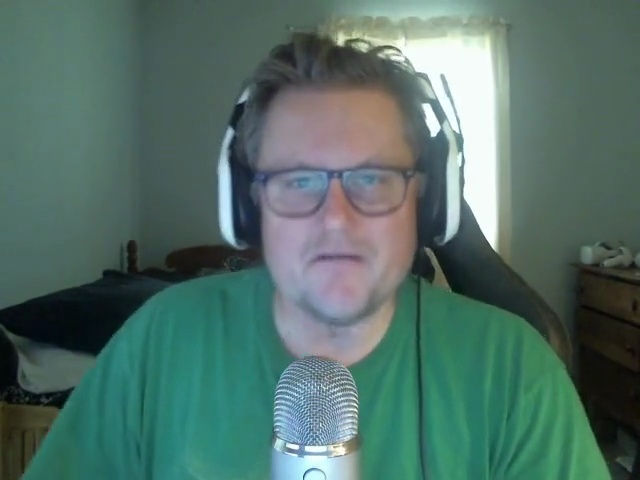} & 
        \includegraphics[width=0.24\linewidth]{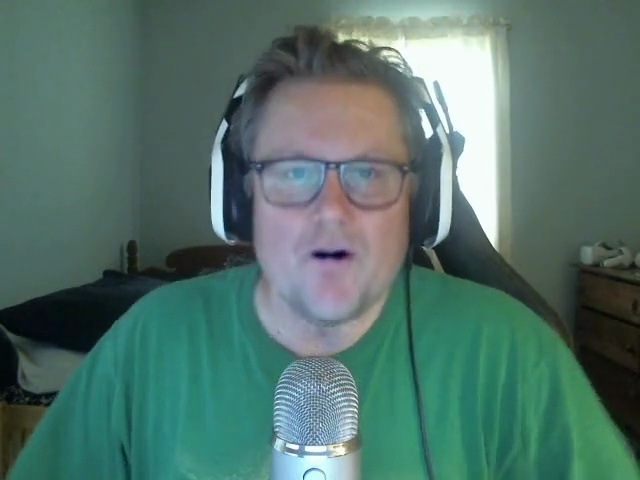} & 
        \includegraphics[width=0.24\linewidth]{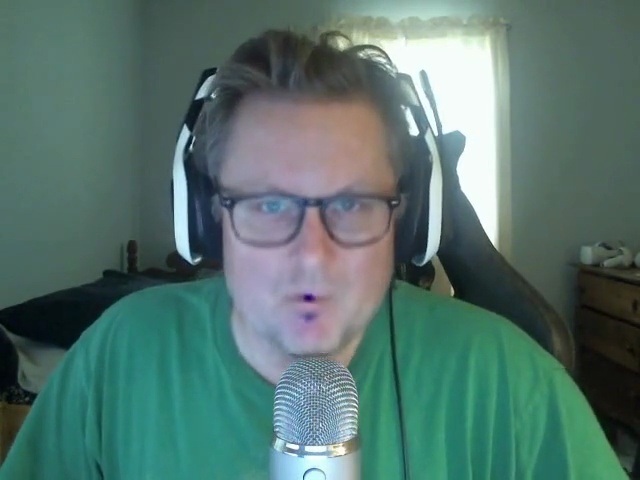} \\
        \includegraphics[width=0.24\linewidth]{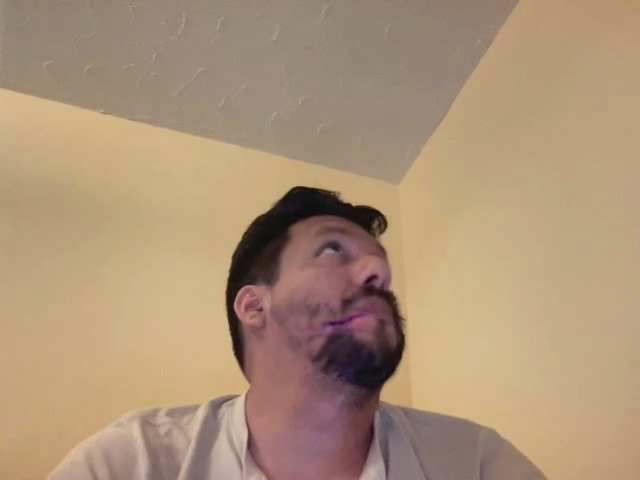} & 
        \includegraphics[width=0.24\linewidth]{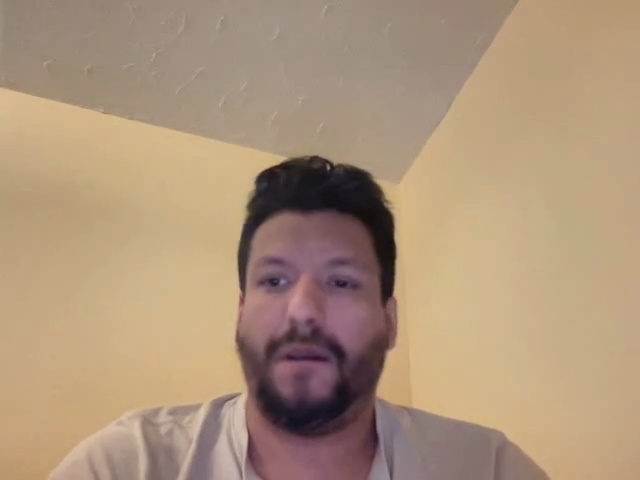} & 
        \includegraphics[width=0.24\linewidth]{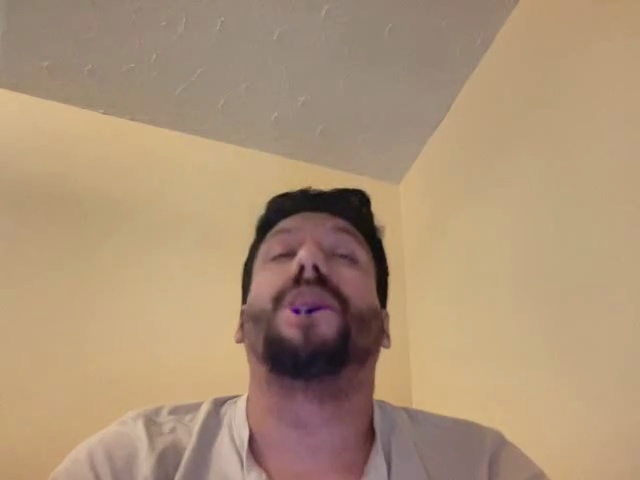} & 
        \includegraphics[width=0.24\linewidth]{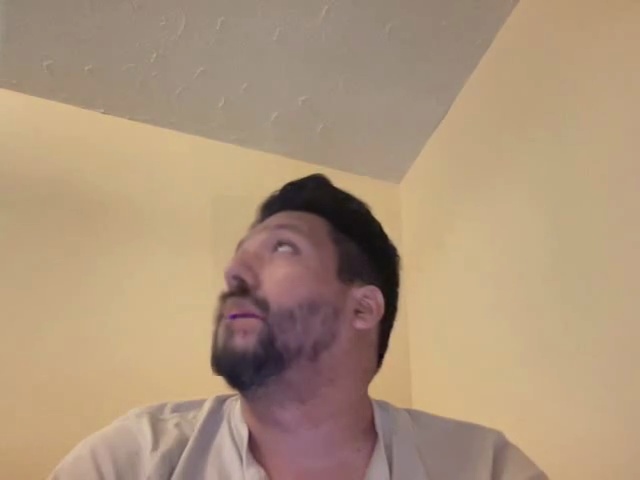} \\
        \includegraphics[width=0.24\linewidth]{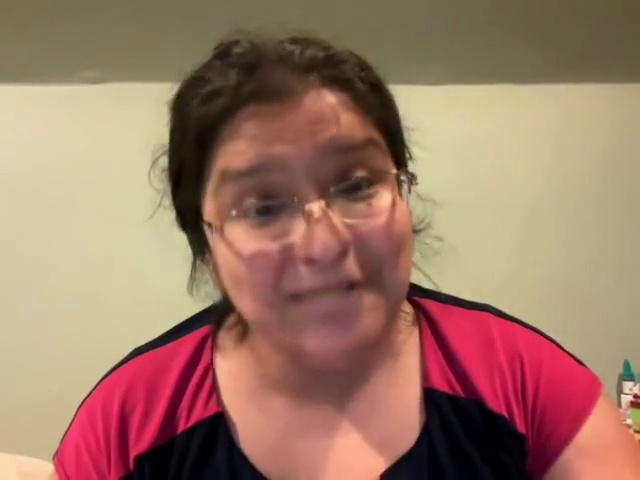} & 
        \includegraphics[width=0.24\linewidth]{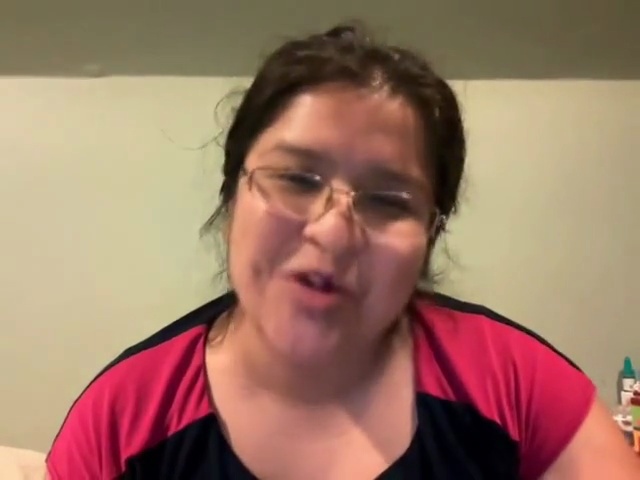} & 
        \includegraphics[width=0.24\linewidth]{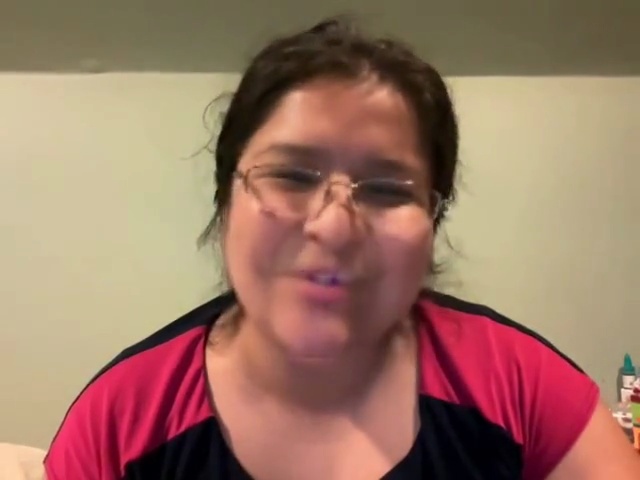} & 
        \includegraphics[width=0.24\linewidth]{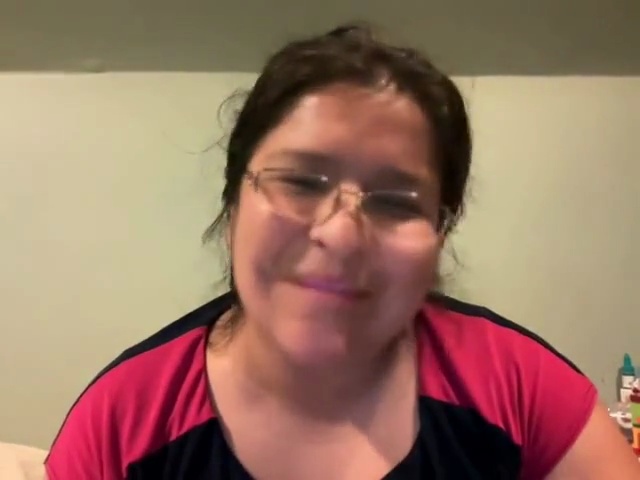}
    \end{tabular}
    }
    \caption{Video frames sampled from three representative examples of \textbf{lip-sync} deepfakes generated using \textbf{Wav2Lip}.}
    \label{fig:frame_examples_wav2lip}
\end{figure}

\newpage


\begin{figure}[H]
    \centering
    \resizebox{0.92\textwidth}{!}{
    \begin{tabular}{c@{\hspace{0cm}}c@{\hspace{0cm}}c@{\hspace{0cm}}c}
        \includegraphics[width=0.24\linewidth]{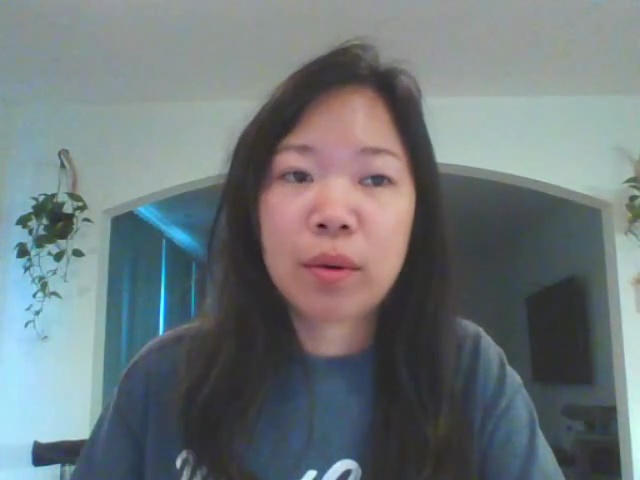} & 
        \includegraphics[width=0.24\linewidth]{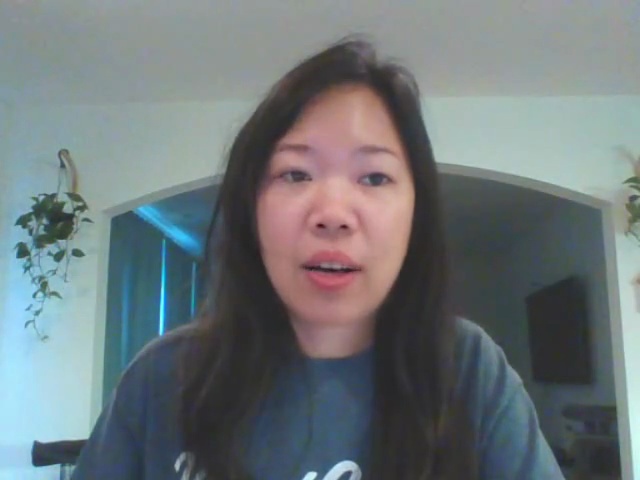} & 
        \includegraphics[width=0.24\linewidth]{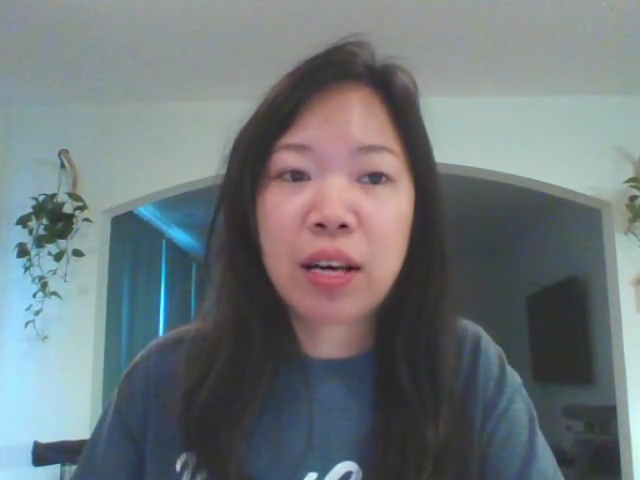} & 
        \includegraphics[width=0.24\linewidth]{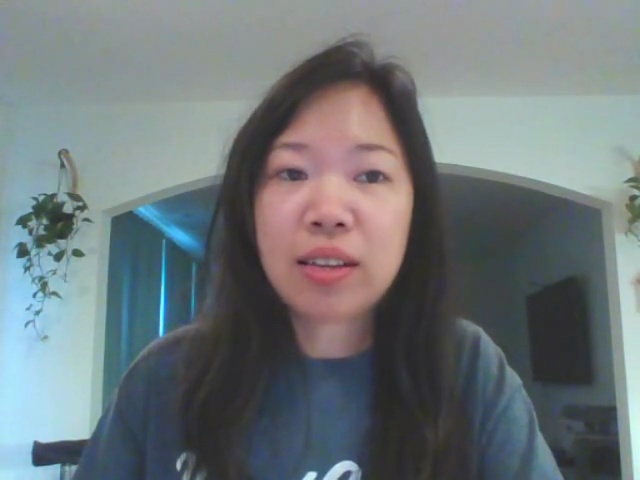} \\
        \includegraphics[width=0.24\linewidth]{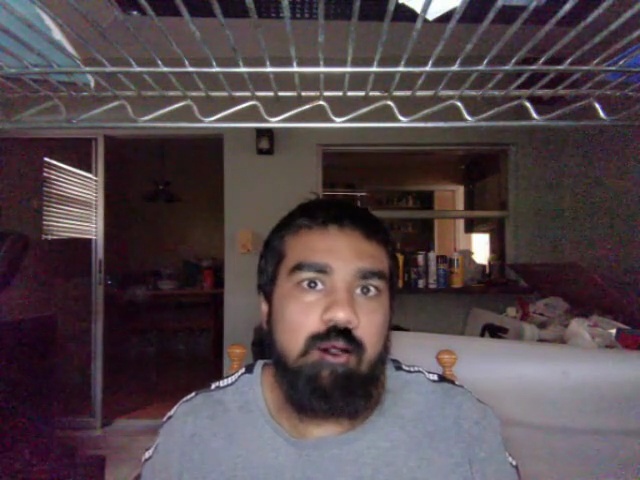} & 
        \includegraphics[width=0.24\linewidth]{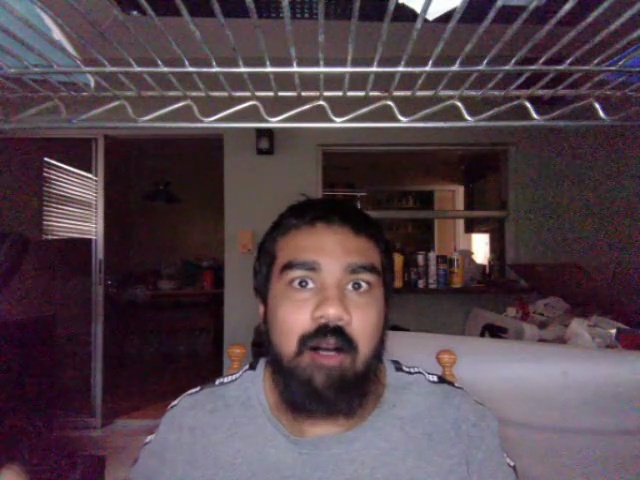} & 
        \includegraphics[width=0.24\linewidth]{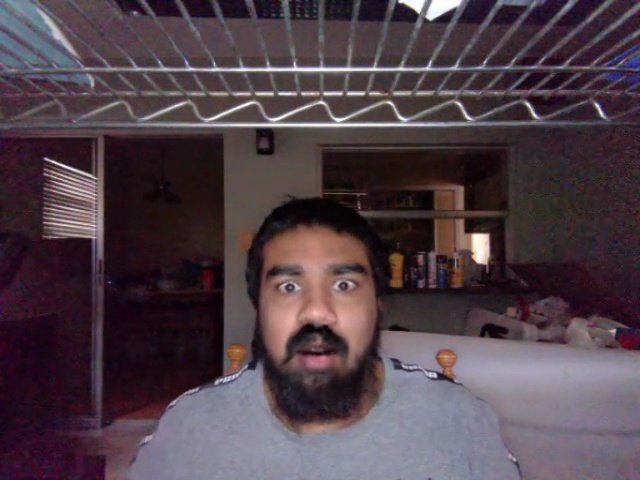} & 
        \includegraphics[width=0.24\linewidth]{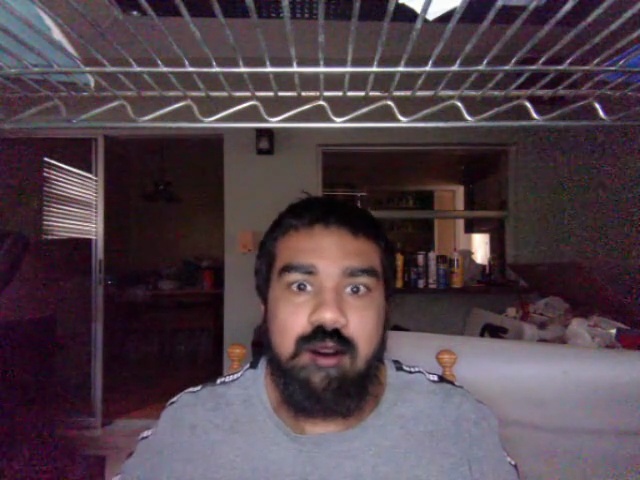} \\
        \includegraphics[width=0.24\linewidth]{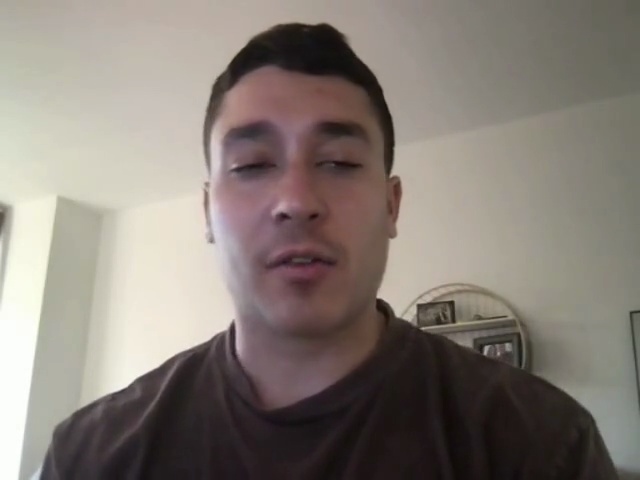} & 
        \includegraphics[width=0.24\linewidth]{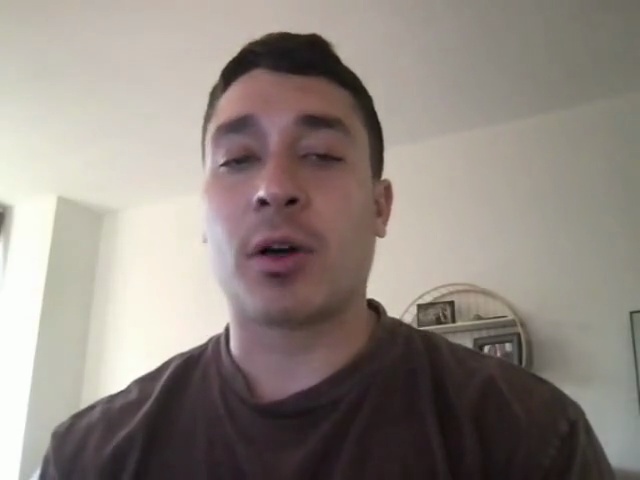} & 
        \includegraphics[width=0.24\linewidth]{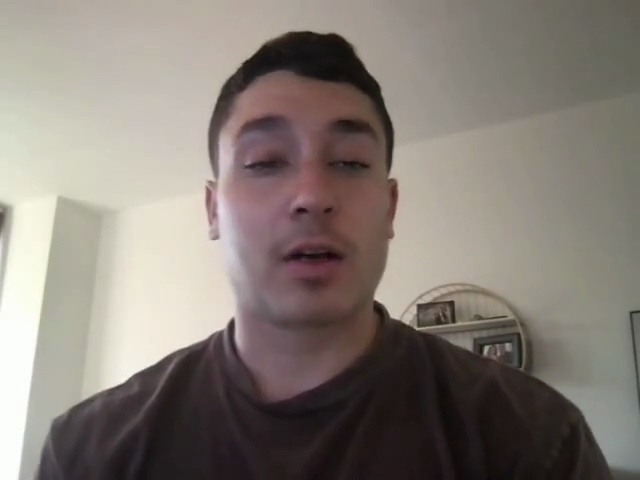} & 
        \includegraphics[width=0.24\linewidth]{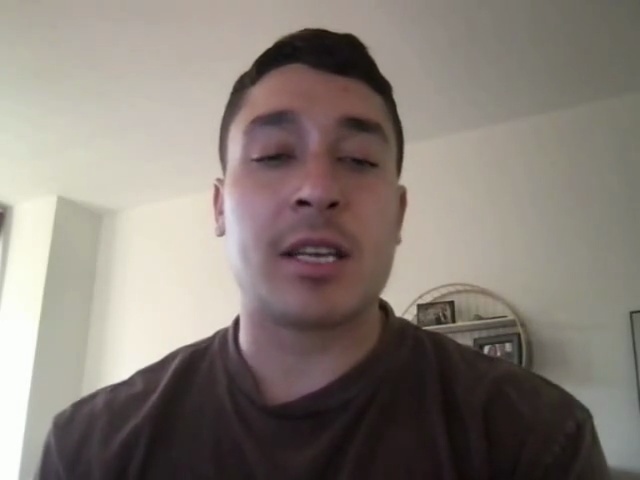}
    \end{tabular}
    }
    \caption{Video frames sampled from three representative examples of \textbf{lip-sync} deepfakes generated using \textbf{VideoRetalking}.}
    \label{fig:frame_examples_videoretalking}
\end{figure}


\begin{figure}[H]
    \centering
    \resizebox{0.92\textwidth}{!}{
    \begin{tabular}{c@{\hspace{0cm}}c@{\hspace{0cm}}c@{\hspace{0cm}}c}
        \includegraphics[width=0.24\linewidth]{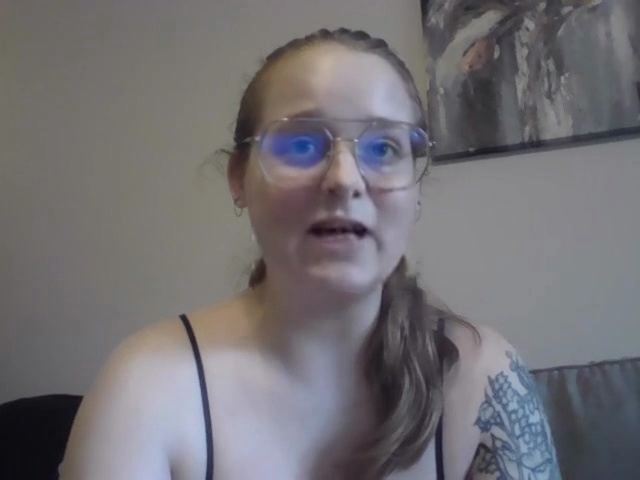} & 
        \includegraphics[width=0.24\linewidth]{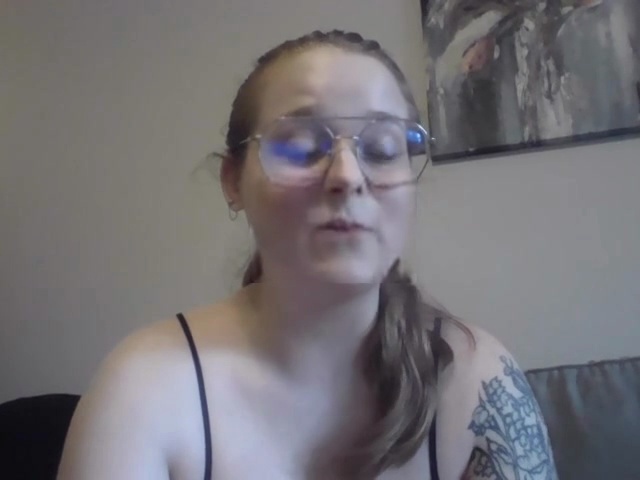} & 
        \includegraphics[width=0.24\linewidth]{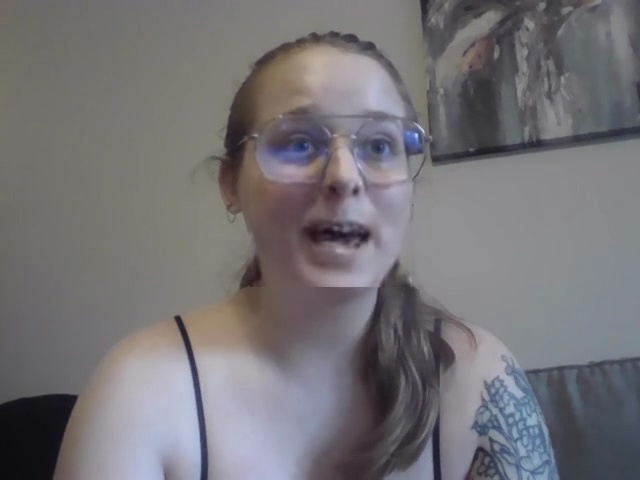} & 
        \includegraphics[width=0.24\linewidth]{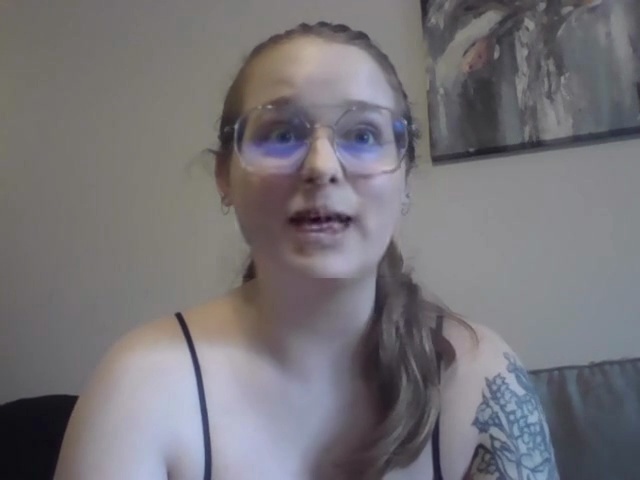} \\
        \includegraphics[width=0.24\linewidth]{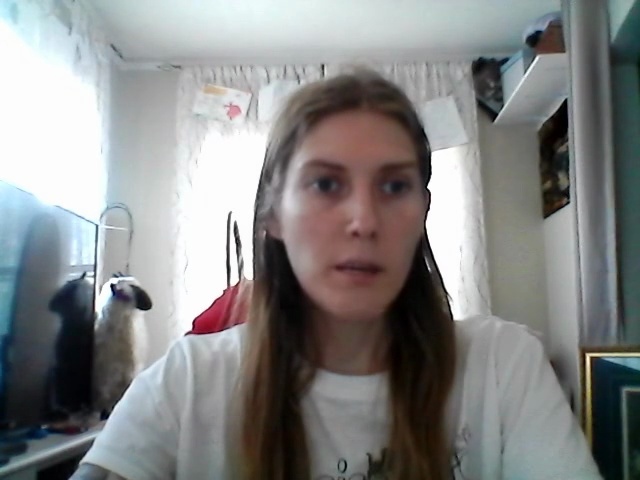} & 
        \includegraphics[width=0.24\linewidth]{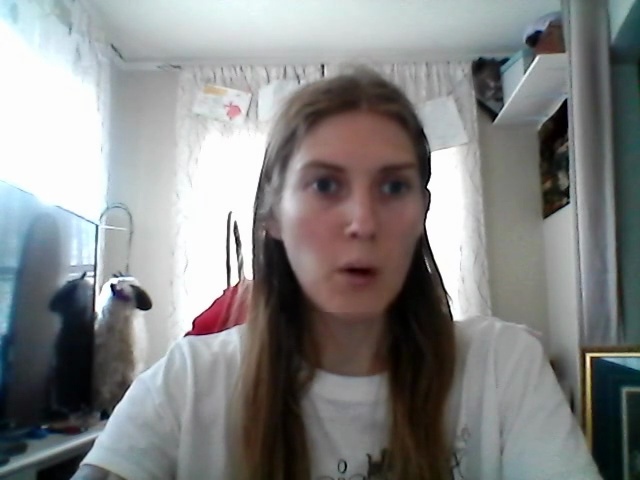} & 
        \includegraphics[width=0.24\linewidth]{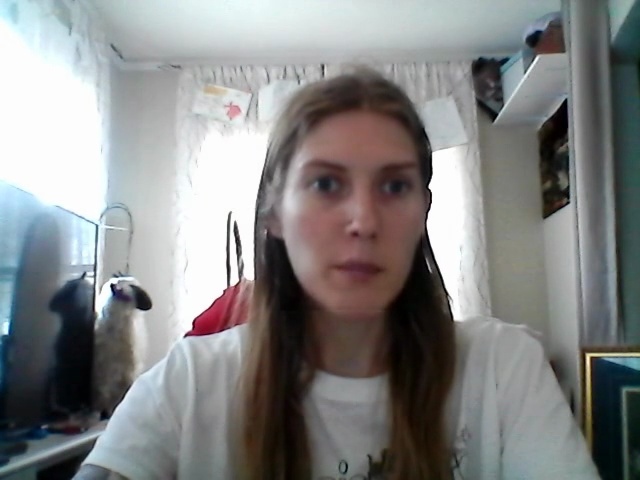} & 
        \includegraphics[width=0.24\linewidth]{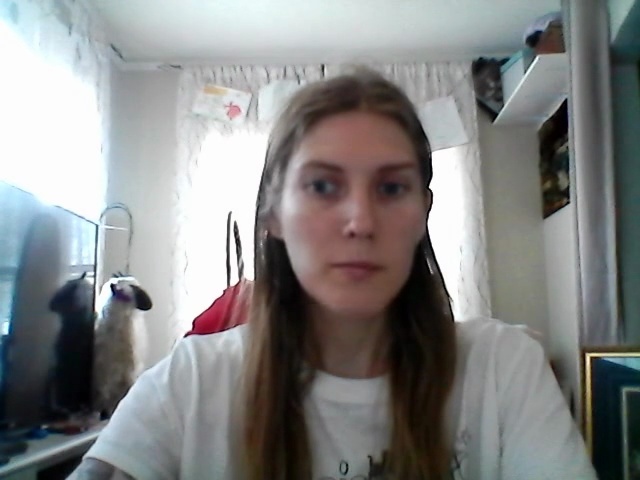} \\
        \includegraphics[width=0.24\linewidth]{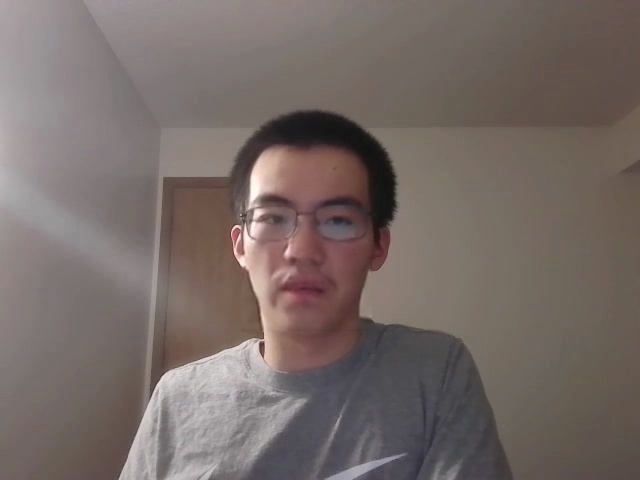} & 
        \includegraphics[width=0.24\linewidth]{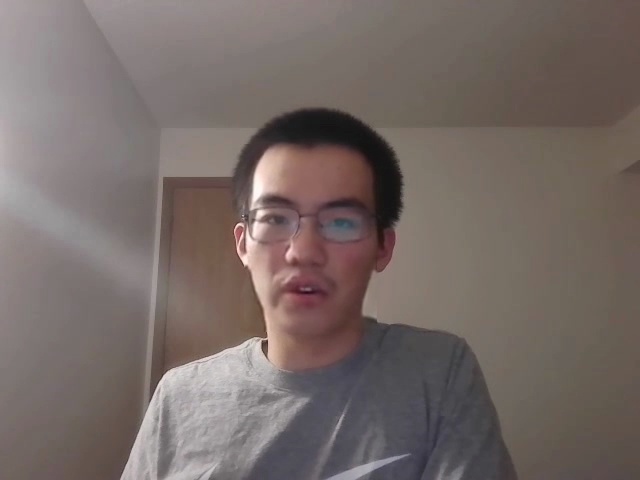} & 
        \includegraphics[width=0.24\linewidth]{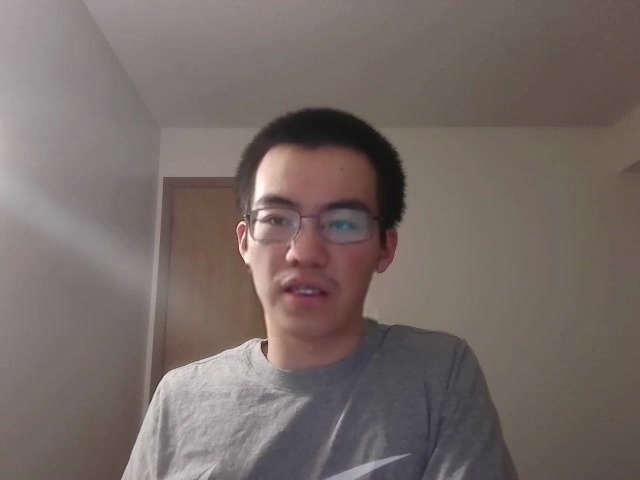} & 
        \includegraphics[width=0.24\linewidth]{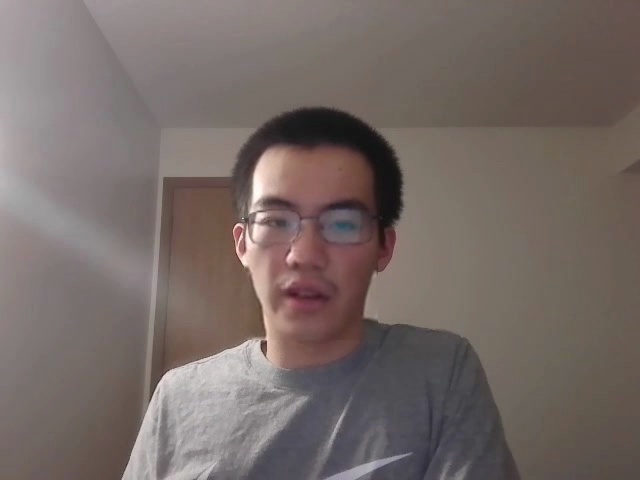}
    \end{tabular}
    }
    \caption{Video frames sampled from three representative examples of \textbf{lip-sync} deepfakes generated using \textbf{Diff2Lip}.}
    \label{fig:frame_examples_diff2lip}
\end{figure}

\newpage


\begin{figure}[H]
    \centering
    \resizebox{0.92\textwidth}{!}{
    \begin{tabular}{c@{\hspace{0cm}}c@{\hspace{0cm}}c@{\hspace{0cm}}c}
        \includegraphics[width=0.24\linewidth]{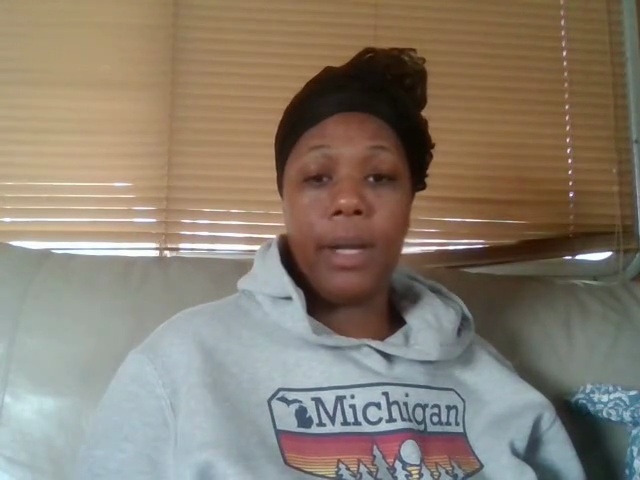} & 
        \includegraphics[width=0.24\linewidth]{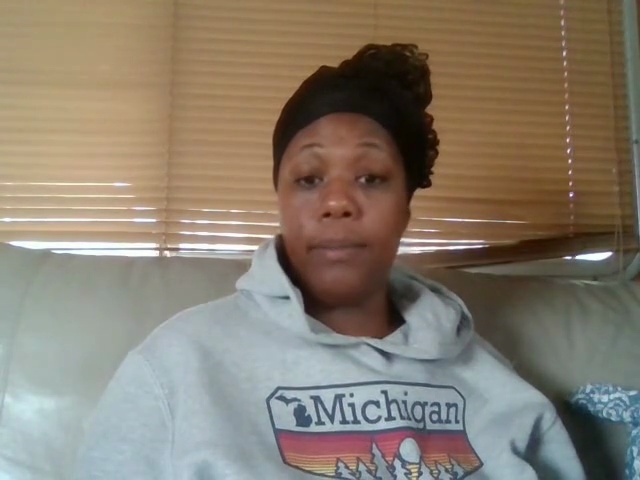} & 
        \includegraphics[width=0.24\linewidth]{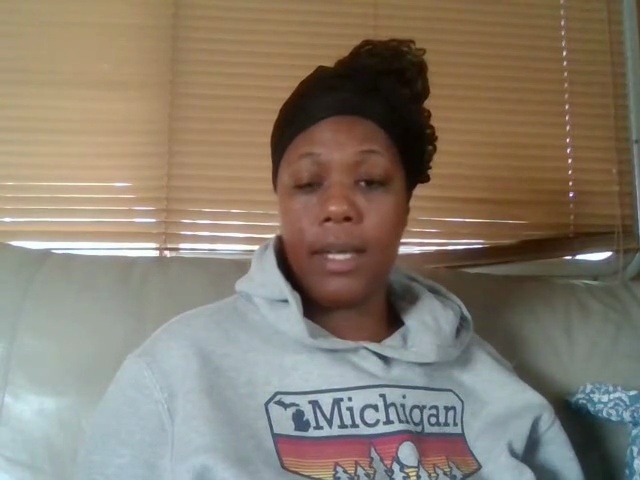} & 
        \includegraphics[width=0.24\linewidth]{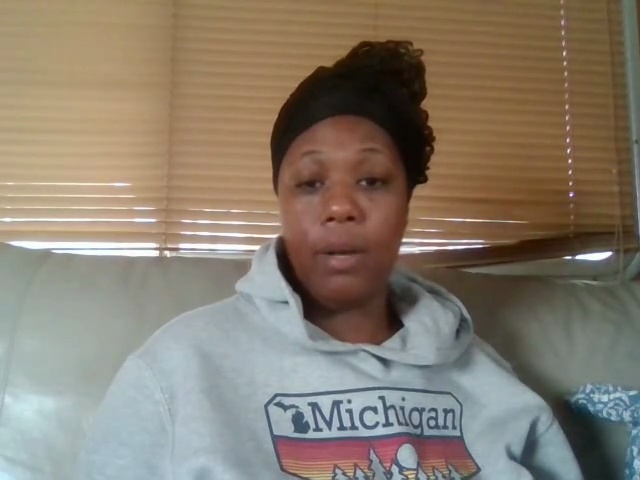} \\
        \includegraphics[width=0.24\linewidth]{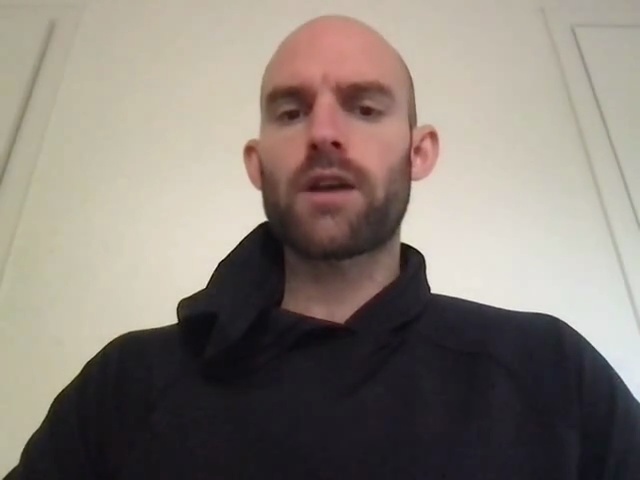} & 
        \includegraphics[width=0.24\linewidth]{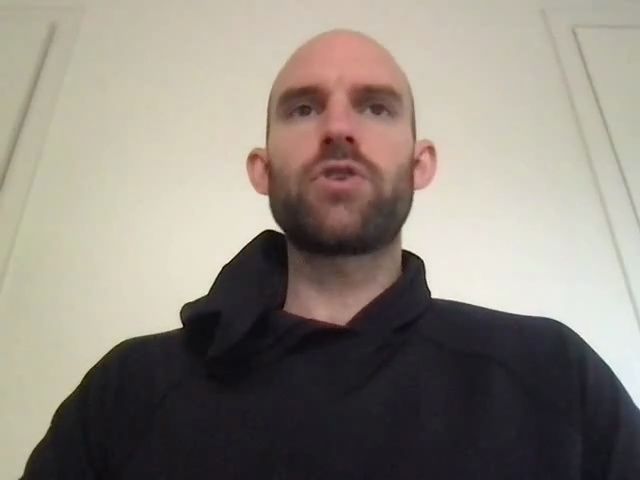} & 
        \includegraphics[width=0.24\linewidth]{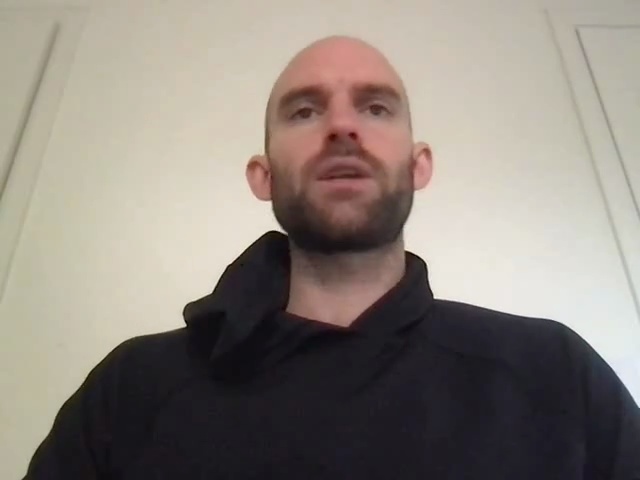} & 
        \includegraphics[width=0.24\linewidth]{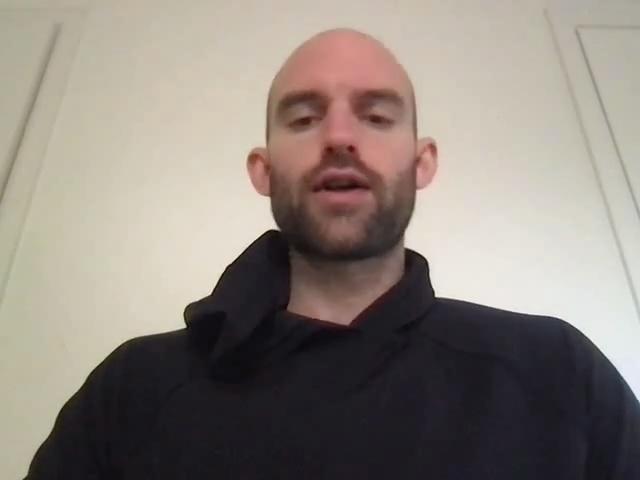} \\
        \includegraphics[width=0.24\linewidth]{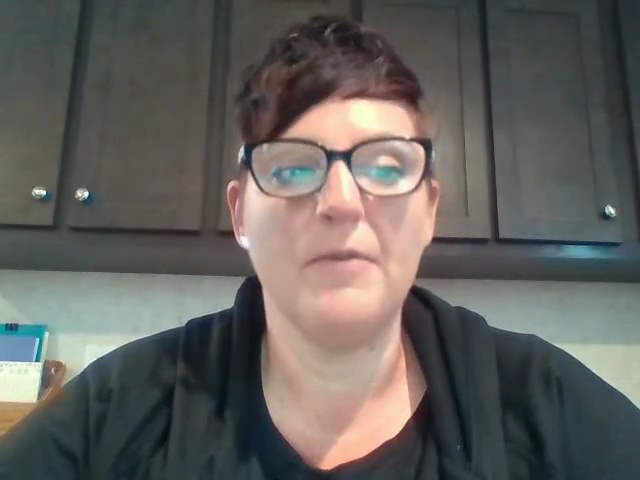} & 
        \includegraphics[width=0.24\linewidth]{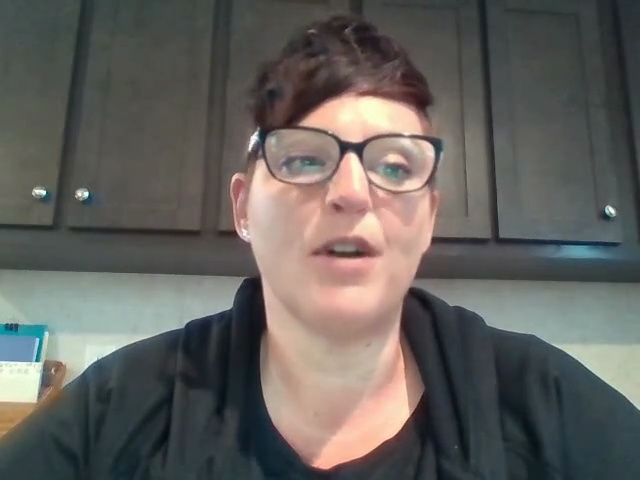} & 
        \includegraphics[width=0.24\linewidth]{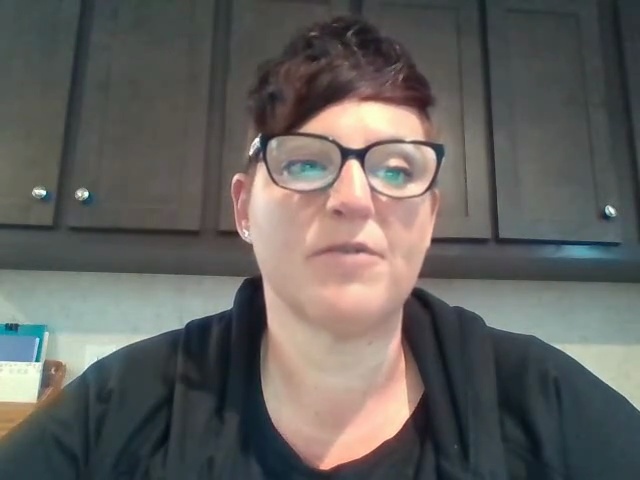} & 
        \includegraphics[width=0.24\linewidth]{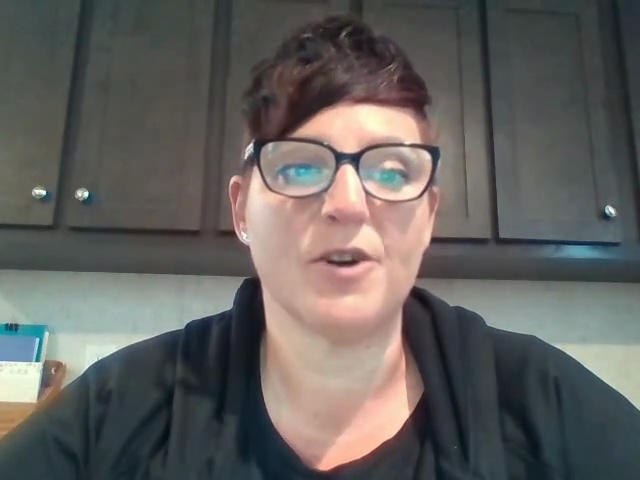}
    \end{tabular}
    }
    \caption{Video frames sampled from three representative examples of \textbf{lip-sync} deepfakes generated using \textbf{LatentSync}.}
    \label{fig:frame_examples_latentsync}
\end{figure}

%
%
%
%


\begin{figure}[H]
    \centering
    \resizebox{0.92\textwidth}{!}{
    \begin{tabular}{c@{\hspace{0cm}}c@{\hspace{0cm}}c@{\hspace{0cm}}c}
        \includegraphics[width=0.24\linewidth]{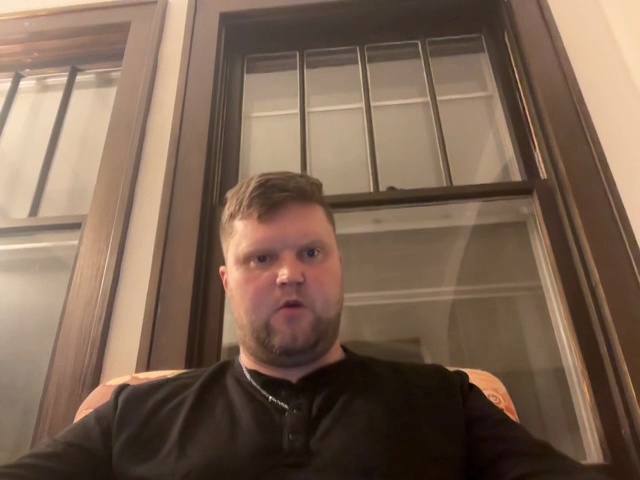} & 
        \includegraphics[width=0.24\linewidth]{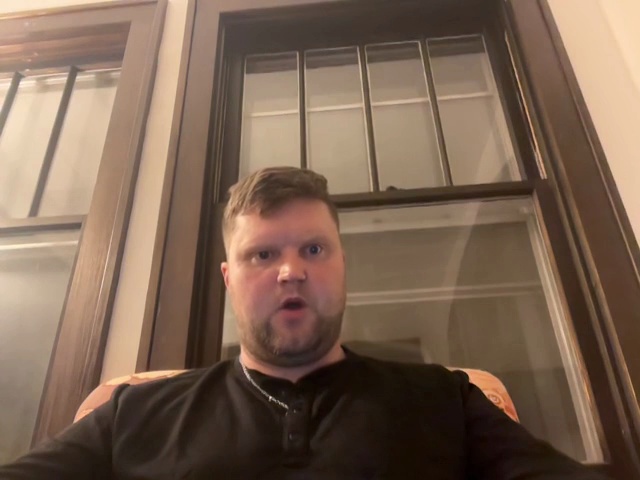} & 
        \includegraphics[width=0.24\linewidth]{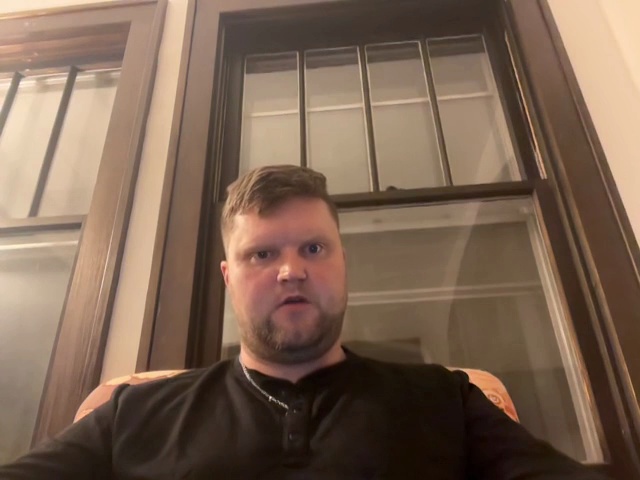} & 
        \includegraphics[width=0.24\linewidth]{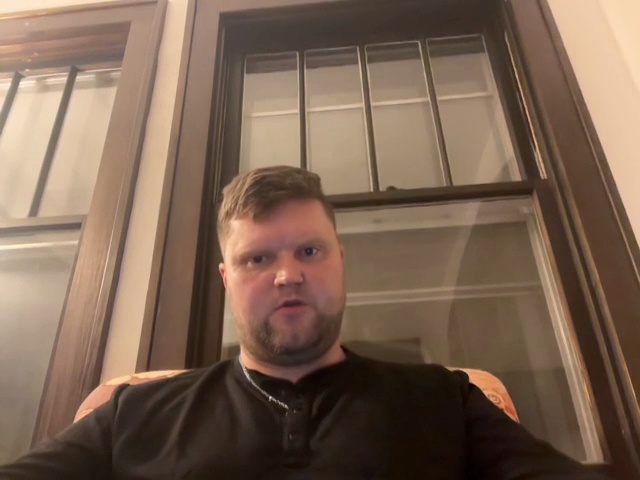} \\
        \includegraphics[width=0.24\linewidth]{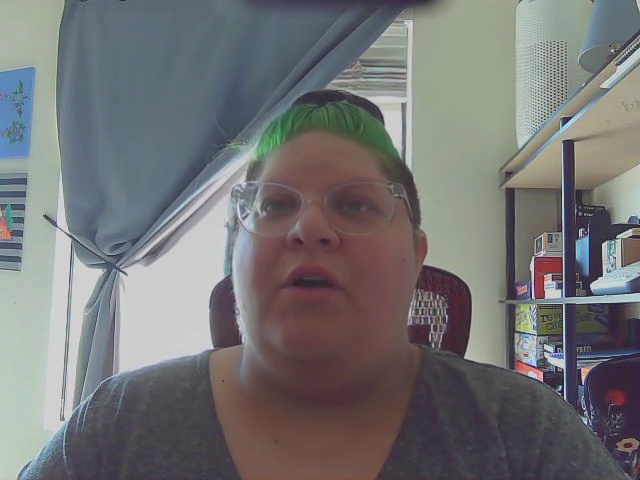} & 
        \includegraphics[width=0.24\linewidth]{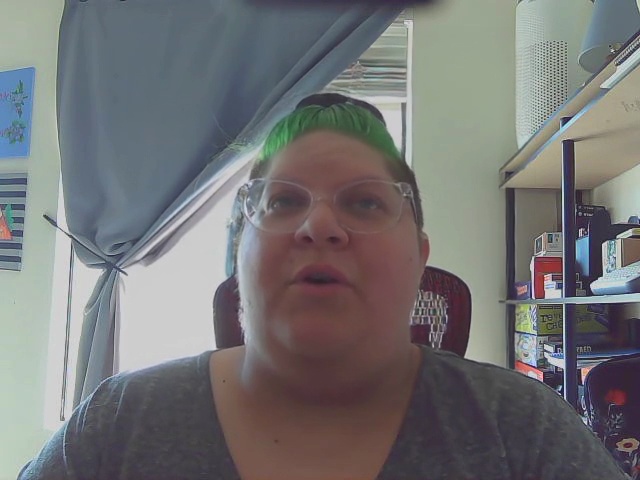} & 
        \includegraphics[width=0.24\linewidth]{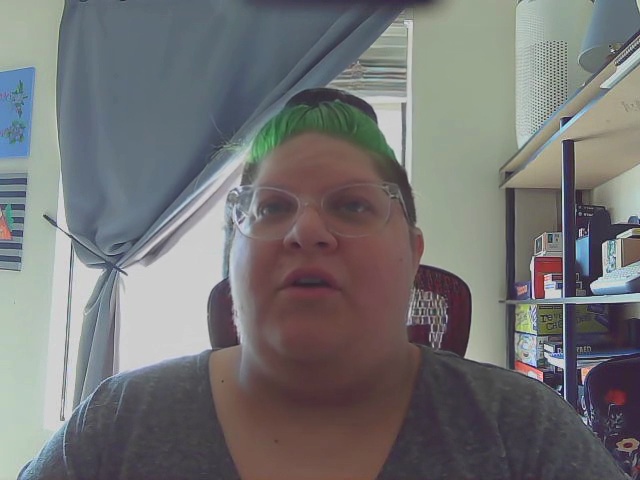} & 
        \includegraphics[width=0.24\linewidth]{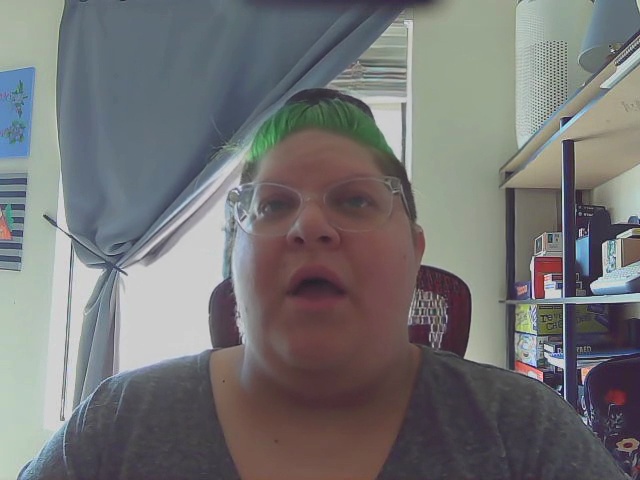} \\
        \includegraphics[width=0.24\linewidth]{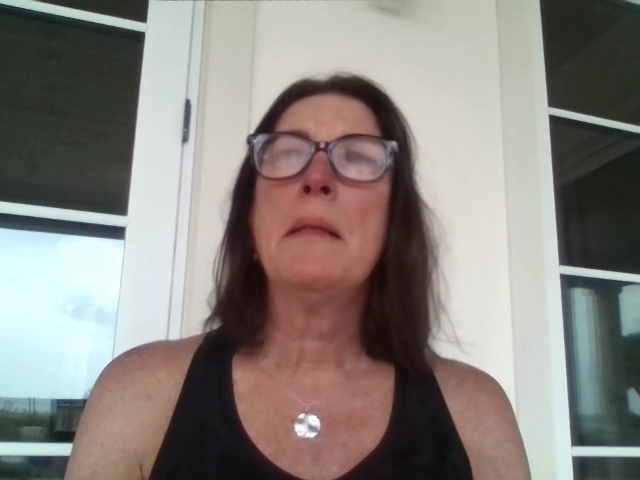} & 
        \includegraphics[width=0.24\linewidth]{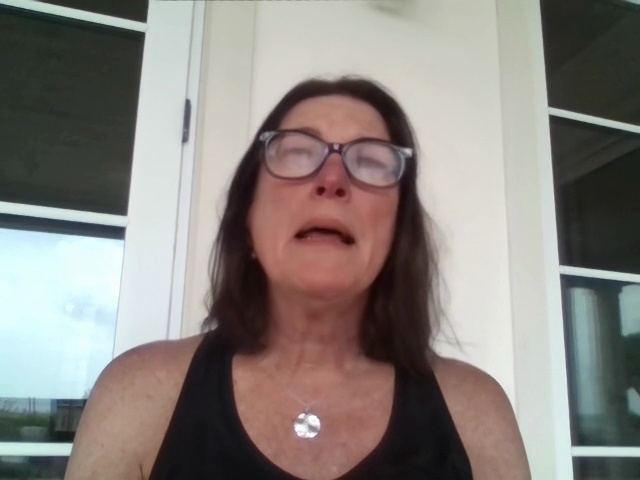} & 
        \includegraphics[width=0.24\linewidth]{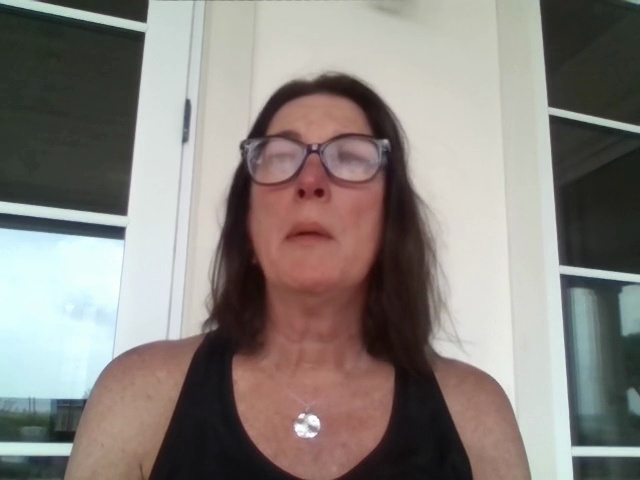} & 
        \includegraphics[width=0.24\linewidth]{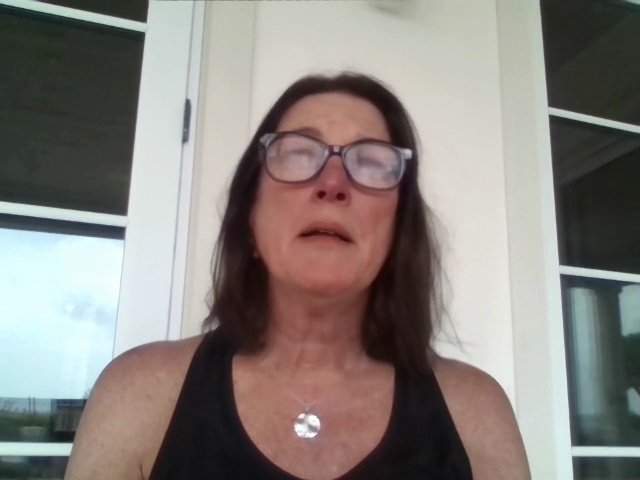}
    \end{tabular}
    }
    \caption{Video frames sampled from three representative examples of \textbf{avatar} deepfakes generated using \textbf{LivePortrait}.}
    \label{fig:frame_examples_liveportrait}
\end{figure}

\newpage


\begin{figure}[H]
    \centering
    \resizebox{0.92\textwidth}{!}{
    \begin{tabular}{c@{\hspace{0cm}}c@{\hspace{0cm}}c@{\hspace{0cm}}c}
        \includegraphics[width=0.24\linewidth]{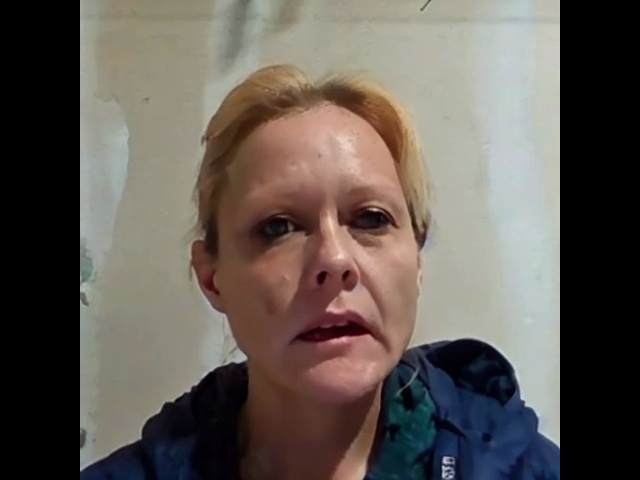} & 
        \includegraphics[width=0.24\linewidth]{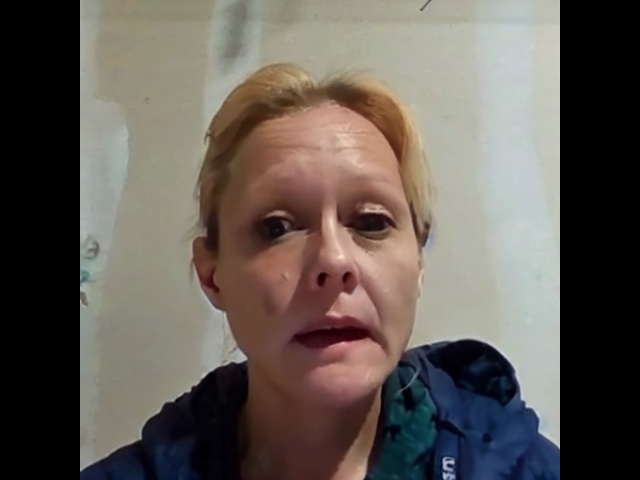} & 
        \includegraphics[width=0.24\linewidth]{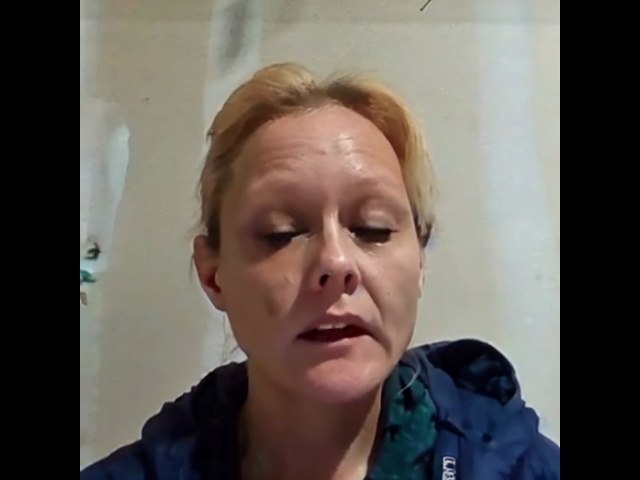} & 
        \includegraphics[width=0.24\linewidth]{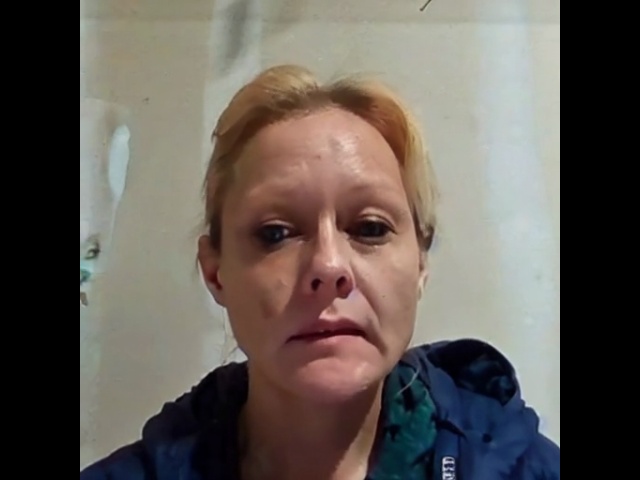} \\
        \includegraphics[width=0.24\linewidth]{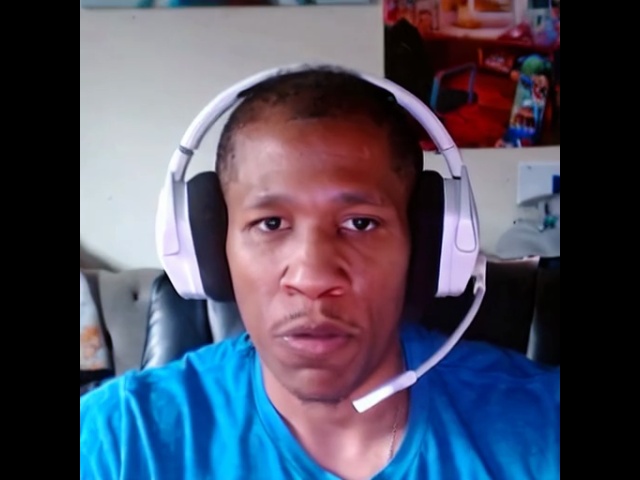} & 
        \includegraphics[width=0.24\linewidth]{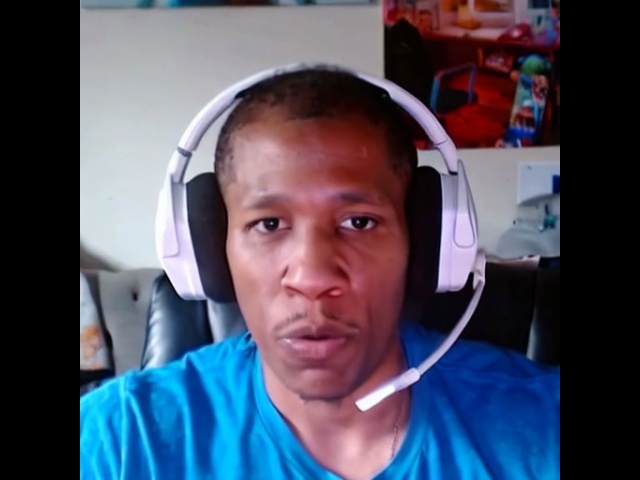} & 
        \includegraphics[width=0.24\linewidth]{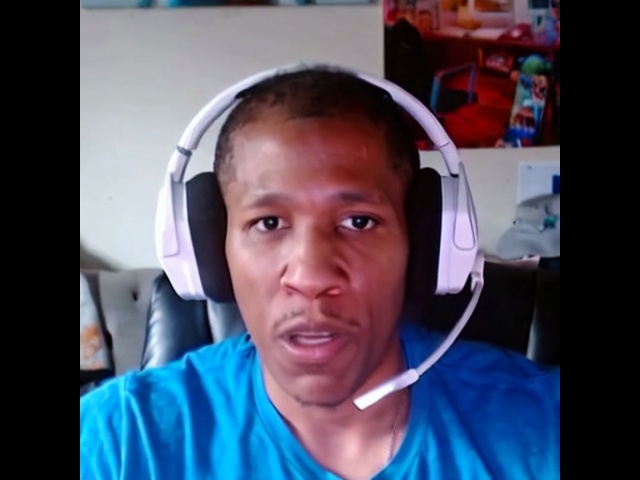} & 
        \includegraphics[width=0.24\linewidth]{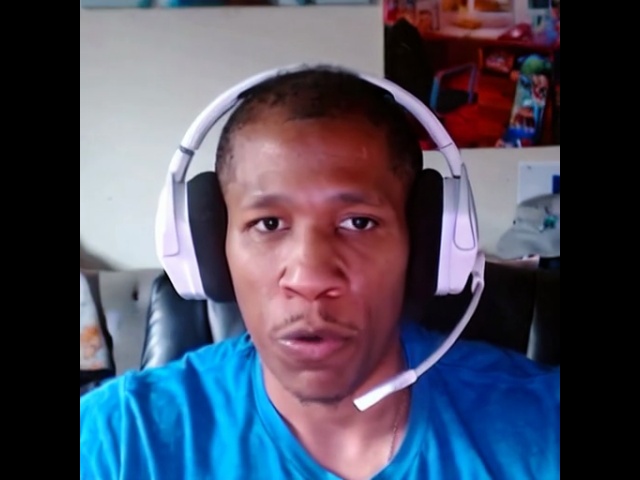} \\
        \includegraphics[width=0.24\linewidth]{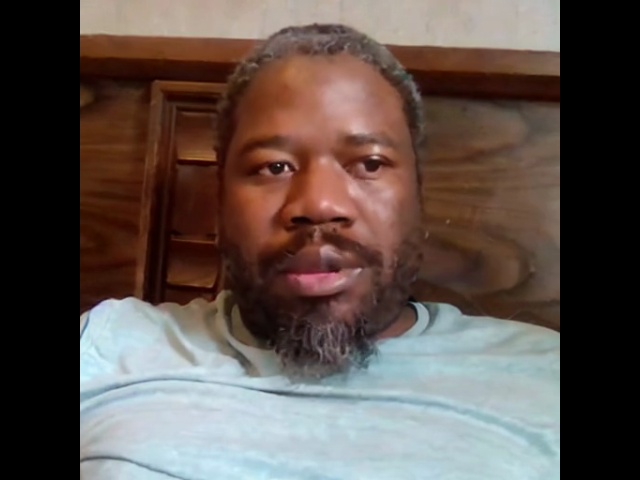} & 
        \includegraphics[width=0.24\linewidth]{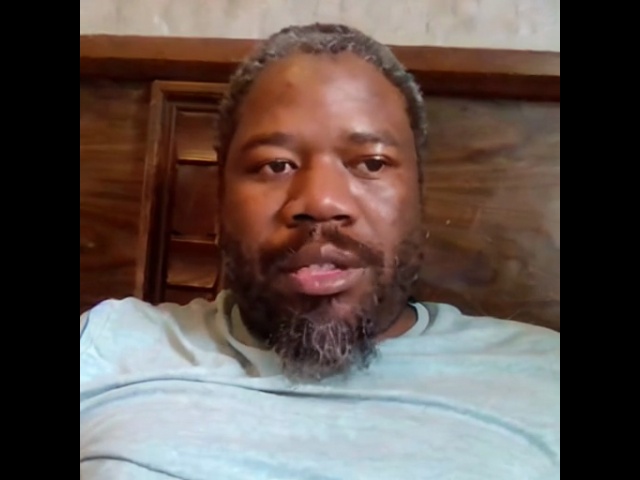} & 
        \includegraphics[width=0.24\linewidth]{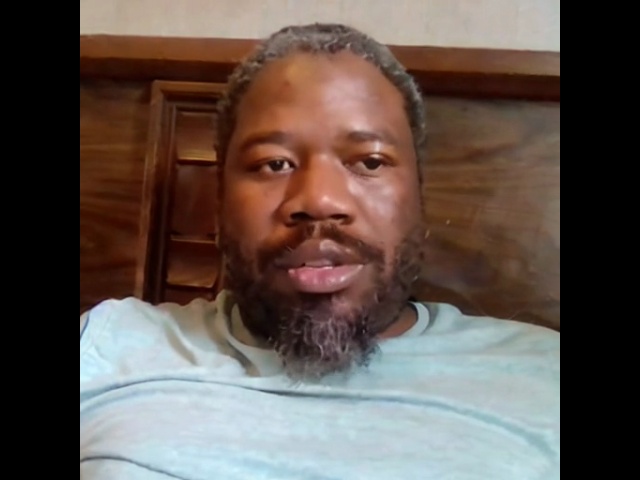} & 
        \includegraphics[width=0.24\linewidth]{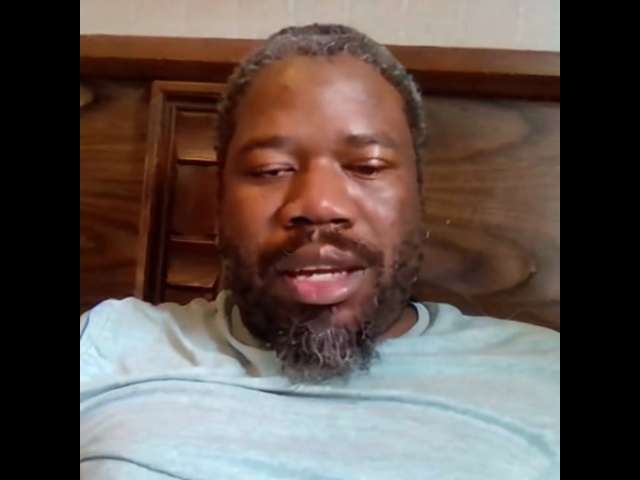}
    \end{tabular}
    }
    \caption{Video frames sampled from three representative examples of \textbf{avatar} deepfakes generated using \textbf{HelloMeme}.}
    \label{fig:frame_examples_hellomeme}
\end{figure}


\begin{figure}[H]
    \centering
    \resizebox{0.92\textwidth}{!}{
    \begin{tabular}{c@{\hspace{0cm}}c@{\hspace{0cm}}c@{\hspace{0cm}}c}
        \includegraphics[width=0.24\linewidth]{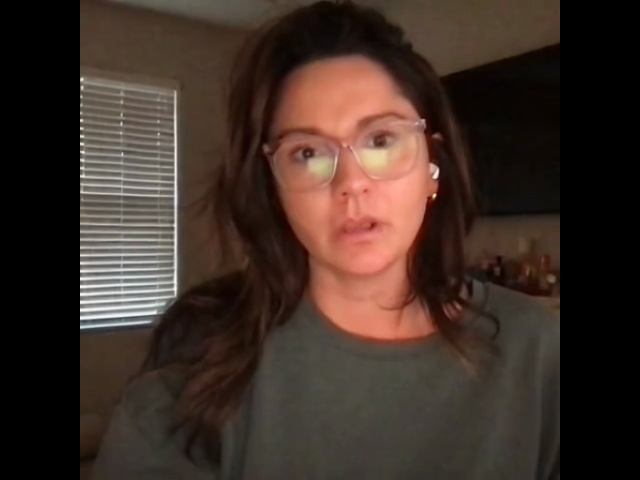} & 
        \includegraphics[width=0.24\linewidth]{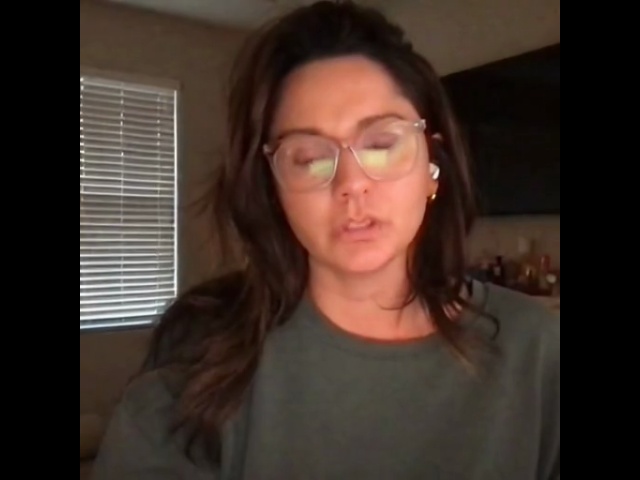} & 
        \includegraphics[width=0.24\linewidth]{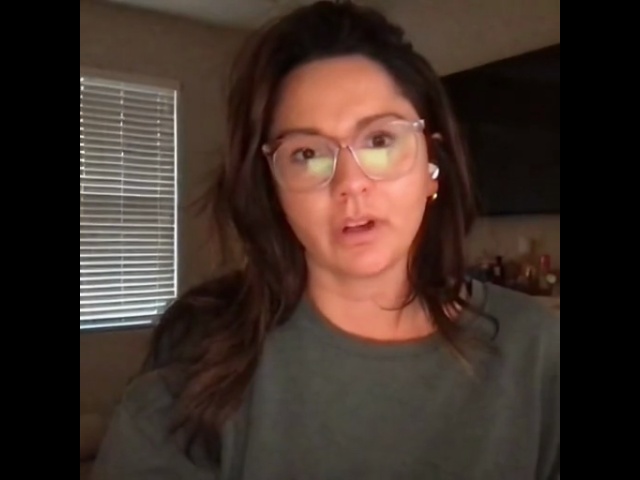} & 
        \includegraphics[width=0.24\linewidth]{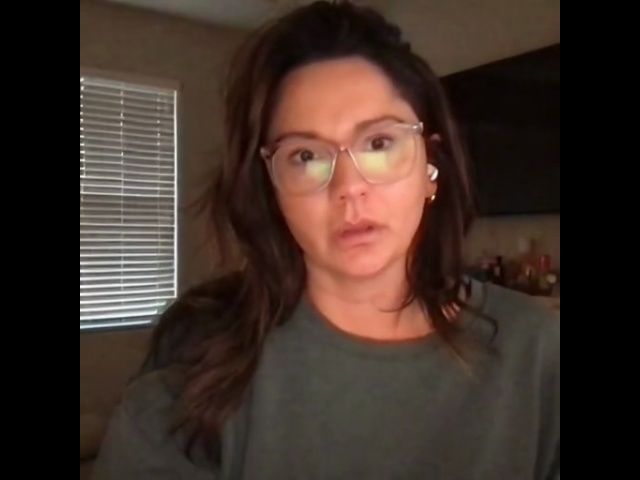} \\
        \includegraphics[width=0.24\linewidth]{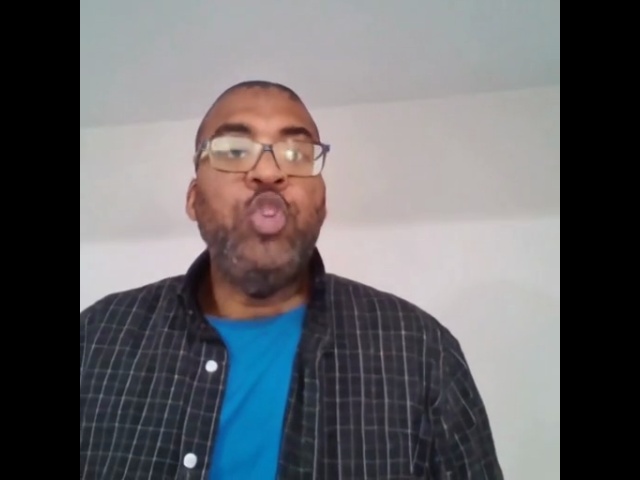} & 
        \includegraphics[width=0.24\linewidth]{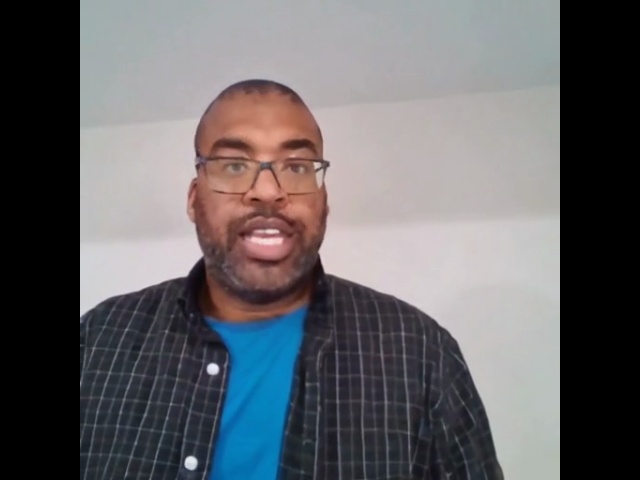} & 
        \includegraphics[width=0.24\linewidth]{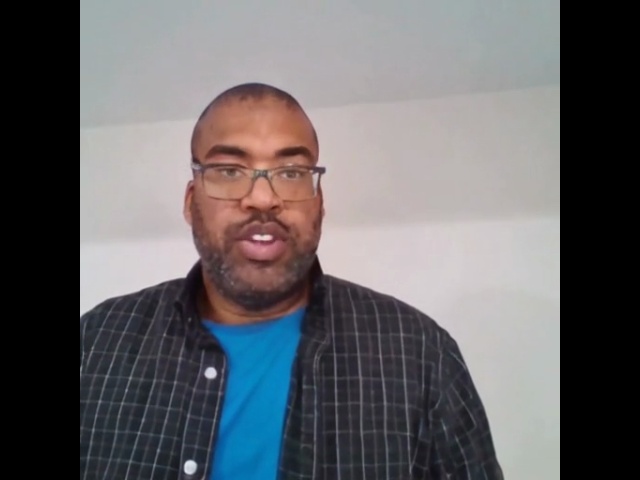} & 
        \includegraphics[width=0.24\linewidth]{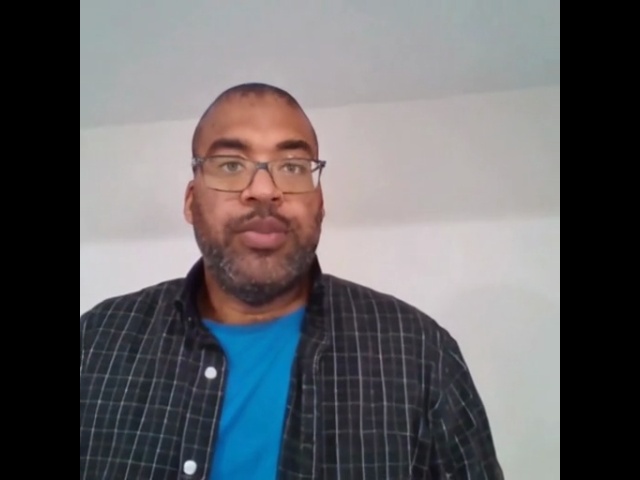} \\
        \includegraphics[width=0.24\linewidth]{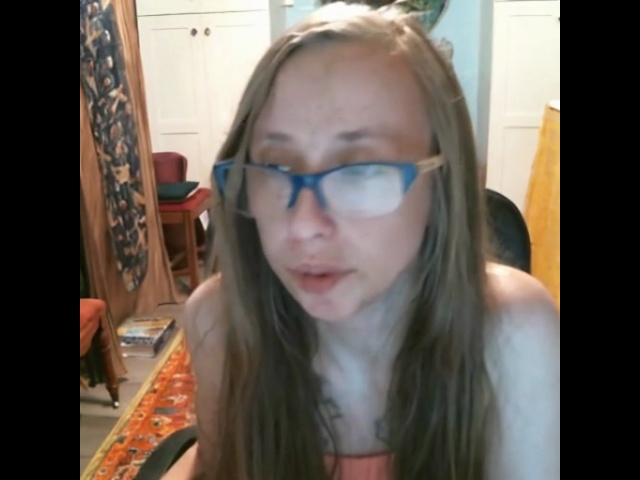} & 
        \includegraphics[width=0.24\linewidth]{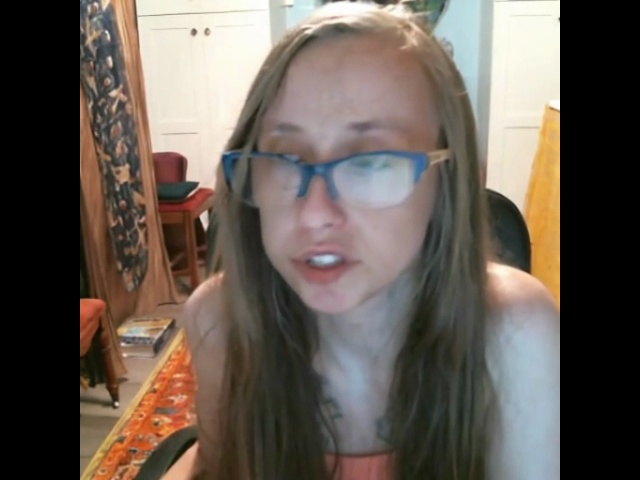} & 
        \includegraphics[width=0.24\linewidth]{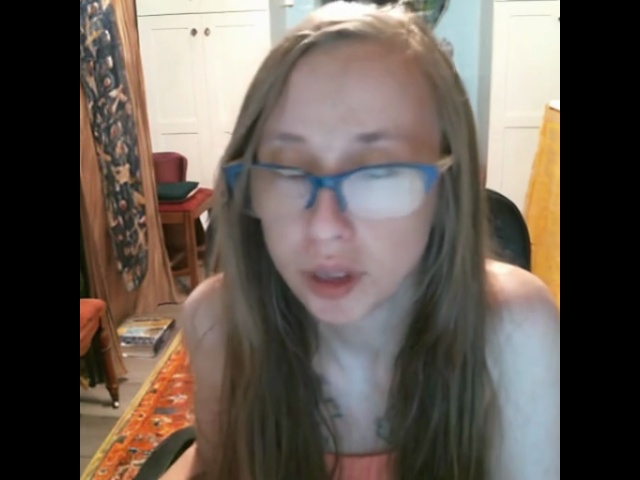} & 
        \includegraphics[width=0.24\linewidth]{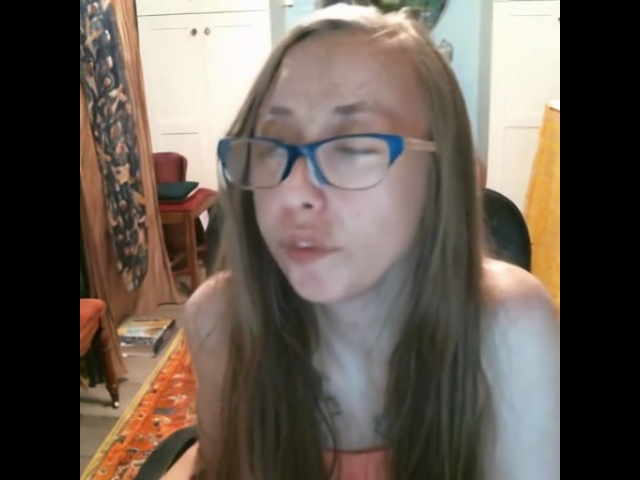}
    \end{tabular}
    }
    \caption{Video frames sampled from three representative examples of \textbf{avatar} deepfakes generated using \textbf{Memo}.}
    \label{fig:frame_examples_memo}
\end{figure}


\section{Visual Identity Pairing Examples}
\label{app:visual_identity_pairing_examples}
\vspace{-\intextsep}
\begin{figure}[H]
    \centering
    \resizebox{1.0\textwidth}{!}{
    \begin{tabular}{c@{\hspace{0cm}}c@{\hspace{0.2cm}}c@{\hspace{0cm}}c}
        \includegraphics[width=0.24\linewidth]{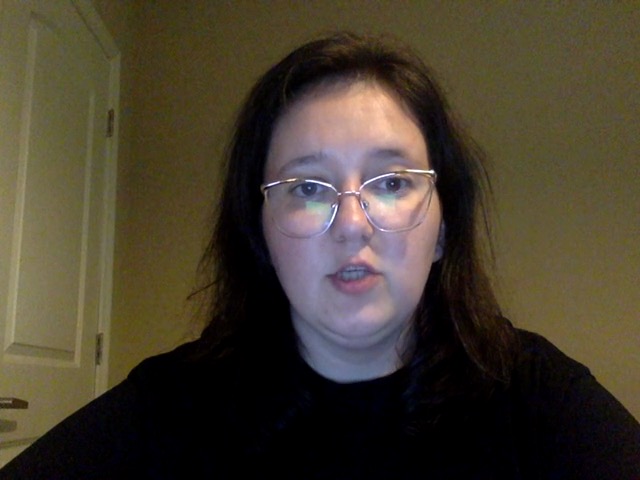} &
        \includegraphics[width=0.24\linewidth]{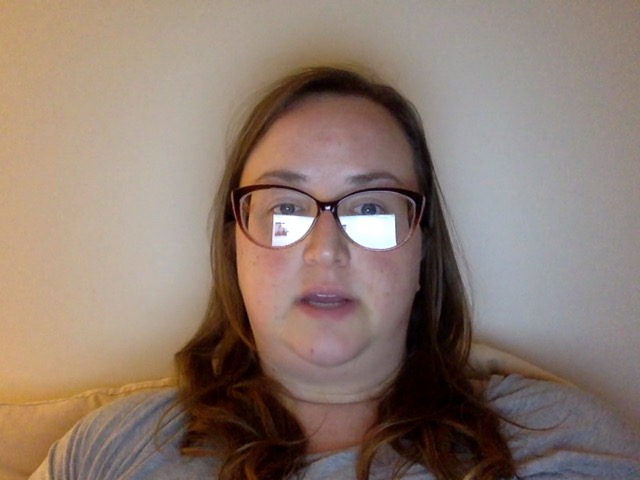} &
        \includegraphics[width=0.24\linewidth]{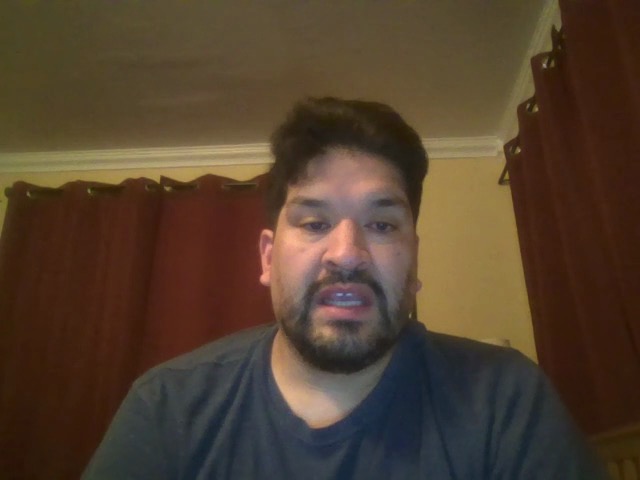} &
        \includegraphics[width=0.24\linewidth]{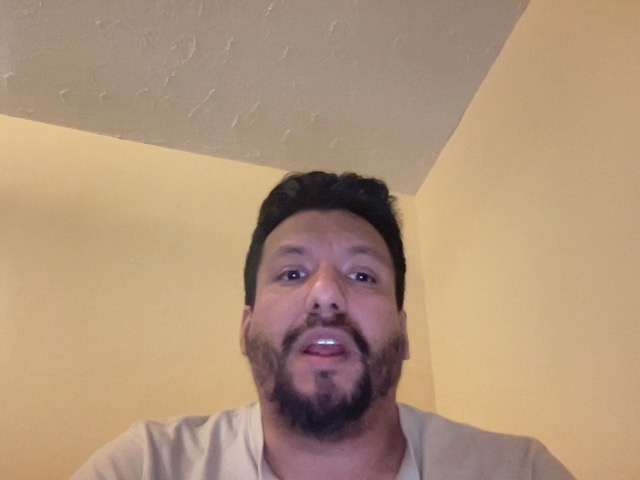} \\
        \includegraphics[width=0.24\linewidth]{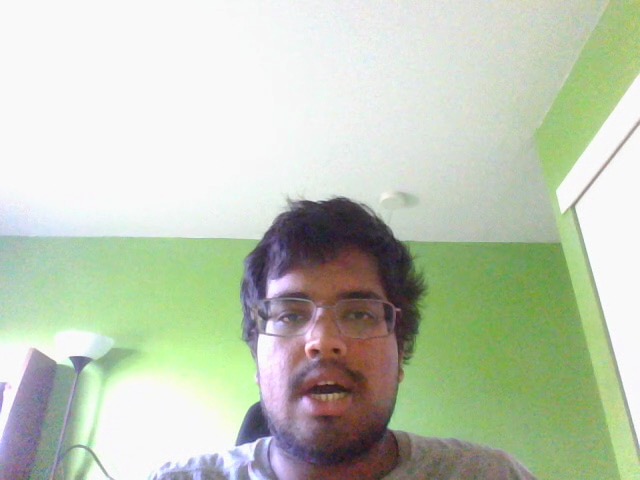} &
        \includegraphics[width=0.24\linewidth]{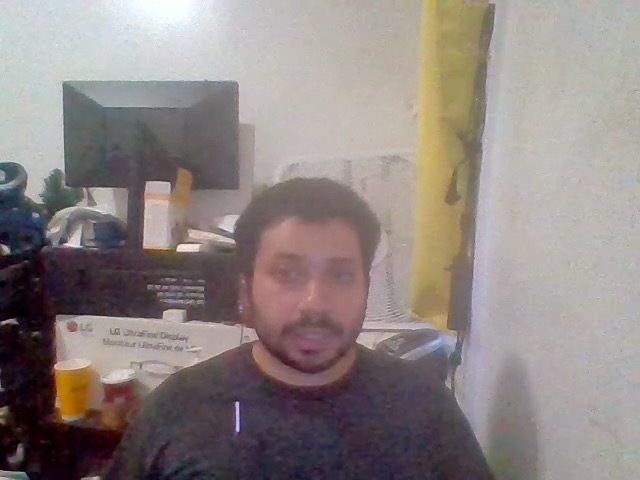} &
        \includegraphics[width=0.24\linewidth]{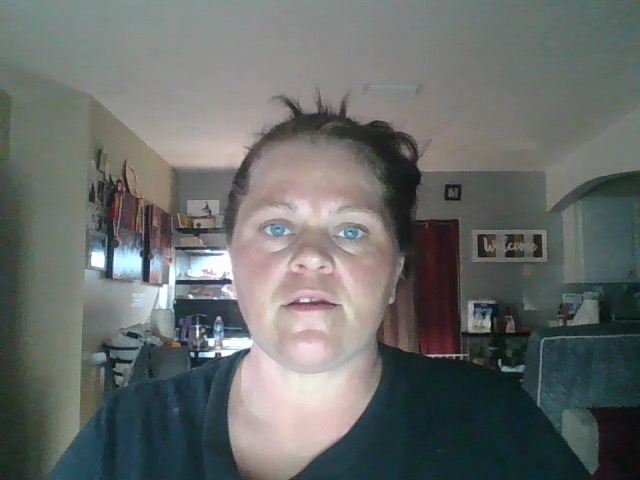} &
        \includegraphics[width=0.24\linewidth]{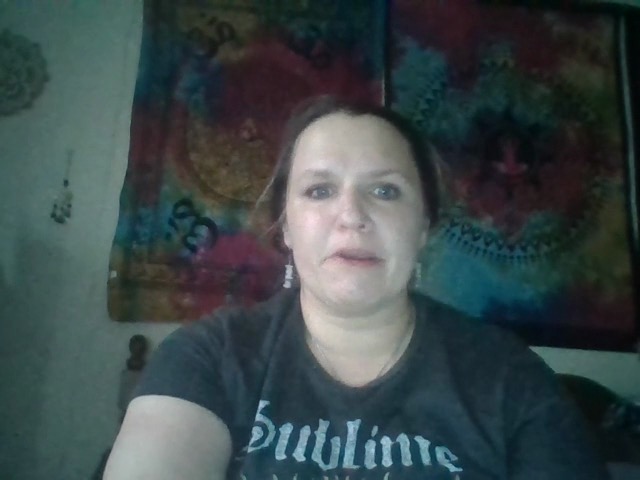} \\
        \includegraphics[width=0.24\linewidth]{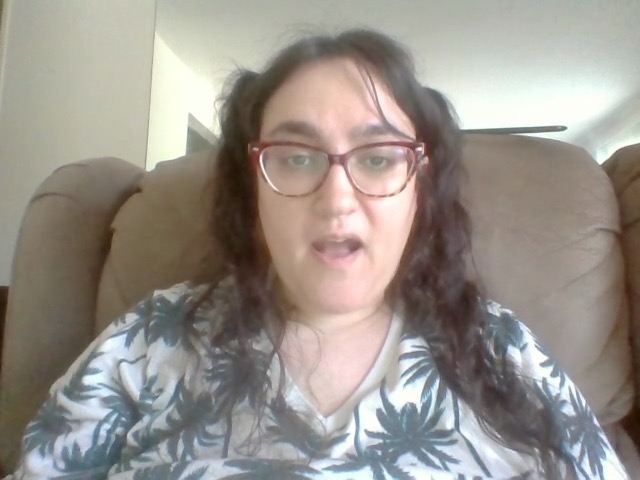} &
        \includegraphics[width=0.24\linewidth]{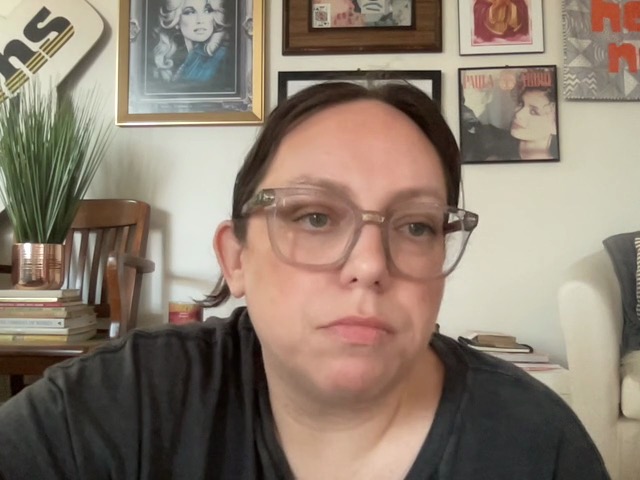} &
        \includegraphics[width=0.24\linewidth]{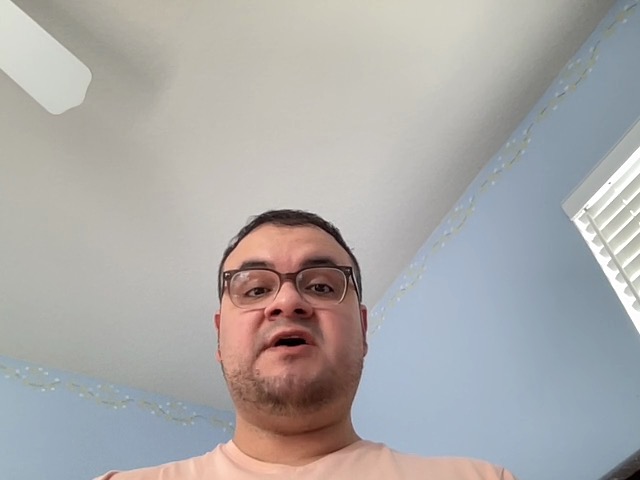} &
        \includegraphics[width=0.24\linewidth]{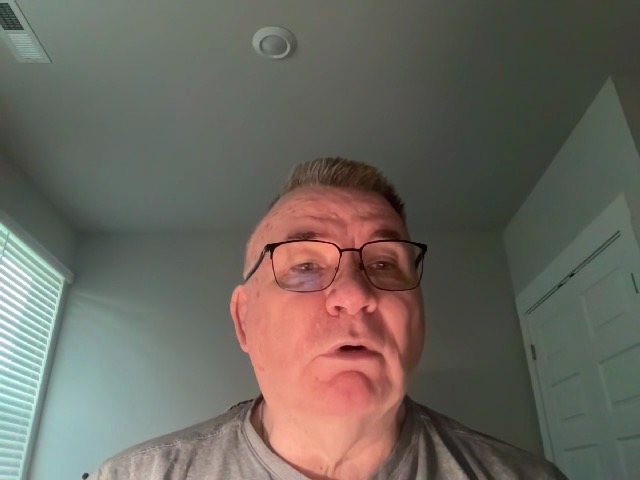} \\
        \includegraphics[width=0.24\linewidth]{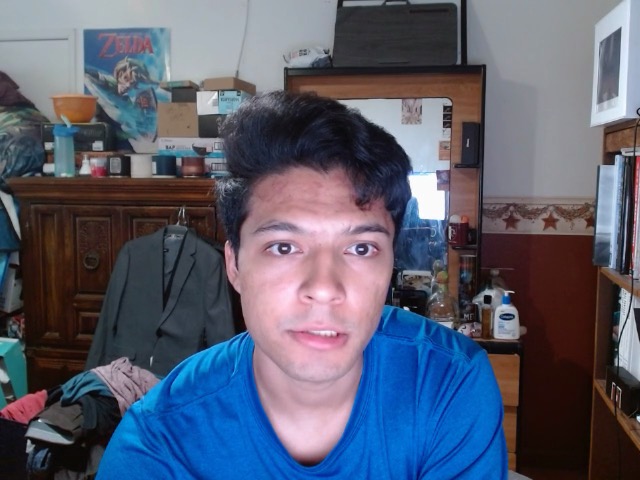} &
        \includegraphics[width=0.24\linewidth]{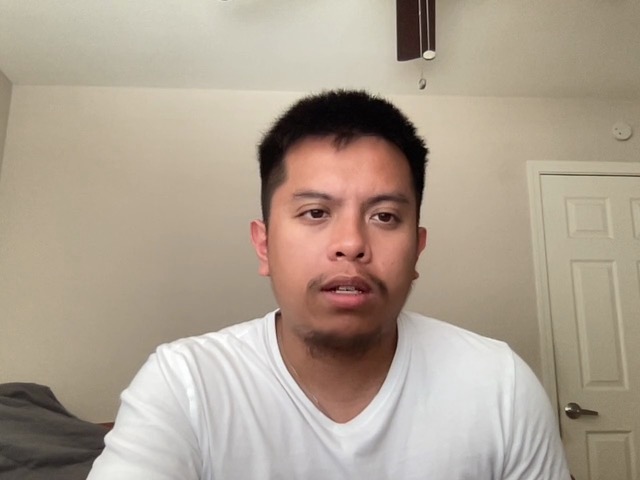} &
        \includegraphics[width=0.24\linewidth]{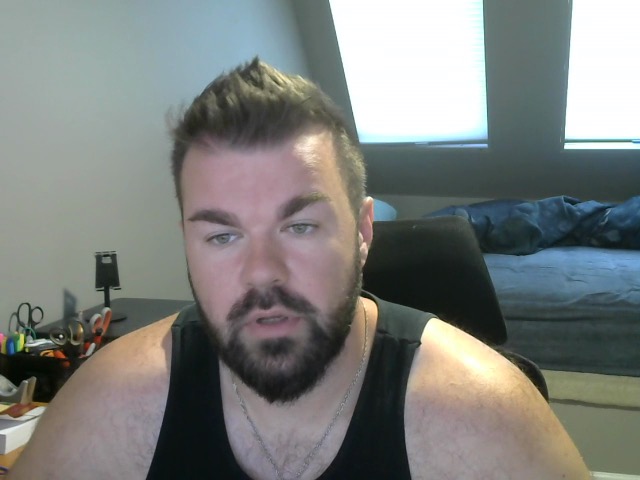} &
        \includegraphics[width=0.24\linewidth]{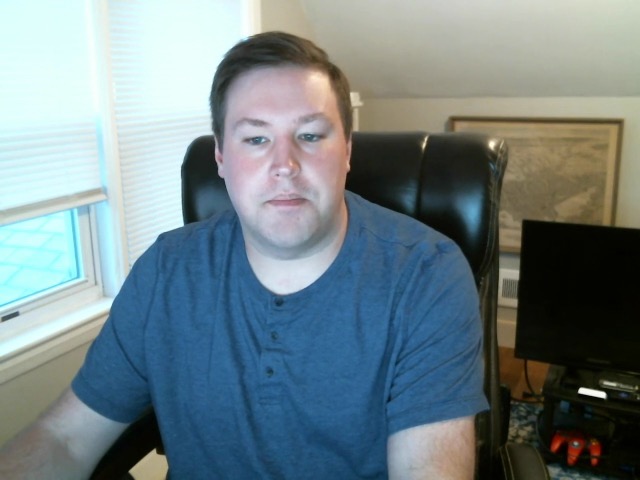}
    \end{tabular}}
    \caption{Additional examples of matched identities from the first release of DeepSpeak data (v1.x). Part 1/2.}
    \label{fig:extra_pairs_v1_a}
\end{figure}

\begin{figure}[H]
    \centering
    \resizebox{1.0\textwidth}{!}{
    \begin{tabular}{c@{\hspace{0cm}}c@{\hspace{0.2cm}}c@{\hspace{0cm}}c}
        \includegraphics[width=0.24\linewidth]{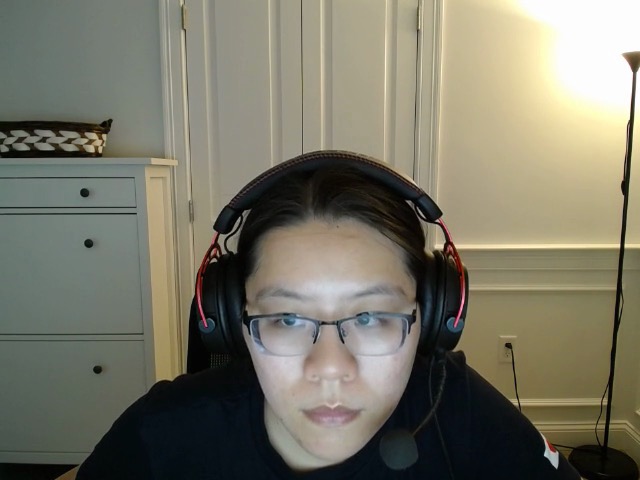} &
        \includegraphics[width=0.24\linewidth]{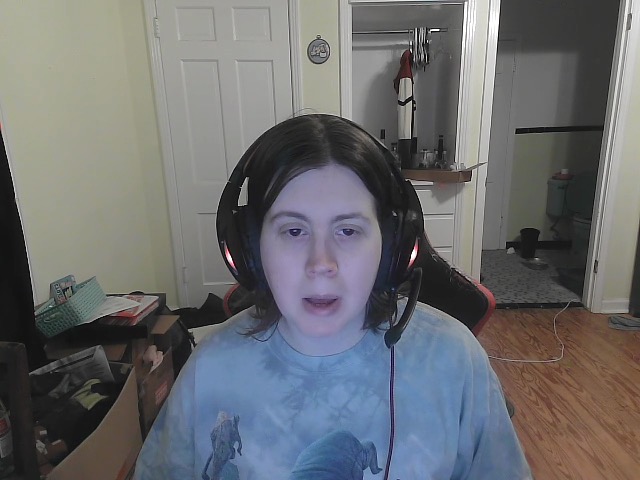} &
        \includegraphics[width=0.24\linewidth]{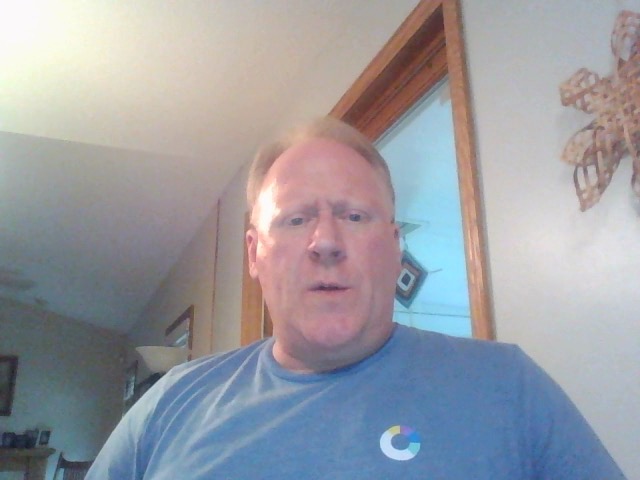} &
        \includegraphics[width=0.24\linewidth]{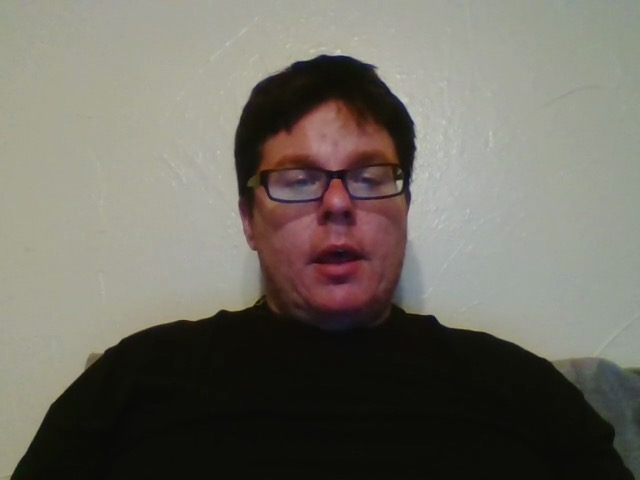} \\
        \includegraphics[width=0.24\linewidth]{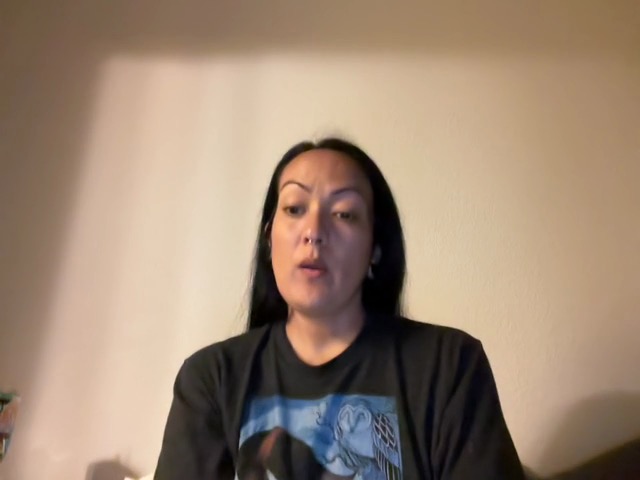} &
        \includegraphics[width=0.24\linewidth]{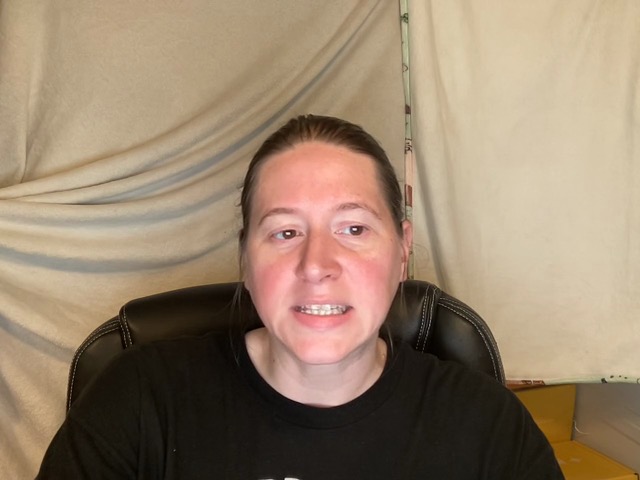} &
        \includegraphics[width=0.24\linewidth]{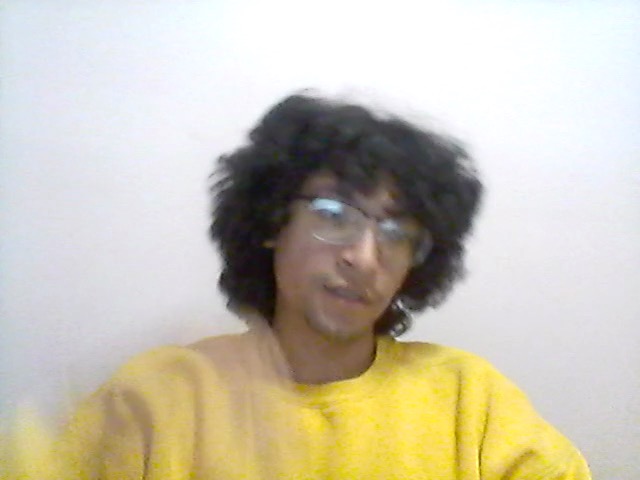} &
        \includegraphics[width=0.24\linewidth]{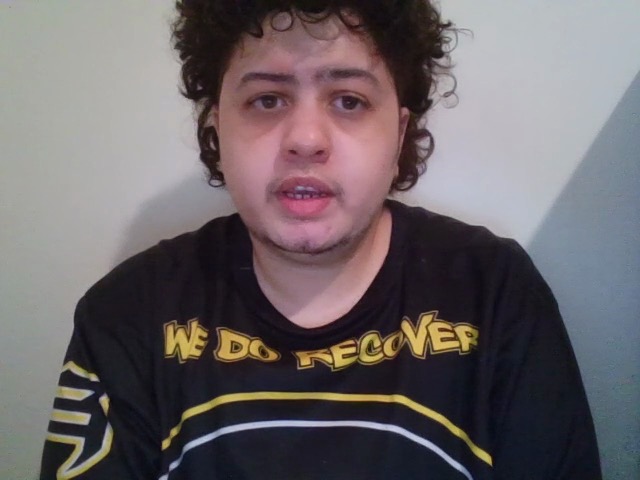} \\
        \includegraphics[width=0.24\linewidth]{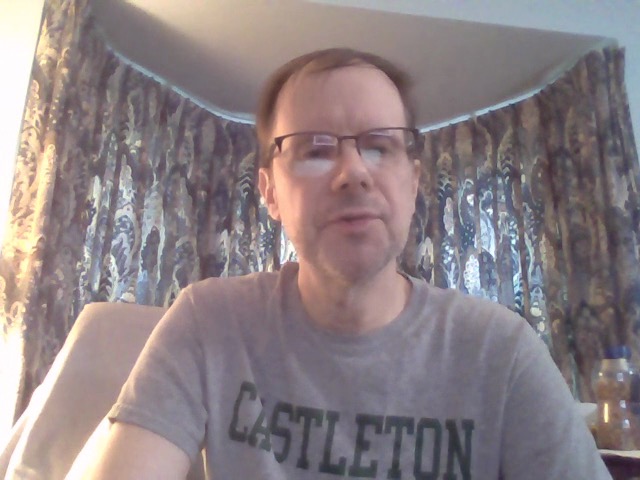} &
        \includegraphics[width=0.24\linewidth]{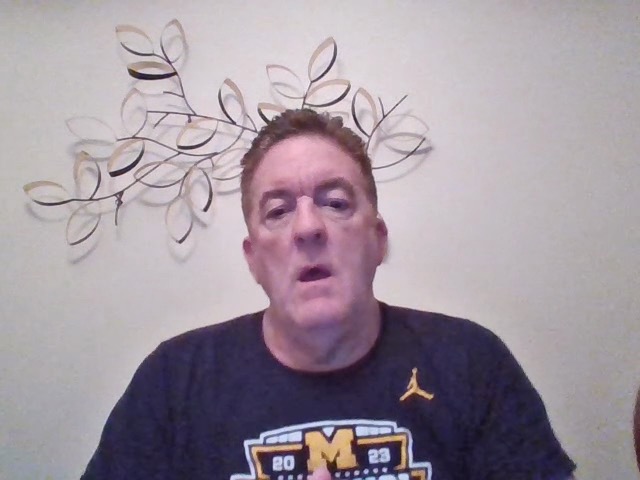} &
        \includegraphics[width=0.24\linewidth]{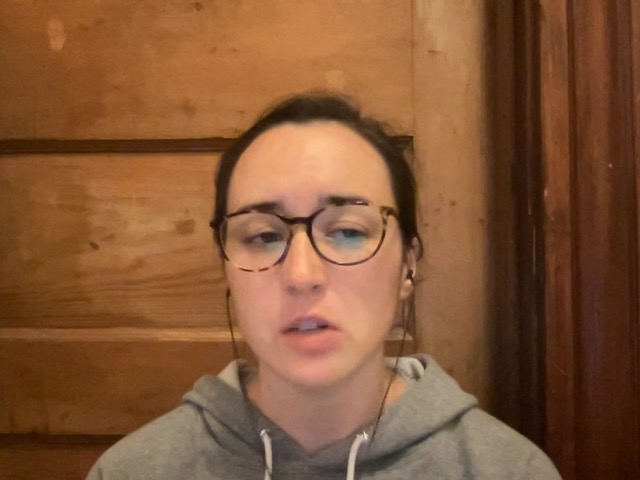} &
        \includegraphics[width=0.24\linewidth]{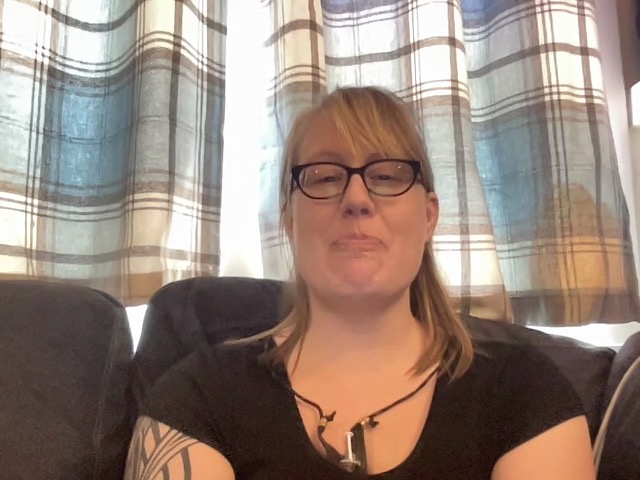} \\
        \includegraphics[width=0.24\linewidth]{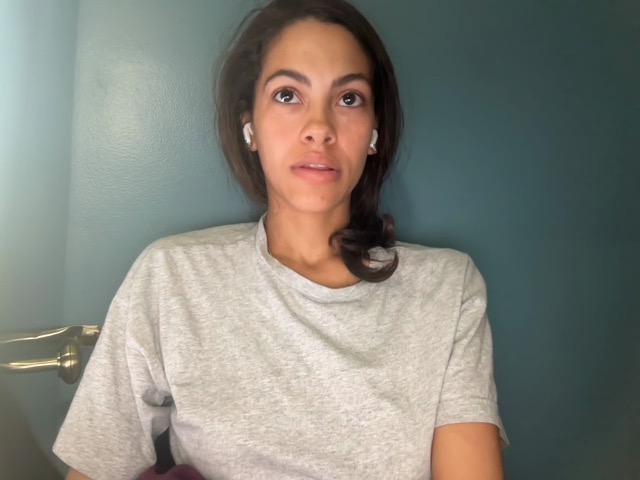} &
        \includegraphics[width=0.24\linewidth]{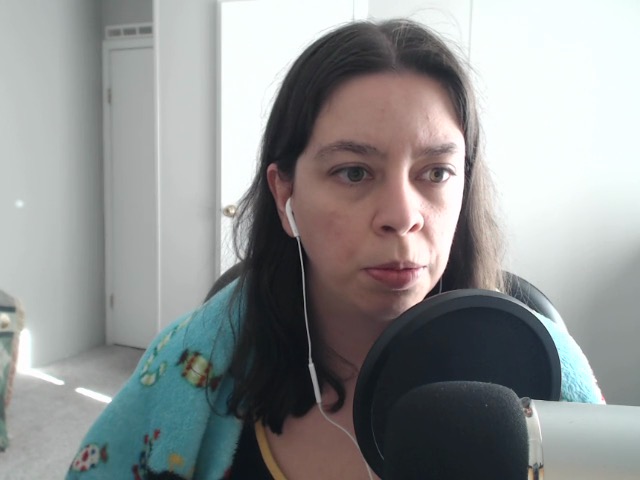} &
        \includegraphics[width=0.24\linewidth]{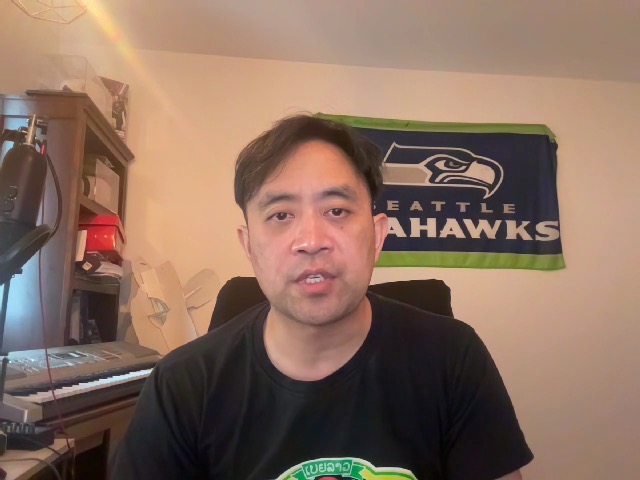} &
        \includegraphics[width=0.24\linewidth]{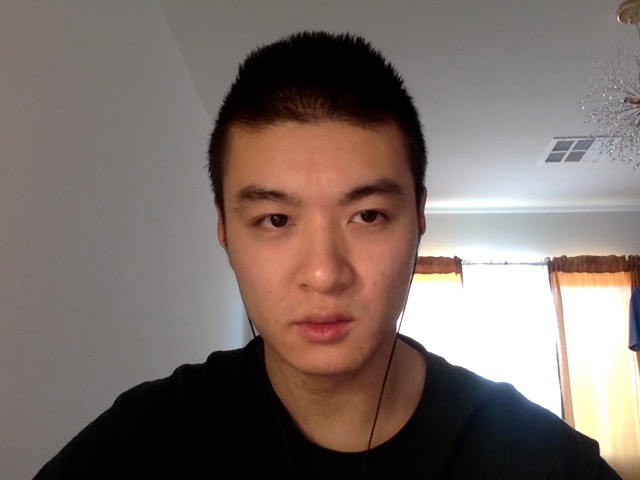}
    \end{tabular}}
    \caption{Additional examples of matched identities from the first release of DeepSpeak data (v1.x). Part 2/2.}
    \label{fig:extra_pairs_v1_b}
\end{figure}

\begin{figure}[H]
    \centering
    \resizebox{1.0\textwidth}{!}{
    \begin{tabular}{c@{\hspace{0cm}}c@{\hspace{0.2cm}}c@{\hspace{0cm}}c}
        \includegraphics[width=0.24\linewidth]{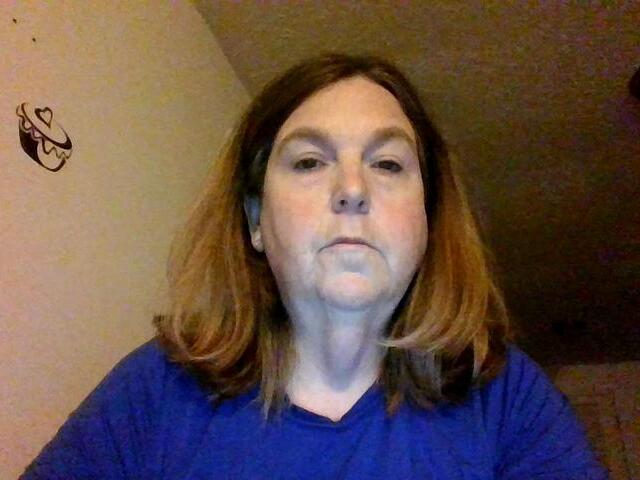} &
        \includegraphics[width=0.24\linewidth]{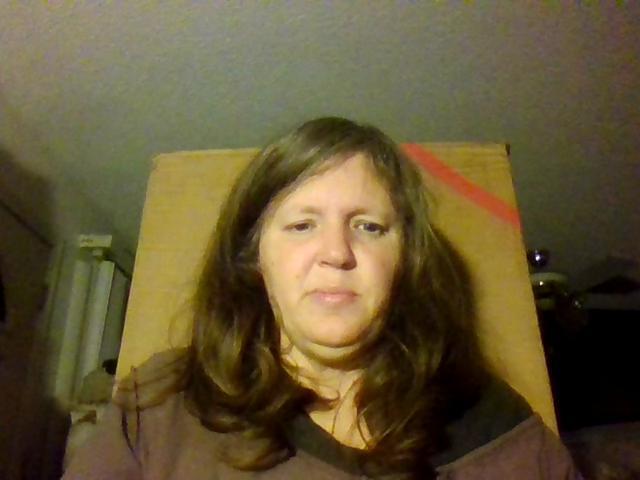} &
        \includegraphics[width=0.24\linewidth]{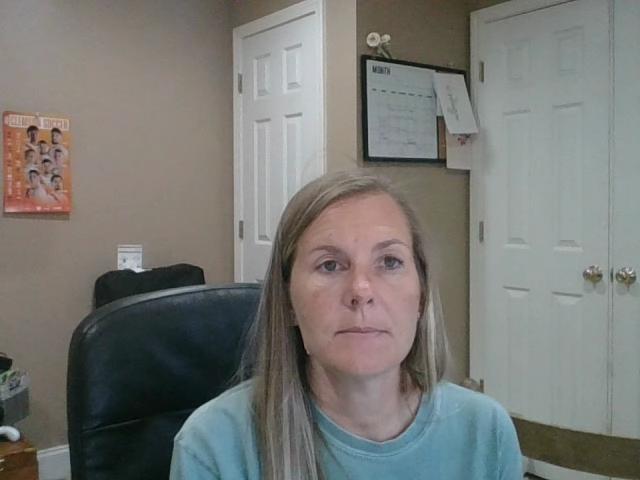} &
        \includegraphics[width=0.24\linewidth]{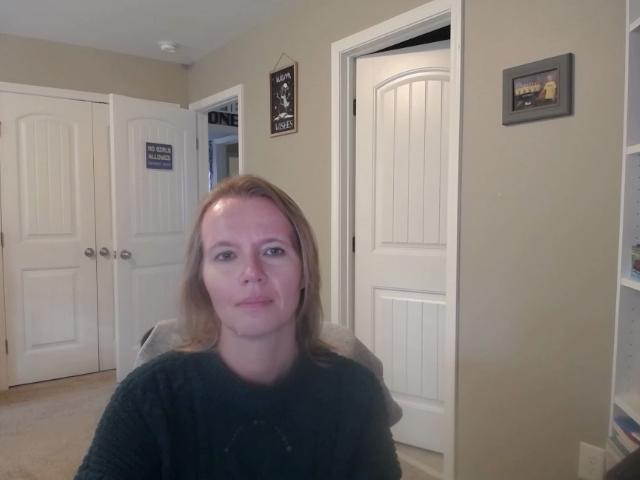} \\
        \includegraphics[width=0.24\linewidth]{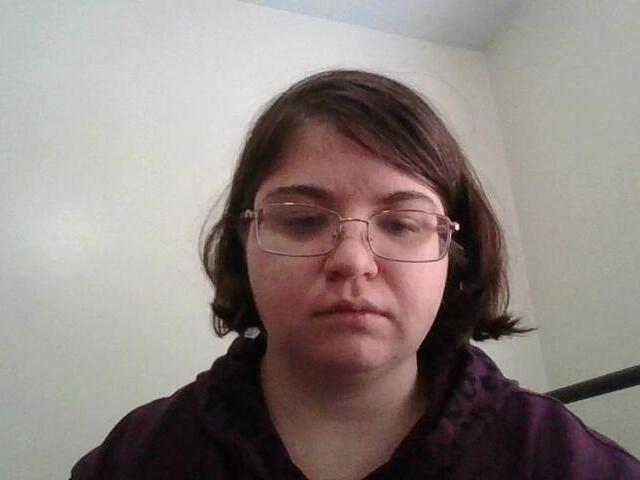} &
        \includegraphics[width=0.24\linewidth]{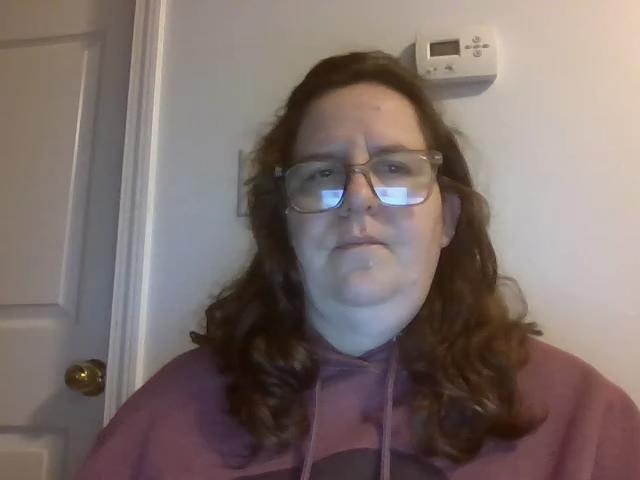} &
        \includegraphics[width=0.24\linewidth]{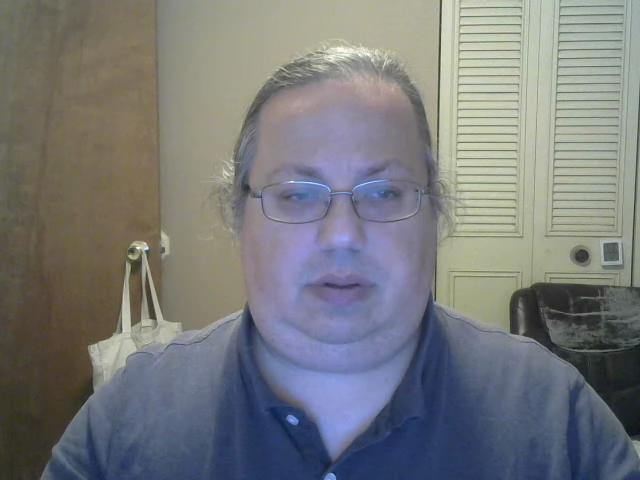} &
        \includegraphics[width=0.24\linewidth]{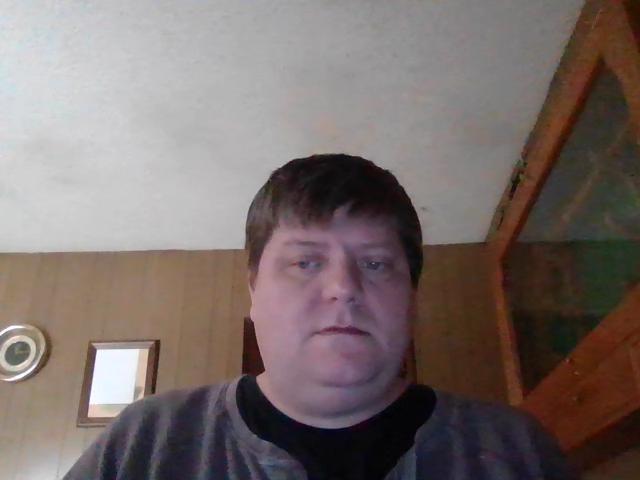} \\
        \includegraphics[width=0.24\linewidth]{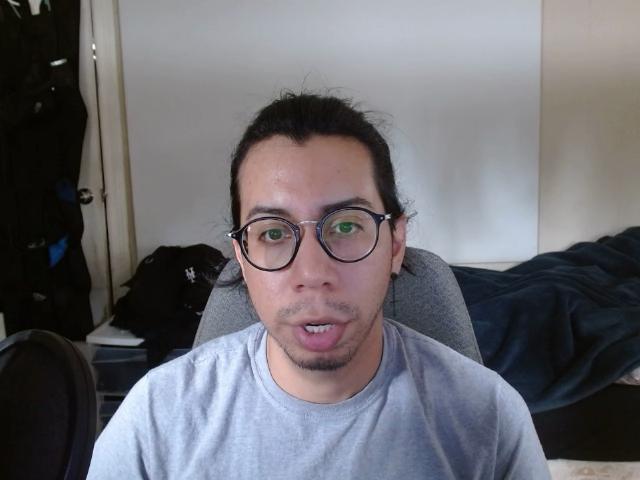} &
        \includegraphics[width=0.24\linewidth]{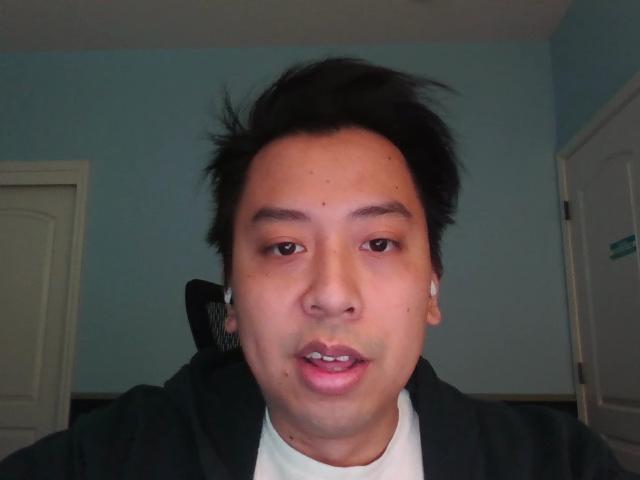} &
        \includegraphics[width=0.24\linewidth]{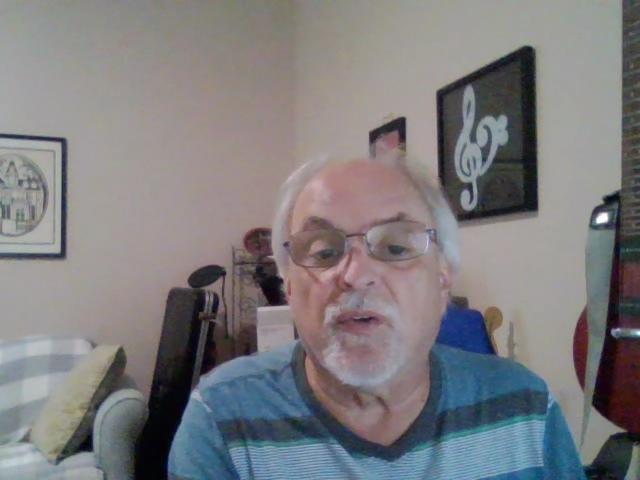} &
        \includegraphics[width=0.24\linewidth]{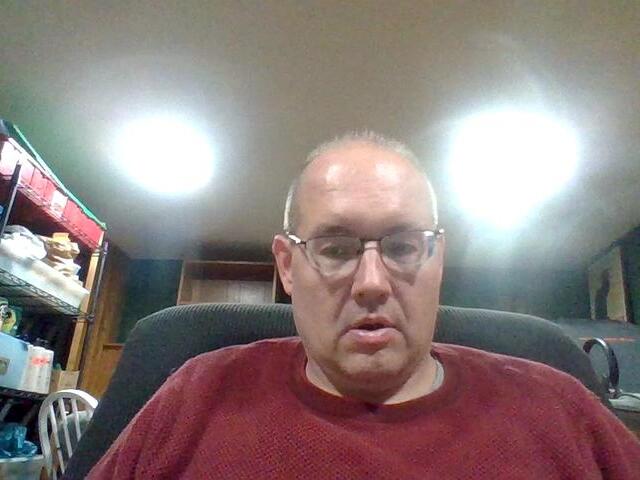} \\
        \includegraphics[width=0.24\linewidth]{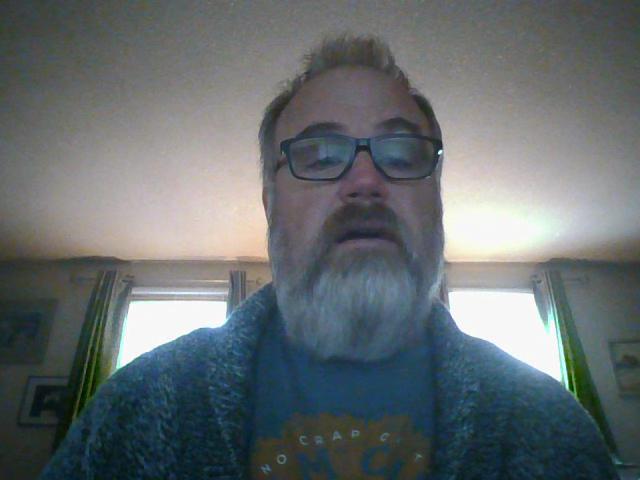} &
        \includegraphics[width=0.24\linewidth]{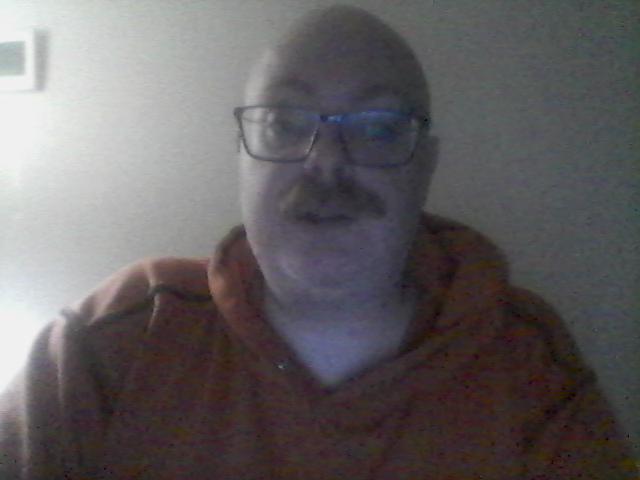} &
        \includegraphics[width=0.24\linewidth]{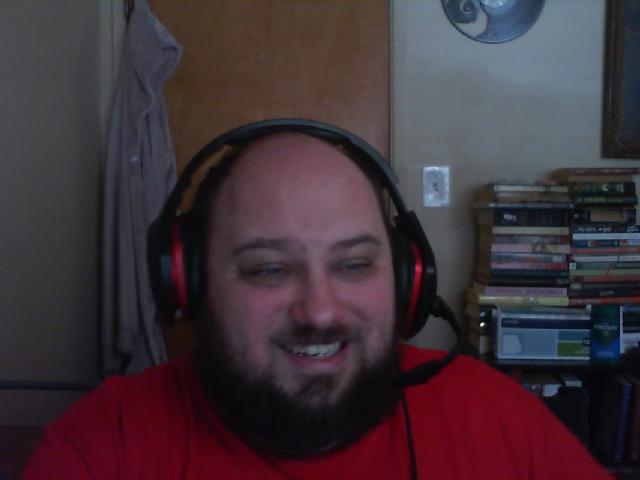} &
        \includegraphics[width=0.24\linewidth]{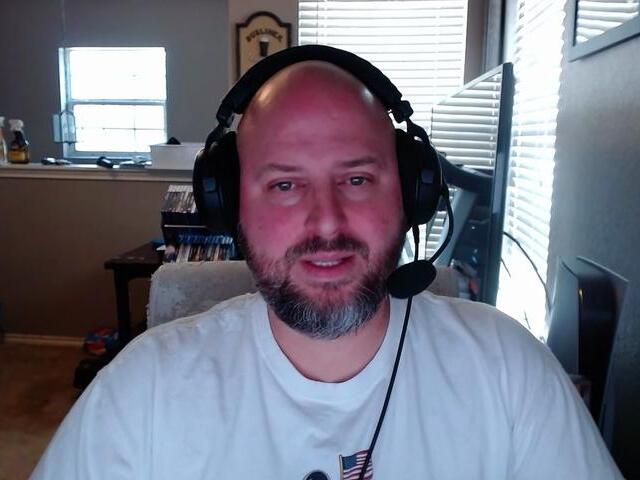}
    \end{tabular}}
    \caption{Additional examples of matched identities from the first release of DeepSpeak data (v2.0). Part 1/2.}
    \label{fig:extra_pairs_v2_a}
\end{figure}

\begin{figure}[H]
    \centering
    \resizebox{1.0\textwidth}{!}{
    \begin{tabular}{c@{\hspace{0cm}}c@{\hspace{0.2cm}}c@{\hspace{0cm}}c}
        \includegraphics[width=0.24\linewidth]{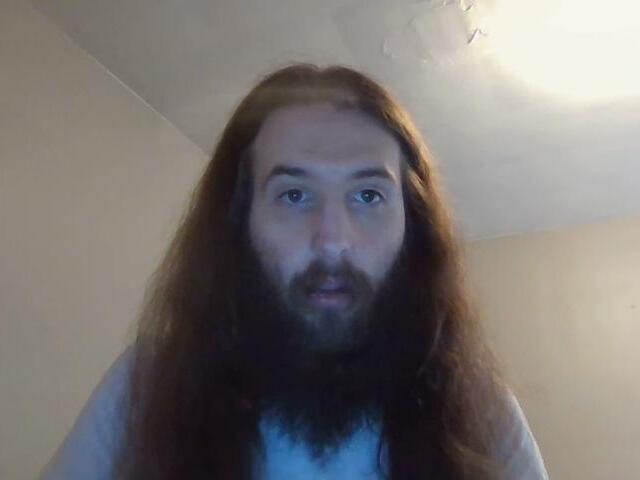} &
        \includegraphics[width=0.24\linewidth]{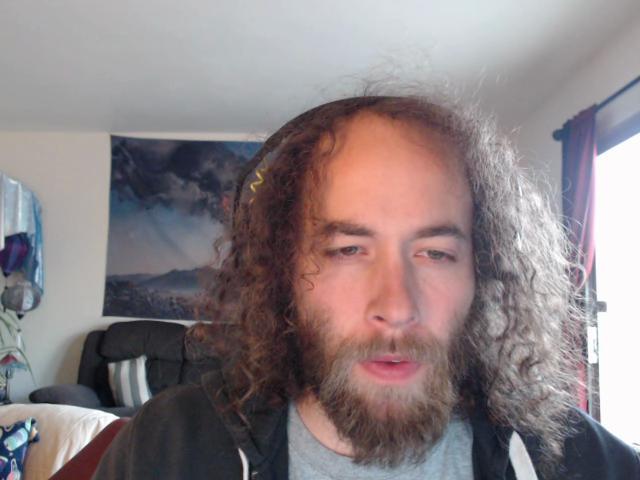} &
        \includegraphics[width=0.24\linewidth]{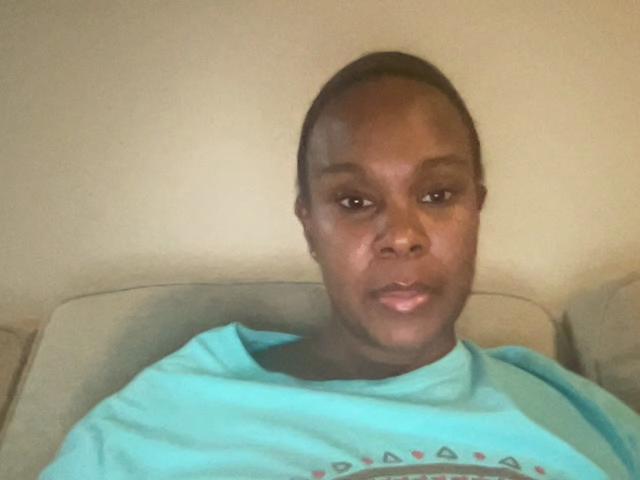} &
        \includegraphics[width=0.24\linewidth]{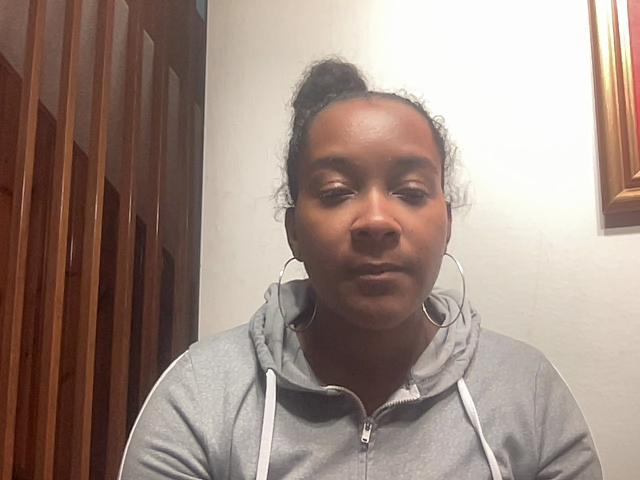} \\
        \includegraphics[width=0.24\linewidth]{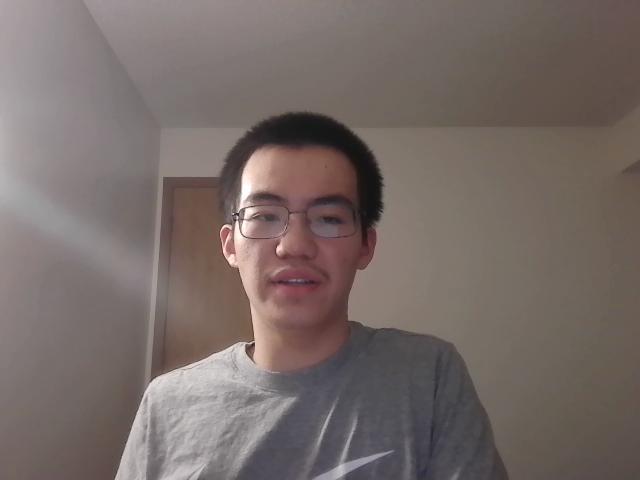} &
        \includegraphics[width=0.24\linewidth]{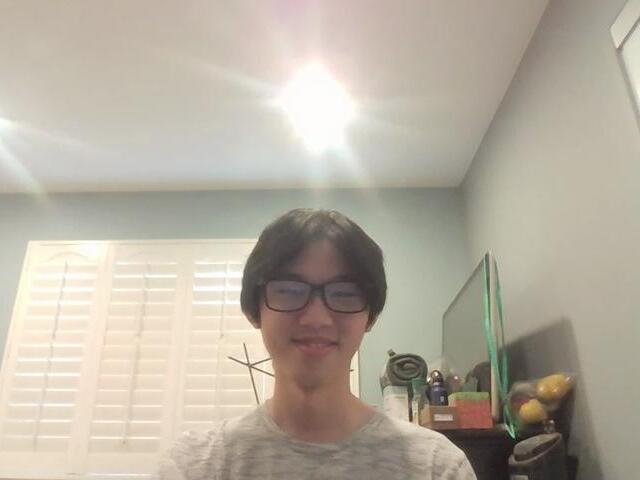} &
        \includegraphics[width=0.24\linewidth]{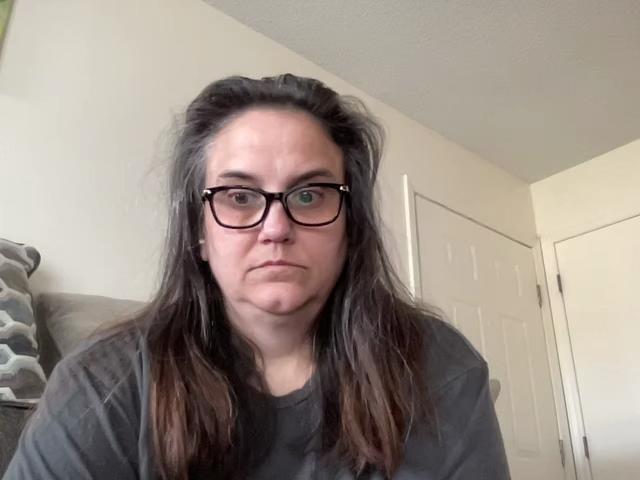} &
        \includegraphics[width=0.24\linewidth]{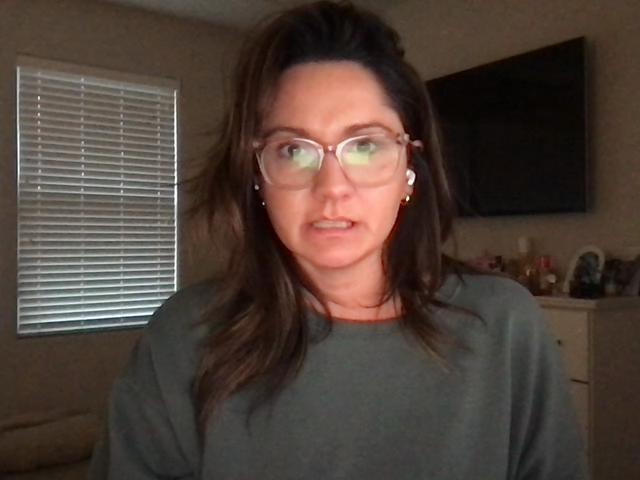} \\
        \includegraphics[width=0.24\linewidth]{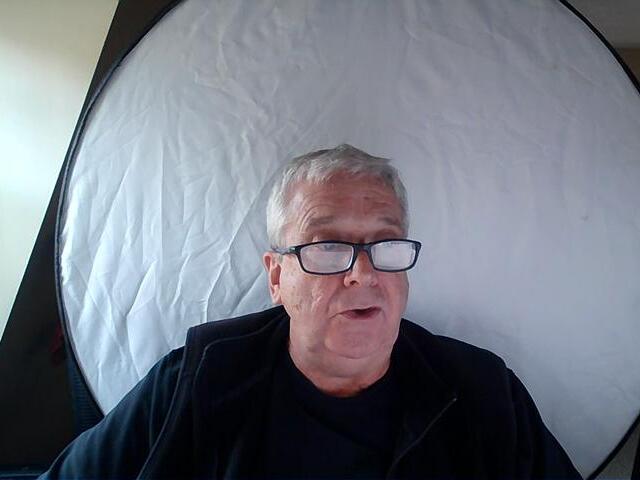} &
        \includegraphics[width=0.24\linewidth]{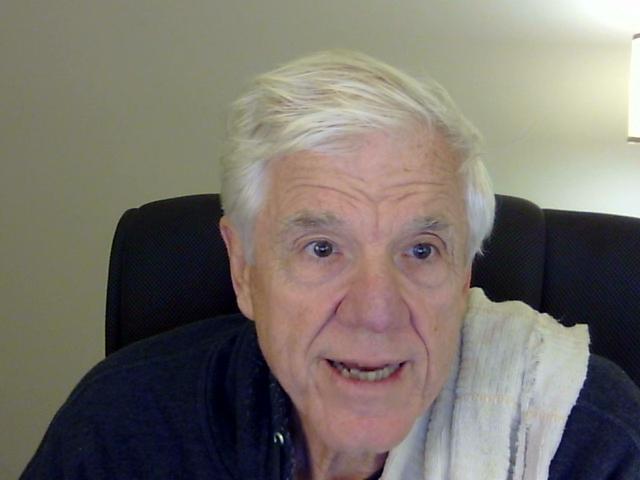} &
        \includegraphics[width=0.24\linewidth]{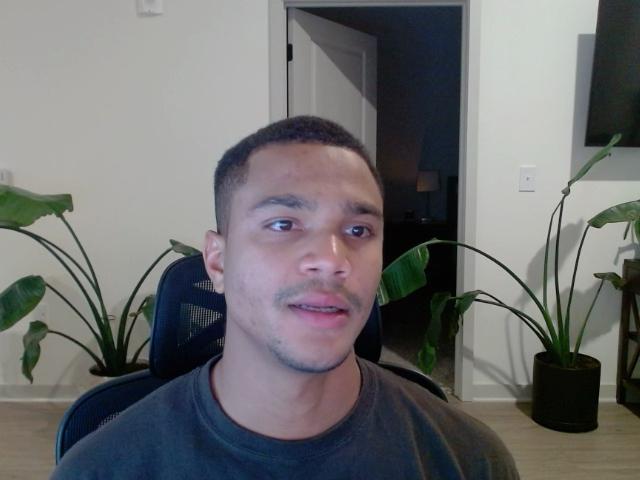} &
        \includegraphics[width=0.24\linewidth]{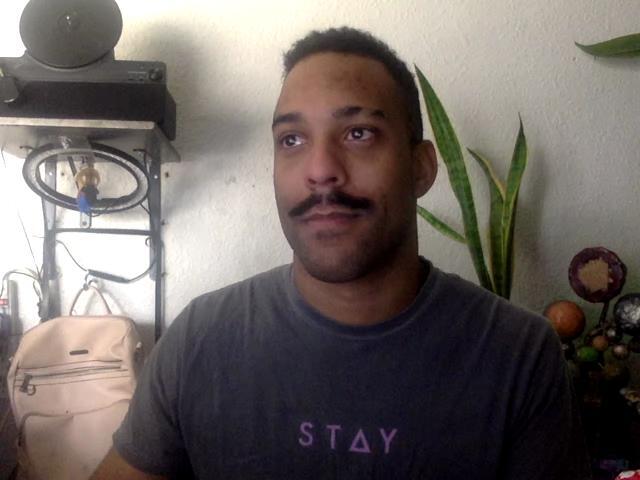} \\
        \includegraphics[width=0.24\linewidth]{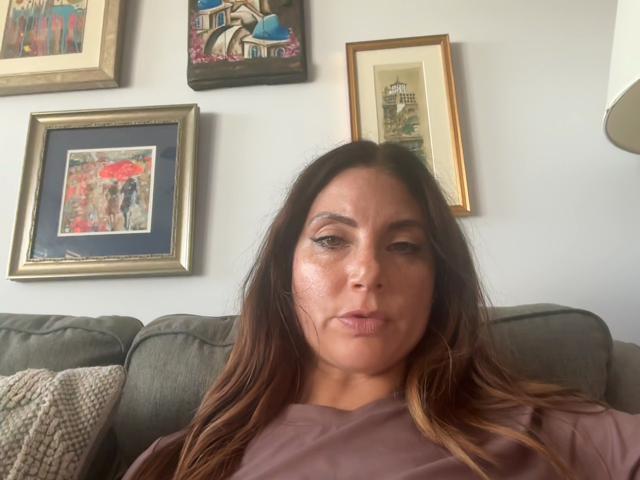} &
        \includegraphics[width=0.24\linewidth]{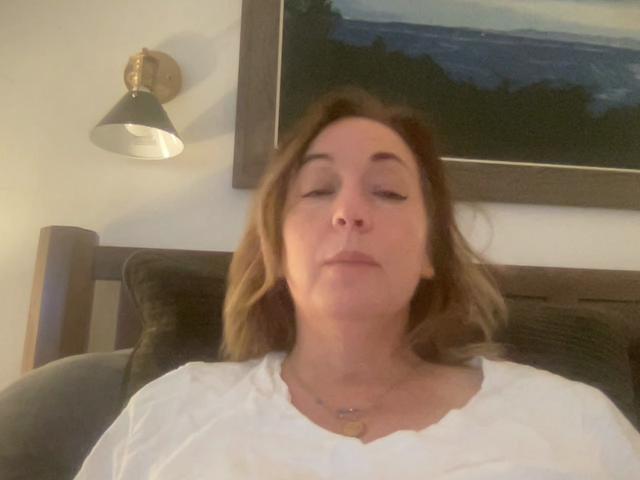} &
        \includegraphics[width=0.24\linewidth]{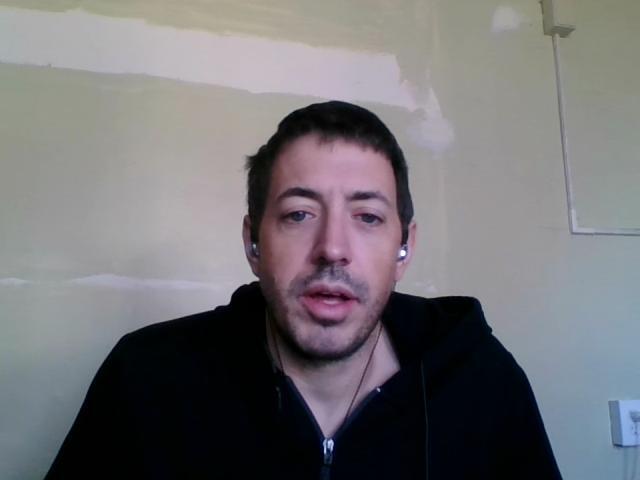} &
        \includegraphics[width=0.24\linewidth]{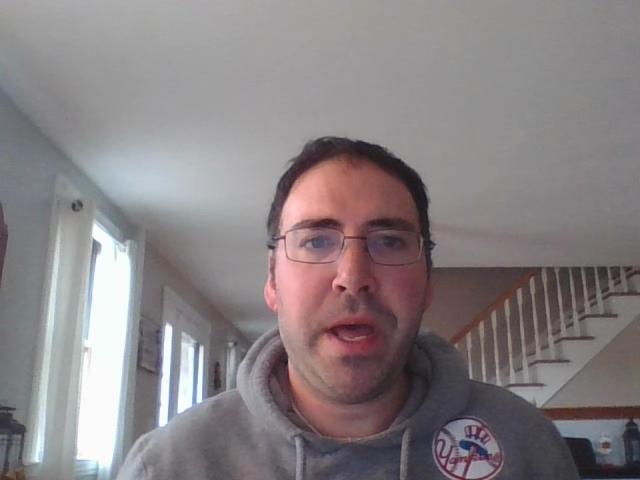}
    \end{tabular}}
    \caption{Additional examples of matched identities from the first release of DeepSpeak data (v2.x). Part 2/2.}
    \label{fig:extra_pairs_v2_b}
\end{figure}

\newpage

\section{Experimental Setup: Data Preprocessing}
\label{app:exp_setup_datapreprocessing}

\subsection{Video Deepfake Detection}

\subsubsection{FreqNet}

\textbf{Original Dataset.} A subset of the GAN-based deepfake dataset compiled by FreqNet's authors, downloaded using the official script (\url{https://github.com/chuangchuangtan/FreqNet-DeepfakeDetection/blob/main/download_dataset.sh}), was used for evaluation. This subset comprised AttGAN, RelGAN, and S3GAN samples.

\textbf{DeepSpeak.} The DeepSpeak videos were split into frames and analyzed individually, consistent with FreqNet's pre-processing.

\subsubsection{GenConViT ED/VAE}

\textbf{Original Dataset.} CelebDF-2, downloaded from its Kaggle clone (\url{https://www.kaggle.com/datasets/reubensuju/celeb-df-v2}), was used for evaluation.

\textbf{DeepSpeak.} The DeepSpeak videos were cropped to the protagonist's face, consistent with GenConViT's pre-processing, as described in the paper. Since the authors did not release a pre-processing script as a part of their official code release, we implemented this script according to its description in the paper. This script is available at \url{anonymized}. 

\subsubsection{LipFD}

\textbf{Original Dataset.} AVLips, downloaded from \url{https://github.com/AaronComo/LipFD?tab=readme-ov-file#-avlips-a-high-quality-audio-visual-dataset-for-lipsync-detection}, was used for evaluation.

\textbf{DeepSpeak.} The DeepSpeak videos were pre-processed into grids with a sequence of five crops of the protagonist's face at the bottom, with a matching spectrogram at the top. This pre-processing was performed using the official script released by LipFD's authors (\url{https://github.com/AaronComo/LipFD/blob/main/preprocess.py}).

\subsection{Audio Deepfake Detection}

\subsubsection{Raw Waveform Models}

\textbf{Original Dataset} AASIST was trained by its authors using the training subset of the ASVSpoof dataset\cite{jung2022aasist}. The authors provide two pretrained benchmark models (RawNet2 and RawGAT-ST) implemented as .py files associate with weights stored in .pth format, all generated using ASVSpoof training data~\url{https://github.com/clovaai/aasist}.

\textbf{DeepSpeak.} For DeepSpeak, the audios samples were used in their full duration as raw waveforms. These were extracted from the original .mp4 files, saved as .wav format, and resampled to 16kHz. 

\subsubsection{Embedding-based Models}

\textbf{TitaNet.} The original TitaNet + classifier model was trained on the TIMIT-ElevenLabs dataset. While the exact embedding extraction and classifier building code are not publicly available, we re-implemented the approach as described in the original work and retrained it on the same dataset. Embeddings were extracted from full-duration audio waveforms, resulting in  192-dimensional vectors per sample which were stored in CSV format and used as input to the downstream classifiers. 

\textbf{Wav2Vec2-xlsr.} For ``pretrained'' models, we trained the Wav2Vec2-xlsr classifier on embeddings extracted from ASVSPoof audios, as we did not find any widely used or well-documented pretrained implementations available. Embeddings were extracted from full-duration audio waveforms, resulting in 1024-dimensional vectors per sample which were stored in CSV format and used as input to the downstream classifiers. Embeddings from the DeepSpeak data were extracted in the same way. 

\textbf{LAION-CLAP.} We trained the LAION-CLAP classifier on ASVSPoof, following the same procedure as used for the Wav2Ve2-xlsr embeddings. Embeddings were extracted from full-duration audio waveforms, resulting in 512-dimensional vectors per sample which were stored in CSV format and used as input to the downstream classifiers. Embeddings from the DeepSpeak data were extracted in the same way.

\section{Experimental Setup: Hyperparameters}
\label{app:exp_setup_hyperparameters}

\subsection{Video Deepfake Detection}

Default hyperparameters values were used when possible. These were taken from the official code repositories: \url{https://github.com/chuangchuangtan/FreqNet-DeepfakeDetection} for FreqNet, \url{https://github.com/erprogs/GenConViT} for GenConViT, and \url{https://github.com/AaronComo/LipFD} for LipFD. Exceptions to default hyperparameter values are marked with $*$. These include the learning rate and weight decay parameters, which sometimes had to be adapted for the models to learn and converge successfully (a small search over neighboring magnitudes of the default value was performed). We also had to change the batch size to accommodate our compute resources.

\begin{table}[h!]
\centering
\begin{tabular}{l|C{1.2cm}C{1.2cm}C{1.2cm}|C{1.2cm}C{1.2cm}C{1.2cm}}
\hline \noalign{\vskip 0.5ex}
\multirow{2}{*}{\textbf{Parameter}} 
& \multicolumn{3}{c|}{\textbf{Full Training}} 
& \multicolumn{3}{c}{\textbf{Fine-tuning}} \\
& {FreqNet} & {GenConViT} & {LipFD} 
& {FreqNet} & {GenConViT} & {LipFD} \\
\hline
Training epochs                 & 10      & 10      & 10      & 10      & 10      & 10 \\
Batch size                      & 32      & 16      & 10      & 32      & 16      & 10 \\
Frame size                      & 256x256 & 224x224 & 500x200 & 256x256 & 224x224 & 500x200 \\
Optimizer                       & Adam    & Adam    & Adam    & Adam    & Adam    & Adam \\
Learning rate (LR)              & 1e-4    & 1e-4    & 1e-5    & 1e-4    & 1e-4    & 1e-5 \\
LR scheduler gamma              & N/A     & 1e-1    & N/A     & N/A     & 1e-1    & N/A \\
LR scheduler step size          & N/A     & 15      & N/A     & N/A     & 15      & N/A \\
Weight decay                    & N/A     & 1e-4    & 1e-4    & N/A     & 1e-4    & 1e-4 \\ \hline
\end{tabular}
\vspace{0.25cm}
\caption{Hyperparameters used for full training and fine-tuning of video deepfake detection models.}
\label{tab:video_hyperparameters}
\end{table}

\subsection{Audio Deepfake Detection}

Details of the hyperparameters used in training the raw waveform models are included in the next subsection. For the embedding-based models, data was subset on a 80/20\% training/testing split, with training data balanced through downsampling the larger class to match the number of audios in the smaller class (in all cases, the ``real'' class was larger than the ``fake'' class due to the fact that only lip-sync deepfakes in DeepSpeak contain fake audio). However, all testing data was used for inference. 

Logistic regression and random forest classifiers were used as the linear and non-linear classification models for real and fake. The following parameters were used by default for both models across all datasets:

\begin{enumerate}
    \item LogisticRegression: max iterations = 1000, with all other parameters as per the defaults in the Scikit-learn LogisticRegression model.
    \item RandomForestClassifier: number of estimators = 100, with all other parameters as per the defaults in the Scikit-learn RandomForestClassifier model.
\end{enumerate}

\newpage
\subsection{Audio Raw Waveform Pretrained Model Configurations}
\lstdefinelanguage{json}{
    basicstyle=\ttfamily\footnotesize,
    breaklines=true,
    frame=single,
    backgroundcolor=\color{gray!5},
    showstringspaces=false,
    literate=
     *{0}{{{\color{blue}0}}}{1}
      {1}{{{\color{blue}1}}}{1}
      {2}{{{\color{blue}2}}}{1}
      {3}{{{\color{blue}3}}}{1}
      {4}{{{\color{blue}4}}}{1}
      {5}{{{\color{blue}5}}}{1}
      {6}{{{\color{blue}6}}}{1}
      {7}{{{\color{blue}7}}}{1}
      {8}{{{\color{blue}8}}}{1}
      {9}{{{\color{blue}9}}}{1}
      {:}{{{\color{black}:}}}{1}
      {,}{{{\color{black},}}}{1}
      {"}{{{\color{red}"}}}{1}
}

\begin{lstlisting}[language=json, caption={AASIST Full Configuration}]
{
  "database_path": "./LA/",
  "asv_score_path": "ASVspoof2019_LA_asv_scores/ASVspoof2019.LA.asv.eval.gi.trl.scores.txt",
  "model_path": "./models/weights/AASIST.pth",
  "batch_size": 24,
  "num_epochs": 100,
  "loss": "CCE",
  "track": "LA",
  "eval_all_best": "True",
  "eval_output": "eval_scores_using_best_dev_model.txt",
  "cudnn_deterministic_toggle": "True",
  "cudnn_benchmark_toggle": "False",
  "model_config": {
    "architecture": "AASIST",
    "nb_samp": 64600,
    "first_conv": 128,
    "filts": [70, [1, 32], [32, 32], [32, 64], [64, 64]],
    "gat_dims": [64, 32],
    "pool_ratios": [0.5, 0.7, 0.5, 0.5],
    "temperatures": [2.0, 2.0, 100.0, 100.0]
  },
  "optim_config": {
    "optimizer": "adam",
    "amsgrad": "False",
    "base_lr": 0.0001,
    "lr_min": 0.000005,
    "betas": [0.9, 0.999],
    "weight_decay": 0.0001,
    "scheduler": "cosine"
  }
}
\end{lstlisting}

\newpage

\begin{lstlisting}[language=json, caption={RawNet2Spoof Full Configuration}]
{
  "database_path": "./LA/",
  "asv_score_path": "ASVspoof2019_LA_asv_scores/ASVspoof2019.LA.asv.eval.gi.trl.scores.txt",
  "model_path": "/home1/irteam/jeeweon/git/AsvSpoofDetection/exp_result/LAmodelRawNet2Spoof_ep100_bs32_lr0.0001/weights/best.pth",
  "batch_size": 32,
  "lr": 0.0001,
  "weight_decay": 0.0001,
  "num_epochs": 100,
  "loss": "CCE",
  "track": "LA",
  "eval_output": "eval_scores_using_best_dev_model.txt",
  "cudnn_deterministic_toggle": "True",
  "cudnn_benchmark_toggle": "False",
  "model_config": {
    "architecture": "RawNet2Spoof",
    "nb_samp": 64600,
    "first_conv": 1024,
    "in_channels": 1,
    "filts": [20, [20, 20], [20, 128], [128, 128]],
    "blocks": [2, 4],
    "nb_fc_node": 1024,
    "gru_node": 1024,
    "nb_gru_layer": 3,
    "nb_classes": 2
  },
  "optim_config": {
    "optimizer": "adam",
    "amsgrad": "False",
    "base_lr": 0.0001,
    "lr_min": 0.000005,
    "betas": [0.9, 0.999],
    "weight_decay": 0.0001,
    "scheduler": "cosine"
  }
}
\end{lstlisting}

\newpage
\begin{lstlisting}[language=json, caption={RawNetGatSpoofST Full Configuration}]
{
  "database_path": "./LA/",
  "asv_score_path": "ASVspoof2019_LA_asv_scores/ASVspoof2019.LA.asv.eval.gi.trl.scores.txt",
  "model_path": "/home1/irteam/jeeweon/git/AsvSpoofDetection/exp_result/LAmodelRawNetGatSpoofST_ep100_bs24_lr0.0001/weights/epoch_12.pth",
  "batch_size": 24,
  "num_epochs": 100,
  "loss": "CCE",
  "track": "LA",
  "eval_output": "eval_scores_using_best_dev_model.txt",
  "cudnn_deterministic_toggle": "True",
  "cudnn_benchmark_toggle": "False",
  "model_config": {
    "architecture": "RawNetGatSpoofST",
    "nb_samp": 64600,
    "first_conv": 128,
    "filts": [70, [1, 32], [32, 32], [32, 64], [64, 64]]
  },
  "optim_config": {
    "optimizer": "adam",
    "amsgrad": "False",
    "base_lr": 0.0001,
    "lr_min": 0.000005,
    "betas": [0.9, 0.999],
    "weight_decay": 0.0001,
    "scheduler": "cosine"
  }
}
\end{lstlisting}

\subsection{Audio Raw Waveform Model Configurations for Training on DeepSpeak}

The model configurations used for training AASIST, RawNet2 and RawGAT-ST on DeepSpeak data were the same as those for the original pretrained models created by the AASIST authors (using ASVSpoof) as detailed in the previous subsection, except with epochs reduced from 100 to 50 for training efficiency. 


\section{Safeguards against Misuse}
\label{app:safeguards}

While deepfake detection technology is broadly beneficial for enhancing the security and safety of individuals, organizations, and societies against harms such as scams, fraud, and disinformation, we acknowledge the potential for misuse of the dataset and outline our safeguards against this below.

We make the data available under license via Hugging Face, where metadata and documentation are publicly visible. However, access to the data itself is restricted: users must request access and briefly describe their intended use. Only projects that aim to improve defenses against deepfakes or support reproducibility studies will be granted access.

By way of safeguards for participants in the data collection, we explicitly obtained consent from all users and informed them that their recordings --- but no other personally identifiable information (PII) --- would be included in a publicly available dataset. Participants were instructed not to include other individuals in their recordings. While we report an overview of the participants in this paper, we do not release individual-level demographic information.

\section{Licensing}
\label{app:licensing}

This section lists licensing terms for all assets employed in this work at the time of submission (May 2025). We refer the reader to the respective publications or code repositories for the most up-to-date licensing terms.

\subsection{DeepSpeak Dataset}

This is a custom asset associated with this paper. Licensing is provided to qualifying academic institutions at no cost under licensing terms available at \url{}. Deepfakes included in DeepSpeak were generated with the following third-party assets (deepfake engines).

\textbf{FaceFusion} is provided under the OpenRAIL-AS license.

\textbf{INSwapper}'s code repository does not specify a license.

\textbf{CodeFormer} is provided under a custom license posted at \url{https://github.com/sczhou/CodeFormer?tab=License-1-ov-file}.

\textbf{SimSwap} is provided under the Creative Commons Attribution-NonCommercial 4.0 International license.

\textbf{RestoreFormer} is provided under the Apache v2.0 license.

\textbf{Wav2Lip}'s code repository does not specify a license.

\textbf{VideoRetalking} is provided under the Apache v2.0 license.

\textbf{Diff2Lip} is provided under the Creative Commons Attribution-NonCommercial 4.0 International license.

\textbf{LatentSync} is provided under the Apache v2.0 license.

\textbf{LivePortrait} is provided under the MIT license.

\textbf{HelloMeme} is provided under the MIT license.

\textbf{Memo} is provided under the Apache v2.0 license.

\subsection{Video Deepfake Detection Experiments}

The following are third-party assets (model code and datasets) used for the video deepfake detection experiments.

\textbf{FreqNet}'s code repository does not specify a license.

\textbf{GenConViT} is provided under the GNU General Public License v3.0.

\textbf{LipFD}'s code repository does not specify a license.

\textbf{Celeb-DF 2}'s data repository does not specify a license.

\textbf{AVLips}'s data repository does not specify a license.

\subsection{Audio Deepfake Detection Experiments}

Both ElevenLabs and PlayAI agreed to grant our team complimentary research access to their commercial voice cloning and Text-to-Speech APIs. For Speechify, the paid Premium commercial tier was used. 

\textbf{ElevenLabs}'s website specifies a commercial license is granted for use of audio created with the API, in accordance with their Terms of Service (~\url{https://help.elevenlabs.io/hc/en-us/articles/13313564601361-Can-I-publish-the-content-I-generate-on-the-platform}). 

\textbf{PlayAI}'s Terms of Service grant users, solely for commercial use, all right, title and interest in and to content generated by the Service based on your User Content (``Output''), subject to any Third Party Terms which may apply to such Output (~
\url{https://play.ai/terms}).

\textbf{Speechify}'s Terms of Service specify a commercial license is granted for audios generated using the paid subscription tier (~\url{https://speechify.com/studio-terms/?srsltid=AfmBOopH_aautZGkvdGqxbBAAld-shNU94UHq8v-xu0wUz1coqYaHJqJ}). 

\section{Compute Resources}
\label{app:compute}

We conducted our experiments on single-node NVIDIA A100 GPU machines. All reported runtimes correspond to this setup, assuming no competing processes. In total, producing the deepfakes and conducting baseline experiments required approximately 1,700 hours of GPU time, excluding environment setup or troubleshooting.

\subsection{DeepSpeak Dataset}

Producing the DeepSpeak video deepfakes required approximately eight weeks of cumulative GPU time, including quality validation and filtering. This process was distributed across multiple GPU machines, with each machine generating a subset of the dataset.

Producing the DeepSpeak audio deepfakes involved approximately one week of GPU time, including audio preprocessing, voice cloning, and Text-To-Speech generation using relevant API calls. The entire process was executed on a single GPU machine.  

\subsection{Video Deepfake Detection Experiments}

\textbf{FreqNet} Training or fine-tuning for $10$ epochs on DeepSpeak takes approximately ten hours. Evaluating the resulting model on testing sets of DeepSpeak and the dataset of GAN-generated deepfakes produces by FreqNet's authors takes less than $15$ minutes. Given one model training, one model fine-tuning, and three evaluation passes, experiments with this architecture required approximately $21$ hours of GPU time.

\textbf{GenConViT} Training or fine-tuning for $10$ epochs on DeepSpeak takes approximately seven hours. Evaluating the resulting model on testing sets of DeepSpeak and Deleb-DF 2 takes less than $15$ minutes. Given two model trainings, two model fine-tunings, and five evaluation passes, experiments with this architecture required approximately $29$ hours of GPU time.

\textbf{LipFD} Training or fine-tuning for $10$ epochs on DeepSpeak takes approximately $30$ hours. Evaluating the resulting model on testing sets of DeepSpeak and AVLips takes under one hour. Given one model training, one model fine-tuning, and three evaluation passes, experiments with this architecture required approximately $63$ hours of GPU time.

\subsection{Audio Deepfake Detection Experiments}

\textbf{Raw waveform models} Training each of AASIST, RawGat-ST and RawNet2 on DeepSpeak data from scratch takes approximately $8$ hours. Evaluating the results each model for inference on DeepSpeak takes under two hours. 

\textbf{Embedding-based models} Embedding generation takes approximately 4 hours per model (TitaNet-L, Wav2Vec2-xlsr and LAION-CLAP) on DeepSpeak training data and approximately 6 hours per model on ASVSpoof training data. Classifier training (logistic regression and random forest) take less than one hour per embedding type on both ASVSpoof and DeepSpeak training data. Inference for both models per testing set of each dataset takes less than one hour. 


\newpage
\section{Survey Materials}
\label{app:survey-materials}

\vspace{-1em} 
\begin{figure}[H]
    \centering
    \resizebox{0.7\textwidth}{!}{
    \fbox{\includegraphics[width=\textwidth]{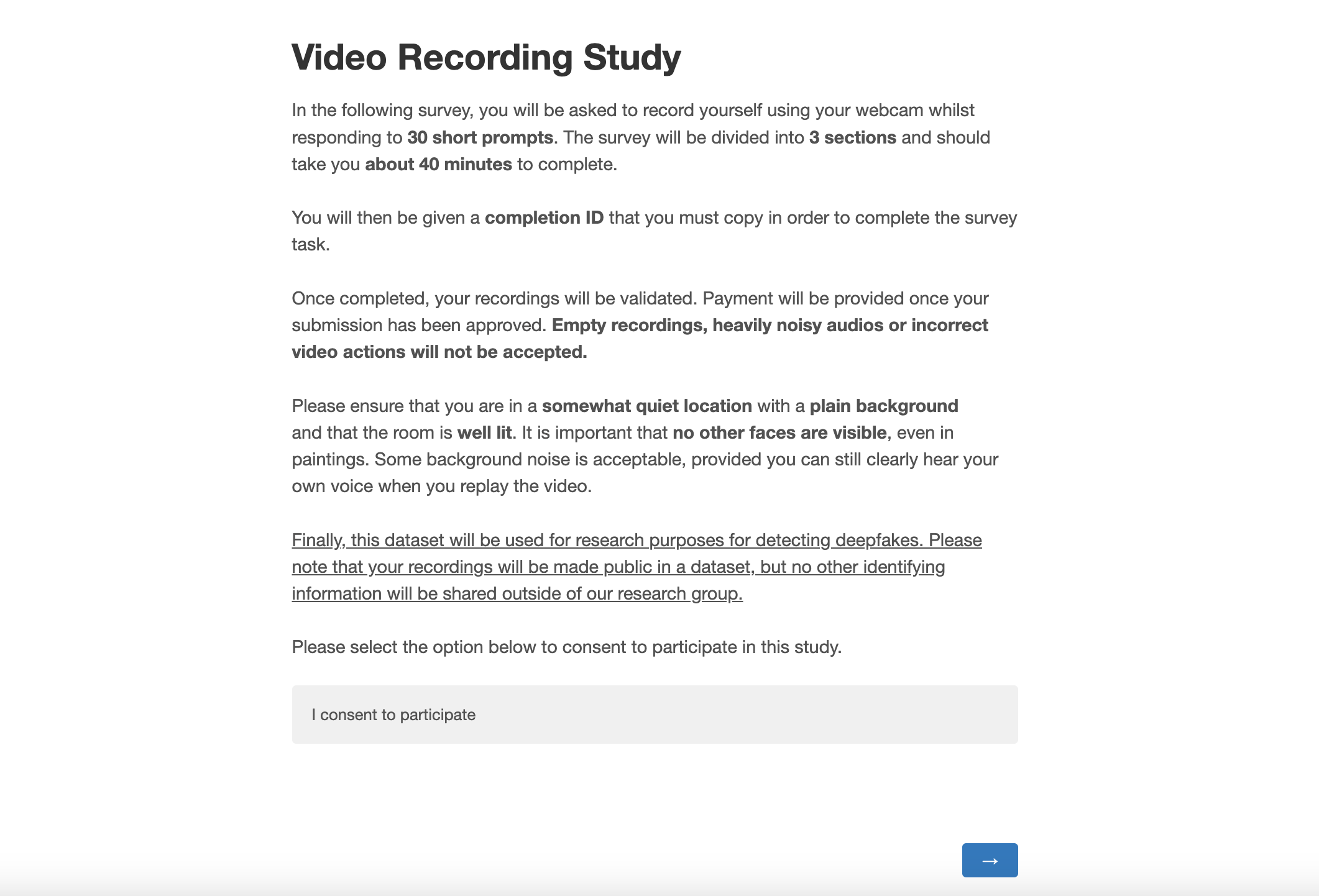}}
    }
    \caption{Screenshot of the introduction and consent page of the data collection study.}
    \label{fig:survey_intro_consent}
\end{figure}

\begin{figure}[H]
    \centering
    \resizebox{0.7\textwidth}{!}{
    \fbox{\includegraphics[width=\textwidth]{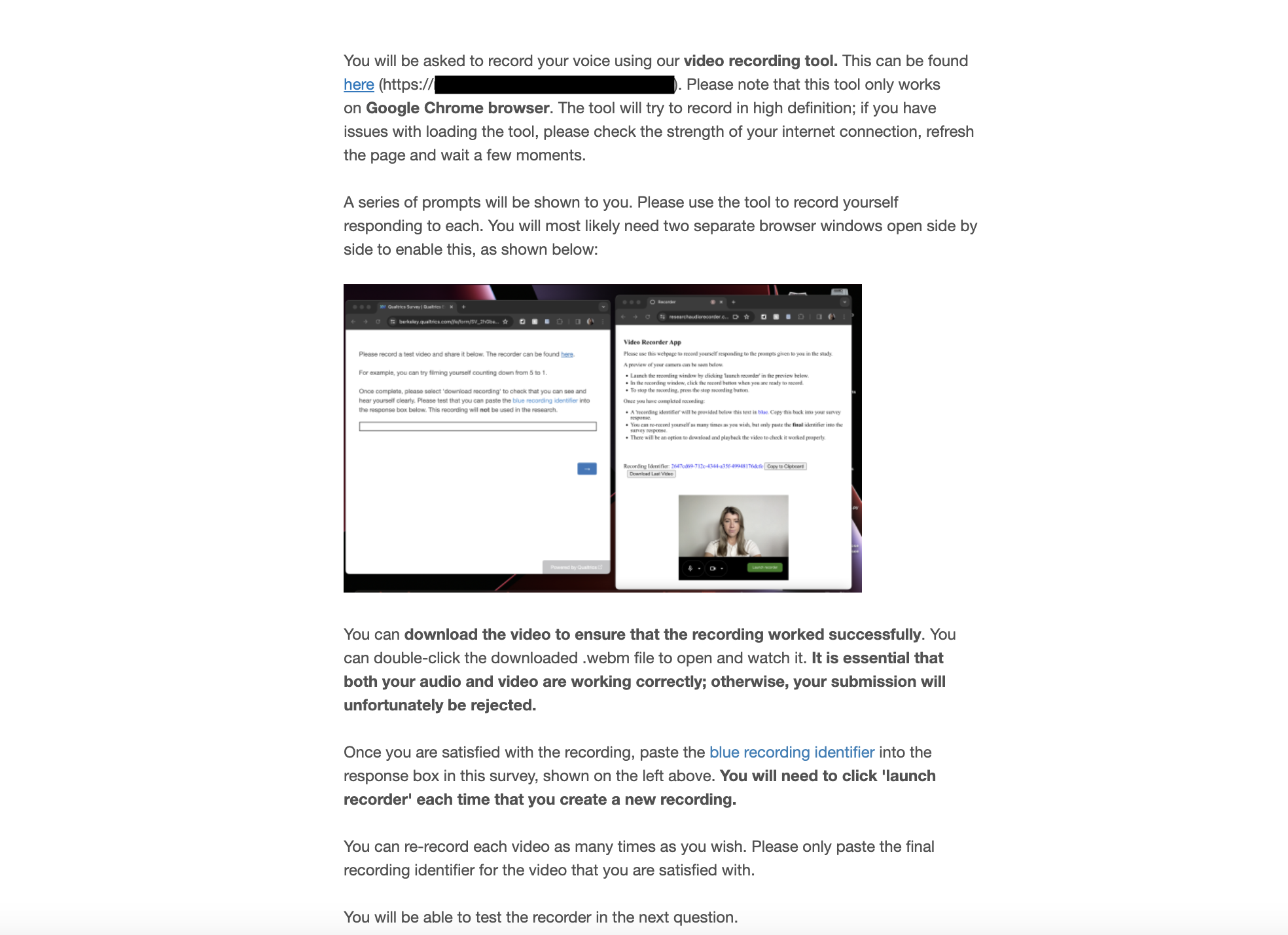}}
    }
    \caption{Screenshot of the overview and recording instructions page of the data collection study.}
    \label{fig:survey_intro_and_recording}
\end{figure}

\begin{figure}[H]
    \centering
    \resizebox{0.7\textwidth}{!}{
    \fbox{\includegraphics[width=\textwidth]{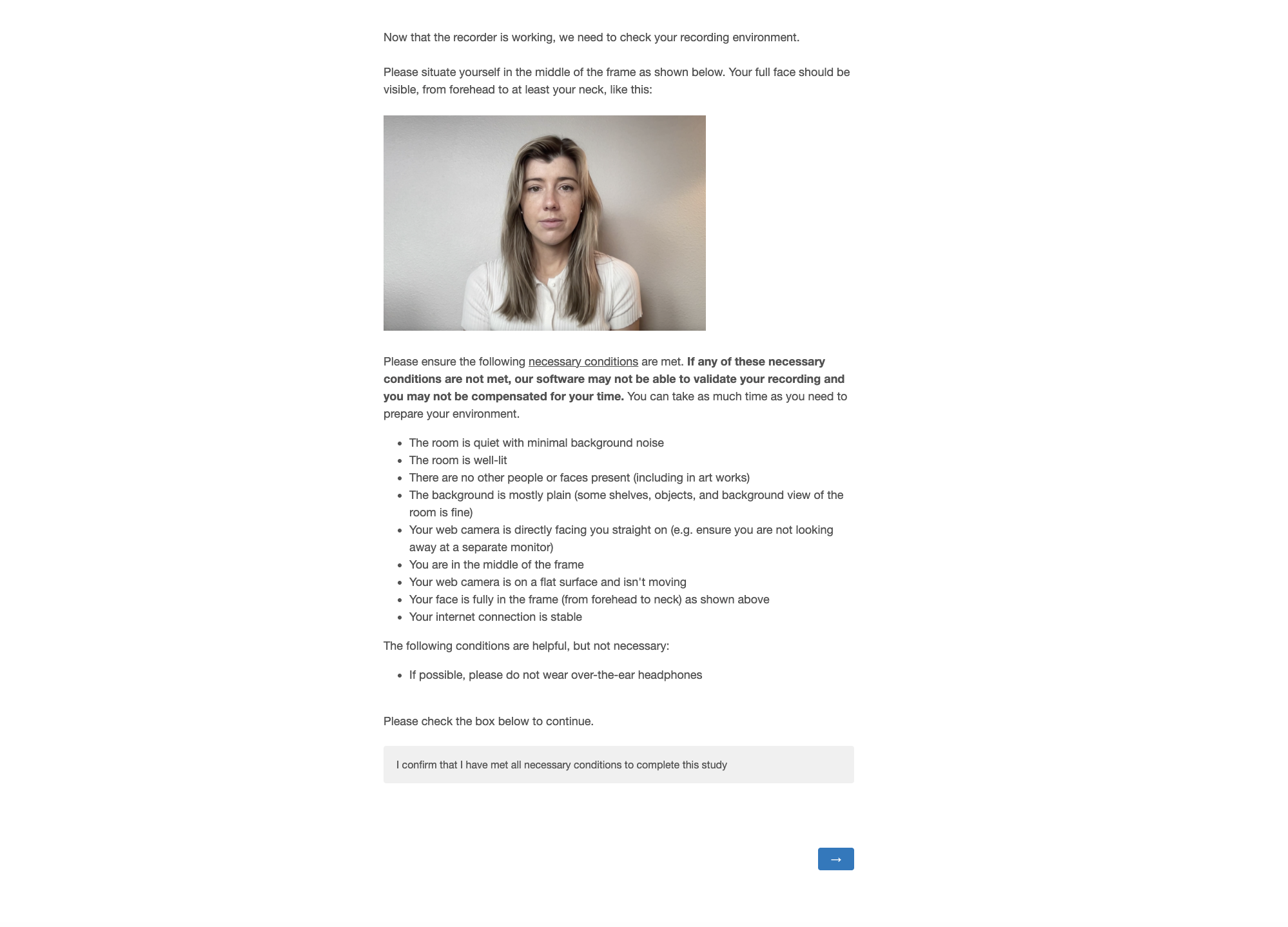}}
    }
    \caption{Screenshot of the environment checks page of the data collection study.}
    \label{fig:survey_environment_test}
\end{figure}

\vspace{-1em} 
\begin{figure}[H]
    \centering
    \resizebox{0.7\textwidth}{!}{
    \fbox{\includegraphics[width=\textwidth]{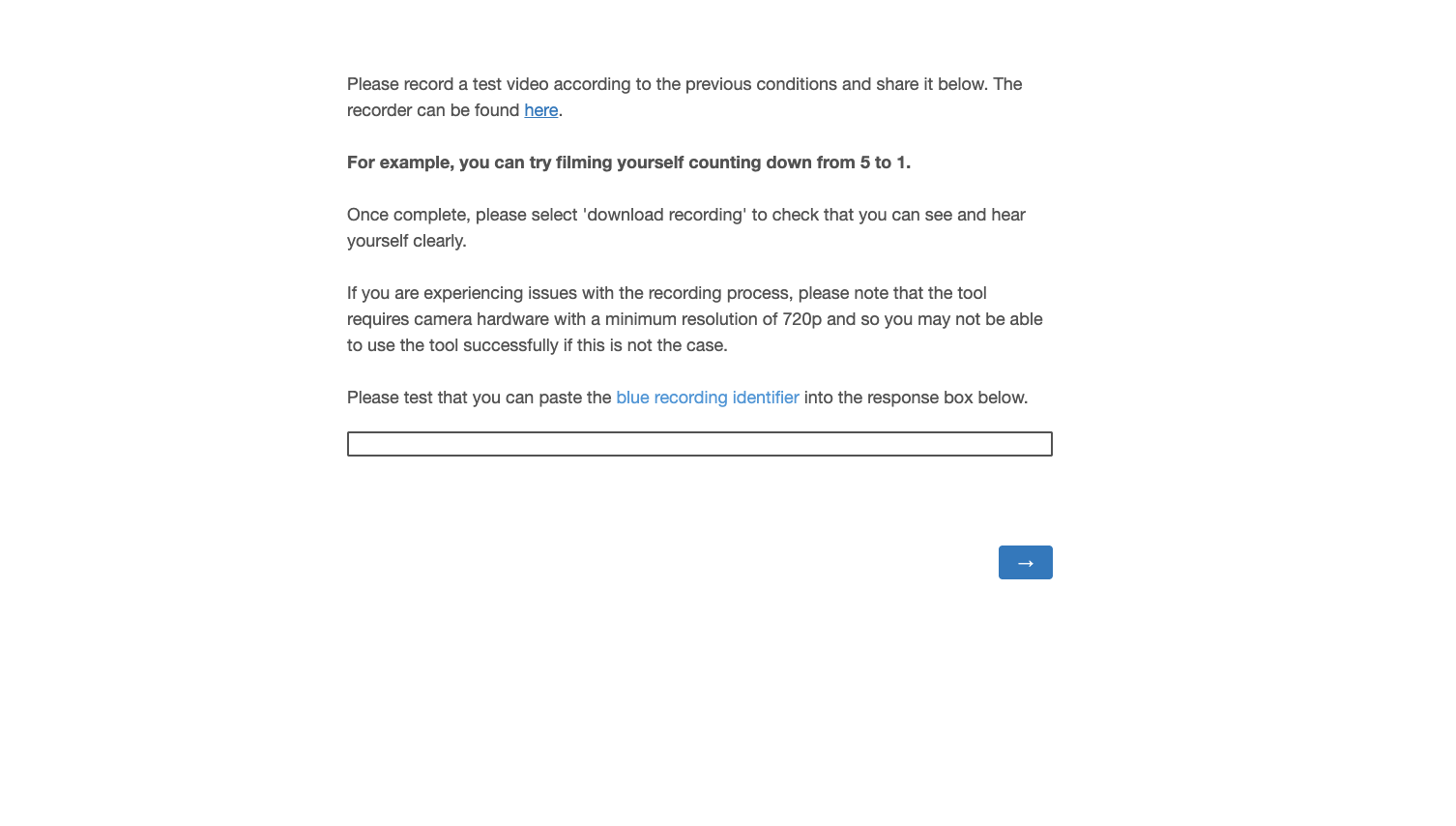}}
    }
    \caption{Screenshot of the recording test page of the data collection study.}
    \label{fig:survey_recording_test}
\end{figure}

\begin{figure}[H]
    \centering
    \resizebox{0.7\textwidth}{!}{
    \fbox{\includegraphics[width=\textwidth]{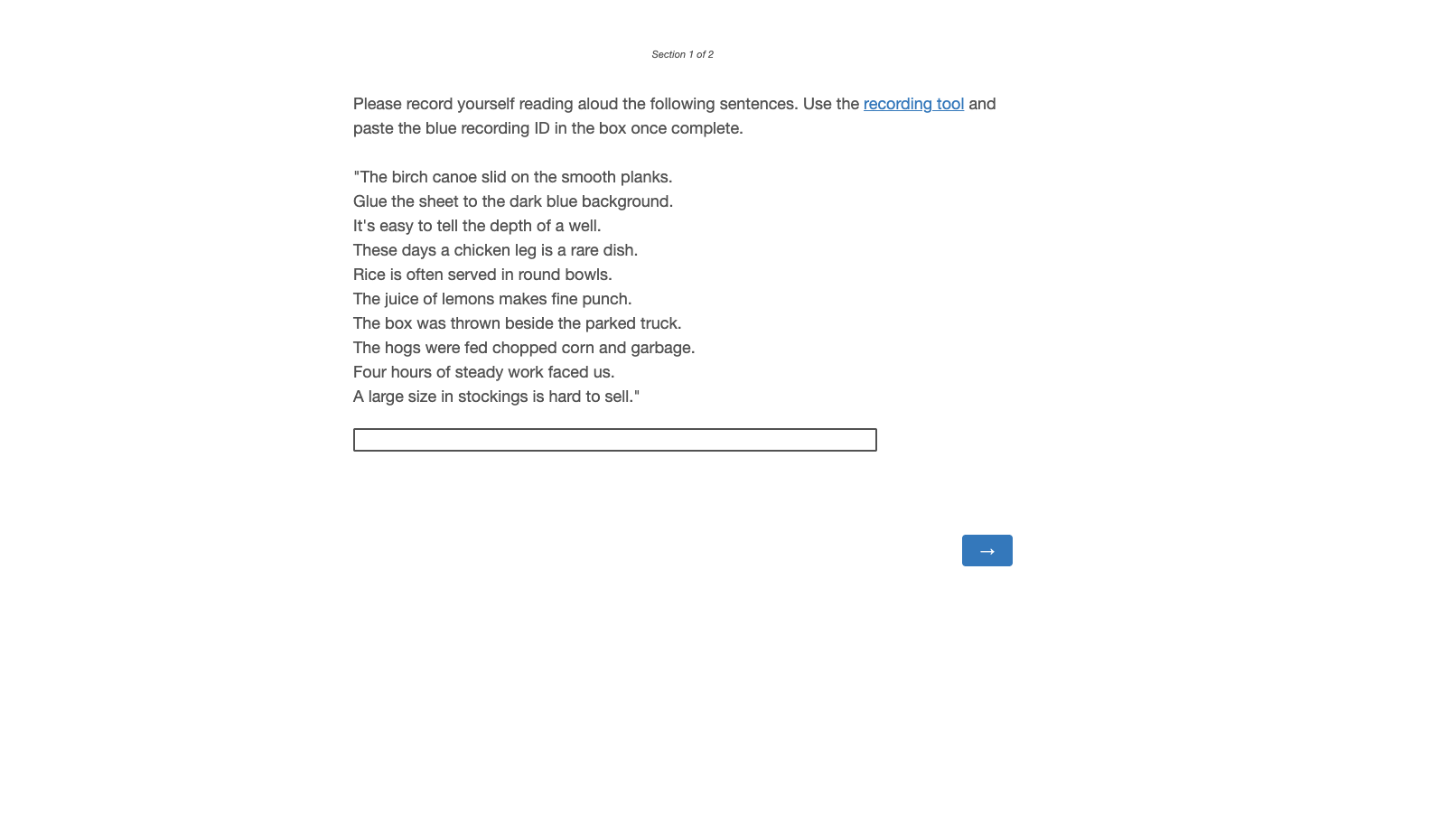}}
    }
    \caption{Screenshot of an example prompt from the data collection study.}
    \label{fig:survey_example_prompt}
\end{figure}

\newpage

\section{Prompts}
\label{app:prompts}

\subsection{Voice Cloning Prompts}
\begin{table}[h!]
    \centering
    \caption{Voice cloning input prompts, consisting of ten consecutive sentences and one continuous paragraph.}
    \footnotesize
    \begin{tabular}{|p{0.9\textwidth}|}
        \hline
        The birch canoe slid on the smooth planks. \\
        Glue the sheet to the dark blue background. \\
        It's easy to tell the depth of a well. \\
        These days a chicken leg is a rare dish. \\
        Rice is often served in round bowls. \\
        The juice of lemons makes fine punch. \\
        The box was thrown beside the parked truck. \\
        The hogs were fed chopped corn and garbage. \\
        Four hours of steady work faced us. \\
        A large size in stockings is hard to sell. \\
        \hline
        Please call Stella. Ask her to bring these things with her from the store: Six spoons of fresh snow peas, five thick slabs of blue cheese, and maybe a snack for her brother Bob. We also need a small plastic snake and a big toy frog for the kids. She can scoop these things into three red bags, and we will go meet her Wednesday at the train station. \\
        \hline
    \end{tabular}
    \vspace{0.25cm}
    \label{tab:voice_cloning_prompts}
\end{table}

\subsection{Standardized Scripted Prompts}
\begin{table}[h!]
    \centering
    \caption{Standardized scripted prompts, consisting of ten separate sentences.}
    \footnotesize
    \begin{tabular}{|p{0.9\textwidth}|}
        \hline
        We apply auditory modeling to computer speech recognition. \\
        \hline
        He sank back sighing and was soon asleep again. \\
        \hline
        The mango and the papaya are in a bowl. \\
        \hline
        Will you tell me why. \\
        \hline
        That experience holds a lesson for us all in regard to birth control today. \\
        \hline
        That is what childhood is he told himself. \\
        \hline
        As a rule part time farmers hire little help. \\
        \hline
        The new birth is miraculous and mysterious. \\
        \hline
        The fear of punishment just didn't bother him. \\
        \hline
        The figure in the corner belched loudly a deep liquid eruption. \\
        \hline
    \end{tabular}
    \vspace{0.25cm}
    \label{tab:standardized_prompts}
\end{table}


\subsection{Unscripted Prompts}
\begin{table}[h!]
    \centering
    \caption{Unscripted prompts.}
    \footnotesize
    \begin{tabular}{|p{0.9\textwidth}|}
        \hline
        \textbf{Unscripted Prompts} \\
        \hline
        What did you do yesterday? \\
        \hline
        Please describe an object that you can see from your current position. \\
        \hline
        What is your favorite type of music and why? \\
        \hline
        Please describe your ideal way to spend a weekend. \\
        \hline
    \end{tabular}
    \vspace{0.25cm}
    \label{tab:unscripted_prompts}
\end{table}

%
%


\newpage

\subsection{Video Action Prompts}
\begin{table}[h!]
    \centering
    \caption{Video action prompts.}
    \footnotesize
    \begin{tabular}{|p{0.9\textwidth}|}
        \hline
        \textbf{Versions 1.0 and 2.0} \\
        \hline
        Lean forwards towards the camera and back to your starting position. \\
        \hline
        1. Look down and to the right, and slowly count out loud to three. \\
        2. Look down and to the middle, and slowly count out loud to three. \\
        3. Look down and to the left, and slowly count out loud to three.\\
        \hline
        1. Look up and to the left, and slowly count out loud to three. \\
        2. Look up and to the middle, and slowly count out loud to three. \\
        3. Look up and to the right, and slowly count out loud to three. \\
        \hline
        Wave your hand back and forth across your face four times while counting out loud. \\
        \hline
        Read each question aloud, followed immediately by your answer. \\
        1. What is my favorite food? \\
        2. What is my favorite movie? \\
        3. What did I have for breakfast? \\
        4. Where would I most like to travel and why? \\
        5. If I could time travel, where in the past or future would I like to go? \\
        \hline
        \textbf{Version 2.0 only} \\
        \hline
        Pretend to yawn, perhaps covering your mouth with your hand. \\
        \hline
        Pretend to laugh for between 2 to 5 seconds. \\
        \hline
        Clap loudly three times, pausing for about 5 seconds between claps. Please make sure your clap makes a loud sound. \\
        \hline
    \end{tabular}
    \vspace{0.25cm}
    \label{tab:standardized_prompts}
\end{table}

\newpage
\subsection{Example frames from action prompts}

\begin{figure}[!h]
    \centering
    \begin{tabular}{c@{\hspace{0.25cm}}c@{\hspace{0.25cm}}c@{\hspace{0.25cm}}c@{\hspace{0.25 cm}}c}
        read & wave hand & turn head right/up & tilt head down & lean in \\
        \includegraphics[width=0.178\textwidth]{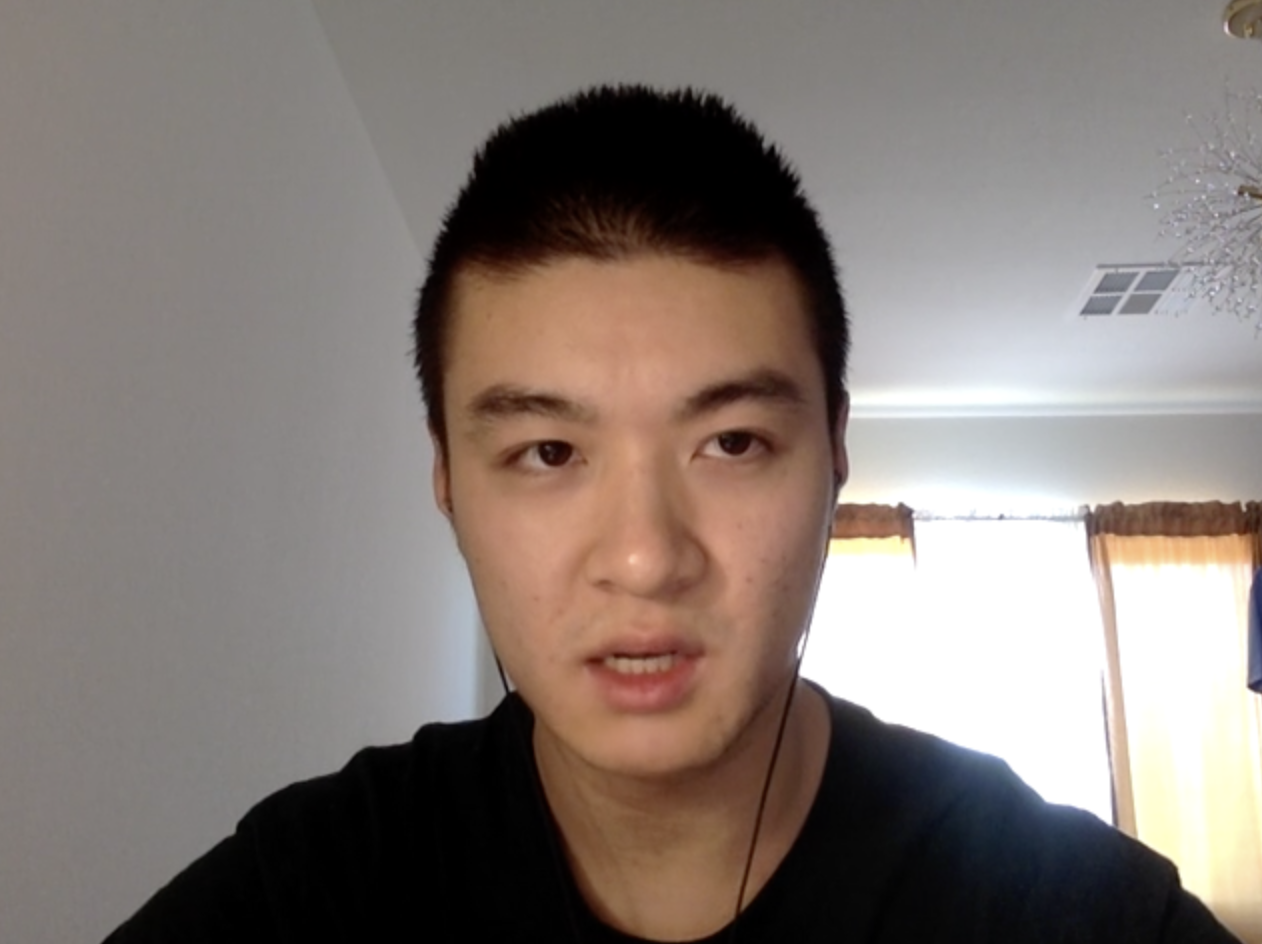} &
        \includegraphics[width=0.178\textwidth]{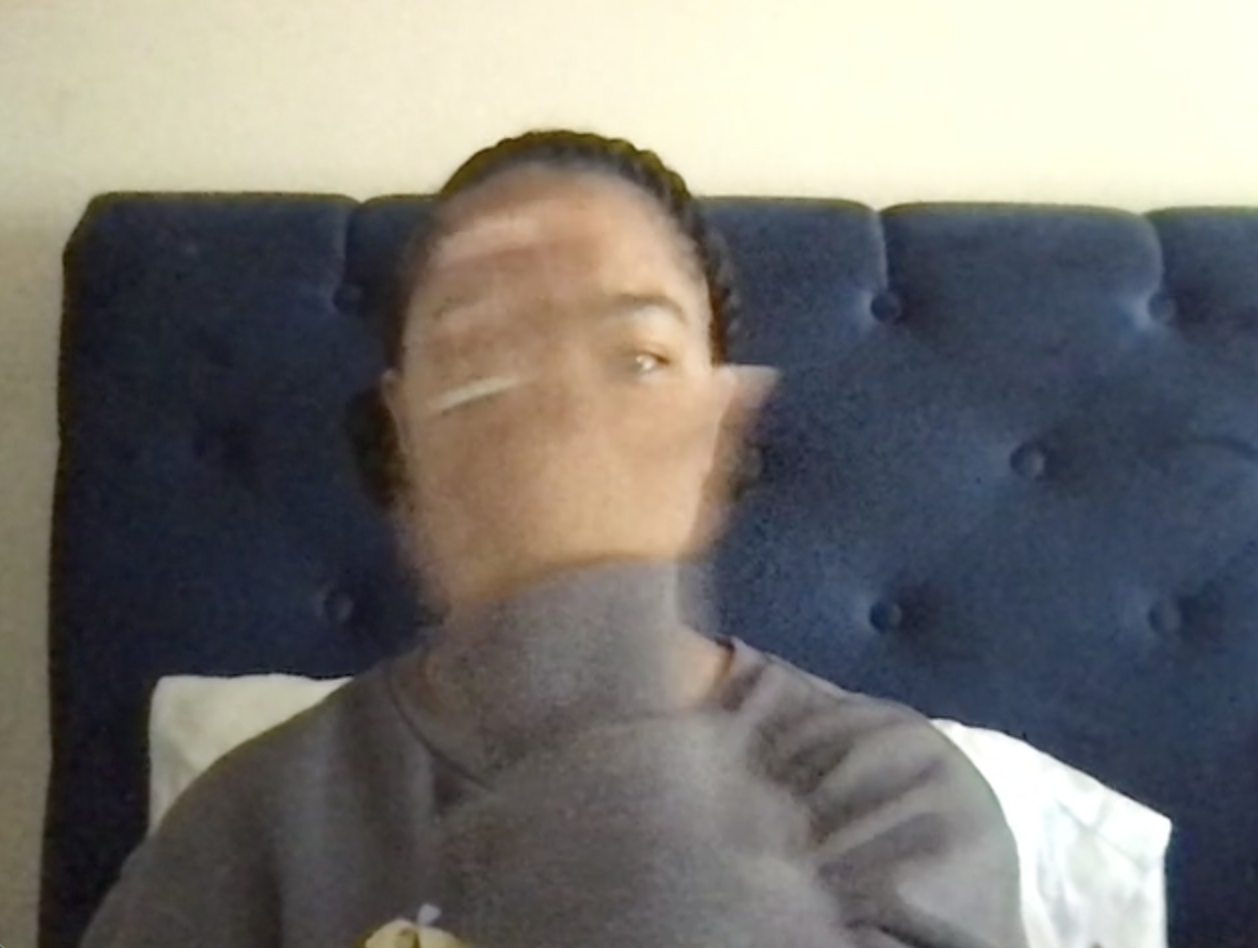} &
        \includegraphics[width=0.178\textwidth]{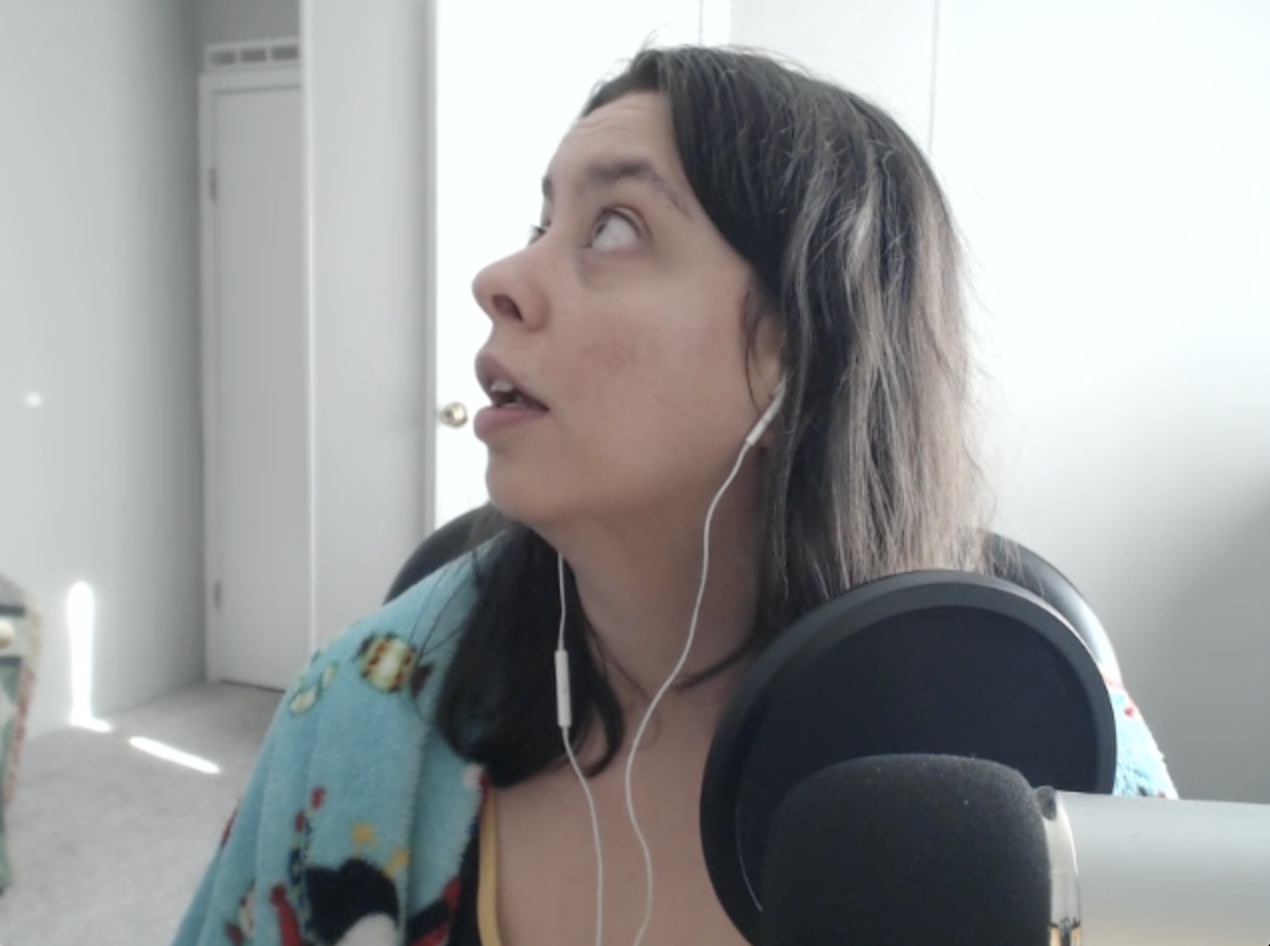} &
        \includegraphics[width=0.178\textwidth]{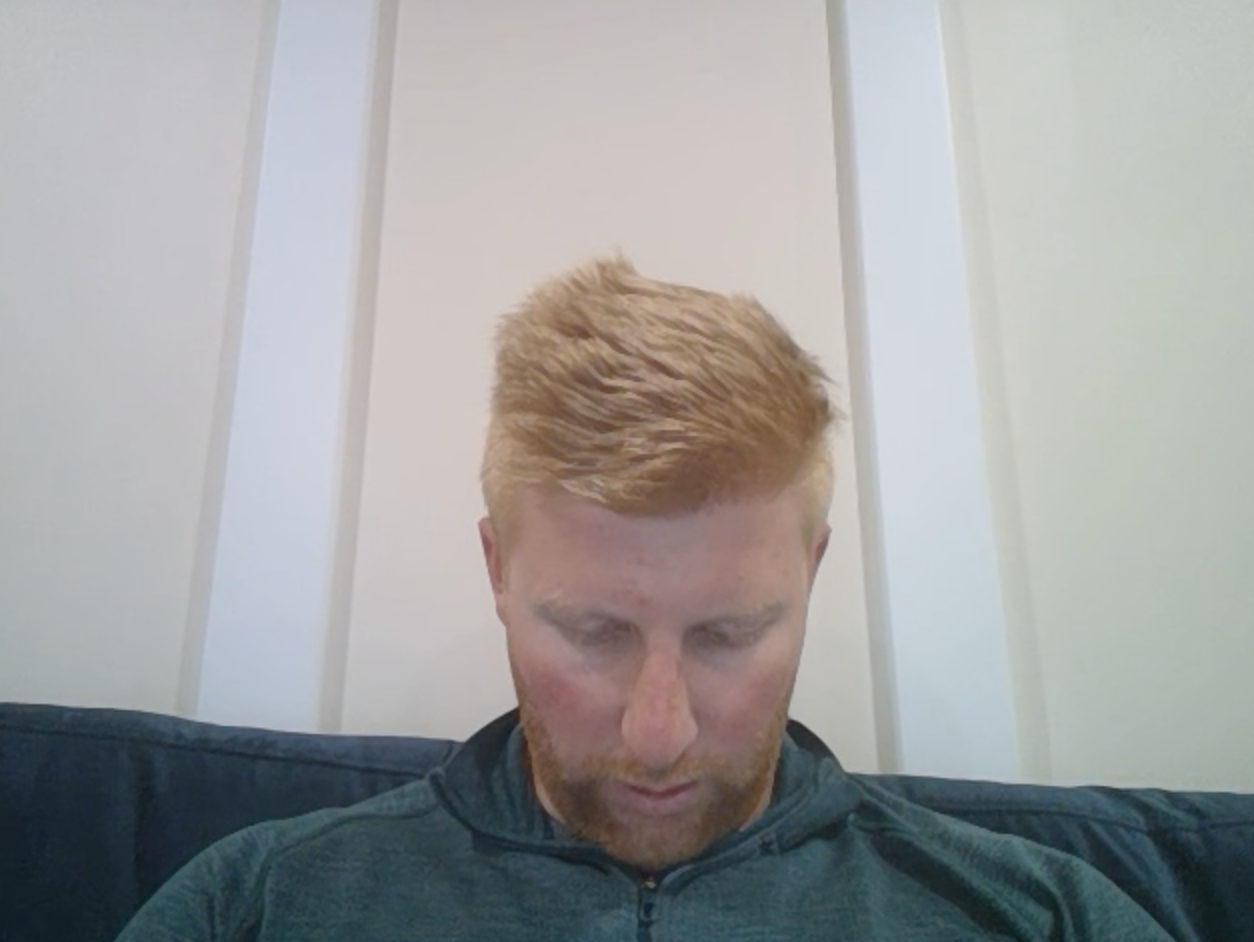} &
        \includegraphics[width=0.178\textwidth]{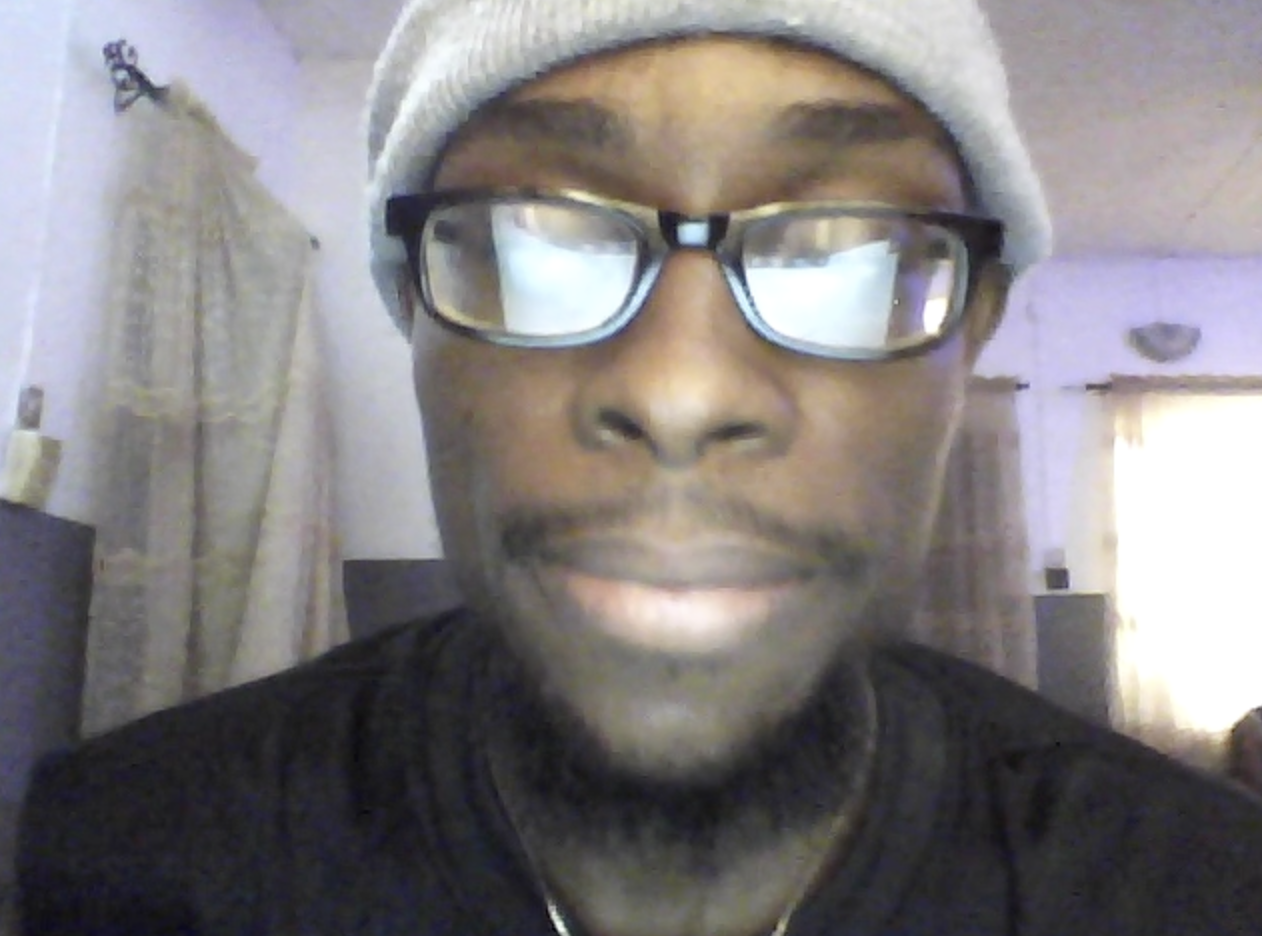}
   \end{tabular}
    \caption{Representative examples of different action prompt types.}
   \label{fig:five_figures}
\end{figure}


\newpage
\subsection{Randomized Scripted Prompts}
\label{tab:three_column_standardized_prompts}

\begin{footnotesize}
\begin{longtable}{|p{0.3\textwidth}|p{0.3\textwidth}|p{0.3\textwidth}|}
\hline
She had your dark suit in greasy wash water all year. & Her classical performance gained critical acclaim. & The diagnosis was discouraging however he was not overly worried. \\
\hline
Rich looked for spotted hyenas and jaguars on the safari. & The football team coach has a watch thin as a dime. & He found an empty bench opened a newspaper and stretched his legs before him. \\
\hline
The sermon emphasized the need for affirmative action. & You're so preoccupied that you've let your faith grow dim. & Gus saw pine trees and redwoods on his walk through sequoia national forest. \\
\hline
Approach your interview with statuesque composure. & The annoying raccoons slipped into phil's garden every night. & Some tore entirely through the whipsawed post oak. \\
\hline
Markets should become more competitive as consumers become more selective. & Pledge to participate in nevada's aquatic competition. & The advertising verse of plymouth variety store never changes. \\
\hline
The eastern coast is a place for pure pleasure and excitement. & She always could sense the shag end of a woolly day. & These planets were much bigger nearly all capable of holding an atmosphere. \\
\hline
He had not covered a hundred yards before a gun crashed from somewhere behind. & Between meetings he helps the president keep track of delegated matters. & A crab challenged me but a quick stab vanquished him. \\
\hline
Just why anybody should wish to start a riot the executive officer didn't know. & They used an aggressive policeman to flag thoughtless motorists. & The source is known so there is no necessity to remove insecticide residues. \\
\hline
A toothpaste tube should be squeezed from the bottom. & A precision transit is set up so that it is lined with respect to true north. & Two gas lamps were no more than a misleading glow. \\
\hline
Those who are not purists used canned vegetables when making stew. & Unless we send out the whole pie their pieces mean nothing. & The easygoing zoologist relaxed throughout the voyage. \\
\hline
His failure to open the store by eight cost him his job. & Smash lightbulbs and their cash value will diminish to nothing. & Etiquette mandates compliance with existing regulations. \\
\hline
Remember to allow identical twins to enter freely. & She always jokes about too much garlic in his food. & The government sought authorization of his citizenship. \\
\hline
Children can consume many fruit candies in one sitting. & Bob bandaged both wounds with the skill of a doctor. & It gives social guidance and direction and makes for programs of social action. \\
\hline
No signs of these no gross hemorrhage of lungs heart brain or stomach. & Herb's birthday occurs frequently on thanksgiving. & All about him stood tombstones his own sensitive great hands had fashioned. \\
\hline
With the spring rains the flow rose rapidly due to infiltration in open sewers. & Scholastic aptitude is judged by standardized tests. & We could barely see the fjords through the snow flurries. \\
\hline
A concurrent effort is needed to make oceanographic data useful on the spot. & This may be of overriding importance in considering military objectives. & A moth zig zagged along the path through otto's garden. \\
\hline
Buying a thoroughbred horse requires intuition and expertise. & His technique is ample and his musical ideas are projected beautifully. & Al received a joint appointment in the biology and the engineering departments. \\
\hline
The high security prison was surrounded by barbed wire. & Count the number of teaspoons of soysauce that you add. & While waiting for chipper she crisscrossed the square many times. \\
\hline
No girl would go this far to fool a man so she could kill him. & Northern liberals are the chief supporters of civil rights and of integration. & That stinging vapor was caused by chloride vaporization. \\
\hline
Whether historically a fact or not the legend has a certain symbolic value. & The library has open shelves even in the unbound periodical stockroom. & Steph could barely handle the psychological trauma. \\
\hline
In every major cloverleaf traffic sometimes gets backed up. & Davy mathews it's disgusting the way you're always eating. & As a precaution the outlaws bought gunpowder for their stronghold. \\
\hline
The entire length of the street could be raked with rifle fire from this barn. & Severe myopia contributed to ron's inferiority complex. & May i order a strawberry sundae after i eat dinner. \\
\hline
And never show my face or my truck around here again. & Any organism that falters or misperceives the signals or weakens is done. & Valley lodge yearly celebrates the first calf born. \\
\hline
Then again there's always that lovely old pastime of hooking or braiding rugs. & A young mouse scampered across the field and disappeared. & What possessed you to tell me a clotheshorse would be a good idea. \\
\hline
Employee layoffs coincided with the company's reorganization. & The antithesis of the ecumenical and the local then no longer exists. & The social and psychological consequences of this continue to affect the area. \\
\hline
Make a paste of brown sugar and mustard and spread lightly over scored surface. & In many of his poems death comes by train a strongly evocative visual image. & In most cases we recognize certain words persons animals or objects. \\
\hline
Cooperation along with understanding alleviate dispute. & Gregory and tom chose to watch cartoons in the afternoon. & Whoever cooperates in finding nan's cameo will be rewarded. \\
\hline
Such legislation was clarified and extended from time to time thereafter. & But none of this could soothe the exacerbated nerves. & He waited until they were inside the elevator and then said now what do we do. \\
\hline
Coffee is grown on steep jungle like slopes in temperate zones. & We like bleu cheese but victor prefers swiss cheese. & My desires are simple give me one informative paragraph on the subject. \\
\hline
Alice i tell you my feet are like chopped beefsteak. & Almost everybody in the senior class is married students say dogmatically. & They all agree that the essay is barely intelligible. \\
\hline
It moved in a silver arc toward his throat then veered downward. & His legs pumped furiously his long black hair streamed out behind him. & If we left one we'd have to wipe it for fingerprints. \\
\hline
The rich should invest in black zircons instead of stylish shoes. & And the surface is driven back in its very surfaceness only by this contrast. & Since a fall or blow might have caused it a cold pack was usually first aid. \\
\hline
Positive results start when it goes towards the hand you use to make your mark. & The preschooler couldn't verbalize her feelings about the emergency conditions. & Are we as safe as we should be from such a disaster. \\
\hline
Before deriving this formula we explain what we mean by problems of this kind. & My sincere wish is that he continues to add to this record he sets here today. & Neither his appetites his exacerbations nor his despair were kin to yours. \\
\hline
The giant redwoods shimmered in the glistening sun. & The cow wandered from the farmland and became lost. & We have become amateur insurance experts and fine feathered yard birds. \\
\hline
Privately he created and magnified an image of himself as a hired assassin. & Both loved the out of doors including mountain climbing and horseback riding. & And their chroniclers are not the dramatic poets but the prose novelists. \\
\hline
According to my interpretation of the problem two lines must be perpendicular. & In the course of its inquiry it took testimony from only seven witnesses. & Seamstresses attach zippers with a thimble needle and thread. \\
\hline
Although they drew light ground fire they saw no signs of activity. & Nonprofit organizations have frequent fund raisers. & Boys and men go along the riverbank or to the alcoves in the top arcade. \\
\hline
The response of reaction is dominated by a concern for what is vanishing. & The fat man has trouble buying life insurance or has to pay higher premiums. & On all sides doors were being slammed in his face. \\
\hline
Which long article was opaque and needed clarification. & To what extent such low density applies to micrometeorites is unknown. & The patient and the surgeon are both recuperating from the lengthy operation. \\
\hline
The haunted house was a hit due to outstanding audio visual effects. & Those answers will be straightforward if you think them through carefully first. & This was chiefly because of the bluish white autofluorescence from the cells. \\
\hline
Biologists use radioactive isotopes to study microorganisms. & At the left is a pair of dressy straw pumps in a light but crisp texture. & He reached out and felt the bath towel hanging on the towel rack over the tub. \\
\hline
Thus far the advances made have been almost entirely along functional lines. & There was a gigantic wasp next to irving's big top hat. & With her sharp tongue she'd have cut his pompousness to ribbons. \\
\hline
However when labor disputes arise its provisions come clearly into play. & Confusion became chaos each succeeding day brought new acts of violence. & His tough honesty condemns him to a solitary and difficult existence. \\
\hline
Later we shall see what happened when an emperor took this idea too literally. & Thirty five they rode at a measured pace through the valley. & Technical writers can abbreviate in bibliographies. \\
\hline
A clearly recognized exception is a statutory merger or consolidation. & Hiding out like this won't get him anything except more trouble or bullet. & I gave them several choices and let them set the priorities. \\
\hline
The water contained too much chlorine and stung his eyes. & An adult male baboon's teeth are not suitable for eating shellfish. & Begin the examination of a site with a good map and aerial photos if possible. \\
\hline
These exclusive documents must be locked up at all times. & Butterscotch fudge goes well with vanilla ice cream. & Index words and electronic switches may be reserved in the following ways. \\
\hline
Thus there is a clearer division of authority administrative and legislative. & This is a significant advance but its import should not be exaggerated. & Todd placed top priority on getting his bike fixed. \\
\hline
He was above all a friend seeker almost pathetic in his eagerness to be liked. & So we note approvingly a fresh sample of unanimity. & No she would not pretend modesty but neither must she be crudely bold. \\
\hline
Roy ignored the spurious data points in drawing the graph. & Whosoever violates our rooftree the legend states can expect maximal sorrow. & A sailboat may have a bone in her teeth one minute and lie becalmed the next. \\
\hline
The new suburbanites worked hard on refurbishing their older home. & He played basketball there while working toward a law degree. & He had accordingly cultivated eccentricity to the point of second nature. \\
\hline
The frightened child was gently subdued by his big brother. & No manufacturer has taken the initiative in pointing out the cost involved. & The desire and ability to read are important aspects of our cultural life. \\
\hline
A complete plan we have made limited application of the parallel ladder plan. & The public is now armed with sophistication and numerous competing media. & The barracuda recoiled from the serpent's poisonous fangs. \\
\hline
He liked to nip ear lobes of unsuspecting visitors with his needle sharp teeth. & Two miles northeast then five miles southwest that sort of thing. & He picked up nine pairs of socks for each brother. \\
\hline
Brush fires are common in the dry underbrush of nevada. & The dead spirits occupied a prominent place in every hope and in every fear. & Behind him billowed a small pungent cloud of smoke. \\
\hline
Suburban housewives often suffer from the gab habit. & Will you please confirm government policy regarding waste removal. & Perhaps this is what gives the aborigine his odd air of dignity. \\
\hline
Thereupon followed a demonstration that tyranny knows no ideological confines. & Research into several cultures has proven her position to be a mistaken one. & She had jumped away from his shy touch like a cat confronted by a sidewinder. \\
\hline
Faces may be made into candles by filling with melted wax and a wick. & In most discussions of this phenomenon the figures are substantially inflated. & C'mon he whispered four levels about three feet down so don't fall. \\
\hline
The sudden solitude had lost its momentary charm and become oppressive. & Vital questions would be quickly answered according to a preprepared agenda. & The mayan neoclassic scholar disappeared while surveying ancient ruins. \\
\hline
They also furnish proof that in modern war message sending must be monitored. & But problems cling to pools as any pool owner knows. & Personal predispositions tend to blunt the ear and in turn the voice as well. \\
\hline
Only rarely is attention given to accurate progress reports and evaluation. & Both have excellent integration of their fiscal tax collection year calendars. & Need for novelty may be a symptom of cultural fatigue and instability. \\
\hline
The mixing head moves back and forth slowly across the width of the receptacle. & First they wanted to clarify a tantalizing bizarre enigma. & Shell shock caused by shrapnel is sometimes cured through group therapy. \\
\hline
The groundhog clearly saw his shadow but stayed out only a moment. & Two cars came over a crest their chrome and glass flashing. & He really crucified him he nailed it for a yard loss. \\
\hline
The trouble is that like many symbols it doesn't seem a very realistic one. & He ripped down the cellophane carefully and laid three dogs on the tin foil. & The flat bottomed boat swung slowly to the pull of the current. \\
\hline
Scientific progress comes from the development of new techniques. & The cigarettes in the clay ashtray overflowed onto the oak table. & The most recent geological survey found seismic activity. \\
\hline
Men believed they could control nature by obeying a moral code. & He spoke briefly sensibly to the point and without oratorical flourishes. & I'll have a scoop of that exotic purple and turquoise sherbet. \\
\hline
But it must be remembered that the plan should not be oriented geographically. & Come on let's hurry down before they lock up for the day. & The boston ballet overcame their funding shortage. \\
\hline
Yet the spirit which lives in community is not identical with the community. & Wine glass heels are to be found in both high and semi heights. & Naturally no woman can ever completely monopolize the sexual initiative. \\
\hline
The local drugstore was charged with illegally dispensing tranquilizers. & Flying standby can be practical if you want to save money. & Assume for example a situation where a farm has a packing shed and fields. \\
\hline
It made no difference that most evidence points to an opposite conclusion. & Rob made hungarian goulash for dinner and gooseberry pie for dessert. & Shall we flip a coin to see which of us goes first. \\
\hline
Jokes cartoons and cynics to the contrary mothers in law make good friends. & Be careful that you keep adequate coverage but look for places to save money. & But fenced or unfenced no pool side is the place for running or horseplay. \\
\hline
Another field had given him fame enough to satisfy any egotist. & The cowboy's humorous name for a cow givin' milk was a milk pitcher. & For wasp stings onion juice obtained by scraping an onion gave quick relief. \\
\hline
He may have a point in urging that decadent themes be given fewer prizes. & Now don't tell me what a good ball player you are. & They know little about their machinery beyond mechanical details. \\
\hline
The legislature met to judge the state of public education. & He is forced to play for little money and must often take another job to live. & Some have walked through pain and sorrow to bring you their message of hope. \\
\hline
The thick elm forest was nearly overwhelmed by dutch elm disease. & Needless to say my art suffered drastically during this turbulent period. & Once you finish greasing your chain be sure to wash thoroughly. \\
\hline
When peeling an orange it is hard not to spray juice. & They stayed at hotels and boardinghouses or at private homes. & Internal national responsibility now a truism need not be documented. \\
\hline
Ironically enough in this instance such personal virtues were a luxury. & Far more frequently overeating is a result of a psychological compulsion. & Solid concrete blocks relatively heavy and dense are used for this shelter. \\
\hline
The overweight charmer could slip poison into anyone's tea. & Along the main thoroughfares hardly a house had not been peppered. & We experience distress and frustration obtaining our degrees. \\
\hline
The lack of heat compounded the tenant's grievances. & Selecting bunks by economic comparison is usually an individual problem. & Yeah seems so don't it the boy laughed hugging her close. \\
\hline
He showed puny men attacked by splendidly tyrannical machines. & He can for example present significant university wide issues to the senate. & The moisture in my eyes is from eyedrops not from tears. \\
\hline
Your voice is delightful he approved with a warm smile. & Differences were related to social economic and educational backgrounds. & He crossed the next meadow and climbed a tree where the jungle trail resumed. \\
\hline
This truth that the moral law is natural has other important corollaries. & If they are not ellipsoids the conclusions will be a reasonable approximation. & But he was very much like his associates in his hatred of camp routine. \\
\hline
His blue eyes sought the shimmering sea of haze ahead. & The single curve line represents a specific formulation in a test example. & He saw a pint sized man with a graying spade beard and an unusually large head. \\
\hline
The word means it won't boil away easily nothing else. & These air or gas bubbles make highly functional thermal barriers. & The proof that you are seeking is not available in books. \\
\hline
A screwdriver is made from vodka and orange juice. & Unit prices for state vehicles are invariably lower than to the general public. & Every single problem touched on thus far is related to good marketing planning. \\
\hline
While one element is announcing progress another is delineating its problems. & A covered container such as a kitchen garbage pail might do as a toilet. & Why the hell didn't you come out when you saw them gang up on me. \\
\hline
Many wealthy tycoons splurged and bought both a yacht and a schooner. & Chip postponed alimony payments until the latest possible date. & A concept of responsibility is in process of articulation and establishment. \\
\hline
A domestic automatic washer that will give equivalent results may be used. & The compounds are divided according to composition into seven categories. & Draw every outer line first then fill in the interior. \\
\hline
Usually they titter loudly after they have passed by. & Chocolate and roses never fail as a romantic gift. & It was as blissful and fulfilling a night as any bride ever experienced. \\
\hline
Yes indeed we too can see a warlike host of infidels encamped against us. & Lifting her skirts she climbed in never relinquishing her grip on his arm. & Again these blocks were set in resin saturated glass cloth and nailed. \\
\hline
Often they are able to get in only because the area is declining economically. & His prescription hot and cold compresses to increase her absorption of water. & Some make beautiful chairs cabinets chests doll houses etc. \\
\hline
For girls the overprotection is far more pervasive. & When mold has more than one design cavity make individual paper patterns. & One upmanship is practiced by both sides in a total war. \\
\hline
The shot reverberated in diminishing whiplashes of sound. & Some observers speculated that this might be his revenge on his home town. & The tragic stage is a platform extending precariously between heaven and hell. \\
\hline
Bruises and black eyes were relieved by application of raw beefsteak. & She knew she was feeling afraid and inwardly laughed at herself. & However this inaugural feast did its sponsors no good whatever. \\
\hline
Superior new material for orthodontic work is another result of research. & He strolled back to the door whistling softly hands still clasped behind him. & Her eyes were glazed as if she didn't hear or even see him. \\
\hline
She found herself able to sing any role and any song which struck her fancy. & In most cases these soils are taken up as liquids through capillary action. & That doctrine has been accepted by many but has it produced good results. \\
\hline
A smile pulled at the lower strip of adhesive tape. & The orchestra was obviously on its mettle and it played most responsively. & Then he would realize they were really things that only he himself could think. \\
\hline
Space charge influences will also decrease at increased voltages. & She served for a number of years without pay beyond her travel and maintenance. & His superiors had also preached this saying it was the way for eternal honor. \\
\hline
In tradition and in poetry the marriage bed is a place of unity and harmony. & He injected more vitality into the score than it has revealed in many years. & Somehow we old timers never figured we would ever retire. \\
\hline
Sprouted grains and seeds are used in salads and dishes such as chop suey. & Its sphere is that of royal courts dynastic quarrels and vaulting ambitions. & Not without good reason has the anatomical been called jocular journalese. \\
\hline
Very peculiar retribution indeed seems to overtake such jokers. & His portrayal of an edgy head in the clouds artist is virtually flawless. & The theory the idea behind our design is modular units or panelization. \\
\hline
You think somebody is going to stand up in the audience and make guilty faces. & His name became synonymous with cold blooded cruelty. & One of the most common of camp maladies was diarrhoea. \\
\hline
No group of officers came in for more spirited denunciation than the doctors. & We seemed to be witnessing the population explosion right in our own backyards. & Being based on so few events these results are of dubious validity. \\
\hline
For me it has more of both elements than the majority of its competitors. & Adults take a long time to convince and you are thwarted if you try to push. & You could also say that in these pamphlets is a relieving quality of maturity. \\
\hline
But considered within technical astronomy a different pattern can be traced. & Neither are beds thanks to air mattresses and sleeping bags. & Rabies were cured or prevented by madstones which the pioneer wore or carried. \\
\hline
Like a fair number of bootleggers he disliked alcohol. & How much and how many profits could a majority take out of the losses of a few. & And he was handsome despite the long thin scar that slanted across his cheek. \\
\hline
She encouraged her children to make their own halloween costumes. & His first glimpse of the ranch house across the brushy swells told him nothing. & They were shown how to advance against an enemy outpost atop a cleared ridge. \\
\hline
Have a test run on the family first to be sure timing and seasoning are right. & As coauthors we presented our new book to the haughty audience. & To keep 'em scattered somewhat and yet herd 'em was called loose herdin'. \\
\hline
We have also seen the power of faith at work among us. & Last year's gas shortage caused steep price increases. & To use these new ways in daily life is the last step. \\
\hline
We now generalize these ideas for general binomial experiments. & An area sheltered from strong winds may be highly desirable for recreation use. & We would lose our export markets and deny ourselves the imports we need. \\
\hline
A cardboard pattern cut to fit inside holder will help to prevent warping. & These were thought to represent regenerating fibers. & The wrinkled mouth laughed revealing astonishingly strong white teeth. \\
\hline
It seems that open season upon veterans' hospitalization is once more upon us. & Why don't they tell me themselves if it bothers them. & Cory attacked the project with extra determination. \\
\hline
Now you know she could've but she isn't that kind of girl. & It had assumed the terrifying inertia of inanimate matter. & It would be well to show the populace how we deal with adulterers. \\
\hline
In an ideological argument the participants tend to thump the table. & The narrow fringe of sadness that ran around it only emphasized the pleasure. & Worse his present crew included five men who had sailed with him before. \\
\hline
Her face seemed to float in an implausibly bright shaft of sunlight. & Sometimes soldiers wrote letters while bullets were whizzing about their heads. & The thought came back the one nagging at him these past four days. \\
\hline
Manual leveling requires an appropriate display of the accelerometer outputs. & The other patrons were taxi drivers and art students and small shopkeepers. & State numbering laws differ from each other in many ways. \\
\hline
But the information on the dynamics of population was often quite misleading. & We will achieve a more vivid sense of what it is by realizing what it is not. & Originals are not necessarily good and adaptations are not necessarily bad. \\
\hline
The odor here was more powerful than that which surrounded the town aborigines. & She drank greedily and murmured thank you as he lowered her head. & In my place you'd follow such advice as you give me. \\
\hline
I took her word for it but is she really going with you. & Sometimes although by no means always these are indeed alkaline. & Mike was of legal age and presumed able to defend himself in the clinches. \\
\hline
As we observe moral law and physical law they appear as being inevitable. & He was busy he said in having someone submit to a monkey gland operation. & Well i guess i ought to dust out that desk anyhow. \\
\hline
Our hypothetical other bum who killed him would have turned out his pockets. & There should be no reason to misinterpret or ignore the intent of this letter. & No amount of ballyhoo will cover up the sordid facts. \\
\hline
We may say of some unfortunates that they were never young. & The word also made him feel hate sincere hate for those so labeled. & Impressions often appear in a symbolic form and cannot be taken at face value. \\
\hline
There are several sources of evidence on the micrometeorite environment. & Ice baths electric shocks lashings wild dogs testicle crushers. & Caution continuous administration is not recommended for lactating cows. \\
\hline
Ideally he knew it should be preceded by concrete progress at lower levels. & The decking is quarter inch mahogany marine plywood. & The battery median grade equivalent was used in data analysis in this study. \\
\hline
There are optimal humidity requirements for various agents when airborne. & It was the story of the rhinoceros fight all over again. & Soil redeposition is evaluated by washing clean swatches with the dirty ones. \\
\hline
What a discussion can ensue when the title of this type of song is in question. & Meanwhile fishermen took advantage of them to pull up whoppers. & The season between spring and summer belongs to life in its carefree aspect. \\
\hline
Opaque cantaloupe and transparent wood brown were used. & Wooded stream valleys in the folds of earth would be saved. & He didn't tell her the truth he now freely admitted to himself. \\
\hline
Now the problem is presented piecemeal and sometimes contradictorily. & Questions came to me from all sides about my world citizenship activities. & He did not however settle back into acquiescence with things as they were. \\
\hline
For roast insert meat thermometer diagonally so it does not rest on bone. & Again the analyticity of the two curves guarantees that such intervals exist. & Lips pursed mournfully he stared down at its crazily sagging left side. \\
\hline
During one reading an image appeared of a prisoner in irons. & Is there any word you would like to offer in your own defense. & Not very well behaved is she to run out on a play mate. \\
\hline
We did not accept the diagnosis at once but gradually we are coming to. & Another brand of indefinite reference arises out of the use of the double verb. & The failure to keep these two usages distinct presents hazards to the reader. \\
\hline
If you destroy confidence in banks you do something to the economy he said. & New self deceiving rags are hurriedly tossed on the too naked bones. & Somewhere birds were sweetly calling were answered. \\
\hline
This process is especially difficult since gyro drifting is typically random. & Keep your seats boys i just want to put some finishing touches on this thing. & The world is constantly changing what was new yesterday is obsolescent today. \\
\hline
Although my shot killed his horse he rolled off the bale on top of me. & It latches when you close it so stay as long as you like. & Other interpretations present the music as an essentially intimate creation. \\
\hline
She had no way of knowing in advance whether an opportunity for murder existed. & No more startling contrast to a system of sullen satellites could be imagined. & They make gin saws and deal in parts supplies and some used gin machinery. \\
\hline
A chaw of tobacco put on an open wound was both antiseptic and healing. & No antigen was detectable in certain dark spherical areas in most cells. & It was like finally getting into one's own nightmares to punish one's dreams. \\
\hline
One species of ambiguity tries to baffle by interweaving repetition. & Though brief it has a sharp dramatic edge and great poignancy. & The continuing modernization of these forces is a costly but necessary process. \\
\hline
Ran away on a black night with a lawful wedded man. & If you use parking attendants can they be replaced by automatic parking gates. & A monstrous shadow fell across the illuminated wall distorted and indefinable. \\
\hline
It will accommodate firing rates as low as half a gallon an hour. & One of these is the solidarity and the confidential relationship of marriage. & Other morphological physical and optical property values are also given. \\
\hline
She took it grudgingly her dark eyes baleful as they met his. & The highest rated non supervisory engineering title is research engineer. & What had been the ambassador's suite was now jagged walls of blackened brick. \\
\hline
He is not talking in the main about probabilities risks and danger in general. & A kerosene shampoo seems a heroic treatment but it did the job. & They should live in modest circumstances avoiding all conspicuous consumption. \\
\hline
Measured performance characteristics for this experimental tube will be listed. & One of the most desirable features for a park are beautiful views or scenery. & The smell is sexual but so powerfully so that a civilized nose must deny it. \\
\hline
Death reminds man of his sin but it reminds him also of his transience. & He merely said any good decorator these days can make you a tasteful home. & Thinking the evidence insufficient to get a conviction he later released him. \\
\hline
Conservatism and traditionalism seem implied by what has just been said. & Thirty five military and civilian students received laboratory training. & They did not know who they were or know their own worth. \\
\hline
Pretty girls among them with blonde hair and pert faces. & He stared at the far morning expecting a pendulum to swing across the horizon. & Also make sure thermometer does not touch the revolving spit or hit the coals. \\
\hline
If it ever got behind me the beep turned to a buzz. & We flew in rickety planes so overloaded that we wondered why they didn't crash. & Tetanus could be avoided by pouring warm turpentine over a wound. \\
\hline
So if anybody solicits by phone make sure you mail the dough to the above. & But they would reconsider it they assured him if he would rewrite it. & Clapping spurs to the bronc he set off at a sharp canter with growing alarm. \\
\hline
She took it with her wherever she went she chose it. & The old woman arose stiffly and led me to a clearing where a small hut stood. & There was no confirmation of such massive assaults from independent sources. \\
\hline
The walls bulged the floor trembled the windowpanes rattled. & Sewing brings numbness writes what makes my hands numb when sewing. & For sweet sour sauce cook onion in oil until soft. \\
\hline
Running around in the moonlight almost naked and slugging a man with a rock. & He remembered the last time he had eaten actual eggs from an actual pan. & The simplest kind of separate system uses a single self contained unit. \\
\hline
You could burn down this whole mountainside with a fire that size. & There are many such competently anonymous performances among the earlier poems. & If a concessionaire runs the cafeteria keep an eye out for quality and price. \\
\hline
Ralph prepared red snapper with fresh lemon sauce for dinner. & This staff deserves a lot of credit working down here under real obstacles. & Tiny bodies dropped onto a dry leaf made a pile as big as a small apple. \\
\hline
No chemical fertilizers and poisonous insecticides and fungicides are used. & The total of these three volumes is the final combustion chamber volume. & We were off the road gleaming barbed wire pulling taut. \\
\hline
Yet it exists and has an objective reality which can be experienced and known. & They inhabit a secret world centered on go codes and gold phones. & Latest models serve hot meals at reasonable prices and at a profit to you. \\
\hline
Both eventualities are possible logically but practically they are impossible. & The system may break down soon so save your files frequently. & But why is it necessary to reproduce the retinal image within the brain. \\
\hline
He daydreamed on the rock while she swam and splashed around. & Bathing the itching parts with kerosene gave relief and also killed the pests. & There is little doubt that the students benefit from vocational education. \\
\hline
The gunman nodded slipping the picture into his breast pocket saying nothing. & They were not yet prepared to accept it as irremediable. & With any luck at all he could easily find a flowerpot. \\
\hline
Evidence that other sources of financing are unavailable must be provided. & We send shovels cement nails and corrugated iron for roofs. & The avocado should have a give to it as you hold it when it is ripe. \\
\hline
Prompted by a guilty urge he had disobeyed the order of a man he respected. & He will say that our country is even now a homogeneous community. & This girl soon drops the bourgeoisie psychiatrist who disapproves of her life. \\
\hline
He slipped outside hugging the walls of buildings and dodging into doorways. & And men also used vacuum cleaners in both rooms sucking dust up once more. & The dimensions of these waves dwarf all our usual standards of measurement. \\
\hline
Promptly at seven he would clatter out of the court with twelve in the tallyho. & Now a distinguished old man called on nine divinities to come and join us. & He hoped they would put in somewhere way way down in the earth. \\
\hline
The population can thereby replenish itself and actually grow larger. & They offered no opinions volunteered nothing betrayed no emotions. & These needs usually concern the reduction of guilt and some relief of tension. \\
\hline
Microorganisms are often responsible for the rapid spoilage of foods. & One of the problems associated with the expressway stems from the basic idea. & Only incomplete imperfect things move towards what they lack. \\
\hline
We'll both be blowing town tomorrow so we won't be moving in on you. & It required an energy he no longer possessed to be satirical about his father. & No question ruffles him or causes him to hesitate. \\
\hline
Flaxseed poultices and mustard plasters still are used by some persons. & All that time rifle barrels were pointing unwaveringly at his head and body. & Steam baths writes do steam baths have any health value. \\
\hline
In time she presents her aristocratic husband with a coal black child. & The straight line would symbolize its uniqueness the circle of universality. & Even then if she took one step forward he could catch her. \\
\hline
Would have been easy to identify as opium by its odor. & This is a problem that goes considerably beyond questions of salary and tenure. & Quince seed gum is the main ingredient in wave setting lotions. \\
\hline
Half slyly he enjoyed seeing her stoop to lift the things. & We hope that he will execute it in a manner that will entitle him to credit. & A site may also be attractive just through the beauty of its trees and shrubs. \\
\hline
Her face turned pink with pleasure and a smothered cough. & Would a camera club be useful in taking pictures pertinent to plant safety. & Family loyalties and cooperative work have been unbroken for generations. \\
\hline
Would a blue feather in a man's hat make him happy all day. & He reached once more into the carpet bag and brought up a package of wieners. & There was a grunt curiously inarticulate like that of an animal in pain. \\
\hline
Insulate weatherstrip double glaze to the maximum. & He wanted to show the town what happened to anyone who tried to start trouble. & Ants carry the seeds so better be sure that there are no ant hills nearby. \\
\hline
You should firmly insist that no bobby pins or hair pins be worn in the water. & In some measure they depend upon the structure of individual personality. & We congratulate the entire membership on its record of good legislation. \\
\hline
On unoccupied roadway the bottle shattered into a small amber flash. & The name fell with lazy affectionate remembrance from her lips. & When he finally did he had to duck his head quickly away as the pitch came in. \\
\hline
Afraid you'll lose your job if you don't keep your mouth shut. & But to the infuriation of scientists for no known reason not all of them did. & Meanwhile the enemy will capitalize on our fears if he can. \\
\hline
Further it has its work cut out stopping anarchy where it is now garrisoned. & It has multiple implications and possible headaches for your marketing program. & One more muddleheaded play like that one and they'd be leading him away. \\
\hline
Traffic frequently has failed to measure up to engineers' rosy estimates. & Ambiguity arises when the pronoun it carries a twofold reference. & However the aircraft which we have today are tied to large soft airfields. \\
\hline
We know that actors can learn to portray a wide variety of character roles. & The ward was a small one four beds kept reserved for female alcoholics. & The sculptor looked at him bugeyed and amazed angry. \\
\hline
But to continue to divorce advanced students from reality is inexcusable. & But this doesn't detract from its merit as an interesting if not great film. & Then may i ask where these muddy foot prints came from. \\
\hline
In the lighted interior he saw other men and women struggling into their wraps. & Her debut over perhaps the earlier scenes will emerge equally fine. & He straightened up alert now as the buffalo hunter came closer. \\
\hline
The crisis later on when debts seemed about to overwhelm me. & Something else distracted him yet there was no sound only tomblike silence. & Let the orthodontist decide the proper time to start treatment he urges. \\
\hline
Forty seven states assign or provide vehicles for employees on state business. & This he added brought about petty jealousies and petty personal grievances. & It is possible to make a few generalizations about the six giants themselves. \\
\hline
This has been attributed to helium film flow in the vapor pressure thermometer. & As his feet slowed he felt ashamed of the panic and resolved to make a stand. & Perhaps one bored holes in the stone with some kind of an electric gadget. \\
\hline
The keelson made of two three inch widths is next installed. & Cleaned cloth must be protected against the redeposition of dispersed soil. & Brief snips of actual events were shown parades dances street scenes. \\
\hline
His black hat with its wide brim high crown and fur trim rode high. & Get copper or earthenware mugs that keep beer chilled or soup hot. & Two clotted balls the color of mucus rolled between fiery lids. \\
\hline
Pretend ham make criss cross gashes on one side of skinless frankfurters. & The platform accelerometers must be slightly modified for this procedure. & They are not true because scientists or prophets say they are true. \\
\hline
The batting average of one success out of seven increased to one out of three. & And his relatively small hands and feet gave him an almost delicate appearance. & But from the start they had two important ingredients sincerity and realism. \\
\hline
From it spokes of order and degree led to the outward rim of the common man. & Resolved that the anti slavery sentiment is becoming ripe for resolute action. & But if she wasn't interested she'd just go back to the same life she'd left. \\
\hline
He looked over at him lying there asleep and he felt a wave of revulsion. & In the winter hibachi in the kitchen or grill over the logs of the fireplace. & Higher toll rates also are helping boost revenues. \\
\hline
He would offer no theory to account for her murder. & He drove sensual patterns off carefully shaving his long upper lip. & More often these offices are restricted to the gathering of empirical data. \\
\hline
In these damp circumstances he was an odds on bet to develop pneumonia. & The problem of solidarity and morale again involves the concept of values. & Maybe you and me will girlie but these two ain't goin' nowhere. \\
\hline
Was it a hysterical release from the long strain of vigilance of those weeks. & Both the conditions and the complicity are documented in considerable detail. & Of particular importance is the study of the actions of drugs in this respect. \\
\hline
Advantages a farm provides a wholesome and healthful environment for children. & His eyes were dark fluid fearful and he gave a sigh as my knife went in. & It does not indicate loose management ineffective controls or poor policy. \\
\hline
These curves were derived by an analysis of extensive skywave measurement data. & Don't they still call you junior as though you're about ten years old. & Fall slowly forward onto the hands and let the body down to rest on the floor. \\
\hline
Except for those minutes in her room he had lost touch with her as a reality. & Replace it with the statue of one or another of the world's famous dictators. & The need for reupholstering redecorating repainting becomes more infrequent. \\
\hline
The beatniks crave a sexual experience in which their whole being participates. & So somebody else knew what would happen to her father's money if she died. & If the other pilots were worried they did not show it. \\
\hline
Our first necessity at the very outset of war is post attack reconnaissance. & To some extent predispositions are shaped by exposure to group environments. & In earlier years the preservation of food was essentially related to survival. \\
\hline
Was she just naturally sloppy about everything but her physical appearance. & Sometimes strong stress serves to focus an important secondary relationship. & Hired hard lackeys of the warmongering capitalists. \\
\hline
He holds that goodness and badness lie in feelings of approval or disapproval. & He murmured to himself with firmness no surrender. & We scour literature for them here we find stored the wisdom of great minds. \\
\hline
And these hardy travelers are not unappreciated today. & The revolution now under way in materials handling makes this much easier. & With those paintings of big constructions crashing down he felt he could stop. \\
\hline
He didn't figure her at all and if he found out a woman it'd be bad. & It offered to surrender its right to exclusive trade but asked an indemnity. & He may not rise to the heights but he can get by and eventually be retired. \\
\hline
Her position covers a number of daily tasks common to any social director. & A flash illumined the trees as a crooked bolt twigged in several directions. & Biological warfare is considered to be primarily a strategic weapon. \\
\hline
They moved toward the skiffs with shocking eagerness elbowing and shoving. & Radio reception is cut down by the shielding necessary to keep out radiation. & Then he fled not waiting to see if she minded him or took notice of his cry. \\
\hline
Avowed atheists or freethinkers are so rare as to be a curiosity. & This big flexible voice with uncommon range has been superbly disciplined. & She greeted her husband's colleagues with smiling politeness offering nothing. \\
\hline
Come sit he repeated motioning to the piled hay bags over the pig leavings. & Before that we lumber dealers were working almost single handed on the problem. & She smiled and teeth gleamed in her beautifully modeled olive face. \\
\hline
Meanwhile three great terrible forces were coagulating and crystallizing. & Residential associations struggle to insulate themselves against intrusions. & You'd think her stomach would've got used to it in three weeks. \\
\hline
Not good looking but self confident and wise so that it made her attractive. & Roleplaying used for analysis follows these general steps leading to training. & To prepare mustard cream blend mustard with enough water to make a thin paste. \\
\hline
After another long pause he asked how many people know who they are. & Within a system however the autonomy of each member library is preserved. & The old shop adage still holds a good mechanic is usually a bad boss. \\
\hline
Beer generally fermented from barley is an old alcoholic beverage. & When you're less fatigued things just naturally look brighter. & Quite often honeybees form a majority on the willow catkins. \\
\hline
They were both walking towards each other unhurried. & He must rearrange matters so that two performers do not bump into each other. & Wet also were the marine's fatigues and the face had an oily film. \\
\hline
They weren't as well paid as they should have been. & It could take place tomorrow night or it might occur months from now. & When we left washington his son tad was ill and mrs lincoln hysterical. \\
\hline
The long and ever increasing column of sportsmen is now moving into a new era. & The bacteria formed typical activated sludge floc. & The filtered air benefits allergies asthma sinus hay fever. \\
\hline
Are planning and strategy development emphasized sufficiently in your company. & His talk turns to what he calls the mess or sometimes this buzzing confusion. & Conceivably the submarine defense problem can be solved by sufficient forces. \\
\hline
Two women who had been chattering like parrots were struck dumb. & In news items a man is less often shot in the body or head than in the suburbs. & This is going to be a language lesson and you can master it in a few minutes. \\
\hline
He doesn't want her to look frowningly at him or speak to him angrily. & But briefly the topping configuration must be examined for its inferences. & A vigorous program existed in skiing skating sports and overnight hiking. \\
\hline
This is the second year your mother's donated my fishing tackle to the bazaar. & When did women begin to assert themselves sexually. & The good man chortled appreciatively and decided the trip was worth his time. \\
\hline
This explains some group ends and provides a justification of their primacy. & Computers are being used to keep branch inventories at more workable levels. & Large injection molded letters are also available for sign installations. \\
\hline
The bristles are soft enough to massage the gums and not scratch the enamel. & There are canoes ideal for fishing in protected waters or for camping trips. & They'll move around that rock all day following the shade. \\
\hline
It takes a great deal of sophisticated thought to get the impact of this fact. & His problem concerns longitudes latitudes and angular velocities. & Did you know he is advertising his ham radio equipment for sale this weekend. \\
\hline
The wheel of social life spun around the royal or aristocratic centre. & But this esoteric doctrine was lost in the shuffle to acquire special powers. & We do not arrive at spatial images by means of the sense of touch by itself. \\
\hline
There are people who travel long distances to assure my continued existence. & The same shelter could be built into an embankment or below ground level. & Everything in the final analysis reduced itself to sexual symbolism. \\
\hline
Is a relaxed home atmosphere enough to help her outgrow these traits. & The elementary school child grows gradually in his ability to work in groups. & We're lost and burning up already she bit out tensely. \\
\hline
Probably around midnight give or take an hour either way. & She saw me and sat down beside me three feet away. & But i'm so sunburned that every move i make is agony. \\
\hline
Closer still regular barricades of barbed wire hung on timber supports. & Cattle which died from them winter storms were referred to as the winter kill. & Hiring the wife for one's company may win her tax aided retirement income. \\
\hline
Paperweight may be personalized on back while clay is leather hard. & Blowers should be operated periodically on a regular schedule. & There is definitely some ligament damage in his knee. \\
\hline
Another stock vaudeville gag ran mother is home sick in bed with the doctor. & He enlisted a staff of loyal experts and of many zealous volunteers. & We often say of a person that he looks young for his age or old for his age. \\
\hline
It cost us a hundred thousand dollars and thirty days lost time to fix them. & Or borrow some money from someone and go home by bus. & Almonds and pistachio nuts are not so high in oil but are rich in protein. \\
\hline
Castor oil made from castorbeans has gone out of style as a medicine. & He went on to personal bequests a list of names largely unknown to him. & This is no assignment for a frivolous girl she assures him. \\
\hline
The stepmother almost without exception has been presented as a cruel ogress. & You're not living up to your own principles she told my discouraged people. & He'd had no idea how unhappy his sweet peach had been. \\
\hline
Women didn't use white face powder nowadays he recalled. & I always say you've got a wonderful husband miss margaret. & It is one of the rare public ventures here on which nearly everyone is agreed. \\
\hline
The larvae kept warm by the queen are full grown in about ten days. & She always did before and showed the utmost confidence in whatever we advised. & As he had longed to be he became the echo of a saga. \\
\hline
This viscosity of the material in the drops is of course not negligible. & High so it only bounce harmlessly but loudly off a car's steel roof. & Maybe it's taking longer to get things squared away than the bankers expected. \\
\hline
Can your insurance company aid you in reducing administrative costs. & One can only speak of what is in front of him and that now is simply the mess. & Push ups push ups are essential but few have the strength for them at first. \\
\hline
In simpler terms it amounts to pointing the platform in the proper direction. & From this motor pool personnel develop other meaningful and related data. & This one came a bit high at thirty thousand or more. \\
\hline
Most of our aid will go to those nearing self sufficiency. & The shock therapies act likewise on the hypothalamic balance. & The knifelike pain in his groin nearly brought him down again. \\
\hline
To be passive to be girlishly shy was palpably absurd. & Look for these features which may mean you can save duplicate coverage. & When he left she knew she would never see him again. \\
\hline
Under this law annual grants are given to systems in substantial amounts. & But our stumping tour of the south wasn't all misery. & My beloved ward my perennial gadfly said the whining voice. \\
\hline
His sinuous melody is a sort of naive transcendence of all experience. & Anybody carrying anything that might hide a rifle. & Woe betide the interviewee if he answered vaguely. \\
\hline
Program note reads as follows take hands this urgent visage beckons us. & In either case they do not appreciate the private detective's zeal. & X ray films of the vertebral column showed progression of the demineralization. \\
\hline
Microscopically there was emphysema fibrosis and vascular congestion. & In a new house generous roof overhangs are a logical and effective solution. & Are you utilizing vending machine proceeds to help pay for your program. \\
\hline
This is what necessitates the nonsystematic character of his astronomy. & Individual human strength is needed to pit against an inhuman condition. & Knows the score with a supreme effort he broke it off. \\
\hline
Why couldn't they have dumped him off on someone else. & His successors have adopted the opposite alternative. & Crooked overlapping twisted or widely spaced teeth. \\
\hline
His artistic accomplishments guaranteed him entry into any social gathering. & They played crack the whip a few minutes without mishap. 
 &  \\
\hline
\caption{728 randomized scripted prompts for version 1.0 and 2.0, from which each participant only responded to 10.} \\

\end{longtable}
\end{footnotesize}

\end{document}